\documentclass{article}

\usepackage{hyperref}
\usepackage{lifetime}
\usepackage{url}
\usepackage{tikz}
\usetikzlibrary{decorations.pathmorphing}
\usepackage{xcolor}

\newcommand{\lsbench}{\textsc{RLSbench}}
\newcommand{\rlsbench}{\textsc{RLSbench}}

\definecolor{cobalt}{rgb}{0.0, 0.28, 0.67}
\definecolor{carnelian}{rgb}{0.7, 0.11, 0.11}

\definecolor{bole}{rgb}{0.47, 0.27, 0.23}

\newcommand{\update}[1]{{\color{black}#1}}

\usepackage[accepted]{icml2023}

\usepackage[capitalize,noabbrev]{cleveref}

\begin{document}

\twocolumn[
\icmltitle{
RLSbench: Domain Adaptation Under Relaxed Label Shift
}

\begin{icmlauthorlist}
\icmlauthor{Saurabh Garg}{cmu}
\icmlauthor{Nick Erickson}{aws}
\icmlauthor{James Sharpnack}{aws}
\icmlauthor{Alex J.~Smola}{aws}
\icmlauthor{Sivaraman Balakrishnan}{cmu}
\icmlauthor{Zachary C.~Lipton}{cmu,aws}
\end{icmlauthorlist}

\icmlaffiliation{aws}{Amazon Web Services}
\icmlaffiliation{cmu}{Carnegie Mellon University}

\icmlcorrespondingauthor{Saurabh Garg}{sgarg2@andrew.cmu.edu}

\icmlkeywords{domain adaptation, distribution shift, label shift, large scale study}

\vskip 0.3in
]

\printAffiliationsAndNotice{}  %
\begin{abstract}

Despite the emergence of principled methods 
for domain adaptation under label shift, 
their sensitivity to shifts in class conditional distributions is
precariously under explored. 
Meanwhile, popular deep domain adaptation heuristics
tend to falter when faced with label proportions shifts. 
While several papers modify these heuristics 
in attempts to handle label proportions shifts,
inconsistencies in evaluation standards, datasets, and baselines 
make it difficult to gauge the current best practices.
In this paper, we introduce \textsc{RLSbench}, 
a large-scale benchmark for \emph{relaxed label shift},
consisting of $>$500 distribution shift pairs 
spanning vision, tabular, and language modalities, 
with varying label proportions.  
Unlike existing benchmarks,
which primarily focus on shifts in class-conditional $p(x|y)$, 
our benchmark also focuses on label marginal shifts. 
First, we assess \update{13 popular domain adaptation methods},
demonstrating more widespread failures
under label proportion shifts than were previously known.
Next, we develop an effective two-step meta-algorithm
that is compatible with most domain adaptation heuristics:
(i) \emph{pseudo-balance} the data at each epoch;
and (ii) adjust the final classifier
with target label distribution estimate.
\update{The meta-algorithm 
improves existing domain adaptation heuristics
under large label proportion shifts,
often by 2--10\% accuracy points,
while conferring minimal effect ($<$0.5\%) 
when label proportions do not shift.}
We hope that these findings and the availability of \textsc{RLSbench}
will encourage researchers to rigorously evaluate proposed methods 
in relaxed label shift settings. 
{Code is publicly available at \url{https://github.com/acmi-lab/RLSbench}}.

\end{abstract}

\section{Introduction}

Real-world deployments of machine learning models are 
typically characterized by distribution shift, 
where data encountered in production exhibits statistical differences 
from the training data~\citep{quinonero2008dataset, 
torralba2011unbiased, wilds2021}. 
Because continually labeling data can be prohibitively expensive,
researchers have focused on the unsupervised Domain Adaptation (DA) setting,
where only labeled data from the \emph{source} distribution
and unlabeled from the \emph{target} distribution are available for training.
Absent further assumptions, the DA problem 
is known to be underspecified \citep{ben2010impossibility}
and thus no method is universally applicable.

Researchers have responded to these challenges in several ways.
One approach is to investigate structural assumptions
under which DA problems are well-posed.
Popular examples include covariate shift and label shift,
for which identification strategies and principled methods exist
whenever the source and target distributions
have overlapping support~\citep{shimodaira2000improving, scholkopf2012causal, gretton2009covariate}.
For example, recent research on label shift
has produced effective methods 
that are applicable in deep learning regimes, 
yielding both consistent estimates 
of the target label marginal 
and principled ways to update the resulting classifier~\citep{lipton2018detecting, alexandari2019adapting, azizzadenesheli2019regularized, garg2020labelshift}.
However, such assumptions are typically, 
to some degree, violated in practice.
Even for archetypal cases like shift 
in disease prevalence,
the label shift assumption can be violated. 
For example, over the course of the COVID-19 epidemic,
changes in disease positivity coincided
with shifts in 
treatment protocols, 
the age distribution of the infected population,
and the genetic makeup of the virus itself.

\begin{figure*}[t!]
    \centering 
    \begin{subfigure}[b]{0.58\linewidth}
    \includegraphics[width=\linewidth]{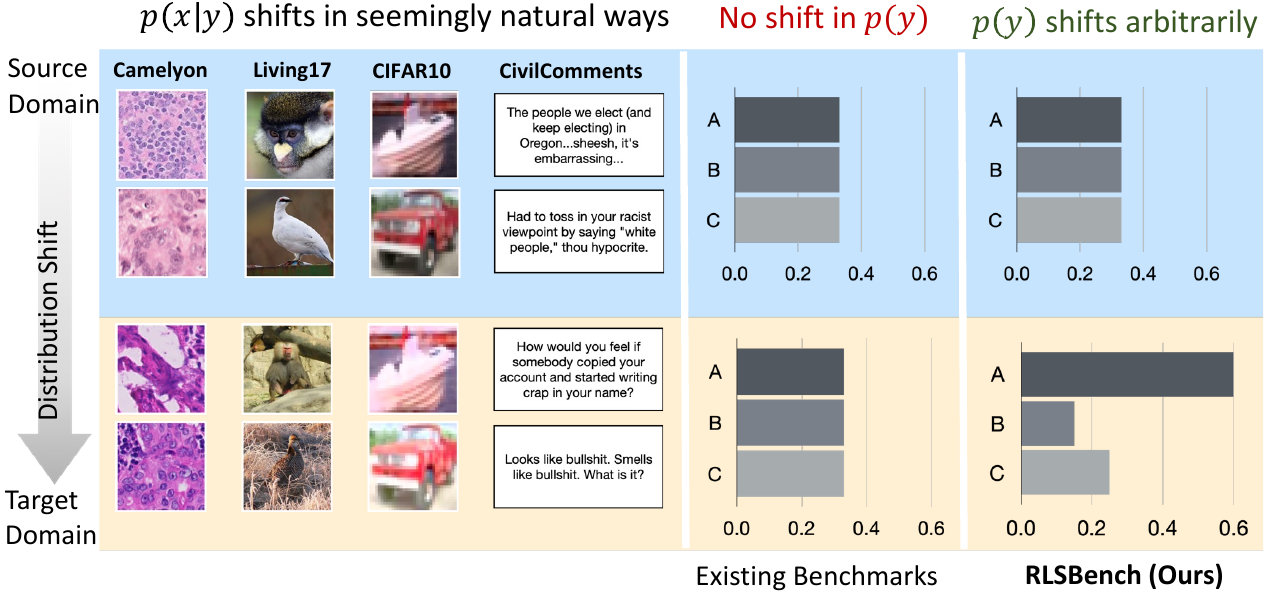} 
    \caption{}
    \end{subfigure}
    \hfill
    \begin{subfigure}[b]{0.39\linewidth}
    \includegraphics[width=\linewidth]{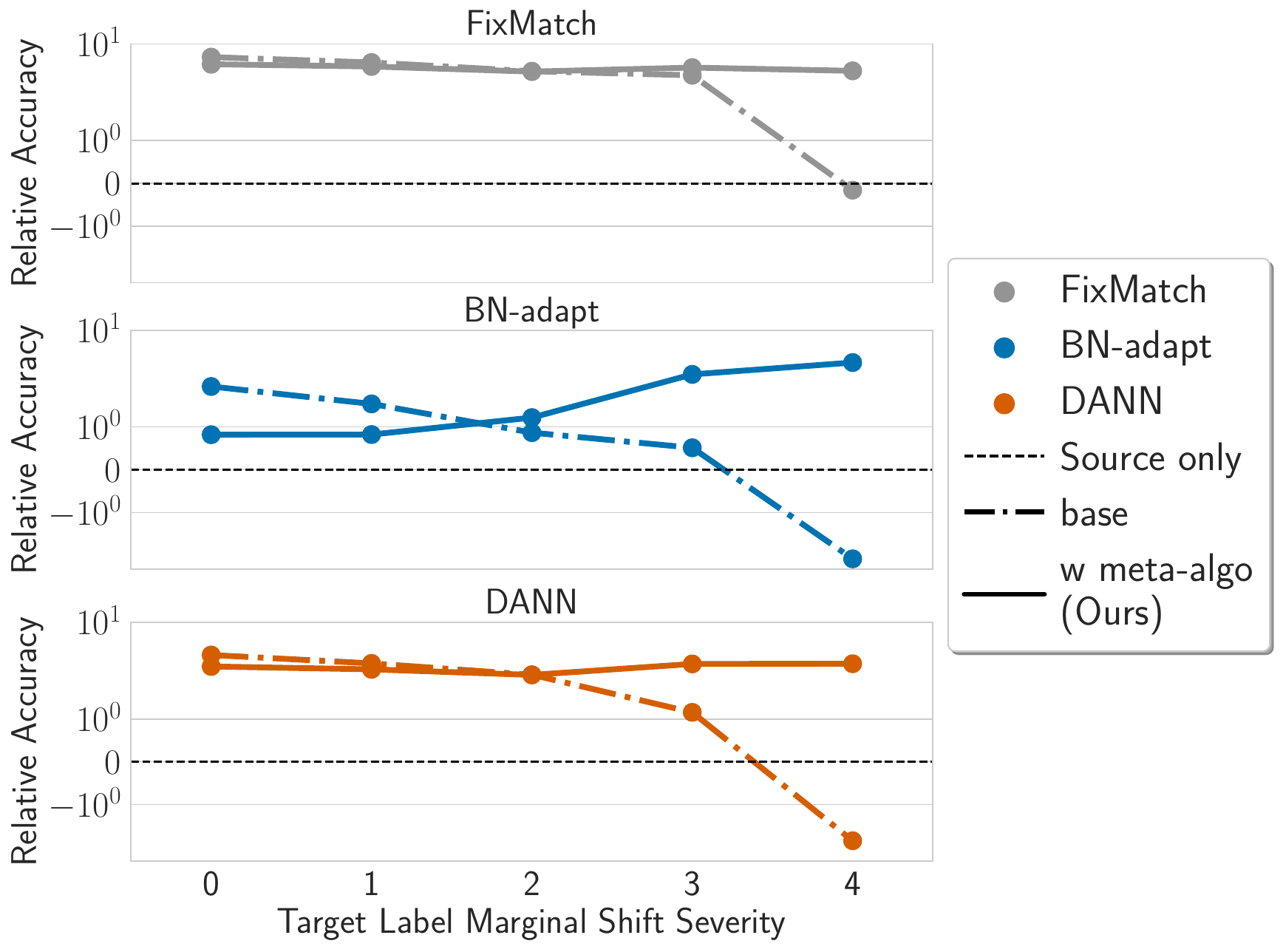}
    \caption{}
    \end{subfigure}
    \vspace{-10pt}
    \caption{\emph{Domain adaptation under Relaxed Label Shift.} \textbf{(a) Overview of RLSbench setup:} Unlike existing benchmarks for which the label marginal $p(y)$ doesn't shift, in RLSbench, $p(y)$ can shift arbitrarily. The class conditionals $p(x|y)$ shift in seemingly natural ways following popular benchmarks. RLSbench draws on 14 multi-domain datasets spanning vision, NLP, and tabular modalities. \textbf{(b) Key results:} As the severity of target label proportion increases, the performance of existing popular DA methods degrades, often dropping below source-only classifiers. DA methods, when paired with our meta-algorithm, significantly improve over a source-only classifier.}
    \label{fig:intro}
\end{figure*}

A complementary line of research focuses on 
constructing benchmark datasets,
in the hopes of finding heuristics   
for incorporating the unlabeled target data 
that result in improvements on the 
kinds of problems that arise in practice. 
Examples of such benchmarks include
OfficeHome~\citep{venkateswara2017deep}, 
Domainnet~\citep{peng2019moment}),
WILDS~\citep{sagawa2021extending}. 
However, most of these benchmark datasets
exhibit little shift in the label distribution $p(y)$
(or none at all).
Consequently, benchmark-driven research 
has produced a variety of heuristic methods
\citep{ganin2016domain, sohn2020fixmatch, wang2021tent, li2016revisiting}
that despite yielding gains in benchmark performance
tend to break when $p(y)$ shifts.
While this vulnerability has previously been demonstrated
for domain-adversarial methods \citep{wu2019domain, zhao2019learning},
we show that this problem is more widespread than previously known.
Several recent papers attempt to address shift in label distribution
compounded by natural variations in $p(x|y)$~\citep{tan2020class, tachet2020domain, prabhu2021sentry}.
However, it can be hard to compare experimental results across papers,
owing to discrepancies in how shifts in $p(y)$ are simulated
and the choice of evaluation metrics.
Moreover, many methods violate the unsupervised contract
by peeking at target validation performance during model selection
and hyperparameter tuning~\citep{wilson2020survey, saito2021tune}. 
\update{In short, there is a paucity of 
comprehensive and fair comparisons between DA methods 
for settings with shifts in label distribution.}

In this paper, we develop \textsc{RLSbench}, the first {standardized} test bed of \emph{relaxed label shift} settings,
where $p(y)$ can shift arbitrarily and the class conditionals 
$p(x|y)$ can shift in seemingly natural ways 
(following the popular DA benchmarks). 
While existing DA benchmarks typically focus on shifts in 
$p(x|y)$, our benchmarks additionally focuses on shifts in label marginals $p(y)$. 
We evaluate a collection of popular DA methods
based on domain-invariant representation learning, 
self-training, and test-time adaptation  
across \update{$14$ multi-domain datasets spanning vision, Natural Language Processing (NLP), and tabular modalities}. 
The different domains in each dataset
present a different shift in $p(x|y)$.
Since these datasets 
exhibit minor to no shift in  
label marginal, we simulate shift in 
target label marginal via stratified sampling with 
varying severity. Overall, 
we obtain \update{$560$ different
source and target distribution shift 
pairs and train $>30k$ models in our testbed.}

\update{Based on our experiments on \textsc{RLSbench}, we make several findings.}
First, we observe that while popular DA methods
often improve over a source-only classifier 
absent shift in target label distribution, %
their performance tends to degrade, 
\update{dropping below source-only classifiers
under severe shifts in target label marginal}.
\update{Next, we develop a meta-algorithm with two simple corrections}: (i) re-sampling the data to balance the 
source and pseudo-balance the target; 
(ii) re-weighting the final classifier 
using an estimate of the target label marginal. 
We observe that in these relaxed label shift settings,
the performance of \update{existing DA methods (e.g. CDANN, FixMatch, and BN-adapt), when paired with our meta-algorithm, significantly improves over a source-only classifier}. 
\update{On the other hand, existing methods specifically proposed 
for relaxed label shift (e.g., IW-CDANN and SENTRY),
often fail to improve over a source-only classifier 
and significantly underperform when compared to
existing DA methods paired with our meta-algorithm.}

Overall, RLSbench
provides a comprehensive and standardized suite
for label distributions shifts,
bringing existing benchmarks one step closer 
to exhibit the sort of diversity
that we should expect to encounter
when deploying models in the wild.
Our findings emphasize the effectiveness of a simple, previously overlooked baseline.
We hope that the $\rlsbench$ and our meta-algorithm (that can be paired with any DA method) provide a framework for 
rigorous and reproducible future research in relaxed label shift scenarios.

\vspace{-5pt}
\section{Preliminaries and Prior Work} \label{sec:related} %
We first setup the notation and formally define the problem. Let $\calX$ be the input space and 
$\calY = \{1,2,\ldots, k\}$ the output space. 
Let $\ProbS, \ProbT : \inpt\times\out \to [0,1]$
be the source and target distributions
and let $\ps$ and $\pt$ denote the corresponding 
probability density (or mass) functions.
Unlike the standard supervised setting, 
in unsupervised DA, 
we possess labeled source data 
$\{(x_1,y_1), (x_2, y_2), \ldots , (x_n, y_n)\}$ 
and unlabeled target data $\{x_{n+1}, x_{n+2}, \ldots, x_{n+m}\}$. 
With $f: \calX \to \Delta^{k-1}$, we denote a predictor  
function which predicts $\wh y = \argmax_y f_y(x)$ on an input $x$. 
For a vector $\vv$, we use $\vv_y$ 
to access the element at index $y$.

In the traditional label shift setting, one assumes
that $p(x|y)$ does not change 
but that $p(y)$ can.
Under label shift, two challenges arise:
(i) estimate the target label marginal $p_t(y)$; 
and (ii) train a classifier $f$ 
to maximize the performance on target domain. 
This paper focuses on the \emph{relaxed label shift} setting. 
In particular, we assume that the label distribution
can shift from source to target arbitrarily 
but that $p(x|y)$ varies between source and target
in some comparatively restrictive way \update{(e.g., shifts arising naturally in the real-world like ImageNet~\citep{russakovsky2015imagenet} to ImageNetV2~\citep{recht2019imagenet})}. 
\update{Mathematically, 
we assume a divergence-based restriction on 
$p(x|y)$. That is, for some small $\epsilon > 0$ and distributional distance $\calD$, we have 
$\max_y  \calD(p_s(x|y), p_t(x|y)) \le \epsilon$
and allow an arbitrary shift in the label marginal $p(y)$.
We discuss several precise instantiations in \appref{app:def}. %
However, in practice, it's hard to empirically verify these distribution distances for small enough $\epsilon$ with finite samples. Moreover, we lack a rigorous characterization of the sense in which those shifts arise in popular DA benchmarks, and since, the focus of our work is on the empirical evaluation with real-world datasets, 
we leave a formal investigation for future work.
}

The goal in DA is 
to adapt a predictor
from a source distribution with labeled data
to a target distribution from which 
we only observe unlabeled examples. 
While prior work addressing relaxed label shift 
has primarily focused on classifier performance, 
we also separately evaluate methods for 
estimating the target label marginal.
This can be beneficial for two reasons.
First, it can shed more light 
into how improving the estimates 
of target class proportion improves target performance. 
Second, understanding how the class proportions are changing 
can be of independent interest.

\subsection{Prior Work}

\textbf{Unsupervised domain adaption~~} 
Two popular settings for which DA is well-posed include 
(i) \emph{covariate shift}~\citep{zhang2013domain,zadrozny2004learning,cortes2010learning,cortes2014domain,gretton2009covariate}
where $p(x)$ can change from source to target but 
$p(y|x)$ remains invariant; 
and (ii) \emph{label shift}~\citep{saerens2002adjusting, lipton2018detecting, azizzadenesheli2019regularized, alexandari2019adapting, garg2020labelshift, zhang2020coping, roberts2022LLS} 
where the label marginal $p(y)$ can change 
but $p(x|y)$ is shared across source and target. 
Principled methods with strong theoretical guarantees exists
for adaptation under these settings 
when target distribution's support 
is a subset of the source support.
\citet{ben2010impossibility, ben2010theory, mansour2009domain, zhao2019learning, wu2019domain, pmlr-v89-johansson19a} 
present theoretical analysis when 
the assumption of contained co-variate support is violated. 
In another line of work, 
\citet{elkan2008learning, pusurvey, garg2021PUlearning, garg2022OSLS}
extend the label shift setting 
to problems where previously unseen classes may
appear in the target and $p(x|y)$ remains invariant among seen classes.  
More recently, a massive literature has emerged
exploring a benchmark-driven heuristic approach~\citep{long2015learning,long2017deep, sun2016deep, sun2017correlation, zhang2019bridging, zhang2018collaborative, ganin2016domain, sohn2020fixmatch}.  
However, rigorous evaluation of DA 
methods is typically restricted to 
these carefully curated 
benchmark datasets where 
their is minor to no shift in
label marginal from source to target.

\textbf{Relaxed Label Shift ~}
Exploring the problem of 
shift in label marginal from 
source to target 
with natural variations in $p(x|y)$, 
a few papers highlighted 
theoretical and empirical failures  
of DA methods based on 
domain-adversarial neural network  
training~\citep{yan2017mind,
wu2019domain,zhao2019learning,pmlr-v89-johansson19a}. 
Subsequently, several papers
attempted to handle
these problems in 
domain-adversarial training~\citep{tachet_domain_2020, 
prabhu2021sentry, liu_adversarial_2021,
tan2020class, manders_adversarial_2019}. 
However, these methods often lack comparisons 
with other prominent DA methods and are evaluated 
under different datasets and model selection criteria. 
To this end, we perform a large scale rigorous comparison 
of popular representative DA methods 
in a standardized evaluation framework.

\textbf{Domain generalization~~} 
In domain generalization, the model is given access to data from
multiple different domains and the goal is to generalize to a previously 
unseen domain at test time~\citep{blanchard2011generalizing, muandet2013domain}. For a survey of different algorithms for 
domain generalization, we refer the reader to \citet{gulrajani2020search}. A crucial distinction here  
is that unlike the domain generalization setting, 
in DA problems, we have access
to unlabeled examples from the test domain.

\textbf{Distinction from previous distribution shift benchmark studies~~}
Previous studies evaluating robustness under distribution shift predominantly focuses on transfer learning and domain generalization settings~\citet{wenzel2022assaying, gulrajani2020search, djolonga2021robustness,wiles2021fine, wilds2021}.
\citet{taori2020measuring, hendrycks2021many} studies the impact of robustness interventions (e.g. data augmentation techniques, adversarial training) on target (out of distribution) performance. 
Notably, \citet{sagawa2021extending} focused on evaluating DA methods on WILDS-2.0.
Our work is complementary to these studies, as we present the first extensive study of 
DA methods under shift in $p(y)$ and natural variations in $p(x|y)$.

\section{\textsc{RLSbench}: A Benchmark for Relaxed Label Shift}\label{sec:LSBench} %
In this section, we introduce $\lsbench$, a suite of datasets 
and DA algorithms that are at the core of our study. 
Motivated by correction methods for the (stricter) label shift setting~\citep{saerens2002adjusting,lipton2018detecting} and 
learning under imbalanced datasets~\citep{wei2021crest, cao2019blearning},
we also present \update{a meta-algorithm with simple corrections compatible with almost any DA method}.

\subsection{Datasets}

\textsc{RLSbench} builds on $14$ multi-domain datasets for classification, including tasks across applications in object classification, satellite imagery, medicine, and \update{toxicity detection}. Across these datasets, we obtain 
a total of \update{$56$ different source and target pairs}. 
More details about datasets are 
in \appref{app:datasets}. 

(i) \textbf{CIFAR-10} which includes the original CIFAR-10~\citep{krizhevsky2009learning}, CIFAR-10-C~\citep{hendrycks2019benchmarking} and CIFAR-10v2~\citep{recht2018cifar}; 
(ii) \textbf{CIFAR-100} including the original dataset and CIFAR-100-C; 
(iii) all four BREEDs datasets~\citep{santurkar2020breeds}, i.e., 
\textbf{Entity13}, \textbf{Entity30}, \textbf{Nonliving26}, \textbf{Living17}. 
BREEDs leverages class hierarchy in ImageNet~\citep{russakovsky2015imagenet} to repurpose original classes to be the subpopulations and define a classification task on superclasses. 
We consider subpopulation shift
and natural shifts induced due to differences in the data collection process of ImageNet, i.e, ImageNetv2~\citep{recht2019imagenet} and a combination of both. 
(iv) \textbf{OfficeHome}~\citep{venkateswara2017deep} which includes four domains: 
art, clipart, product, and real;
(v) \textbf{DomainNet}~\citep{peng2019moment} where we consider four domains: clipart, painting, real, sketch; 
(vi) \textbf{Visda}~\citep{peng2018syn2real, visda2017} which contains three domains: train, val and test;
(vii) \textbf{FMoW}~\citep{wilds2021,christie2018functional} from {WILDS} benchmark which includes three domains: train,  OOD val, and OOD test---with satellite images taken in different geographical regions and at different times;
(viii) \textbf{Camelyon}~\citep{bandi2018detection} from {WILDS} benchmark which includes three domains: train,  OOD val,  and OOD test, for tumor identification with domains corresponding to different hospitals;
\update{
(ix) \textbf{Civilcomments}~\citep{borkan2019nuanced} which includes three domains: train,  OOD val, and OOD test, for toxicity detection with domains corresponding to different demographic subpopulations; 
(x) \textbf{Retiring Adults}~\citep{ding2021retiring} where we consider the ACSIncome prediction task with various domains representing different states and time-period;  
and (xi) \textbf{Mimic Readmission}~\citep{johnson2020mimic, physiobank2000physionet} where the task is to predict readmission risk with various domains representing data from different time-period.} %

\textbf{Simulating a shift in target marginal~~} The above datasets present minor to no shift in label marginal. Hence, we simulate such a shift by altering the target label marginal and keeping the source target distribution fixed (to the original source label distribution). Note that, unlike some previous studies, we do not alter the source label marginal because, in practice, we may have an option to carefully curate the training distribution but might have little to no control over the target label marginal. %

For each target dataset, we have the true labels which allow us to vary the target label distribution.  
In particular, we sample the target label marginal from a Dirichlet distribution with a parameter $\alpha \in \{ 0.5, 1, 3.0, 10\}$ multiplier to the original target marginal. Specifically, \smash{$p_t(y) \sim \textrm{Dir}(\beta)$} where \smash{$\beta_y = \alpha \cdot p_{t,0}(y)$} and $p_{t,0}(y)$ is the original target label marginal.  The Dirichlet parameter $\alpha$ controls the severity of shift in target label marginal. Intuitively, as $\alpha$ decreases, the severity of the shift increases. 
For completeness, we also include the target dataset with the original target label marginal. For ease of exposition, we denote the shifts as \textsc{None} (no external shift) in the set of Dirichlet parameters, i.e.~the limiting distribution as \smash{$\alpha \rightarrow \infty$}. 
After simulating the shift in the target label marginal \update{(with two seeds for each $\alpha$)}, we obtain \update{560} pairs of 
different source and target datasets.

\subsection{Domain Adaptation Methods} \label{subsec:methods}

We implement the following algorithms (a more detailed description of each method is included in \appref{app:methods}):

\textbf{Source only~~} As a baseline, we include model trained with empirical risk minimization~\citep{vapnik1999overview} with cross-entropy loss on the source domain. 
We include source only models trained with and without augmentations. We also include adversarial robust models trained on source data with augmentations (\textbf{Source (adv)}). In particular, we use models adversarially trained against $\ell_2$-perturbations. 

\textbf{Domain alignment methods~~} These methods employ domain-adversarial training schemes aimed to learn invariant representations across different domains~\citep{ganin2016domain, zhang2019bridging, tan2020class}. For our experiments, we include the following \emph{five} methods: 
Domain Adversarial Neural Networks (\textbf{DANN}~\citep{ganin2016domain}),  
Conditional DANN (\textbf{CDANN}~\citep{long2018conditional}, 
\update{Maximum Classifier Discrepancy (\textbf{MCD}~\citep{saito2018maximum})}, %
Importance-reweighted DANN and CDANN (i.e. \textbf{IW-DANN} \& \textbf{IW-CDANN} \citet{tachet2020domain}).

\textbf{Self-training methods~~} These methods ``pseudo-label'' unlabeled examples with the model’s own predictions and then train on them as if they were labeled examples. \update{For vision datasets, these methods} often also use consistency regularization, which encourages the model to make consistent predictions on augmented views of unlabeled examples~\citep{lee2013pseudo, xie2020self, berthelot2021adamatch}. We include the following three algorithms: 
\textbf{FixMatch}~\citep{sohn2020fixmatch},
\textbf{Noisy Student}~\citep{xie2020selfb},
Selective Entropy Optimization via Committee Consistency (\textbf{SENTRY}~\citep{prabhu2021sentry}). \update{For NLP and tabular dataset, where we do not have strong augmentations defined, we consider \textbf{PseudoLabel} algorithm~\citep{lee2013pseudo}.} %

\textbf{Test-time adaptation methods~~} These methods take a source model and adapt a few parameters (e.g. batch norm parameters, etc.) on the unlabeled target data with an aim to improve target performance. 
We include: 
\textbf{CORAL}~\citep{sun2016return} or Domain Adjusted Regression ({DARE}~\citep{rosenfeld2022domain}), 
BatchNorm adaptation (\textbf{BN-adapt}~\citep{li2016revisiting, schneider2020improving}), 
Test entropy minimization (\textbf{TENT}~\citep{wang2021tent}).

\subsection{Meta algorithm to handle target label marginal shift}

\begin{algorithm}[t!]
  \caption{Meta algorithm to handle label marginal shift}
  \label{alg:LSBench}
  \begin{algorithmic}[1]
  \INPUT Source training and validation data: $(X_S, Y_S)$ and $(X^\prime_S, Y^\prime_S)$, unlabeled target training and validation data: ${X}_T$ and $X^\prime_T$, classifier $f$, and DA algorithm $\calA$  
    \STATE $\wt X_S, \wt Y_S \gets$ SampleClassBalanced$(X_S, Y_S)$ \\\hfill \textcolor{blue}{\COMMENT{\texttt{Balance source data}}}
    \FOR{$t = 1$ to $T$}
        \STATE $\wh Y_T \gets \argmax_y f_y(X_T)$ %
        \STATE $\wt X_T \gets $ SampleClassBalanced$(X_T, \wh Y_T)$ \\\hfill \textcolor{blue}{\COMMENT{\texttt{Pseudo-balance target data}}}
        \STATE Run an epoch of $\calA$ to update $f$ on balanced source data $\{\wt X_S, \wt Y_S\}$ and target data  $\{\wt X_T\}$  
    \ENDFOR
    \STATE 
    $\wh p_t(y) \gets$ EstimateLabelMarginal$(f, X^\prime_S, Y^\prime_S, X^\prime_T)$
    \STATE $f^\prime_j \gets \mfrac{\wh p_t(y=j) \cdot f_j}{\sum_k \wh p_t(y=k) \cdot f_k}$ for all $j \in \calY$ \\\hfill \textcolor{blue}{\COMMENT{\texttt{Re-weight classifier}}} 
    \vspace{3pt}
  \OUTPUT Target label marginal $\wh p_t(y)$ and classifier $f^\prime$ 
\end{algorithmic}
\end{algorithm}

Here we discuss two simple general-purpose corrections that we 
implement
in our framework. 
First, note that, as the severity of shift 
in the target label marginal 
increases, the performance of DA methods can falter 
as the training is done over source and target 
datasets with different class proportions.
Indeed, failure of domain adversarial 
training methods (one category of 
deep DA methods)
 has been theoretically and empirically shown in the 
literature~\citep{wu2019domain, zhao2019learning}. 
In our experiments, we show that a failure due to a shift in label distribution is not limited to domain adversarial training methods,  
but is common with all the popular DA methods (\secref{sec:results}). 

\textbf{Re-sampling ~~} To handle label imbalance in standard 
supervised learning, re-sampling the data  
to balance the class marginal 
is a known successful strategy~\citep{chawla2002smote, 
buda2018systematic, cao2019learning}. 
In relaxed label shift, 
we seek to handle the imbalance 
in the target data 
(with respect to the source label marginal), 
where we do not have access 
to true labels. 
We adopt an alternative strategy 
of leveraging pseudolabels 
for target data to perform
pseudo class-balanced re-sampling\footnote{A different strategy could be to re-sample target pseudolabel marginal to match source label marginal. 
For simplicity, we choose 
to balance source label marginal and target pseudolabel marginal.}~\citep{zou2018unsupervised, wei2021crest}. 
For relaxed label shift problems, 
\citep{prabhu2021sentry} employed 
this technique with their committee 
consistency objective, SENTRY. 
However, they did not explore 
re-sampling based correction
for existing DA techniques. 
Since this technique can be used 
in conjunction with any DA methods, 
we employ this re-sampling technique 
with existing DA methods and 
find that re-sampling benefits all 
DA methods, often improving over SENTRY 
in our testbed (\secref{sec:results}). 

\textbf{Re-weighting ~~} With re-sampling, we can hope to train the 
classifier $\wh f$ on a mixture of balanced source 
and balanced target datasets
in an ideal case. 
However, this still leaves open the problem of 
adapting the classifier $\wh f$ to the original target 
label distribution which is not available. 
If we can estimate the target label marginal, 
we can post-hoc adapt the classifier $\wh f$ with a simple
re-weighting correction~\citep{lipton2018detecting, alexandari2019adapting}.
To estimate the target label marginal, 
we turn to techniques developed under 
the stricter label shift assumption 
(recall, the setting where $p(x|y)$ 
remains domain invariant).
These approaches leverage off-the-shelf classifiers 
to estimate target marginal and provide $\calO(1/\sqrt{n})$ 
convergence rates under the label shift condition with mild assumptions on the classifier~\citep{lipton2018detecting, azizzadenesheli2019regularized, garg2020labelshift}.

While the relaxed label shift scenario
violates the conditions required 
for consistency of label shift estimation 
techniques, we nonetheless employ 
these techniques and empirically
evaluate efficacy of these methods 
in our testbed. In particular, 
to estimate the target label marginal, we experiment with: 
(i) RLLS~\citep{azizzadenesheli2019regularized};
(ii) MLLS~\citep{alexandari2019adapting}; and
(iii) \emph{baseline estimator} that simply averages 
the prediction of a classifier $f$ 
on unlabeled target data. 
We provide precise details about 
these methods in \appref{app:label_shift_estimation}. 
Since these methods leverage off-the-shelf classifiers, 
classifiers obtained with any
DA methods can be used in conjunction 
with these estimation methods.

\textbf{Summary~~} {Overall, in \algoref{alg:LSBench}, we illustrate 
how to incorporate the re-sampling and 
re-weighting correction 
with existing DA techniques.} 
\figref{fig:RS-RW_method} in \appref{sec:illustration} illustrates the method.   
Algorithm $\calA$ can be any DA method and in Step 7,
we can use any of the three methods listed above 
to estimate the target label marginal. 
We instantiate \algoref{alg:LSBench} 
with several algorithms from \secref{subsec:methods}
in \appref{app:methods}. 
\update{Intuitively, in an ideal scenario when 
the re-sampling step in our meta-algorithm
perfectly corrects for label imbalance between source and target, 
we expect DA methods 
to adapt classifier $f$ to $p(x|y)$ shift. 
The re-weighting step in our meta-algorithm can then adapt the classifier 
$f$ to the target label marginal $p_t(y)$.} 
We emphasize that in our work, we \emph{do not} 
claim to propose these corrections. 
But, to the best of our knowledge, 
our work is the first to combine these 
two corrections together 
and
perform extensive experiments 
across diverse datasets.

\subsection{Other choices for realistic evaluation}

For a fair evaluation and
comparison across different datasets and 
DA
algorithms, we re-implemented 
all the algorithms %
with consistent design choices whenever applicable.  
We also make 
several additional implementation choices, described below. 
We defer the additional details to \appref{app:setup_details}.

\textbf{Model selection criteria and hyperparameters~~} 
Given that we lack validation 
i.i.d data from the target distribution,
model selection in DA problems \emph{can not} follow 
the standard workflow used in supervised training. 
Prior works often omit details on how to choose 
hyperparameters leaving open a possibility of 
choosing hyperparameters using the test set which can 
provide a false and unreliable sense of 
improvement. Moreover, inconsistent 
hyperparameter selection strategies can 
complicate fair evaluations mis-associating 
the improvements to the algorithm under study.

In our work, we use source hold-out 
performance to pick 
the best hyperparameters.
First, 
for 
$\ell_2$ regularization and learning rate, we perform a sweep over
random hyperparameters to maximize the performance of source 
only model on the hold-out source data. Then for each dataset, we 
keep these hyperparameters fixed across DA algorithms. 
For DA methods specific hyperparameters,  
we use the same hyperparameters 
across all the methods 
incorporating the suggestions 
made in corresponding papers.  
Within a run, we use hold out performance on the source 
to pick the early stopping point. In appendices, we report 
\emph{oracle} performance by choosing the early stopping point with 
target accuracy.

\textbf{Evaluation criteria~~} To evaluate the target label
marginal estimation, we report $\ell_1$ error 
between the estimated label distribution and true label distribution. 
To evaluate the classifier performance on target data, we report 
performance of the (adapted) classifier on a hold-out partition of target data.

\textbf{Architectural and pretraining details~~} We experiment
with different architectures (e.g., DenseNet121, Resenet18, Resnet50, \update{DistilBERT, MLP and Transformer}).
\update{We experiment with randomly-initialized models and Imagenet, and DistillBert pre-trained models}. %
Given a dataset, we use the same architecture across 
different DA algorithms. 

\textbf{Data augmentation~~} Data augmentation is a standard ingredient to train vision models which can approximate some of the variations between domains. Unless stated otherwise, we train all the vision datasets using the standard strong augmentation technique: random horizontal flips, random crops, augmentation with Cutout~\citep{devries2017improved}, and RandAugment~\citep{cubuk2020randaugment}.   
To understand help with data augmentations alone, we also experiment with source-only models trained without any data augmentation. \update{For tabular and NLP datasets, we do not use any augmentations.}

\section{Main Results} \label{sec:results}

\begin{figure*}[t!]
    \centering 
    \begin{subfigure}[b]{\linewidth}
    \includegraphics[width=\linewidth]{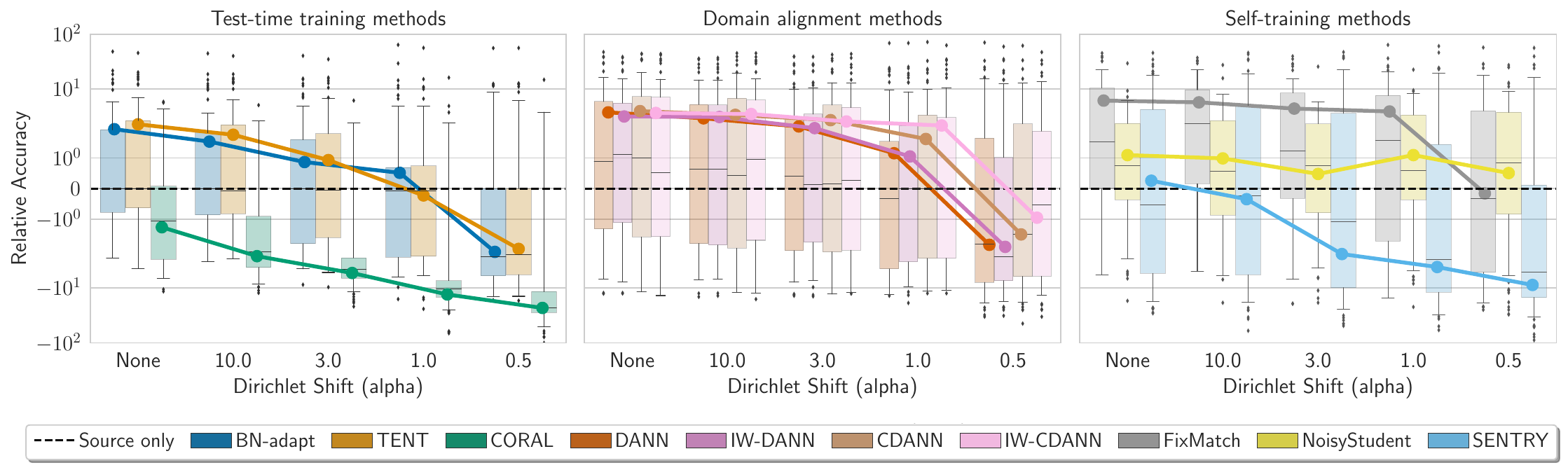}
    \caption{\update{Performance of DA methods relative to source-only training with increasing severity of target label marginal shift}}
    \end{subfigure}
    \begin{subfigure}[b]{\linewidth}
    \includegraphics[width=\linewidth]{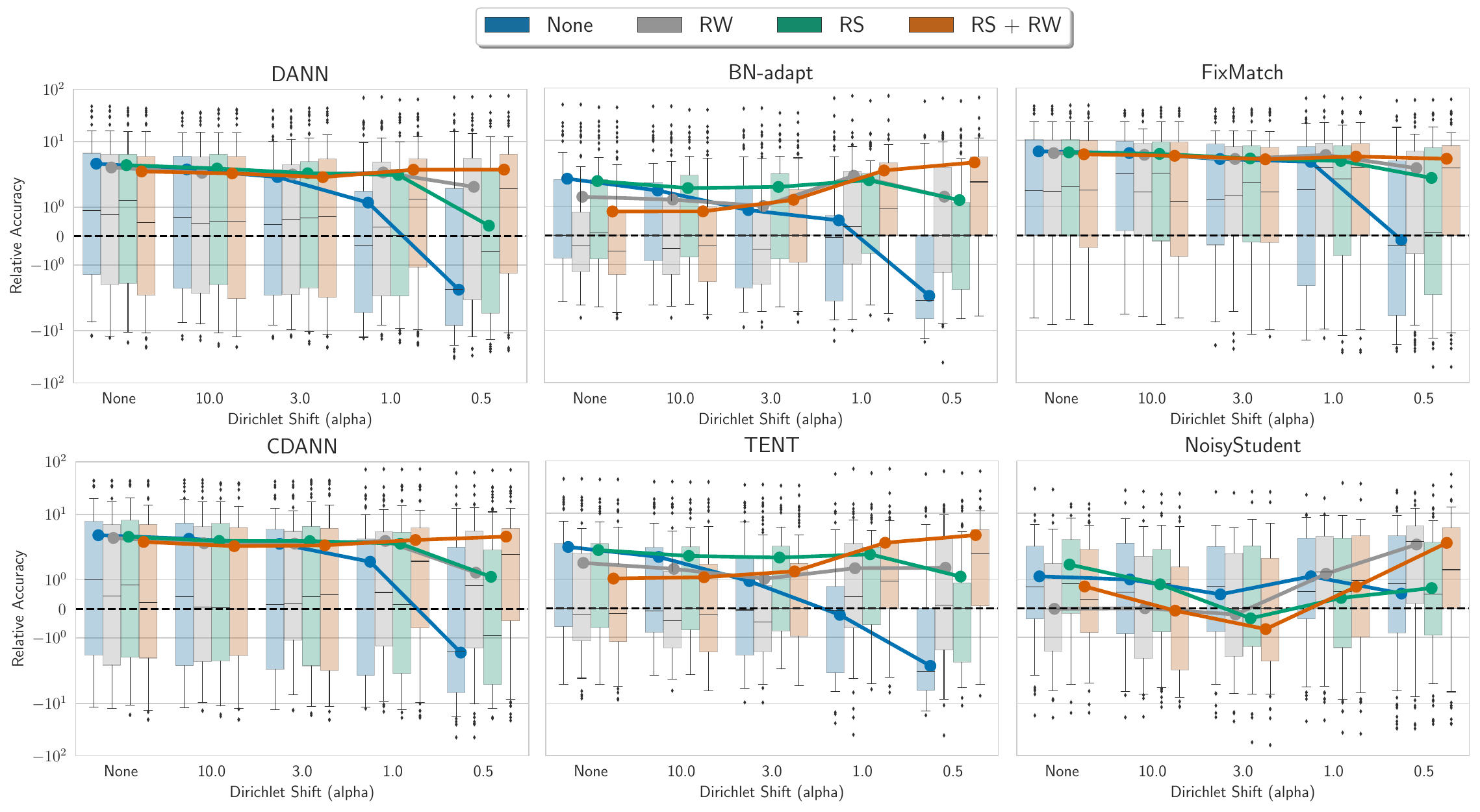}
    \caption{\update{Performance of DA methods relative to source-only training when paired with our meta-algorithm (RS and RW corrections)}}
    \end{subfigure}
    \caption{\update{\emph{Performance of different DA methods relative to a source-only model across all distribution shift pairs in vision datasets grouped by shift severity in label marginal}. 
    For each distribution shift pair and DA method, we plot the relative accuracy of the model trained with that DA method by subtracting the accuracy of the source-only model. Hence, the black dotted line at 0 captures the performance of the source-only model. 
    Smaller the Dirichlet shift parameter, the more severe is the shift in target class proportion. 
    \textbf{(a)} Shifts with $\alpha = \{\textsc{None}, 10.0, 3.0\}$ have little to no impact on different DA methods whereas the performance of all DA methods degrades when $\alpha \in \{1.0, 0.5\}$ often falling below the performance of a source-only classifier (except for Noisy Student).
    \textbf{(b)} RS and RW (in our meta-algorithm) together significantly improve aggregate performance over no correction for all DA methods. While RS consistently helps (over no correction) across different label marginal shift severities, RW hurts slightly for BN-adapt, TENT, and NoisyStudent when shift severity is small. However, for severe shifts ($\alpha \in \{3.0, 1.0, 0.5\}$) RW significantly improves performance for all the methods.
    Parallel results on tabular and language datasets in \appref{app:NLP_tabular_results}. Detailed results with all methods on individual datasets in
    \appref{app:individual_dataset}. A more detailed description of the plotting technique in \appref{app:description}.}}
    \label{fig:DA_vision}
\end{figure*}

\begin{figure*}[t!]
 \centering
    \includegraphics[width=0.32\linewidth]{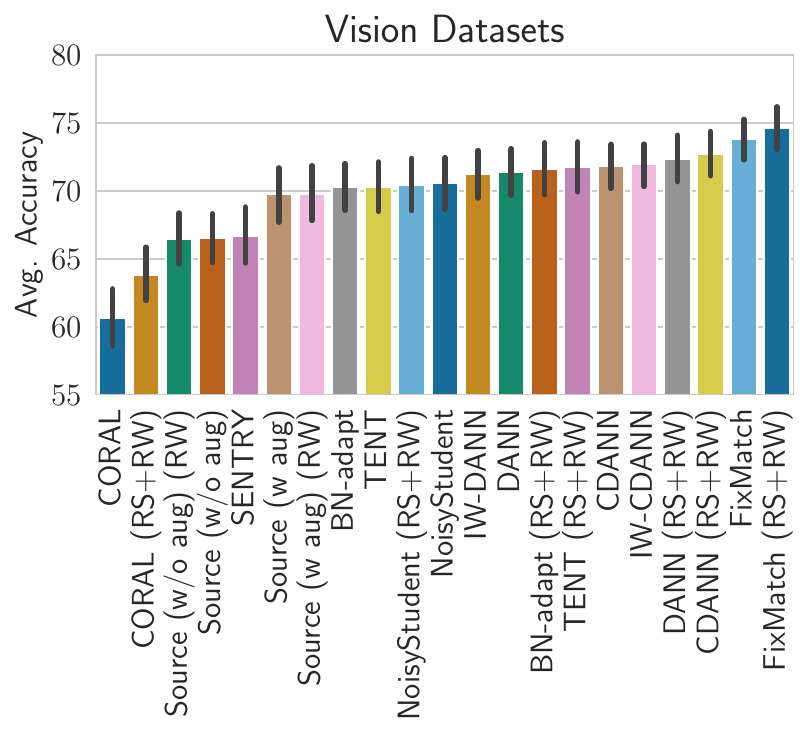}
    \includegraphics[width=0.32\linewidth]{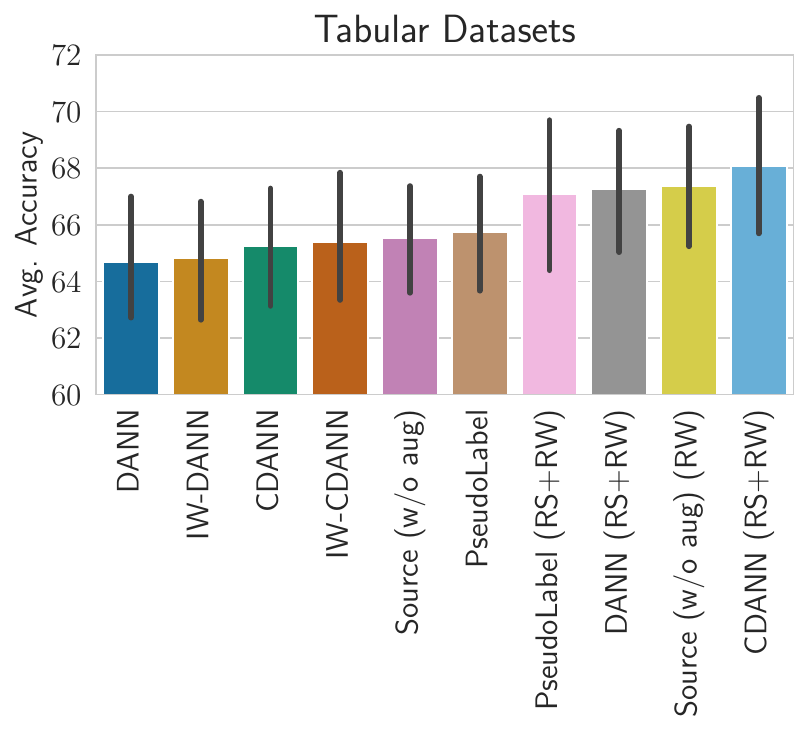}
    \includegraphics[width=0.32\linewidth]{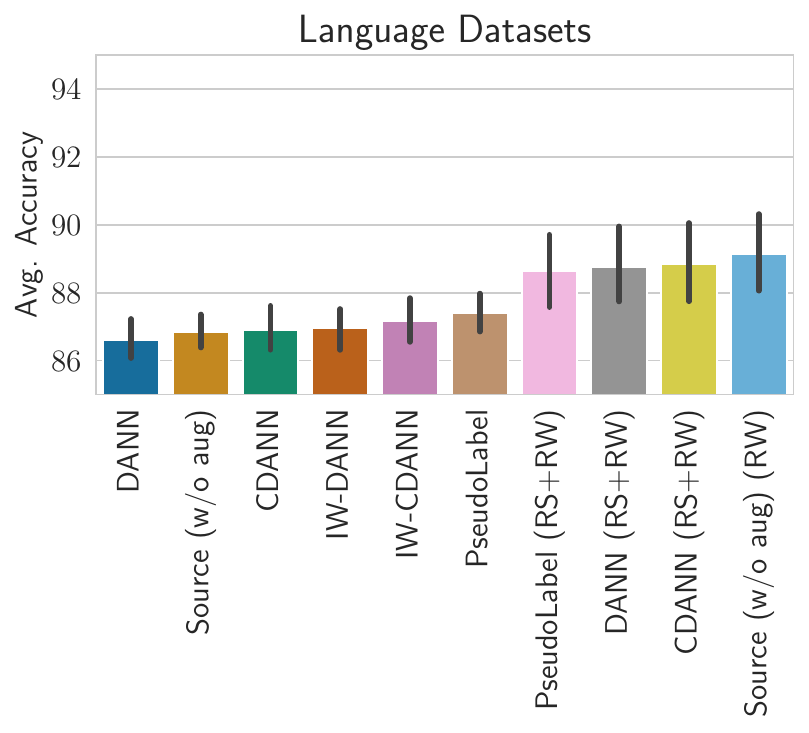} 
    \caption{\update{\emph{Average accuracy of different DA methods aggregated across all distribution pairs in each modality.} 
    Parallel results with all methods on individual datasets in
    \appref{app:individual_dataset}.}}
    \label{fig:aggregate_results}
\end{figure*}

\begin{figure*}[t!]
 \centering
    \includegraphics[width=0.3\linewidth]{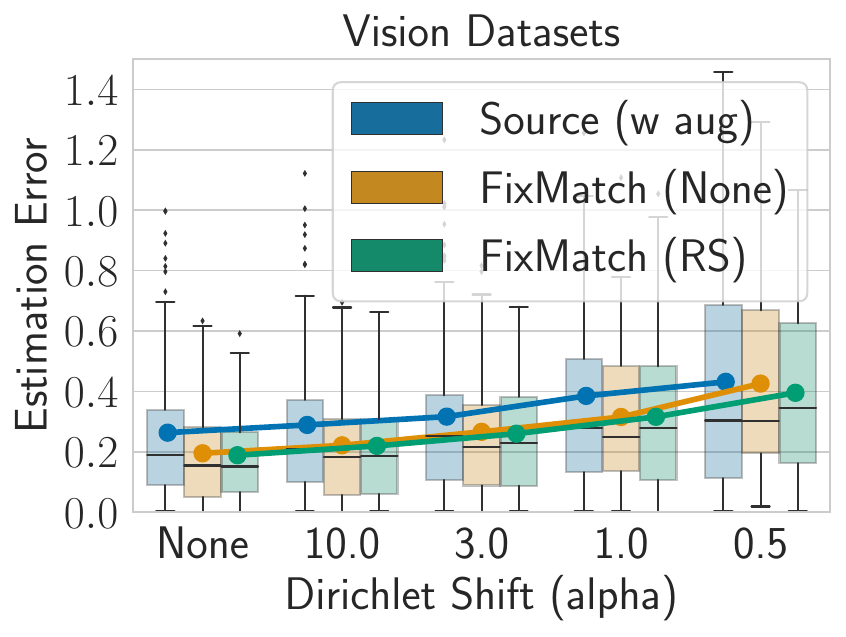}
    \includegraphics[width=0.29\linewidth]{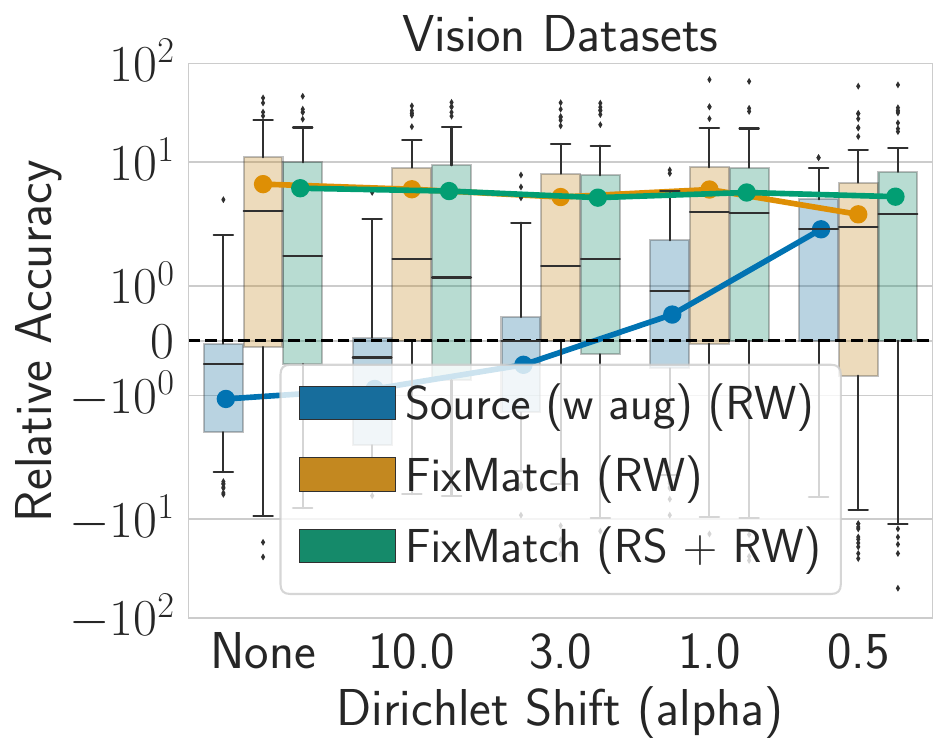}\ \ \  \vrule width 1pt\ \ 
    \includegraphics[width=0.3\linewidth]{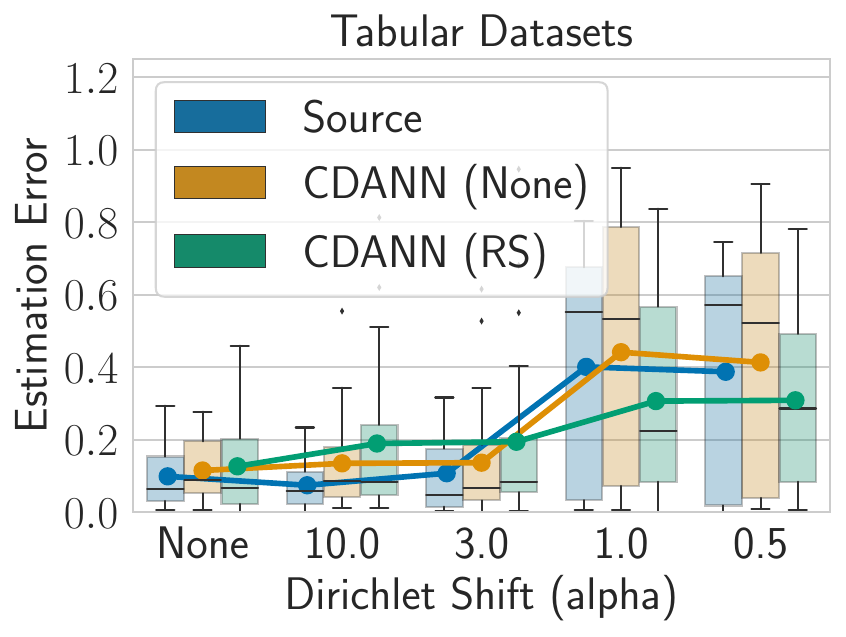}
    \caption{\update{{\emph{Target label marginal estimation ($\ell_1$) error and accuracy with RLLS and classifiers obtained with different DA methods.}} 
    \textbf{(Left)} Across all shift severities in vision datasets, RLLS with classifiers obtained with DA methods improves over RLLS with a source-only classifier. \textbf{(Right)} For tabular datasets, RLLS with classifiers obtained with DA methods improves over RLLS with a source-only classifier for severe target label marginal shifts. Plots for each DA method and all datasets are in \appref{app:target_marginal_estimation}. 
    }}
    \label{fig:results_ls_estimation}
\end{figure*}

\update{We present aggregated results on vision datasets in our testbed in \figref{fig:DA_vision}. In \appref{app:NLP_tabular_results}, we present aggregated results on NLP and tabular datasets.
We include results on each dataset in \appref{app:individual_dataset}.} Note that we do not include RS results with a source only model as it is trained only on source data and we observed no differences with just balancing the source data  (as for most datasets source is already balanced) in our experiments. 
\update{Unless specified otherwise, we use source validation performance as the early stopping criterion.}
Based on running our entire $\lsbench$ suite, we distill our findings into the following takeaways.  

\textbf{Popular deep DA methods without any correction falter.~} While DA methods often improve over a source-only classifier for cases when the target label marginal shift is absent or low, the performance of these methods (except Noisy Student) drops below the performance of a source-only classifier when the shift in target label marginal is severe (i.e., when $\alpha = 0.5$ \update{in Fig. \ref{fig:DA_vision}a,  \ref{fig:tabular}a, and \ref{fig:NLP}a}). 
\update{On the other hand,  DA methods when paired with RS and RW correction, significantly improve over a source-only model even when the shift in target label marginal is severe (\update{Fig. \ref{fig:DA_vision}b,  \ref{fig:tabular}b, and \ref{fig:NLP}b})}.

\textbf{Re-sampling to pseudobalance target often helps all DA methods \update{across all modalities}.~}  When the 
shift in target label marginal is absent or very small (i.e., $\alpha \in \{ \textsc{None}, 10.0\}$ in \update{Fig. \ref{fig:DA_vision}b,  \ref{fig:tabular}b, and \ref{fig:NLP}b}), we observe no (significant) differences in performance with re-sampling. 
However, as the shift severity in target label marginal increases (i.e., $\alpha \in \{3.0, 1.0, 0.5\}$ in \update{Fig. \ref{fig:DA_vision}b,  \ref{fig:tabular}b, and \ref{fig:NLP}b}), we observe that re-sampling typically improves all DA methods in our testbed.

\textbf{\update{Benefits of post-hoc re-weighting of the classifier depends on shift severity and the underlying DA algorithm.}~} 
\update{For domain alignment methods (i.e. DANN and CDANN) and self-training methods, in particular FixMatch and PseudoLabel, we observe that RW correction typically improves (over no correction) significantly when the target label marginal shift is severe (i.e., $\alpha \in \{ 3.0, 1.0, 0.5\}$ in  Fig. \ref{fig:DA_vision}b,  \ref{fig:tabular}b, and \ref{fig:NLP}b) and
has no (significant) effect when the shift in target label marginal is absent or very small (i.e., $\alpha \in \{ \textsc{None}, 10.0\}$ in Fig. \ref{fig:DA_vision}b,  \ref{fig:tabular}b, and \ref{fig:NLP}b). 
For BN-adapt, TENT, and NoisyStudent, RW correction can slightly hurt when target label marginal shift is absent or low (i.e., $\alpha \in \{ \textsc{None}, 10.0\}$ in \figref{fig:DA_vision}b) but continues to improve significantly when the target label marginal shift is severe (i.e., $\alpha \in \{ 3.0, 1.0, 0.5\}$ in \figref{fig:DA_vision}b).  
Additionally, we observe that in specific scenarios of the real-world shift in $p(x|y)$  
(e.g., subpopulation shift in BREEDs datasets, camelyon shifts, and replication study in CIFAR-10 which are benign relative to other vision dataset shifts in our testbed), RW correction does no harm to performance for BN-adapt, TENT, and NoisyStudent even 
when the target label marginal shift is less severe or absent (refer to datasets in \appref{app:individual_dataset}).}

\textbf{\update{DA methods paired with our meta-algorithm often improve over source-only classifier but no one method consistently performs the best.}~} 
First, we observe that our source-only numbers are better than previously published results. Similar to previous studies~\citep{gulrajani2020search}, this can be attributed to improved design choices (e.g. data augmentation, hyperparameters) which we make consistent across all methods. 
\update{While there is no consistent method that does the best across datasets, overall, FixMatch with RS and RW (our meta-algorithm) performs the best for vision datasets. For NLP datasets, source-only with RW (our meta-algorithm) performs the best overall. For tabular datasets, CDANN with RS and RW (our meta-algorithm) performs the best overall (\figref{fig:aggregate_results}).}

\textbf{\update{Existing DA methods when paired with our meta-algorithm significantly outperform other DA methods specifically proposed for relaxed label shift.~}}
\update{We observe that, with consistent experimental design 
across different methods, existing DA methods
with RS and RW corrections  
often improve over previously proposed methods
specifically aimed to tackle relaxed label shift, i.e., 
IW-CDANN, IW-DANN, and SENTRY (\figref{fig:comparison}).
For severe target label marginal shifts, 
the performance of IW-DANN, IW-CDANN, and SENTRY often falls below that of the source-only model.
Moreover, while the importance weighting  
(i.e., IW-CDANN and IW-DANN) improves over 
CDANN and DANN resp. (Fig. \ref{fig:DA_vision}a, \ref{fig:tabular}a and \ref{fig:NLP}a), 
RS and RW corrections significantly 
outweigh those improvements (\figref{fig:comparison}).
} 

\update{\textbf{BN-adapt and TENT with our meta-algorithm are simple and strong baselines.~}} 
For models with batch norm parameters, BN-adapt (and TENT) with RS and RW steps is a computationally efficient and strong baseline. We observe that while the performance of BN-adapt (and TENT) can drop substantially when the target label marginal shifts (i.e., $\alpha \in \{1.0, 0.5\}$ in \figref{fig:DA_vision}(a)), RS and RW correction improves the performance often improving BN-adapt (and TENT) over all other DA methods when the shift in target label marginal is extreme (i.e., $\alpha = 0.5$ in \figref{fig:DA_vision}(b)). 

\textbf{DA methods yield better target label marginal estimates, and hence larger accuracy improvements 
 with re-weighting, than source-only classifiers.}
Recall that we experiment with target label marginal estimation methods 
that leverage off-the-shelf classifiers to obtain an estimate. 
\update{We observe that estimators leveraging DA classifiers 
tend to perform better than using source-only classifiers for tabular and vision datasets}
\update{(\figref{fig:results_ls_estimation}).
For NLP, we observe that DA classifier and source-only classifier have performance (with source-only often performing slightly better). }
\update{Correspondingly, as one might expect, better estimation
yields greater accuracy improvements 
when applying our RW correction. 
In particular, RW correction with DA methods 
improves over the source-only classifier for vision and tabular datasets
and vice-versa for NLP datasets.
(\figref{fig:results_ls_estimation})}.

\textbf{Early stopping criterion matters.~}
\update{We observe a consistent $\approx$2\% and $\approx$8\% accuracy difference on vision  and tabular datasets respectively with all methods (\figref{fig:oracle_selection}). On NLP datasets, while the early stopping criteria have $\approx$2\% accuracy difference when RW and RS corrections are not employed, the difference becomes  negligible when these corrections are employed (\figref{fig:oracle_selection}). These results highlight that subsequent works should describe the early stopping criteria used within their evaluations.}

\textbf{Data augmentation helps.~} Corroborating findings from previous studies in other settings~\citep{gulrajani2020search, sagawa2021extending}, \update{we observe that data augmentation can improve the performance of a source-only model on vision datasets in relaxed label shift scenarios (refer to result on each dataset in \appref{app:individual_dataset}). Thus, whenever applicable, subsequent methods should use data augmentations.}

\section{Conclusion} %
\label{sec:discussion}

Our work is the first large-scale study investigating 
methods under the relaxed label shift scenario. 
Relative to works operating strictly 
under the label shift assumption,
\textsc{RLSbench} provides an opportunity for sensitivity analysis,
allowing researchers to measure the robustness of their methods
under various sorts of perturbations to the class-conditional distributions. 
Relative to the benchmark-driven deep domain adaptation literature, 
our work provides a comprehensive and standardized suite
for evaluating under shifts in label distributions,
bringing these benchmarks one step closer 
to exhibit the sort of diversity
that we should expect to encounter
when deploying models in the wild.
On one hand, the consistent improvements observed 
from label shift adjustments are promising.
At the same time, given the underspecified nature of the problem,
practitioners must remain vigilant and take performance 
on any benchmark with a grain of salt, 
considering the various ways that it might (or might not)
be representative of the sorts of situations 
that might arise in their application of interest.

In the future, we hope to 
extend $\lsbench$ 
to datasets from real applications in consequential domains
such as healthcare and self-driving, 
where label marginals  
and class conditionals  
can be expected to shift across locations and over time. 
We also hope to incorporate 
self-supervised methods that learn representations
by training on a union of unlabeled data 
from source and target via proxy tasks 
like reconstruction~\citep{gidaris2018unsupervised, he2022masked} and 
contrastive learning~\citep{caron2020unsupervised, chen2020simple}. 
While re-weighting predictions using 
estimates of the target label distribution yields significant gains,
the remaining gap between our results and oracle performance
should motivate future work geared towards improved estimators.
Also, we observe that the success of target label marginal estimation techniques depends on the nature of the shifts in $p(x|y)$.
Mathematically characterizing the behavior of 
label shift estimation techniques 
when the label shift assumption is violated would be an important contribution.

\section*{Reproducibility Statement}

Our code with all the results will be released on github. {\url{https://github.com/acmi-lab/RLSbench}}.
We implement our \textsc{RLSbench} library in PyTorch~\citep{paszke2017automatic} 
and provide an infrastructure to 
run all the experiments to generate corresponding results. We have stored all models and logged all hyperparameters to facilitate reproducibility.
In our appendices, we provide additional details on datasets and experiments.  In \appref{app:datasets}, we 
describe datasets and in \appref{app:setup_details}, we provide hyperparameter details.

\subsection*{Acknowledgments}

We thank Amrith Setlur, Pratyush Maini, and Aditi Raghunathan  for providing feedback on an earlier draft of RLSbench. 
We also thank Xingjian Shi and Weisu Yin for their initial help with running the large-scale experiments.  
SG acknowledges Amazon Graduate Fellowship and  JP Morgan AI Ph.D. Fellowship for their support.

\bibliography{iclr2023_conference}
\bibliographystyle{iclr2023_conference}

\newpage

\appendix

\onecolumn

\section*{Appendix}

\section{{Description of Plots}}
\label{app:description}

{For each plot in \figref{fig:DA_vision}, we obtain all the distribution shift pairs with a specific alpha 
(i.e., the value on the x-axis). Then for each distribution shift pair (with a specific alpha value), 
we obtain \emph{relative performance} by 
subtracting the performance of a source-only model trained 
on the source dataset of that distribution shift pair from the performance 
of the model trained on that distribution shift pair with the DA algorithm of interest.
Thus for each alpha and each DA method, we obtain $112$ relative performance values. We 
draw the box plot and the mean of these relative performance values. }

{For (similar-looking) plots, we use the same technique throughout the paper. 
The only thing that changes is the group of points over which aggregation is performed.}

\section{{Tabular and NLP Results Omitted from the Main Paper}}
\label{app:NLP_tabular_results}

\subsection{{Tabular Datasets}}
\begin{figure*}[h!]
    \centering 
    \begin{subfigure}[b]{\linewidth}
    \centering 
    \includegraphics[width=0.7\linewidth]{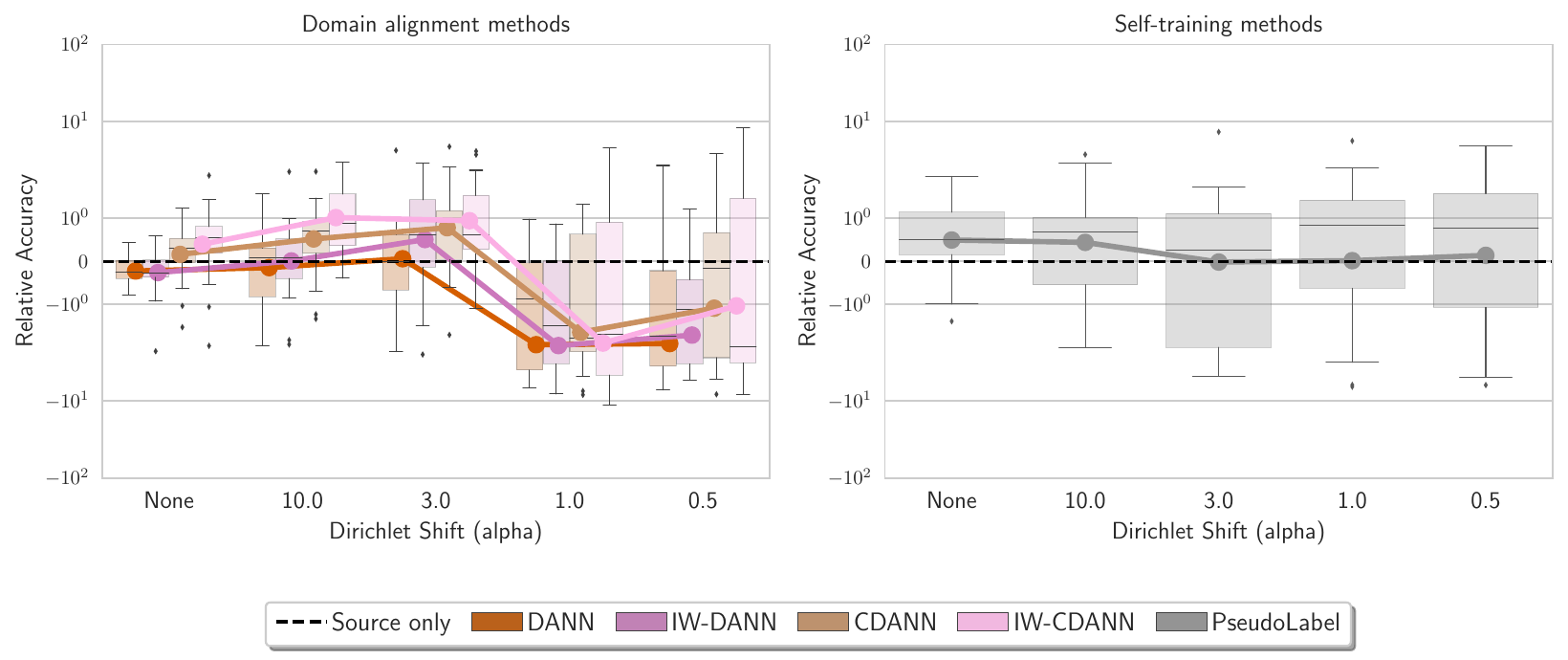}
    \caption{{Performance of DA methods relative to source-only training with increasing target label marginal shift}}
    \end{subfigure}

    \begin{subfigure}[b]{\linewidth}
     \includegraphics[width=\linewidth]{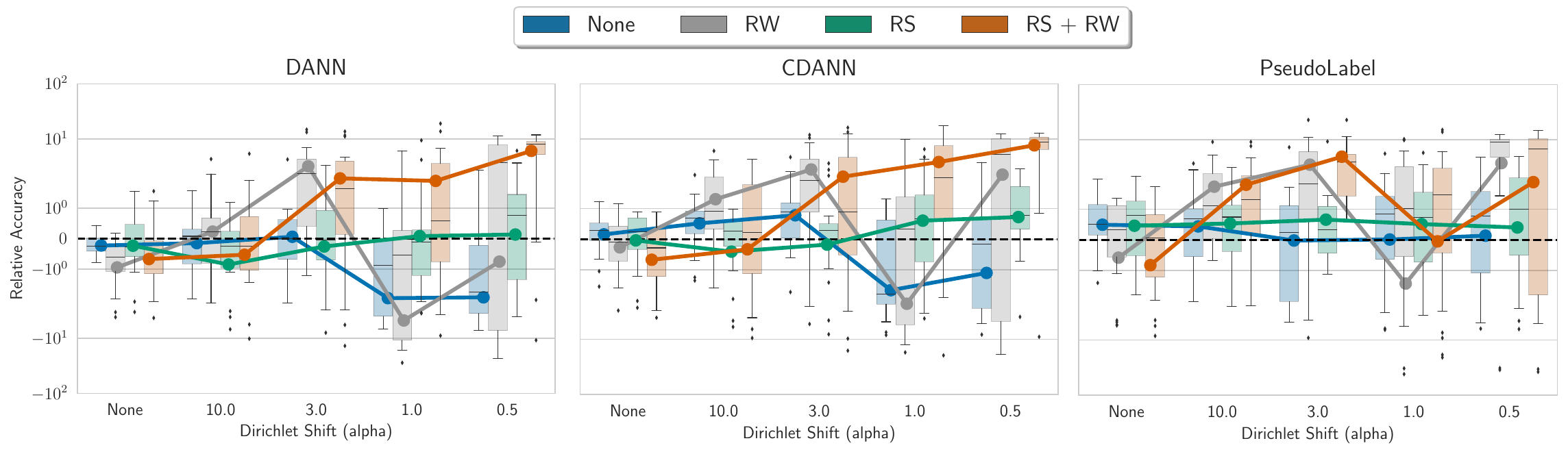}
    \caption{{Relative performance of DA methods when paired with our meta-algorithm (RS and RW corrections)}}
    \end{subfigure}
    \caption{{\emph{Performance of different DA methods relative to a source-only model across all distribution shift pairs in tabular datasets grouped by shift severity in label marginal}. 
    For each distribution shift pair and DA method, we plot the relative accuracy of the model trained with that DA method by subtracting the accuracy of the source-only model. Hence, the black dotted line at 0 captures the performance of the source-only model. 
    Smaller the Dirichlet shift parameter, the more severe is the shift in target class proportion. 
    \textbf{(a)} Shifts with $\alpha = \{\textsc{None}, 10.0, 3.0\}$ have little to no impact on different DA methods whereas the performance of all DA methods degrades when $\alpha \in \{1.0, 0.5\}$ often falling below the performance of a source-only classifier.
    \textbf{(b)} RS and RW (in our meta-algorithm) together significantly improve aggregate performance over no correction for all DA methods. While RS consistently helps (over no correction) across different label marginal shift severities, RW hurts slightly when shift severity is small. However, for severe shifts ($\alpha \in \{3.0, 1.0, 0.5\}$) RW significantly improves performance for all the methods. Results with all methods on individual datasets in
    \appref{app:individual_dataset}.
    }}
    \label{fig:tabular}
\end{figure*}

\subsection{{NLP Datasets}}
\begin{figure}[H]
    \centering 
    \begin{subfigure}[b]{\linewidth}
    \centering 
    \includegraphics[width=0.7\linewidth]{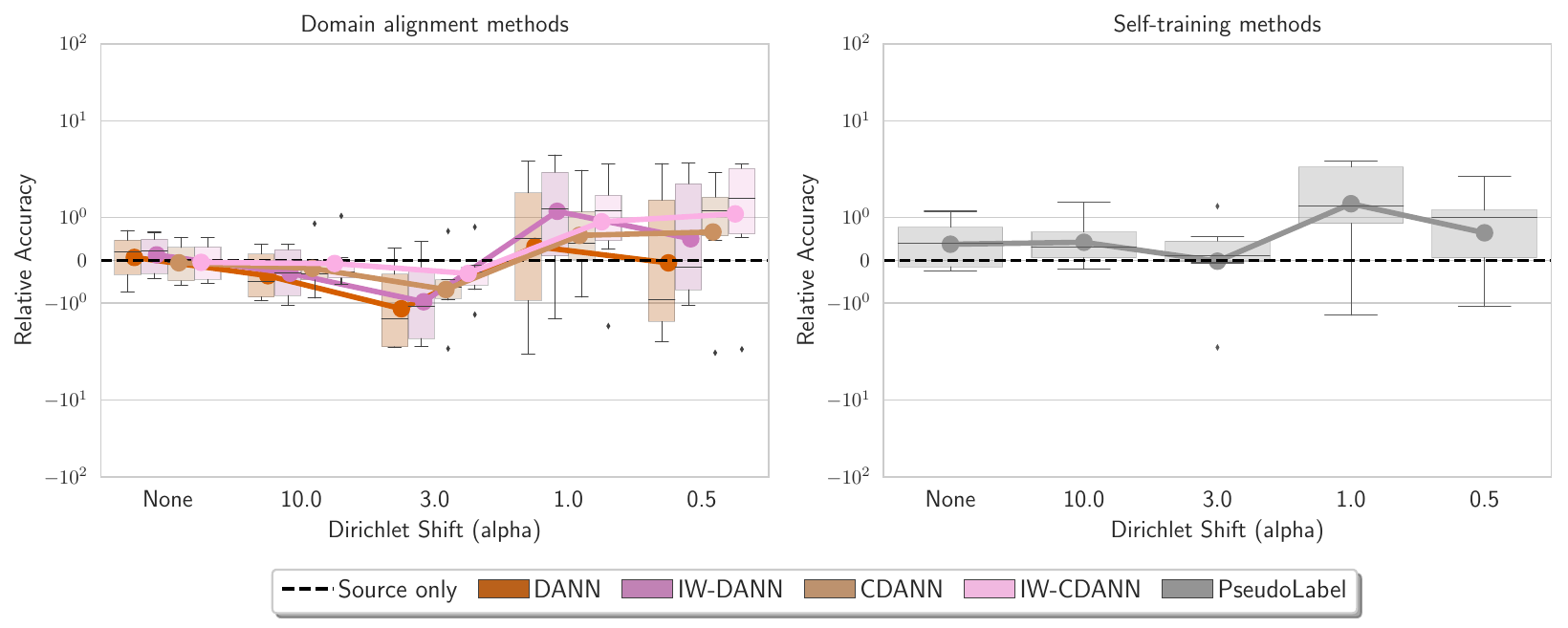}
    \caption{{Performance of DA methods relative to source-only training with increasing target label marginal shift}}
    \end{subfigure}

    \begin{subfigure}[b]{\linewidth}
    \includegraphics[width=\linewidth]{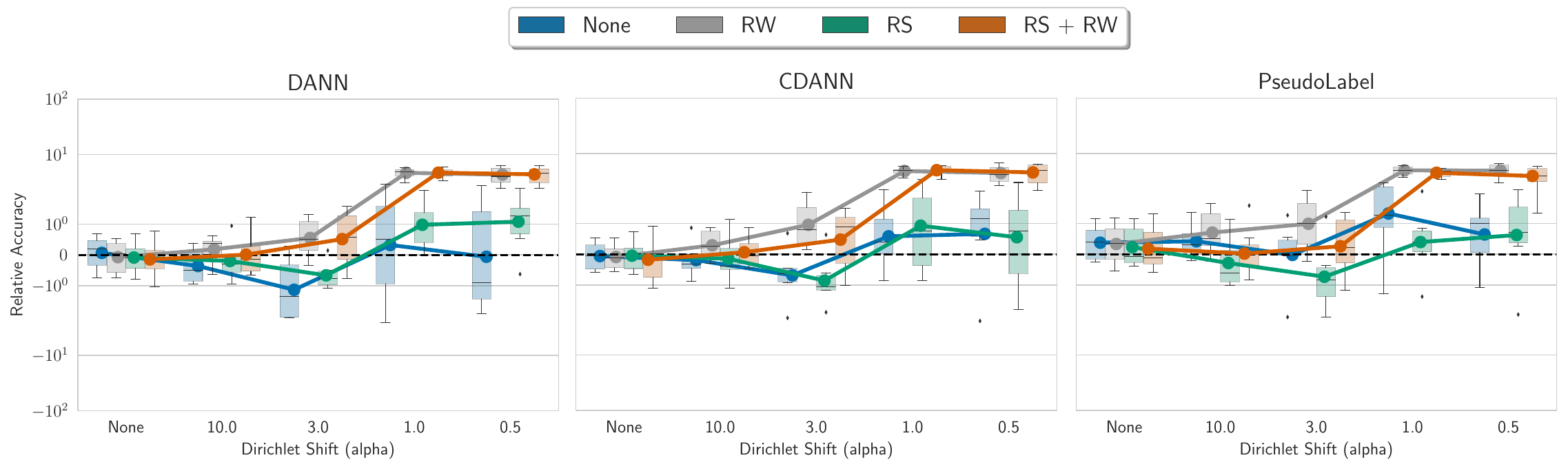}
    \caption{{Relative performance of DA methods when paired with our meta-algorithm (RS and RW corrections)}}
    \end{subfigure}
    \caption{{\emph{Performance of different DA methods relative to a source-only model across all distribution shift pairs in NLP datasets grouped by shift severity in label marginal}. 
    For each distribution shift pair and DA method, we plot the relative accuracy of the model trained with that DA method by subtracting the accuracy of the source-only model. Hence, the black dotted line at 0 captures the performance of the source-only model. 
    Smaller the Dirichlet shift parameter, the more severe is the shift in target class proportion. 
    \textbf{(a)} Performance of DANN and IW-DANN methods degrades with increasing severity of target label marginal shift often falling below the performance of a source-only classifier (except for Noisy Student). Performance of PsuedoLabel, CDANN, and IW-CDANN show less susceptibility to increasing severity in target marginal shift.
    \textbf{(b)} RS and RW (in our meta-algorithm) together significantly improve aggregate performance over no correction for all DA methods. While RS consistently helps (over no correction) across different label marginal shift severities, RW hurts slightly for BN-adapt, TENT, and NoisyStudent when shift severity is small. However, for severe shifts ($\alpha \in \{3.0, 1.0, 0.5\}$) RW significantly improves performance for all the methods. Detailed results with all methods on individual datasets in  \appref{app:individual_dataset}.
    }}
        \label{fig:NLP}
\end{figure}

\section{{Comparison between IW-CDANN, IW-DANN, and SENTRY with Existing DA methods paired with our Meta-Algorithm}}
\label{app:existing_comparison}

\begin{figure}[ht]
    \centering
    \includegraphics[width=0.48\linewidth]{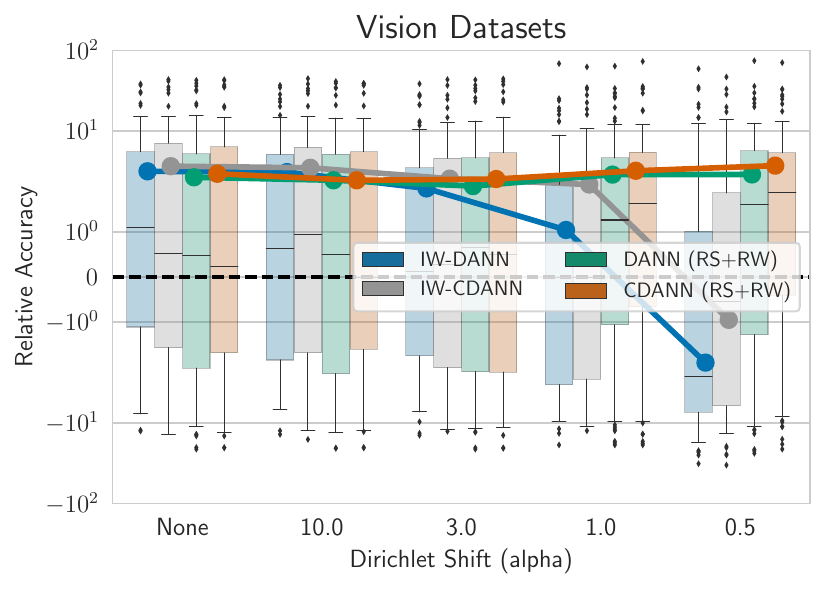}
    \includegraphics[width=0.48\linewidth]{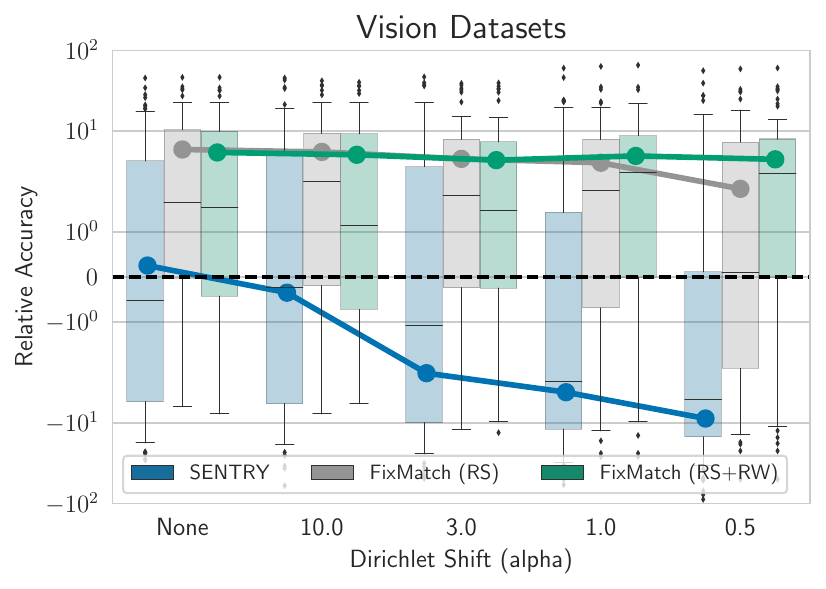}
    \includegraphics[width=0.48\linewidth]{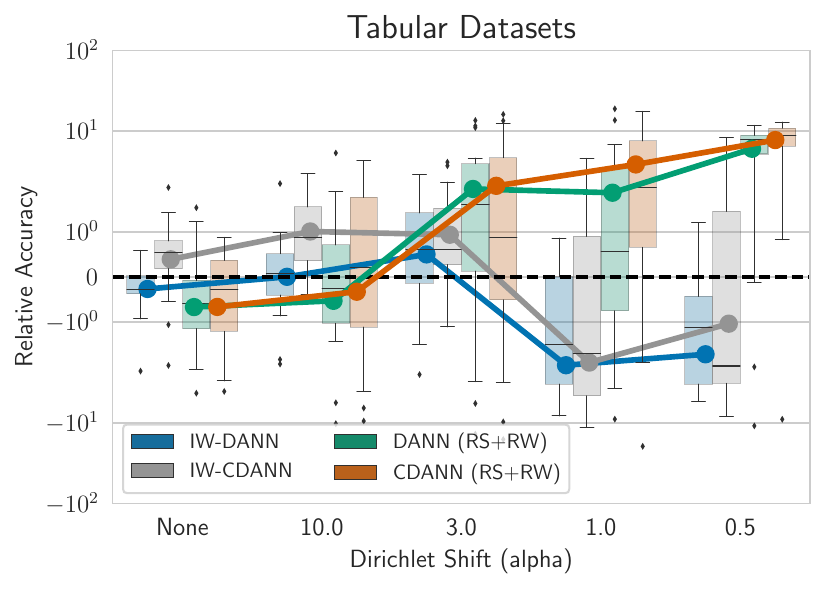}
    \includegraphics[width=0.48\linewidth]{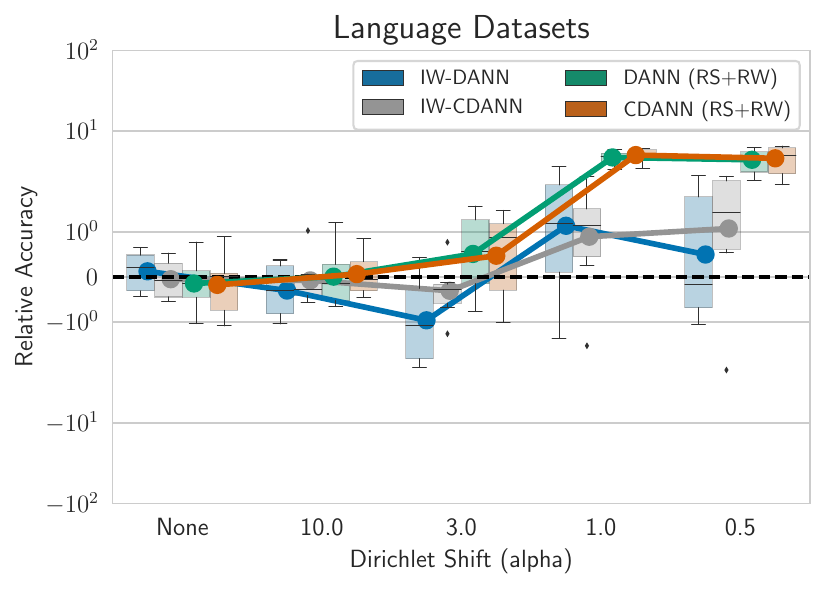}
    \caption{\update{\emph{Comparison of existing DA methods paired with our RS and RW correction 
    and DA methods specifically proposed for relaxed label shift problems.} 
    Across vision and tabular datasets, we observe the susceptibility of IW-DAN, IW-CDAN, and SENTRY with increasing severity of target label marginal shifts. In particular, for severe target label marginal shifts, the performance of IW-DAN, IW-CDAN, and SENTRY often falls below that of the source-only model. However, existing DA techniques when paired with RS + RW correction significantly improve over the source-only model. For NLP, datasets we observe similar behavior but with relatively less intensity.}}
    \label{fig:comparison}
\end{figure}

\update{\figref{fig:comparison} shows the relevant comparison.}

\update{\textbf{Note.}}
On Officehome dataset, we observe 
a slight discrepancy
between SENTRY results with our runs 
and numbers 
originally reported in the paper~\citep{prabhu2021sentry}.
We find that this is due to 
differences in batch size used in original 
work versus in our runs (which we kept the same
for all the algorithms). 
In \appref{app:sentry_more}, we report 
SENTRY results with the updated 
batch size. With the new batch size, 
we reconcile SENTRY results but also 
observe a significant improvement 
in FixMatch results. 
We refer reader to \appref{app:sentry_more} 
for a more detailed discussion.

\section{Dataset Details} 
\label{app:datasets}
In this section, we provide additional details about the datasets used in our benchmark study.

\begin{figure}[t!]
    \centering
    \includegraphics[width=0.8\textwidth]{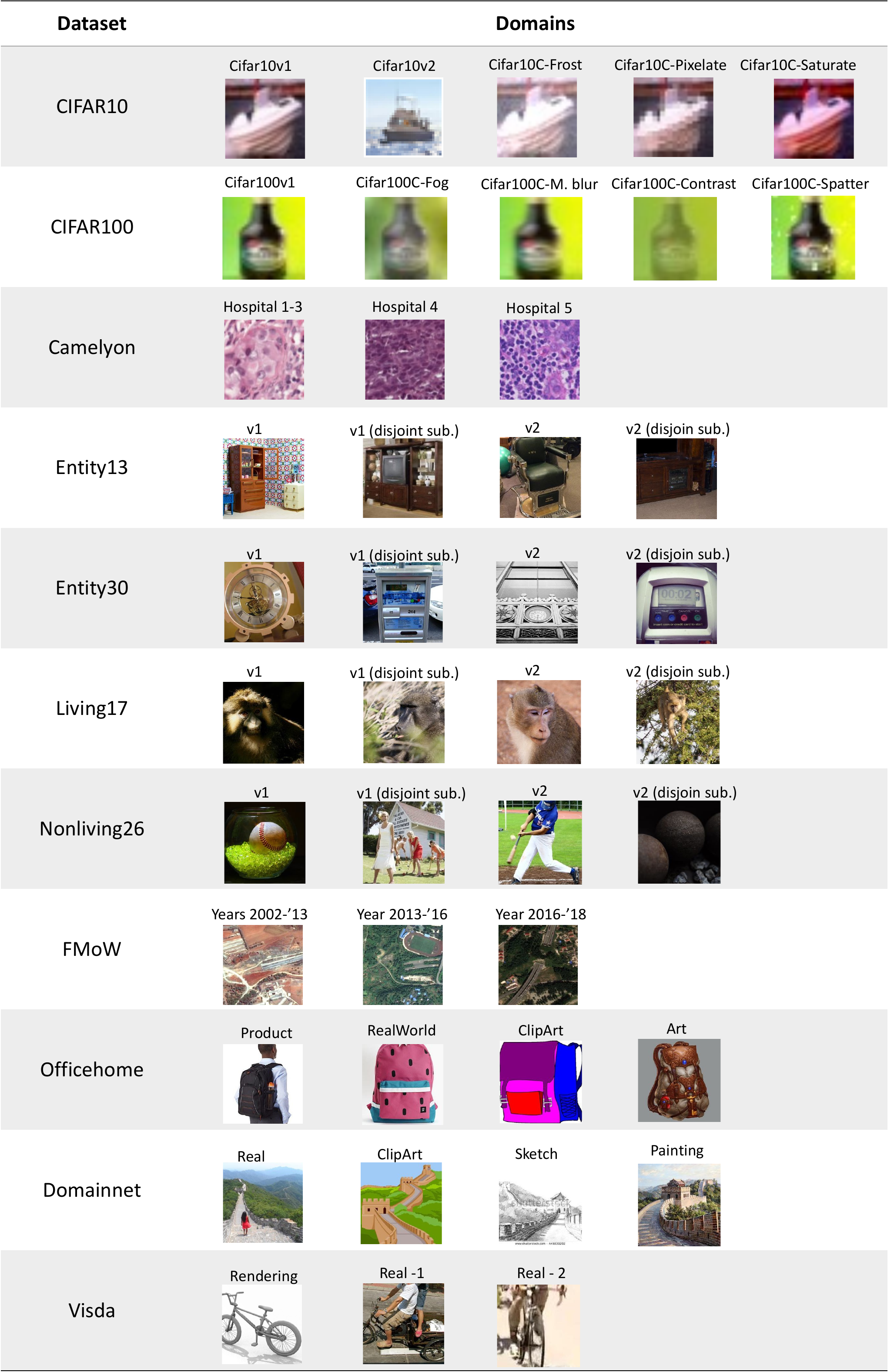}
    \caption{Examples from all the domains in each \update{vision} dataset. }
    \label{fig:LSBench_imgs}
\end{figure}

\begin{table}[ht]
    \begin{adjustbox}{width=0.9\columnwidth,center}
    \centering
    \small
    \tabcolsep=0.12cm
    \renewcommand{\arraystretch}{1.5}
    \begin{tabular}{lcccc}
    \toprule    
    Dataset && Source & & Target  \\
    \midrule
    CIFAR10 &&  CIFAR10v1  && \thead{CIFAR10v1, CIFAR10v2, CIFAR10C-Frost (severity 4),\\ CIFAR10C-Pixelate (severity 5), CIFAR10-C Saturate (severity 5)} \\ 
    CIFAR100 && CIFAR100 && \thead{CIFAR100, CIFAR100C-Fog (severity 4), \\ CIFAR100C-Motion Blur (severity 2), CIFAR100C-Contrast (severity 4), \\ CIFAR100C-spatter (severity 2)} \\
    Camelyon && \thead{Camelyon \\ (Hospital 1--3)} && Camelyon (Hospital 1--3), Camelyon (Hospital 4), Camelyon (Hospital 5)\\ 
    FMoW && FMoW (2002--'13) && FMoW (2002--'13), FMoW (2013--'16), FMoW (2016--'18)\\
    Entity13 && \thead{Entity13 \\(ImageNetv1 \\ sub-population 1)} && \thead{Entity13 (ImageNetv1 sub-population 1), \\ Entity13 (ImageNetv1 sub-population 2), \\  Entity13 (ImageNetv2 sub-population 1),\\ Entity13 (ImageNetv2 sub-population 2) } \\
    Entity30 && \thead{Entity30 \\(ImageNetv1 \\ sub-population 1)} && \thead{Entity30 (ImageNetv1 sub-population 1), \\ Entity30 (ImageNetv1 sub-population 2), \\  Entity30 (ImageNetv2 sub-population 1), \\ Entity30 (ImageNetv2 sub-population 2) } \\
    Living17 && \thead{Living17 \\(ImageNetv1 \\ sub-population 1)} && \thead{Living17 (ImageNetv1 sub-population 1), \\ Living17 (ImageNetv1 sub-population 2), \\  Living17 (ImageNetv2 sub-population 1), \\ Living17 (ImageNetv2 sub-population 2) } \\ 
    Nonliving26 && \thead{Nonliving26 \\(ImageNetv1 \\ sub-population 1)} && \thead{Nonliving26 (ImageNetv1 sub-population 1),\\ Nonliving26 (ImageNetv1 sub-population 2), \\  Nonliving26 (ImageNetv2 sub-population 1), \\ Nonliving26 (ImageNetv2 sub-population 2) }  \\ 
    Officehome && Product && Product, Art, ClipArt, Real\\
    DomainNet && Real && Real, Painiting, Sketch, ClipArt  \\
    Visda && \thead{Synthetic\\(originally referred \\to as train)} && \thead{Synthetic, Real-1 (originally referred to as val), \\ Real-2  (originally referred to as test)} \\
    Civilcomments && Train &&  Train, Val and Test (all formed by disjoint partitions of online articles) \\
    Mimic Readmissions && \thead{Mimic Readmissions \\ (year: 2008)} && \thead{Mimic Readmissions (year: 2008), Mimic Readmissions (year: 2009), \\ Mimic Readmissions (year: 2010), Mimic Readmissions (year: 2011), \\ Mimic Readmissions (year: 2012), Mimic Readmissions (year: 2013)}\\
    Retiring Adults && \thead{Retiring Adults \\ (year: 2014 \\ states: ['MD', 'NJ', 'MA'])} && \thead{Retiring Adults (year: 2015; states: ['MD', 'NJ', 'MA']), \\ Retiring Adults (year: 2016; states: ['MD', 'NJ', 'MA']), \\ Retiring Adults (year: 2017; states: ['MD', 'NJ', 'MA']), \\ Retiring Adults (year: 2018; states: ['MD', 'NJ', 'MA'])}\\
    \bottomrule 
    \end{tabular}  
    \end{adjustbox}  
    \caption{Details of the datasets considered in our $\rlsbench$.}\label{table:dataset}
 \end{table}
 
\begin{itemize}[leftmargin=*]
    \item \textbf{CIFAR10 {} {}} We use the original CIFAR10 dataset~\citep{krizhevsky2009learning} as the source dataset. For target domains, we consider (i) synthetic shifts (CIFAR10-C) due to common corruptions~\citep{hendrycks2019benchmarking}; and (ii) natural distribution shift, i.e., CIFAR10v2~\citep{recht2018cifar,torralba2008tinyimages} due to differences in data collection strategy. 
    We randomly sample 3 set of CIFAR-10-C datasets. Overall, we obtain 5 datasets (i.e., CIFAR10v1, CIFAR10v2, CIFAR10C-Frost (severity 4), CIFAR10C-Pixelate (severity 5), CIFAR10-C Saturate (severity 5)).

    \item \textbf{CIFAR100 {} {}} Similar to CIFAR10, we use the original CIFAR100 set as the source dataset. For target domains we consider synthetic shifts (CIFAR100-C) due to common corruptions. We sample 4 CIFAR100-C datasets, overall obtaining 5 domains (i.e., CIFAR100, CIFAR100C-Fog (severity 4), CIFAR100C-Motion Blur (severity 2), CIFAR100C-Contrast (severity 4), CIFAR100C-spatter (severity 2) ).

    \item \textbf{FMoW {} {}} In order to consider distribution shifts faced in the wild, we consider FMoW-WILDs~\citep{wilds2021,christie2018functional} from \textsc{Wilds} benchmark, which contains satellite images taken  in different geographical regions and at different times. We use the original train as source and OOD val and OOD test splits as target domains as they are collected over different time-period. Overall, we obtain 3 different domains.

    \item \textbf{Camelyon17 {} {}} Similar to FMoW, we consider tumor identification dataset from the wilds benchmark~\citep{bandi2018detection}. We use the default train as source and OOD val and OOD test splits as target domains as they are collected across different hospitals. Overall, we obtain 3 different domains.

    \item \textbf{BREEDs {} {}} We also consider {BREEDs} benchmark~\citep{santurkar2020breeds}  in our setup to assess robustness to subpopulation shifts. {BREEDs} leverage class hierarchy in ImageNet to re-purpose original classes to be the subpopulations and defines a classification task on superclasses. 
    We consider distribution shift due to subpopulation shift which is induced by directly making the subpopulations present in the training and test distributions disjoint. 
    {BREEDs}  benchmark contains 4 datasets \textbf{Entity-13}, \textbf{Entity-30},
    \textbf{Living-17}, and \textbf{Non-living-26}, each focusing on 
    different subtrees and levels in the hierarchy. We also consider natural shifts due to differences in the data  collection process of ImageNet~\citep{russakovsky2015imagenet},  e.g, ImageNetv2~\citep{recht2019imagenet} and a combination of both. 
    Overall, for each of the 4 BREEDs datasets (i.e., {Entity-13}, {Entity-30},
    {Living-17}, and {Non-living-26}), we obtain four different domains. We refer to them as follows: BREEDsv1 sub-population 1 (sampled from ImageNetv1), BREEDsv1 sub-population 2 (sampled from ImageNetv1), BREEDsv2 sub-population 1 (sampled from ImageNetv2), BREEDsv2 sub-population 2 (sampled from ImageNetv2). 
    For each BREEDs dataset, we use BREEDsv1 sub-population A as source and the other three as target domains.

    \item \textbf{OfficeHome {} {}} We use four domains (art, clipart, product and real) from OfficeHome dataset~\citep{venkateswara2017deep}. We use the product domain as source and the other domains as target.
    
    \item \textbf{DomainNet {} {}} We use four domains (clipart, painting, real, sketch) from the Domainnet dataset~\citep{peng2019moment}. We use real domain as the source and the other domains as target.
    
    \item \textbf{Visda {} {}} We use three domains (train, val and test) from the Visda dataset~\citep{peng2018syn2real}. While `train' domain contains synthetic renditions of the objects, `val' and `test' domains contain real world images. To avoid confusing, the domain names with their roles as splits, we rename them as `synthetic', `Real-1' and `Real-2'. We use the synthetic (original train set) as the source domain and use the other domains as target. 
    \item \update{\textbf{Civilcomments}~\citep{borkan2019nuanced} from the wilds benchmark which includes three domains: train,  OOD val, and OOD test, for toxicity detection with domains corresponding to different demographic subpopulations. The dataset has subpopulation shift across different demographic groups as the dataset in each domain is collected from a different partition of online articles.}
    \item \update{\textbf{Retiring Adults}~\citep{ding2021retiring} where we consider the ACSIncome prediction task with various domains representing different states and time-period; We randomly select three states and consider dataset due to shifting time across those states. Details about precise time-periods and states are in \tabref{table:dataset}.}
    \item \update{\textbf{Mimic Readmission}~\citep{johnson2020mimic, physiobank2000physionet} where the task is to predict readmission risk with various domains representing data from different time-period. Details about precise time-periods are in \tabref{table:dataset}. } 
\end{itemize}

We provide scripts to setup these datasets with single command in our code. To investigate the performance of different methods under the stricter label shift setting, we also include a hold-out partition of source domain in the set of target domains. For these distribution shift pairs where source and target domains are i.i.d. partitions, we obtain the stricter label shift problem. We summarize the information about source and target domains in \tabref{table:dataset}.

\textbf{Train-test splits~~} We partition each source and target dataset into $80\%$ and $20\%$ i.i.d. splits. We use $80\%$ splits for training and $20\%$ splits for evaluation (or validation). We throw away labels for the $80\%$ target split and only use labels in the $20\%$ target split for final evaluation. 
The rationale behind splitting the target data is to use a completely unseen batch of data for evaluation. This avoids evaluating on examples where a model potentially could have overfit. 
over-fitting to unlabeled examples for evaluation. In practice, if the aim is to make predictions on all the target data (i.e., transduction), we can simply use the (full) target set for training and evaluation.

\section{Illustration of Our Proposed Meta=algorithm} \label{sec:illustration}
\begin{figure}[H]
    \centering
    \includegraphics[width=0.48\linewidth]{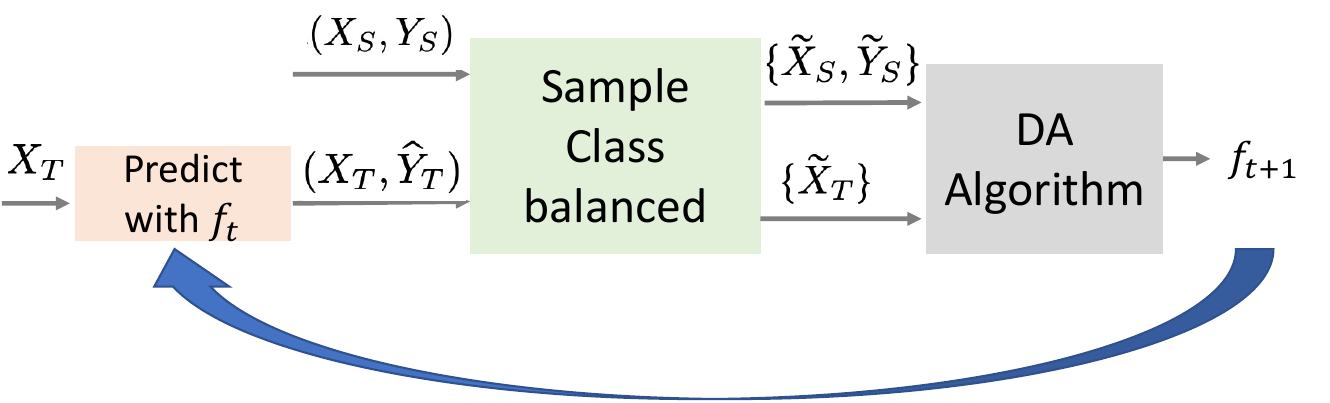}
    \hfill
    \vrule
    \hfill 
    \includegraphics[width=0.48\linewidth]{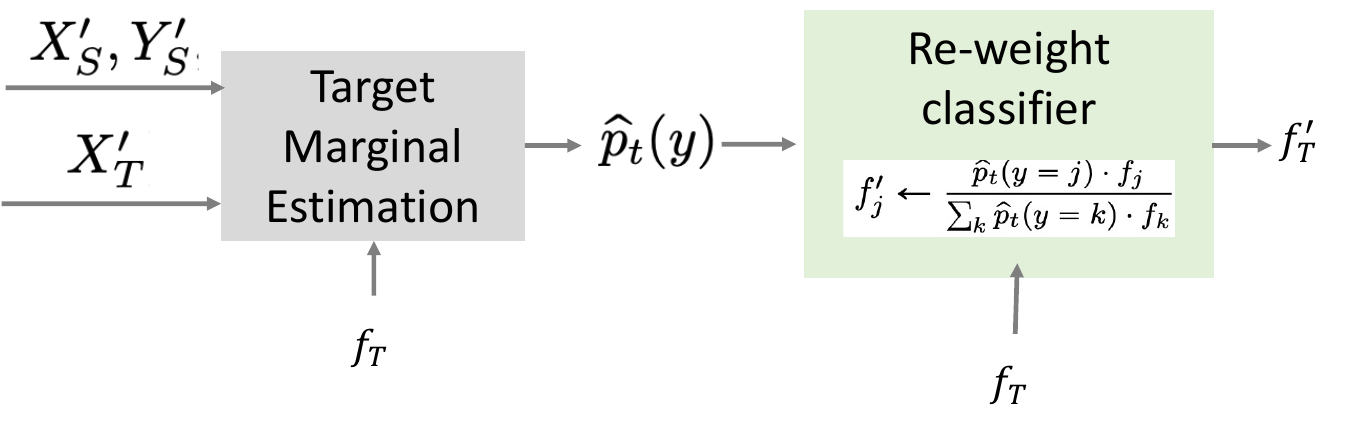}
    \caption{\textbf{(left)} Illustration of RS method at every iteration. \textbf{(right)} Illustration of post-hoc reweighting of the classifier with RW method.}
    \label{fig:RS-RW_method}
\end{figure}

\section{Methods to estimate target marginal under the stricter label shift assumption}
\label{app:label_shift_estimation}
In this section, we describe the methods proposed to estimate the target label marginal 
under the stricter label shift assumption. Recall that under the label shift assumption, 
$p_s(y)$ can differ from $p_t(y)$ but the class conditional stays the same, i.e., $p_t(x|y) = p_s(x|y)$. 
We focus our discussion on recent methods that leverage 
off-the-shelf classifier to yield consistent estimates under mild assumptions~\citep{lipton2018detecting, azizzadenesheli2019regularized, alexandari2019adapting, garg2020labelshift}. 
For simplicity, we assume we possess labeled source data 
$\{(x_1,y_1), (x_2, y_2), \ldots , (x_n, y_n)\}$ 
and unlabeled target data $\{x_{n+1}, x_{n+2}, \ldots, x_{n+m}\}$. 

\textbf{RLLS~~} First, we discuss \emph{Regularized Learning under Label Shift} (RLLS)~\citep{azizzadenesheli2019regularized} (a variant of \emph{Black Box Shift Estimation} (BBSE, \citet{lipton2018detecting})): 
moment-matching based estimators that leverage 
(possibly biased, uncalibrated, or inaccurate) 
predictions to estimate the shift. 
RLLS solves the following optimization problem to estimate the importance weights $w_t(y)  = \frac{p_t(y)}{p_s(y)}$ as:
\begin{equation}
    \wh w_t^\text{RLLS} = \argmin_{w \in \calW} \norm{\wh C_f w - \wh \mu_f }{2} + \lambda_{\text{RLLS}} \norm{w - 1}{2}\,. 
\end{equation}

where $\calW = \{ w \in \Real^d  | \sum_y w(y) p_s(y) = 1 \text{ and } \forall y \in \calY \quad w(y) > 0 \}$. $\wh C_f$ is empirical confusion matrix of the classifier $f$ on source data and $\wt \mu_f$ is the empirical average of predictions of the classifier $f$ on unlabeled target data. 
With labeled source data data, the empirical confusion matrix can be computed as: $$[\wh C_f]_{i,j} = \frac{1}{n} \sum_{k=1}^n f_i(x_k) \cdot \indict{y_k = j}\,.$$
To estimate target label marginal, we can multiple the estimated importance weights with the source label marginal (we can estimate source label marginal simply from labeled source data).

In our relaxed label shift problem, we use validation source data to compute the confusion matrix and use hold portion of target unlabeled data to compute $\mu_f$.
Unless specified otherwise, we use RLLS to estimate the target label marginal throughout the paper. We choose $\lambda_{\text{RLLS}}$ as suggested in the original paper~\citep{azizzadenesheli2019regularized}.

\textbf{MLLS~~} Next, we discuss Maximum Likelihood Label Shift (MLLS) \citep{saerens2002adjusting, alexandari2019adapting}: an Expectation Maximization (EM) algorithm that 
maximize the likelihood of observed unlabeled target data to estimate target label marginal 
assuming access to a classifier that outputs
the source calibrated probabilities. In particular, MLLS uses the following objective:

\begin{equation}
    \wh w_t^\text{MLLS} = \argmin_{w \in \calW} \frac{1}{m} \sum_{i=1} \log(w^Tf(x_{i+n})) \,,
\end{equation}
where $f$ is the classifier trained on source and $\calW$ is the same constrained set defined above. We can again estimate the target label marginal by simply multiplying the estimated importance weights with the source label marginal.

\textbf{Baseline estimator~~} Given a classifier $f$, we can estimate the target label marginal as simply the average of the classifier output on unlabeled target data, i.e., 
\begin{equation}
    \wh p_t ^\text{baseline} = \frac{1}{m}\sum_{i = 1} f(x_{i+n}) \,. 
\end{equation}

Note that all of the methods discussed before leverage an off-the-shelf classifier $f$. 
Hence, we experiment with classifiers obtained with 
various deep domain adaptation heuristics to estimate the target 
label marginal.

Having obtained an estimate of target label marginal, 
we can simply re-weight the classifier with $\wh p_t$ as $f^\prime_j = \mfrac{\wh p_t(y=j) \cdot f_j}{\sum_k \wh p_t(y=k) \cdot f_k}$ for all $j \in \calY$. Note that, if we train $f$ on a non-uniform source class-balance (and without re-balancing as in Step 1 of \algoref{alg:LSBench}), then we can re-weight the classifier with importance-weights $\wh w_t$ as  $f^\prime_j = \mfrac{\wh w_t(y=j) \cdot f_j}{\sum_k \wh w_t(y=k) \cdot f_k}$ for all $j \in \calY$.

\section{{Theoretical Definition for Relaxed Label Shift}}
\label{app:def}
\update{Domain adaptation problems are, in general, ill-posed~\citep{ben2010impossibility}. 
Several attempts have been made to investigate additional assumptions that render the problem well-posed. 
One such example includes the label-shift setting, where 
$p(x|y)$ does not change 
but that $p(y)$ can.
Under label shift, two challenges arise:
(i) estimate the target label marginal $p_t(y)$; 
and (ii) train a classifier $f$ 
to maximize the performance on the target domain. 
However, these assumptions are 
typically, to some degree, violated in practice.
This paper aims to relax this assumption 
and focuses on \emph{relaxed label shift} setting. 
In particular, we assume that the label distribution
can shift from source to target arbitrarily 
but that $p(x|y)$ varies between source and target
in some comparatively restrictive way \update{(e.g., shifts arising naturally in the real world like ImageNet~\citep{russakovsky2015imagenet} to ImageNetV2~\citep{recht2019imagenet})}. 
}

\update{Mathematically, 
we assume a divergence-based restriction on 
$p(x|y)$, i.e., for some small $\epsilon > 0$ and distributional distance $\calD$, we have 
$\max_y \calD(p_t(x|y), p_t(x|y)) \le \epsilon$
but allowing an arbitrary shift in the label marginal $p(y)$.
Previous works have defined these constraints in different ways~\citep{wu2019domain, tachet2020domain, kumar2020understanding}.}

\update{In particular, we can use Wasserstein-infinity distance to define our constraint.
First, we define Wasserstein given probability measures $p, q$ on $\calX$: 
\begin{align*}
    W_\infty(p,q) = \inf \{ \sup_{x \in \Real^d } \norm{f(x) - x}{2}: f: \Real^d \to \Real^d, f_{\#} p = q \},
\end{align*}
where $\#$ denotes the push forward of a measure, i.e., for every set $S \subseteq \Real^d, p(S) = p(f^{-1}(S))$. Intuitively, $W_\infty$ moves points from the distribution $p$ to $q$ by distance at most $\epsilon$ to match the distributions.  Hence, our $D \defeq \max_y W_\infty(p_s(x|y), p_t(x|y)) \le \epsilon$. Similarly, we can define our distribution constraint in KL or TV distances. 
We can define our constraint in a representation space $\calZ$ obtained by projection inputs $x\in \calX$ with a function $h: \calX \to \calZ$. Intuitively, we want to define the distribution distance with some $h$ that captures all the required information for predicting the label of interest but satisfies a small distributional divergence in the projected space.
However, in practice, it's hard to empirically verify these distribution distances for small enough $\epsilon$ with finite samples. Moreover, we lack a rigorous characterization of the sense in which those shifts arise in popular DA benchmarks, and since, the focus of our work is on the empirical evaluation with real-world datasets, we leave a formal investigation for future work. 
}.

\section{{Target Marginal Estimation and its Effect on Accuracy}}
\label{app:target_marginal_estimation}
\subsection{{Vision Datasets}}

\begin{figure*}[ht]
    \centering 
    \begin{subfigure}[b]{0.87\linewidth}
    \centering 
    \includegraphics[width=0.32\linewidth]{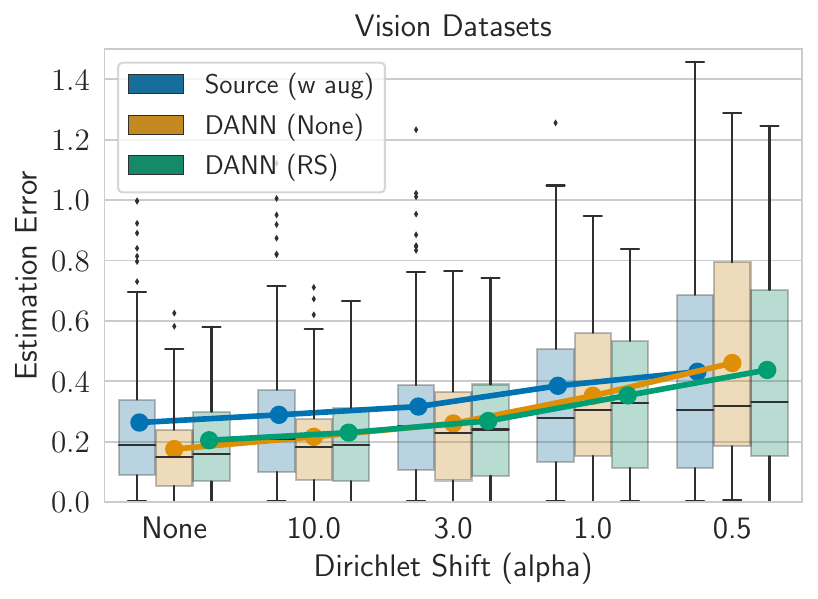}
    \includegraphics[width=0.32\linewidth]{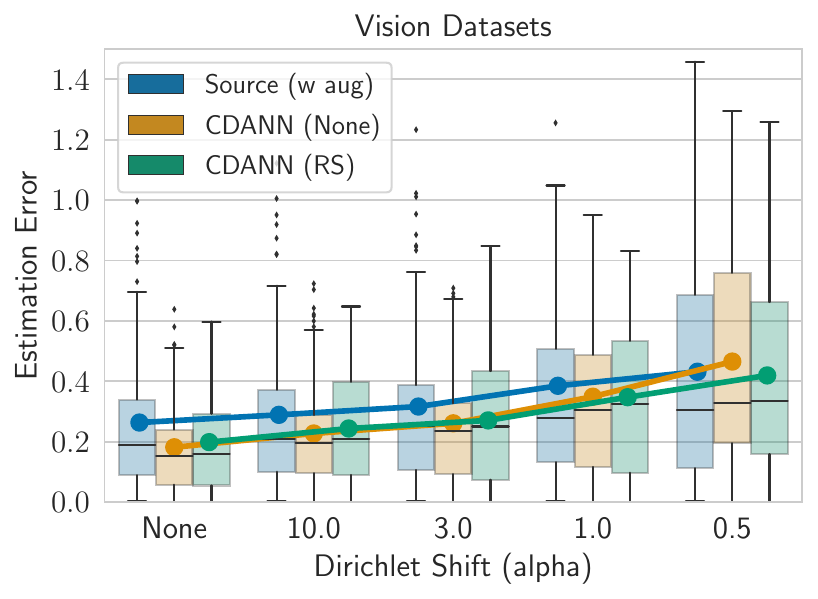}
    \includegraphics[width=0.32\linewidth]{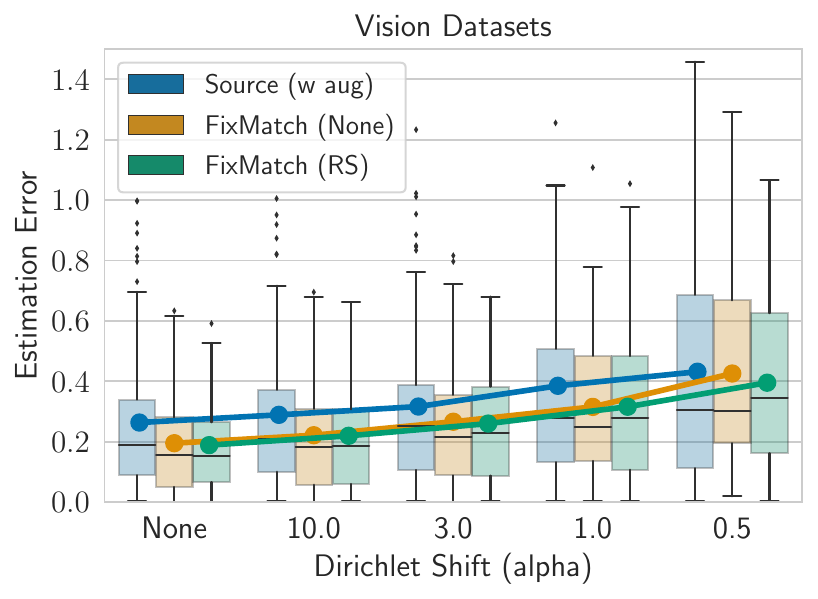}

    \includegraphics[width=0.32\linewidth]{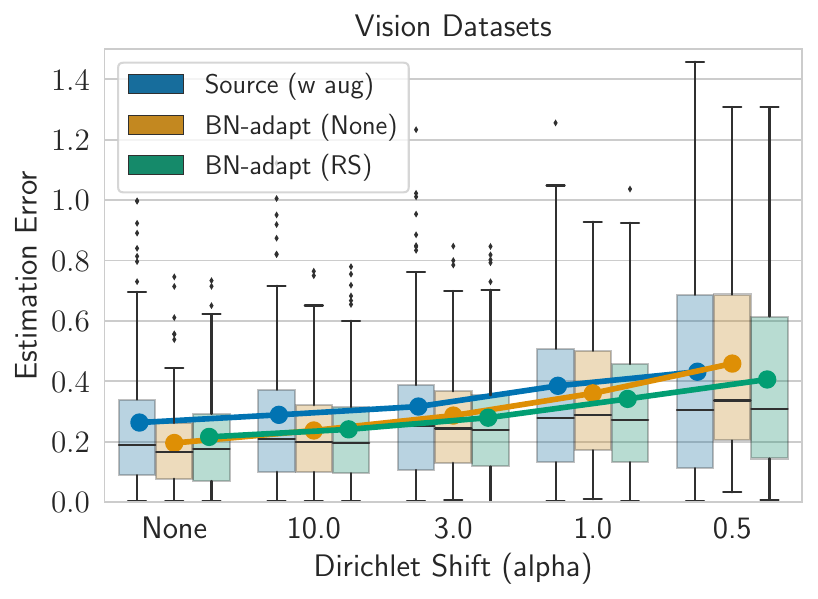}
    \includegraphics[width=0.32\linewidth]{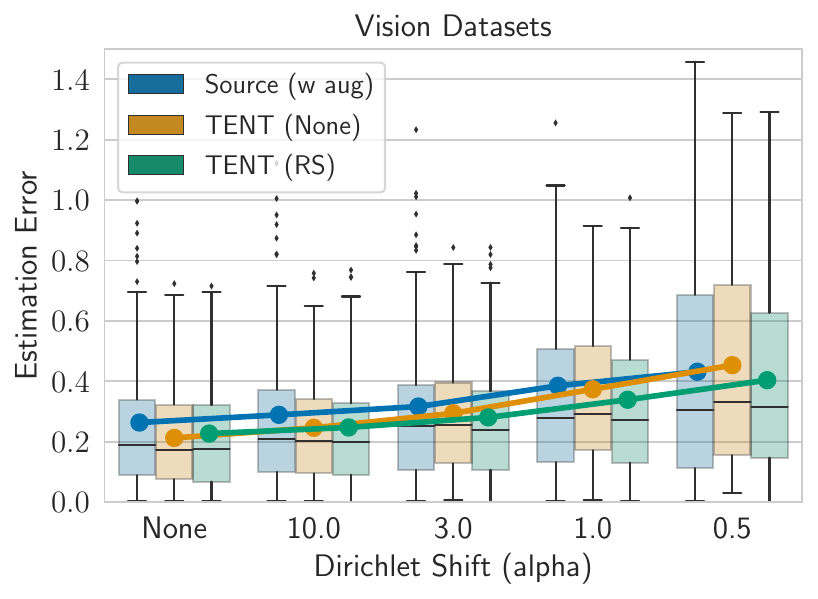}
    \includegraphics[width=0.32\linewidth]{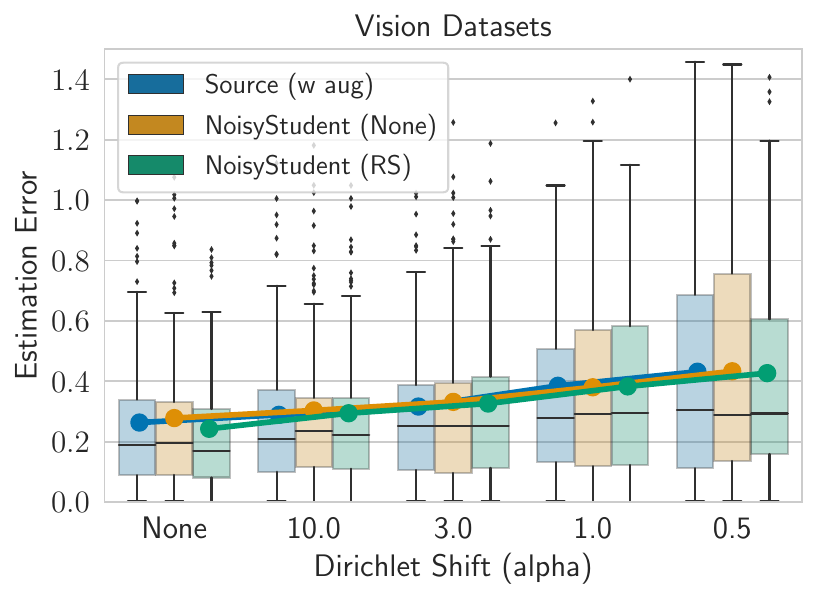}
    \caption{{Target label marginal estimation ($\ell_1$) error with RLLS and classifiers obtained with different DA methods}}
    \end{subfigure}

    \begin{subfigure}[b]{0.87\linewidth}
    \includegraphics[width=0.32\linewidth]{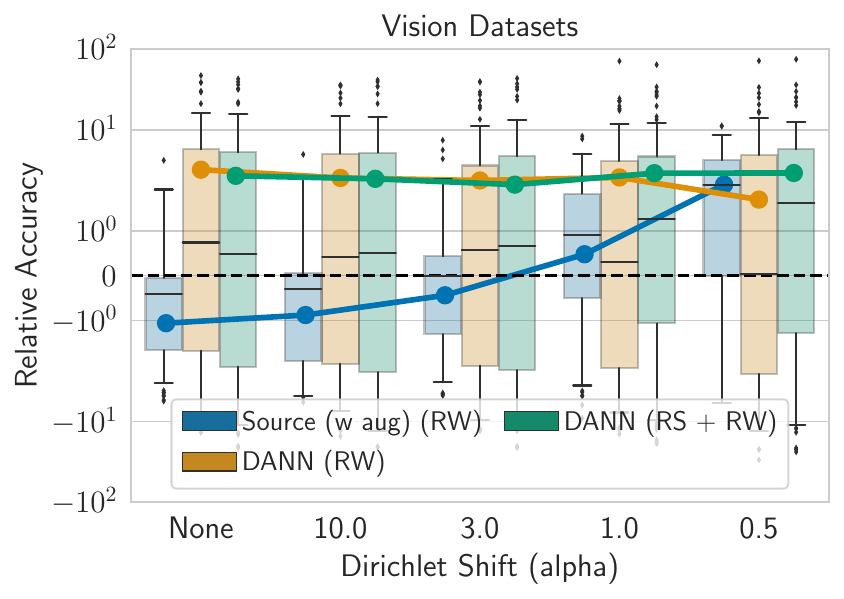}
    \includegraphics[width=0.32\linewidth]{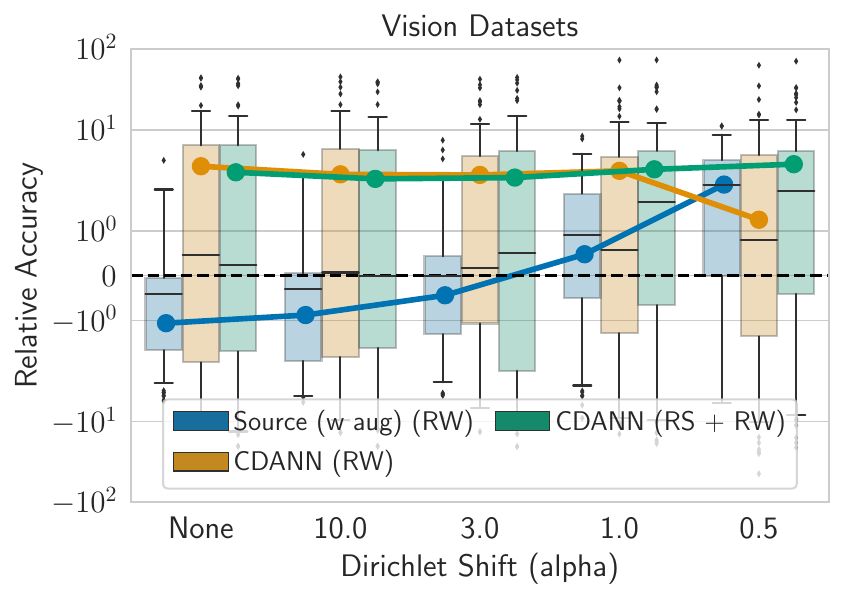}
    \includegraphics[width=0.32\linewidth]{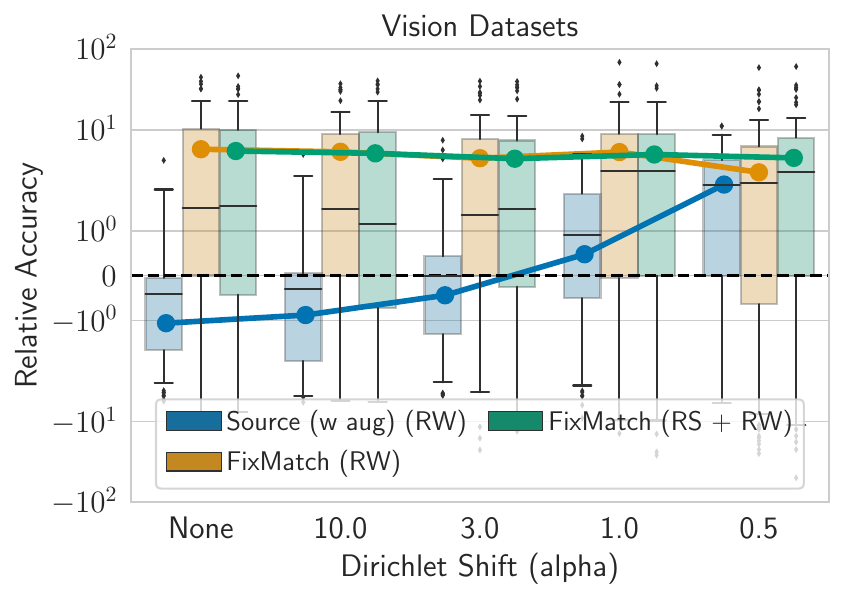}

    \includegraphics[width=0.32\linewidth]{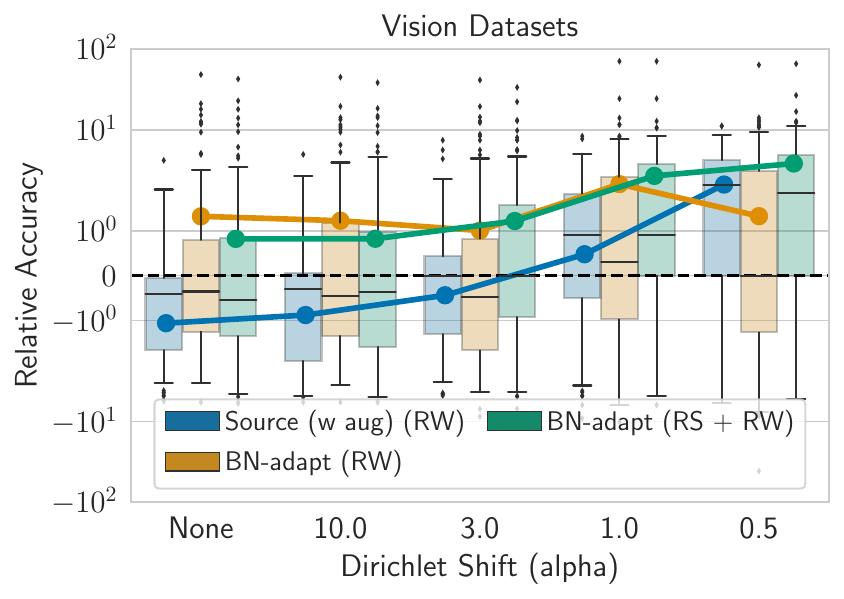}
    \includegraphics[width=0.32\linewidth]{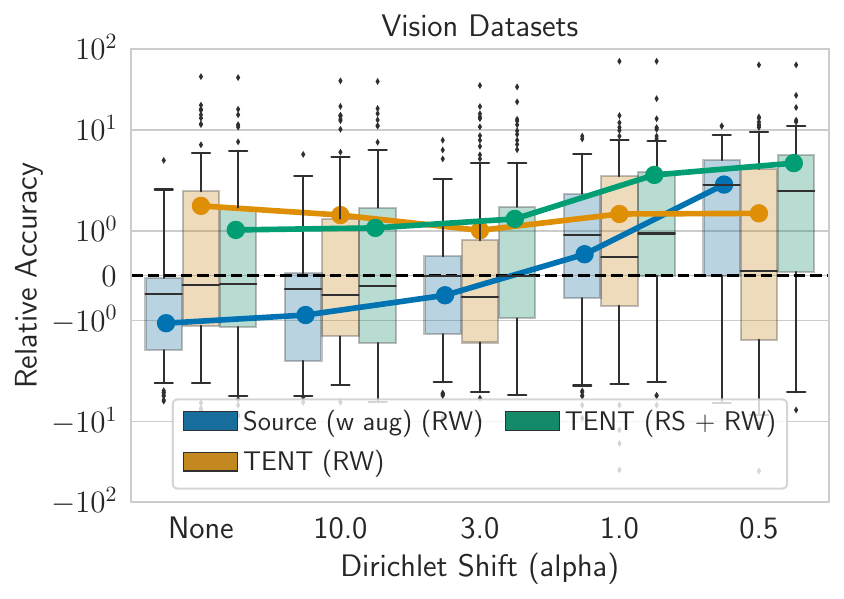}
    \includegraphics[width=0.32\linewidth]{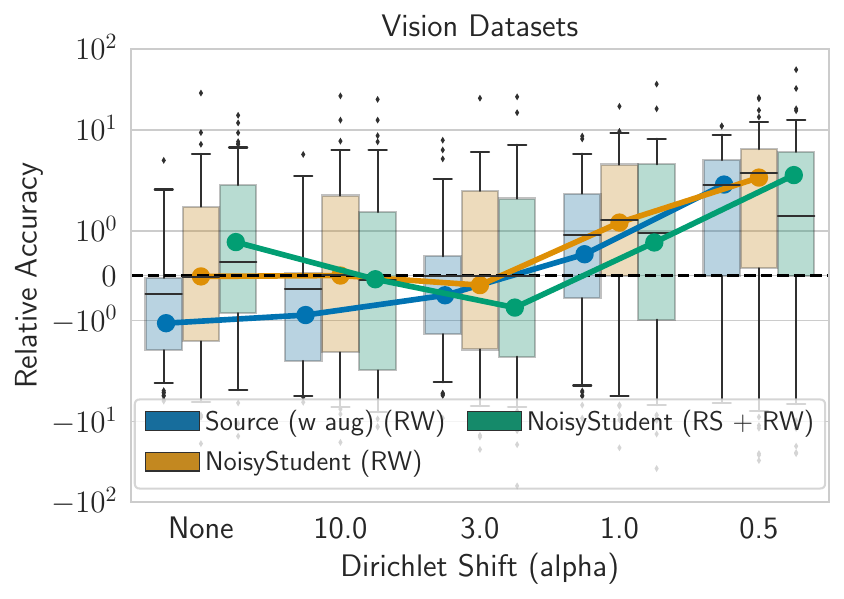}
    \caption{{Relative performance of DA methods when paired with RW corrections}}
    \end{subfigure}
    \caption{{{\emph{Target label marginal estimation ($\ell_1$) error and relative performance with RLLS and classifiers obtained with different DA methods.}}
    Across all shift severities (except for $\alpha = 0.5$) in vision datasets, RLLS with classifiers obtained with DA methods improves over RLLS with a source-only classifier. Correspondingly, we see significantly improved performance with post-hoc RW correction applied to classifiers trained with DA methods as compared to when applied to source-only models.}}
\end{figure*}

\pagebreak
\subsection{{Tabular Datasets}}

\begin{figure}[H]
    \centering 
    \begin{subfigure}[b]{0.87\linewidth}
    \centering 
    \includegraphics[width=0.32\linewidth]{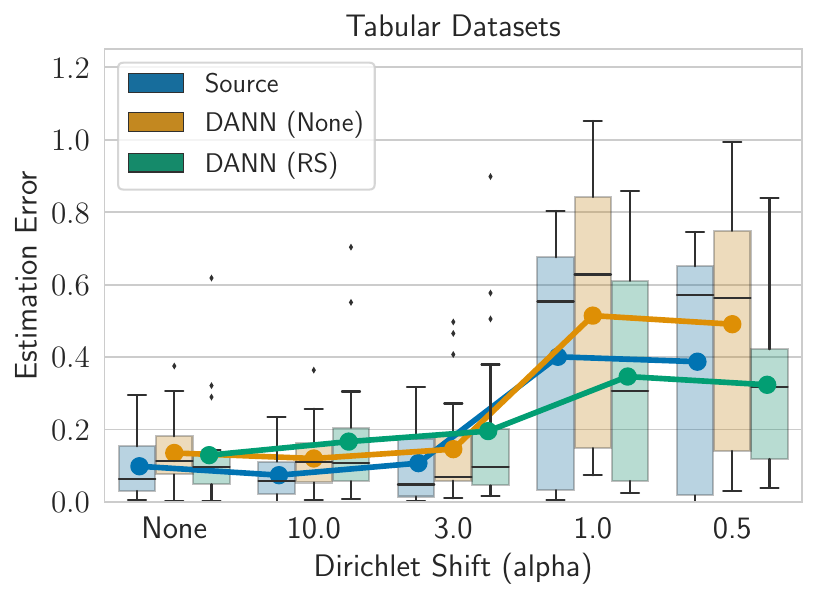}
    \includegraphics[width=0.32\linewidth]{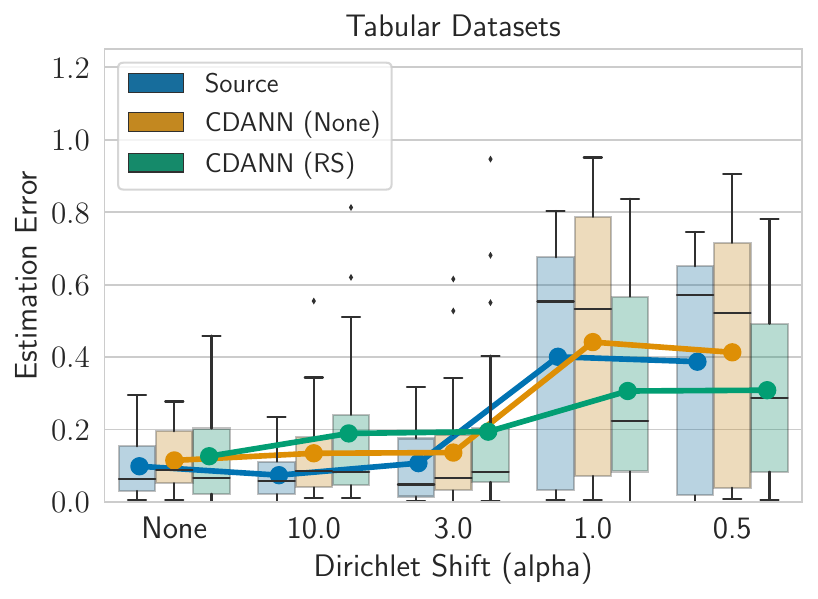}
    \includegraphics[width=0.32\linewidth]{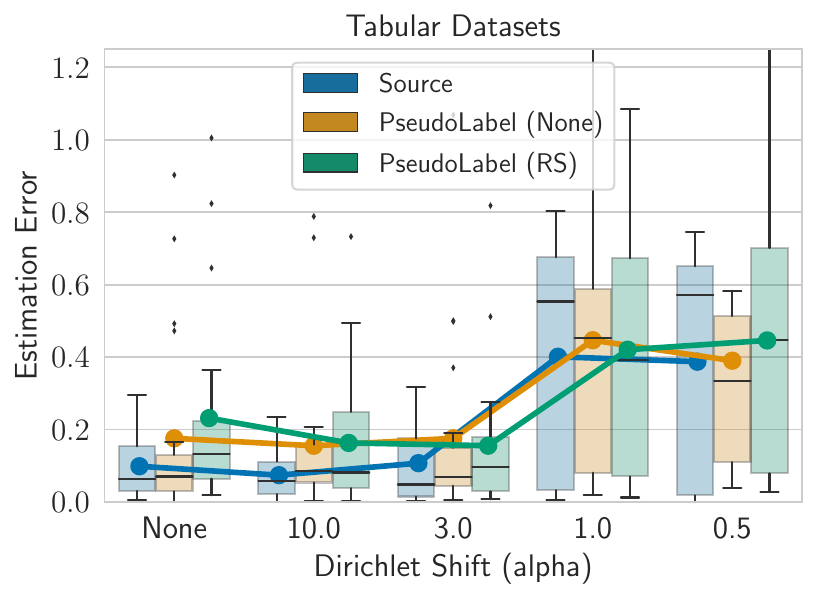}

    \caption{{Target label marginal estimation ($\ell_1$) error with RLLS and classifiers obtained with different DA methods}}
    \end{subfigure}

    \begin{subfigure}[b]{0.87\linewidth}
    \includegraphics[width=0.32\linewidth]{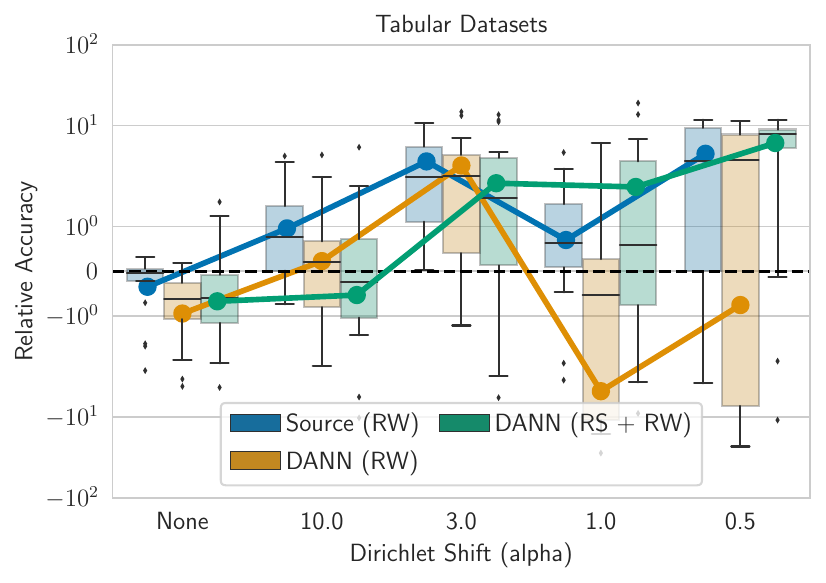}
    \includegraphics[width=0.32\linewidth]{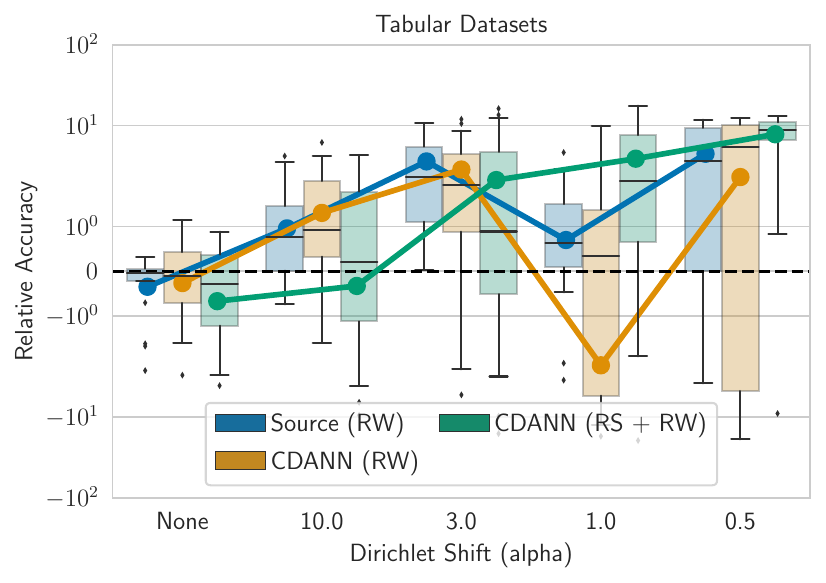}
    \includegraphics[width=0.32\linewidth]{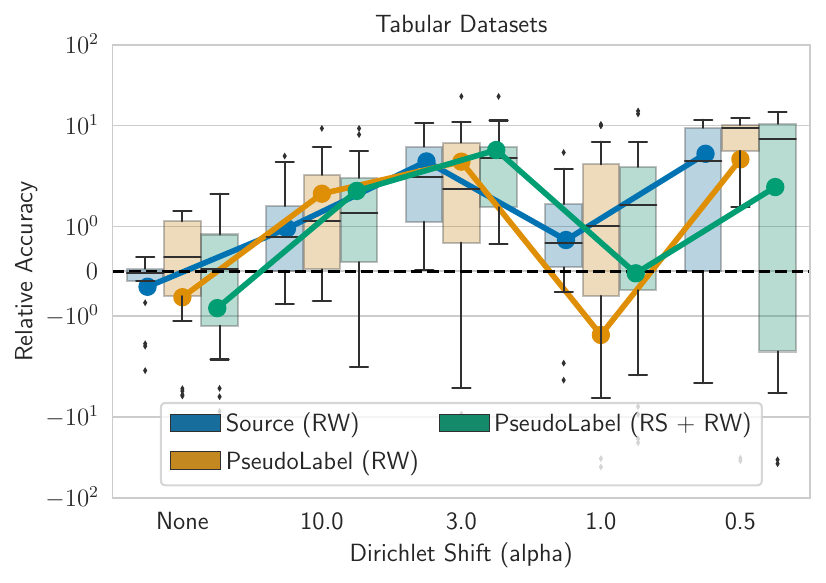}
    \caption{{Relative performance of DA methods when paired with RW corrections}}
    \end{subfigure}
    \caption{{{\emph{Target label marginal estimation ($\ell_1$) error and relative performance with RLLS and classifiers obtained with different DA methods.}}
    For tabular datasets, RLLS with classifiers obtained with DA methods improves over RLLS with a source-only classifier for severe target label marginal shifts. Correspondingly for severe target label marginal shifts, we see improved performance with post-hoc RW correction applied to classifiers trained with DA methods as compared to when applied to source-only models.}}
\end{figure}

\subsection{{NLP Datasets}}

\begin{figure}[H]
    \centering 
    \begin{subfigure}[b]{0.87\linewidth}
    \centering 
    \includegraphics[width=0.32\linewidth]{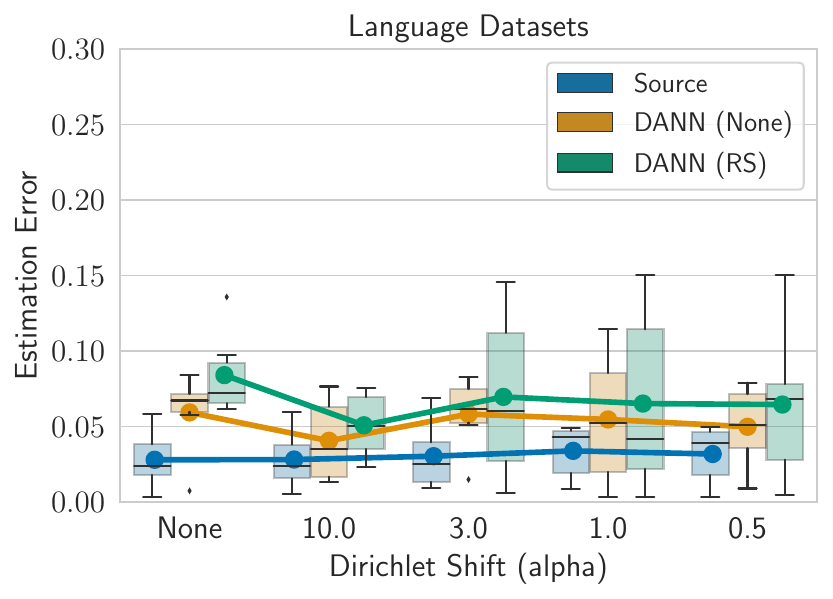}
    \includegraphics[width=0.32\linewidth]{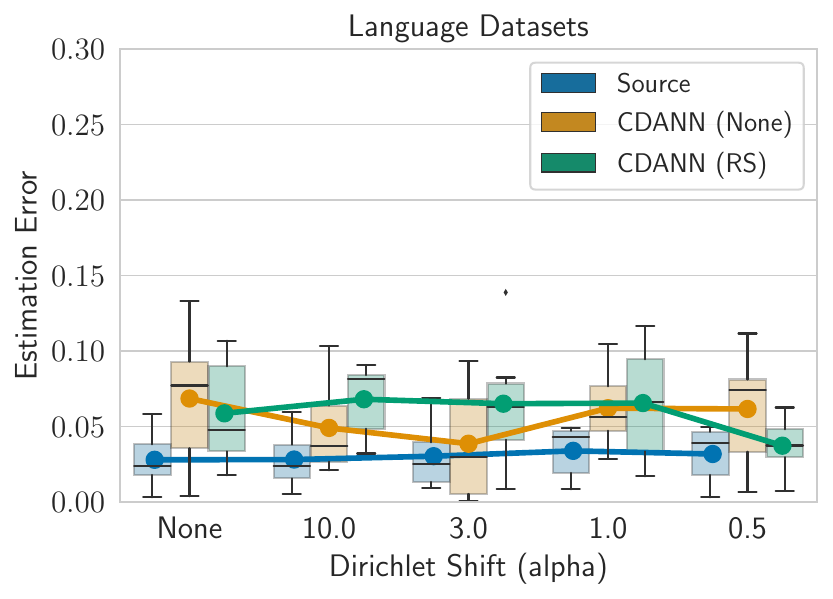}
    \includegraphics[width=0.32\linewidth]{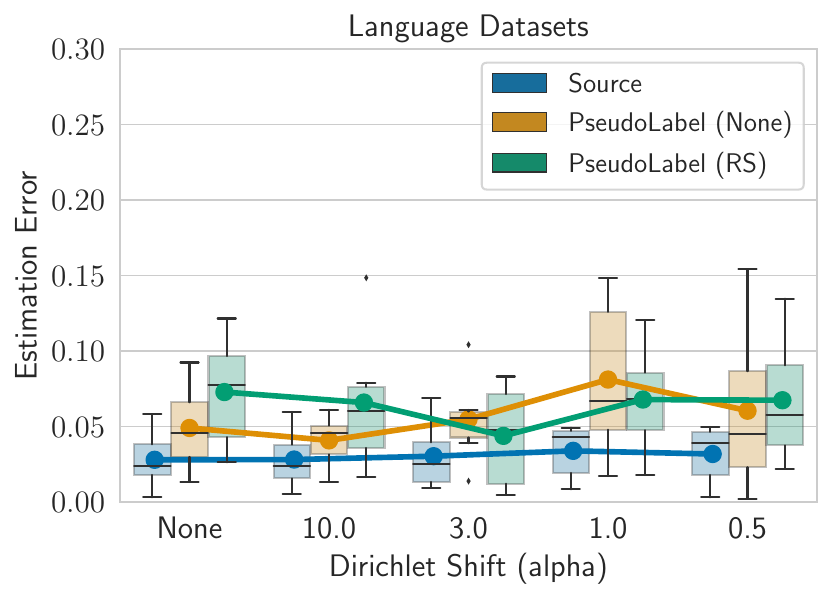}

    \caption{{Target label marginal estimation ($\ell_1$) error with RLLS and classifiers obtained with different DA methods}}
    \end{subfigure}

    \begin{subfigure}[b]{0.87\linewidth}
    \includegraphics[width=0.32\linewidth]{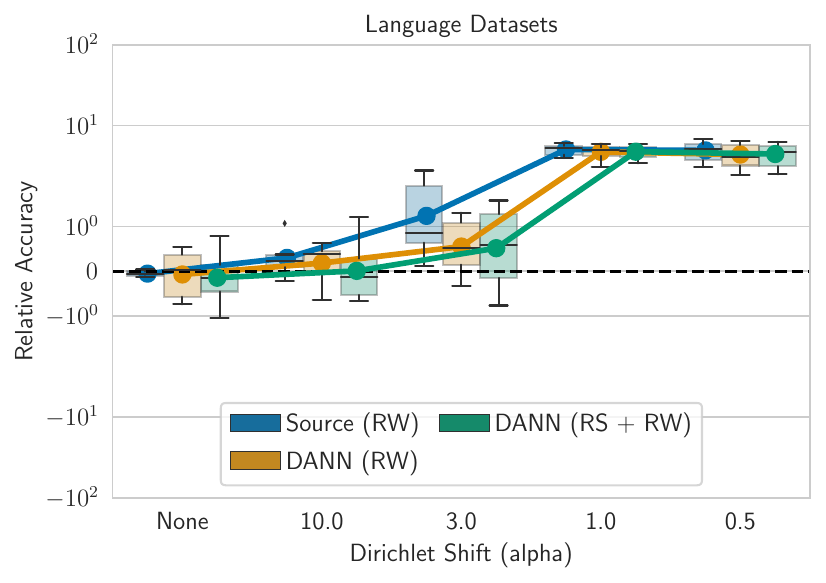}
    \includegraphics[width=0.32\linewidth]{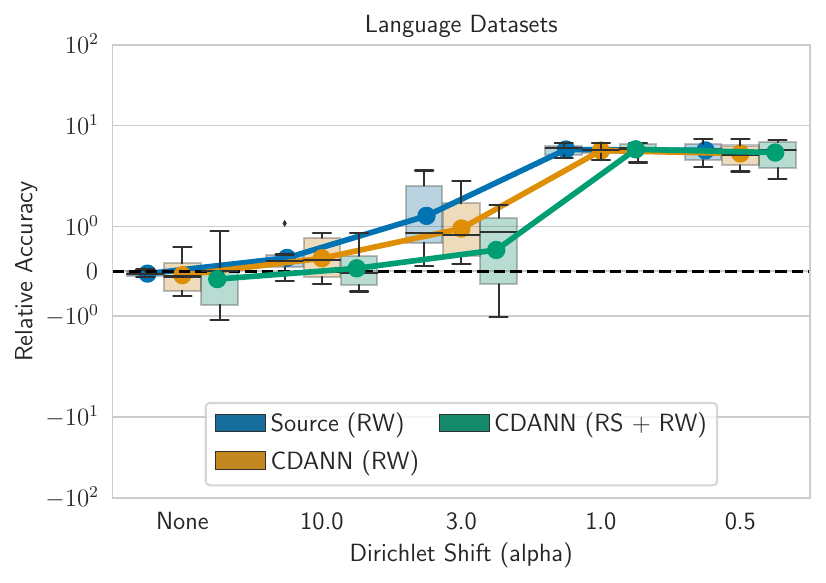}
    \includegraphics[width=0.32\linewidth]{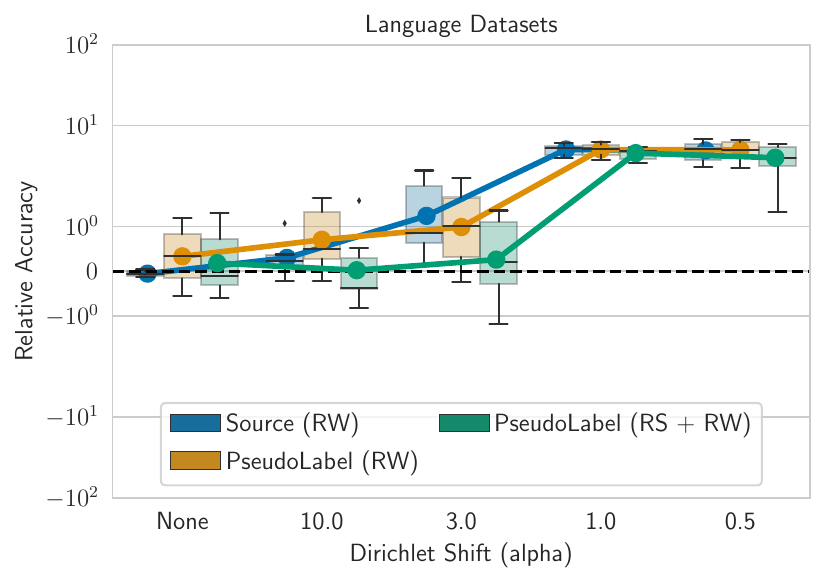}
    \caption{{Relative performance of DA methods when paired with RW corrections}}
    \end{subfigure}
    \caption{{{\emph{Target label marginal estimation ($\ell_1$) error and relative performance with RLLS and classifiers obtained with different DA methods.}}
    For NLP datasets, RLLS with source-only classifiers performs better than RLLS with classifiers obtained with DA methods. Correspondingly, we see improved performance with post-hoc RW correction applied to source-only models over classifiers trained with DA methods.}}
\end{figure}

\clearpage

\subsection{{Comparison of different target label marginal estimation methods}}

\begin{figure*}[ht]
    \centering 
    \includegraphics[width=0.32\linewidth]{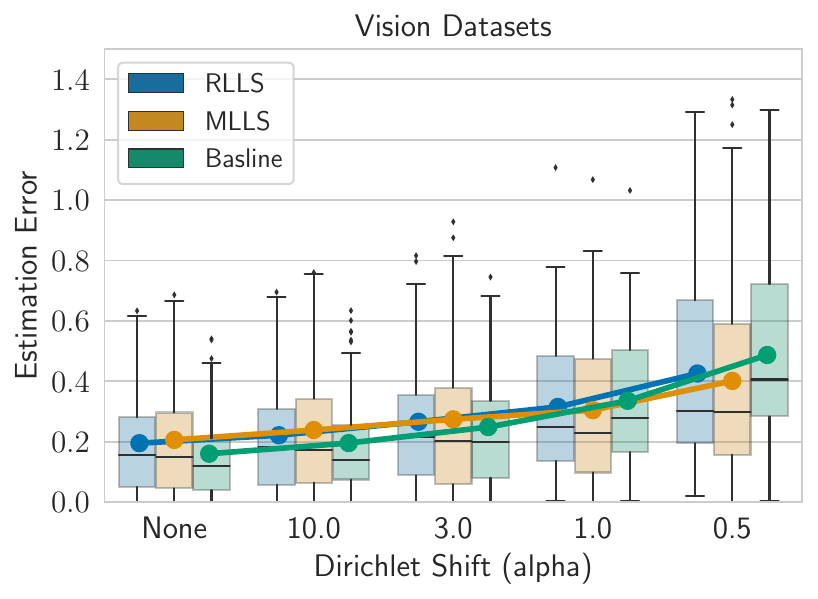}
    \includegraphics[width=0.32\linewidth]{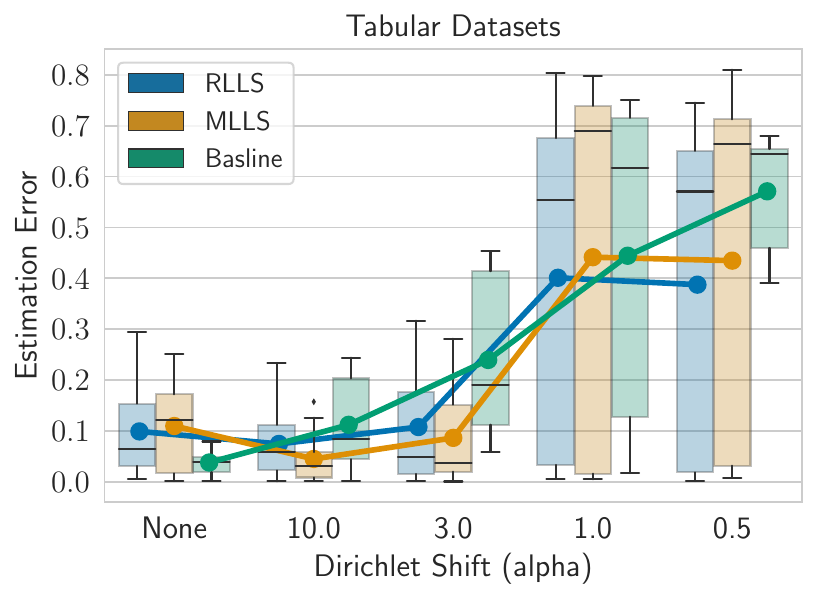}
    \includegraphics[width=0.32\linewidth]{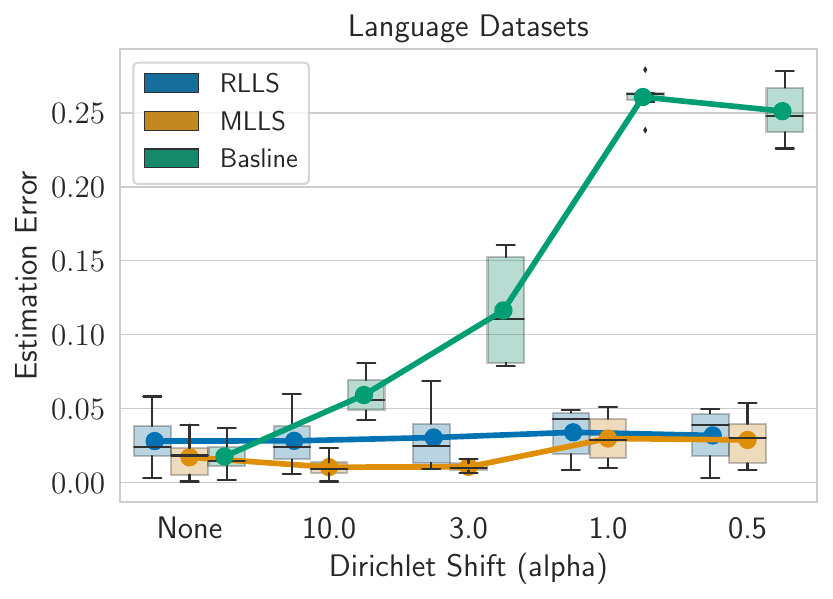}
    
    \caption{{\emph{Comparison of different target label marginal estimation methods.} We plot estimation errors with different methods with the source-only classifier. 
    For all modalities, we observe a trade-off between estimation error with the baseline method and RLLS (or MLLS) method with severity in target marginal shift. 
    }}
    \label{fig:estimation_method_tradeoff}
\end{figure*}

\section{{Results with Oracle Early Stopping Criterion}}
\label{app:oracle_results}
In this section, we report results with oracle early stopping criterion. 
{On vision and tabular datasets}, we observe differences in performance 
when using target performance versus 
source hold-out 
performance for model selection. 
This highlights 
a more nuanced behavior than
the accuracy-on-the-line 
phenomena~\citep{miller2021accuracy, recht2019imagenet}. 
We hope to study this contrasting behavior in
more detail in future work. 

\begin{figure}[H]
 \centering
    \includegraphics[width=0.4\linewidth]{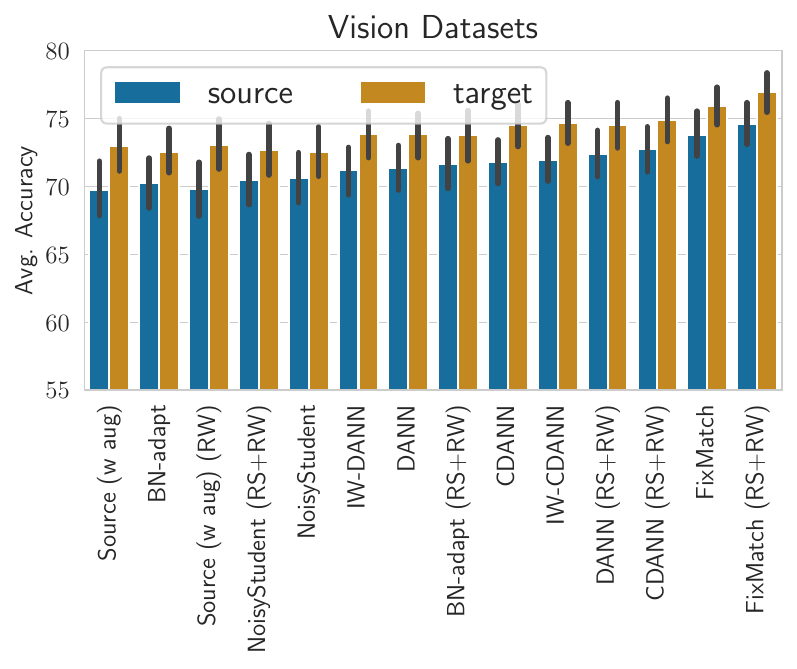}\\
    \includegraphics[width=0.4\linewidth]{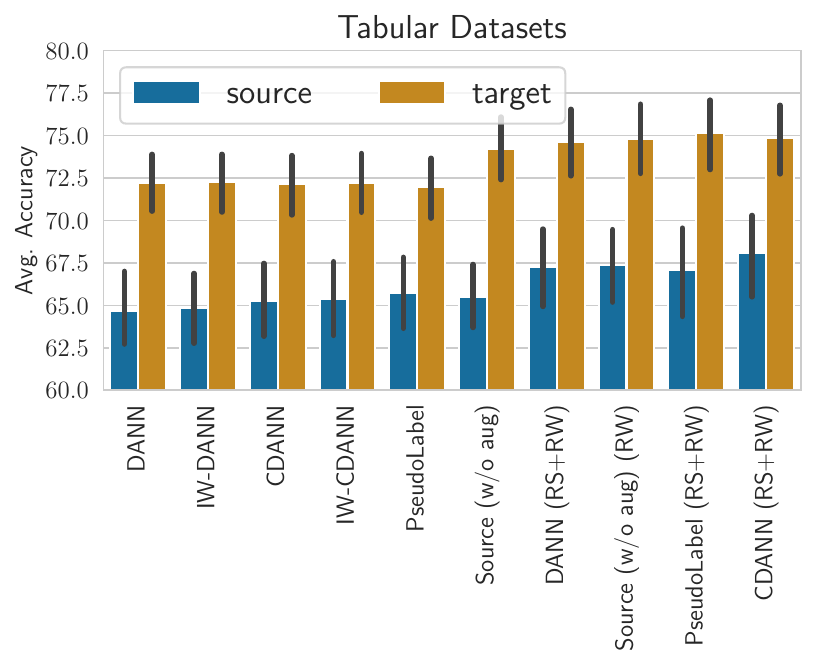}
    \includegraphics[width=0.4\linewidth]{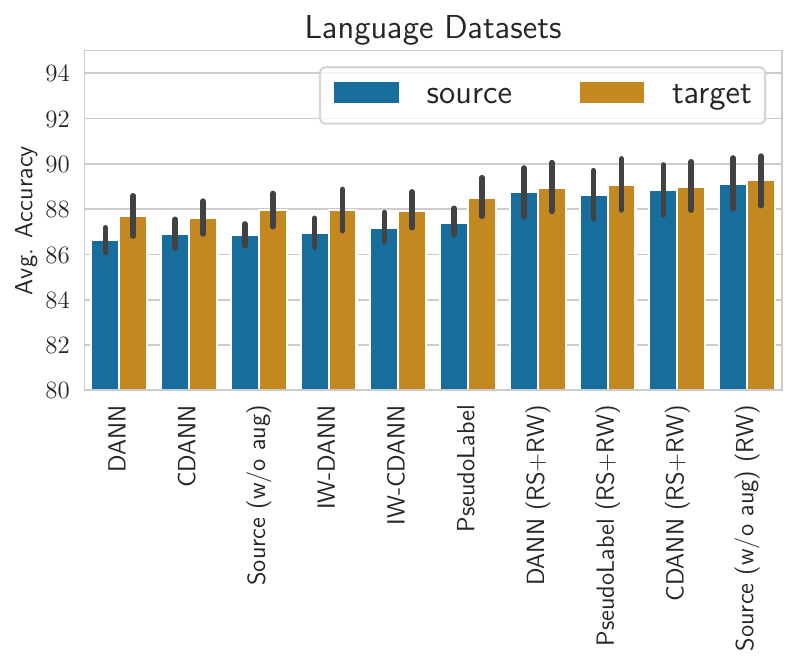}
    \caption{{\emph{Average accuracy of different DA methods aggregated across all distribution pairs in each modality.} 
    We compare the performance with early stopping point obtained with source validation performance and target validation performance.}}
    \label{fig:oracle_selection}
\end{figure}

\begin{figure}[ht]
 \centering
    \includegraphics[width=0.24\linewidth]{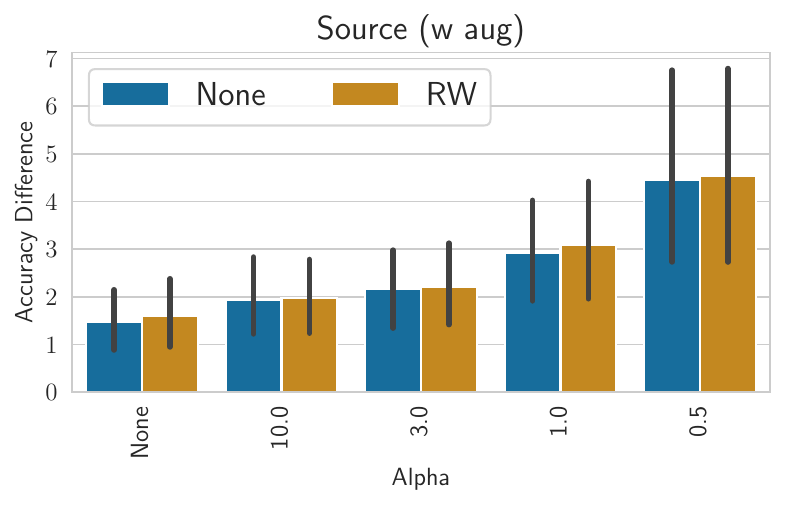}
    \includegraphics[width=0.24\linewidth]{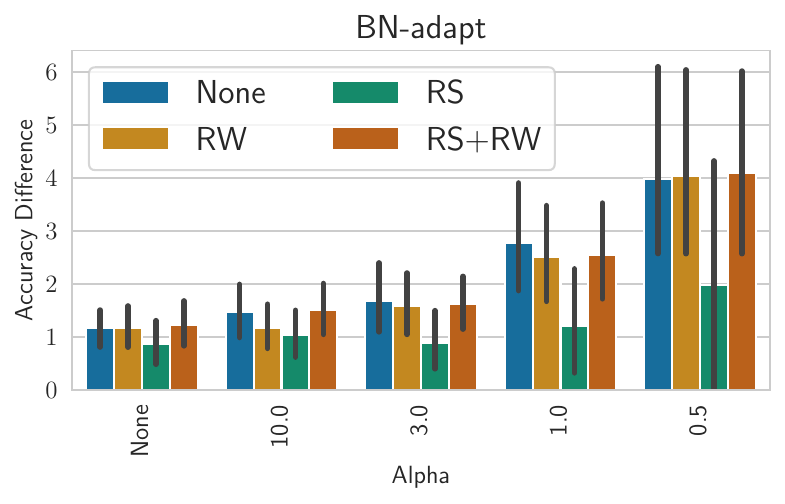}
    \includegraphics[width=0.24\linewidth]{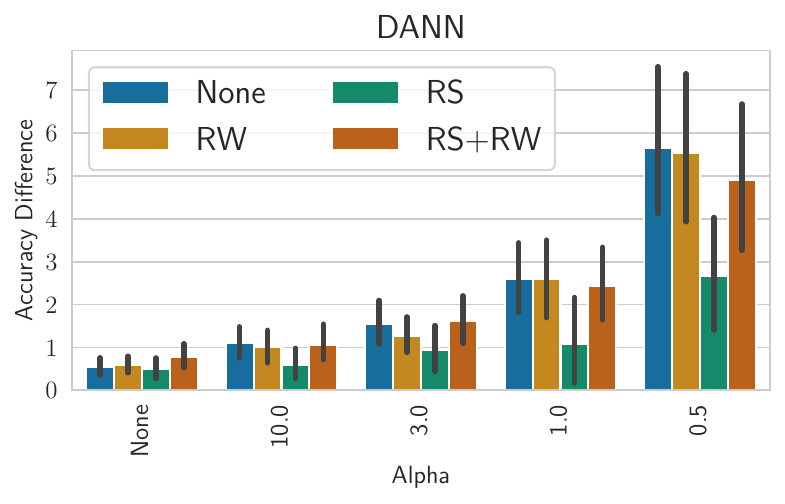}
    \includegraphics[width=0.24\linewidth]{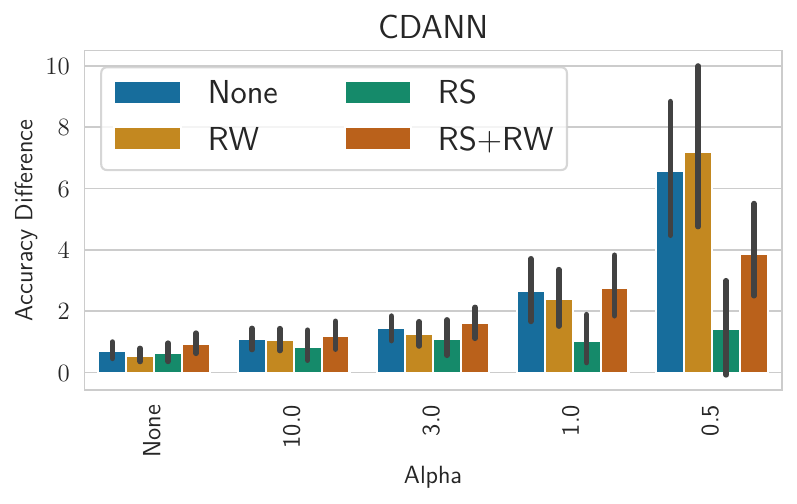}

    \includegraphics[width=0.24\linewidth]{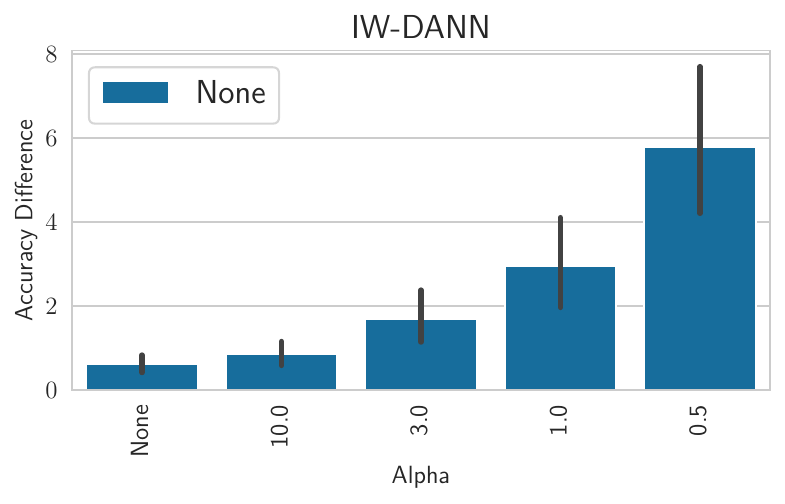}
    \includegraphics[width=0.24\linewidth]{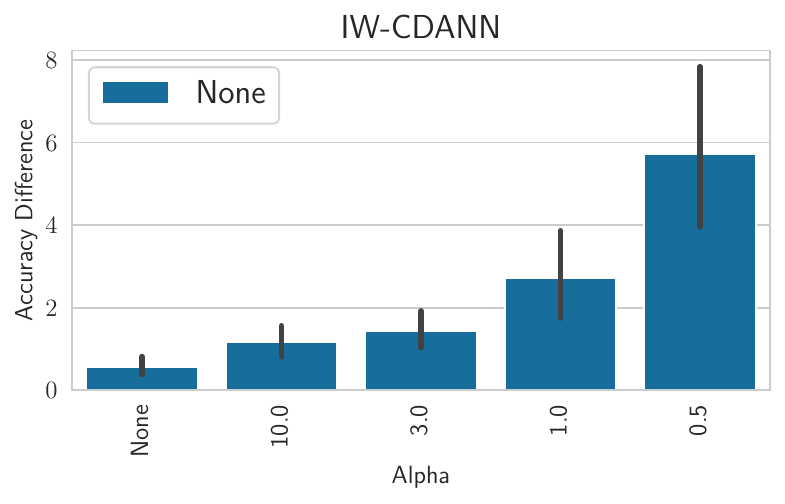}
    \includegraphics[width=0.24\linewidth]{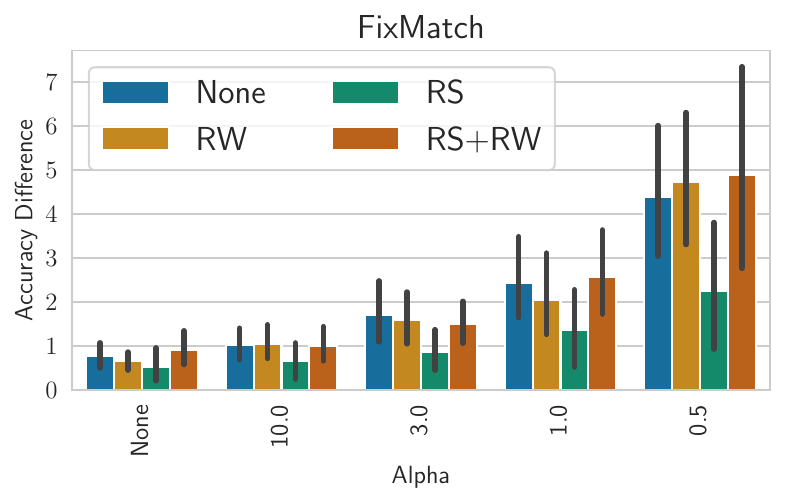}
    \includegraphics[width=0.24\linewidth]{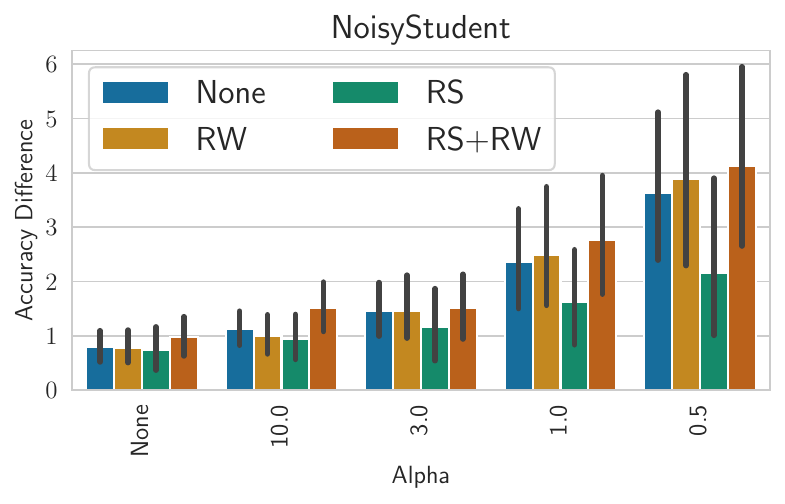}
    \caption{{\emph{Accuracy difference between using source and target performance as early stopping criteria for different DA methods aggregated across all distribution shift pairs in vision datasets.} We observe that as the shift severity increases (i.e., as $\alpha$ decreases), the accuracy difference increases for all the methods.}}
    \label{fig:oracle_selection_ES}
\end{figure}

\begin{figure}[H]
 \centering
    \includegraphics[width=0.24\linewidth]{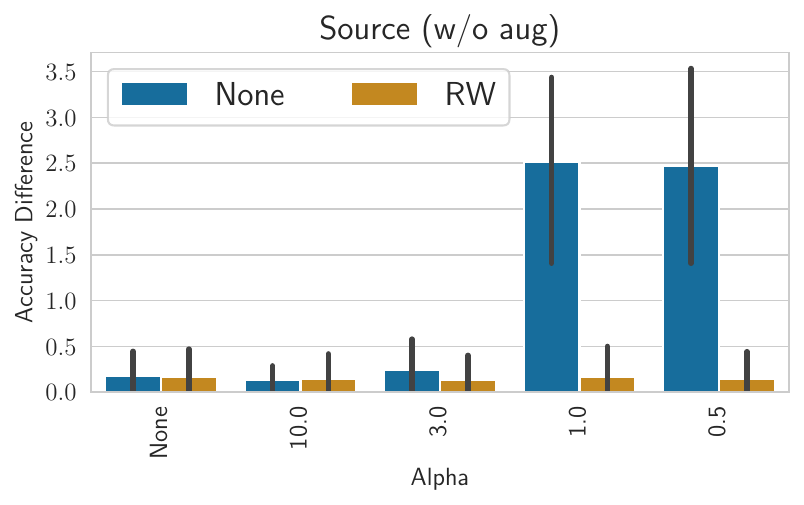}
    \includegraphics[width=0.24\linewidth]{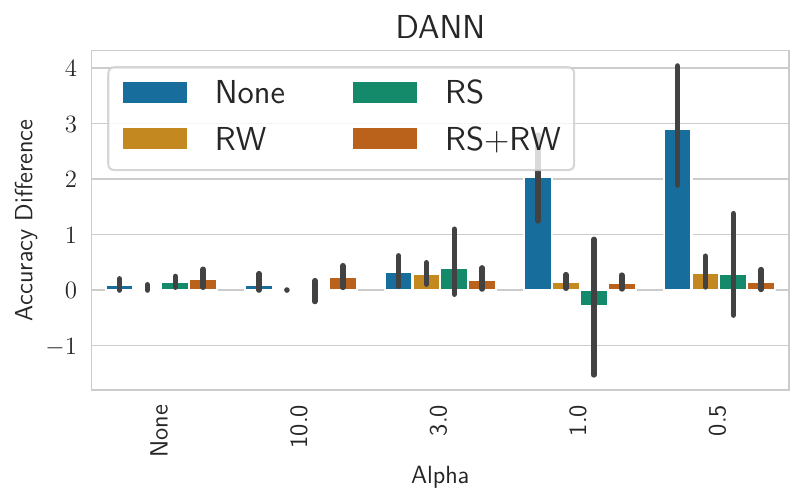}
    \includegraphics[width=0.24\linewidth]{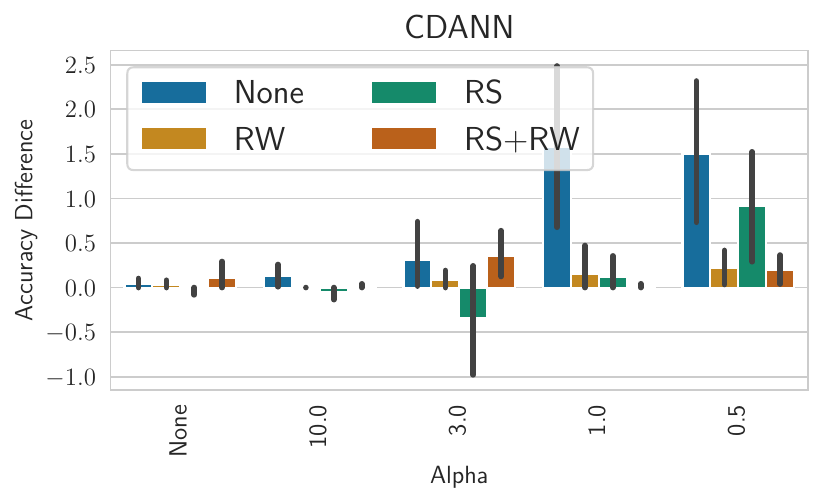} \\

    \includegraphics[width=0.24\linewidth]{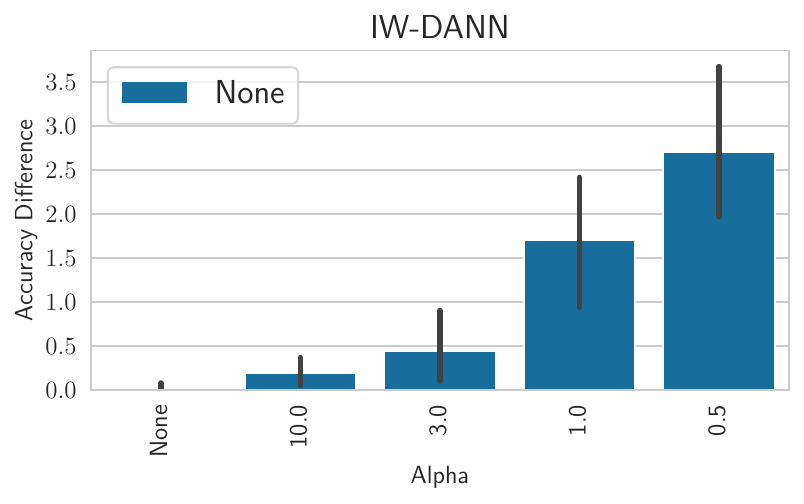}
    \includegraphics[width=0.24\linewidth]{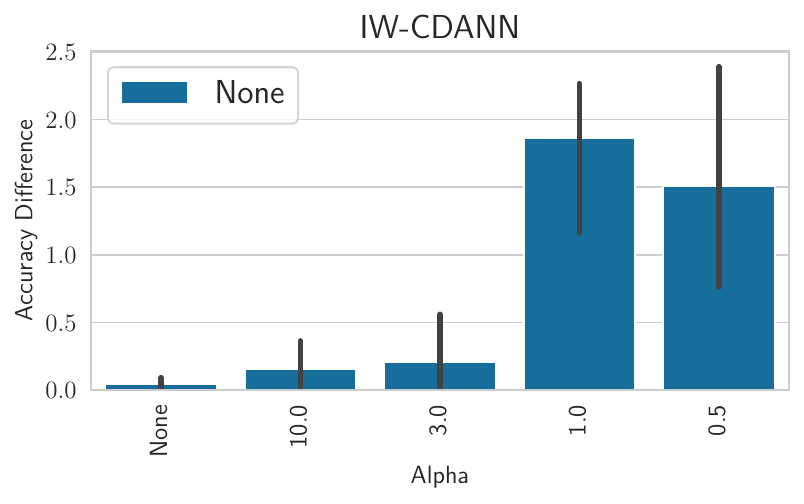}
    \includegraphics[width=0.24\linewidth]{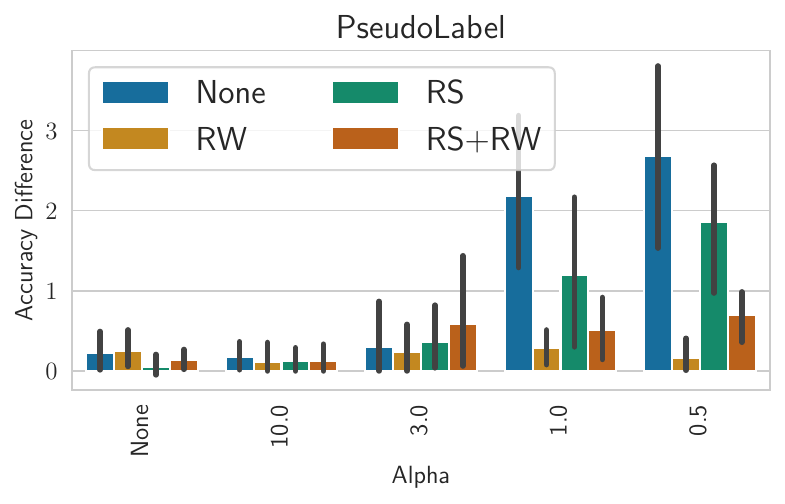}
    \caption{{\emph{Accuracy difference between using source and target performance as early stopping criteria for different DA methods aggregated across all distribution shift pairs in language datasets.} We observe that as the shift severity increases (i.e., as $\alpha$ decreases), the accuracy difference increases for all the methods without any correction. With RS and RW corrections, we observe that the accuracy difference remains relatively constant as the shift severity increases.}}
    \label{fig:oracle_selection_NLP}
\end{figure}

\begin{figure}[H]
 \centering
    \includegraphics[width=0.24\linewidth]{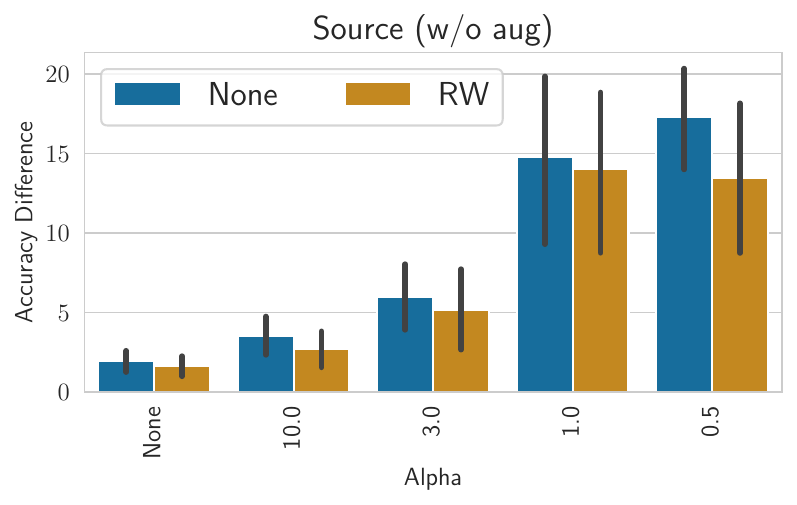}
    \includegraphics[width=0.24\linewidth]{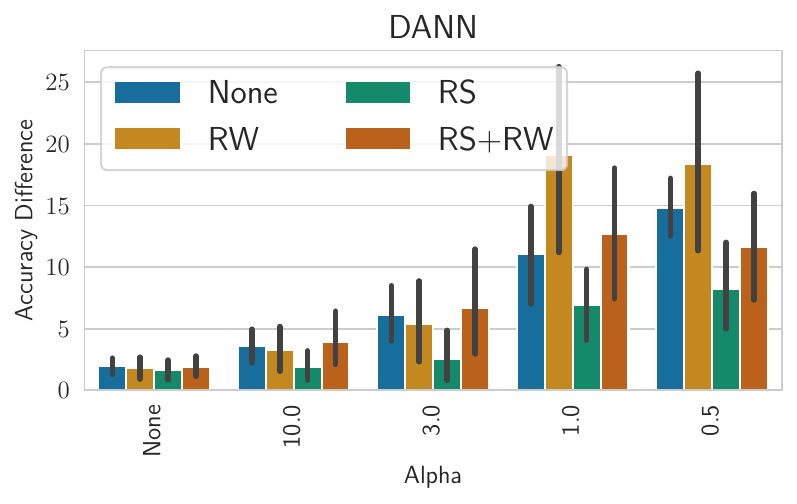}
    \includegraphics[width=0.24\linewidth]{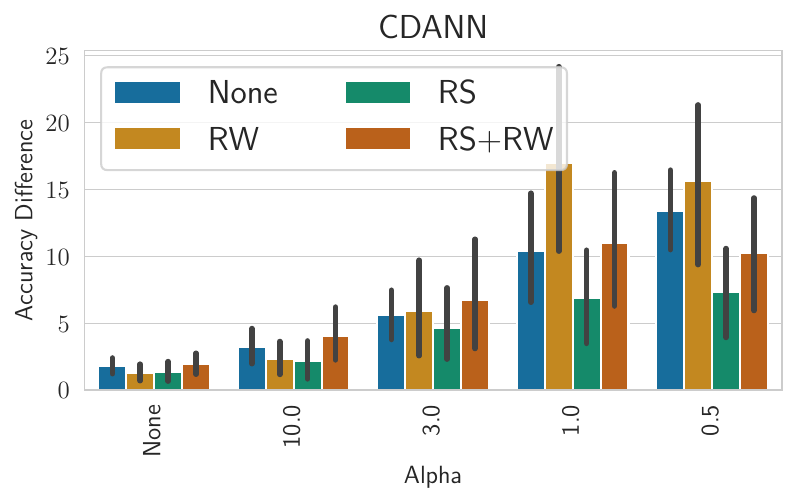} \\

    \includegraphics[width=0.24\linewidth]{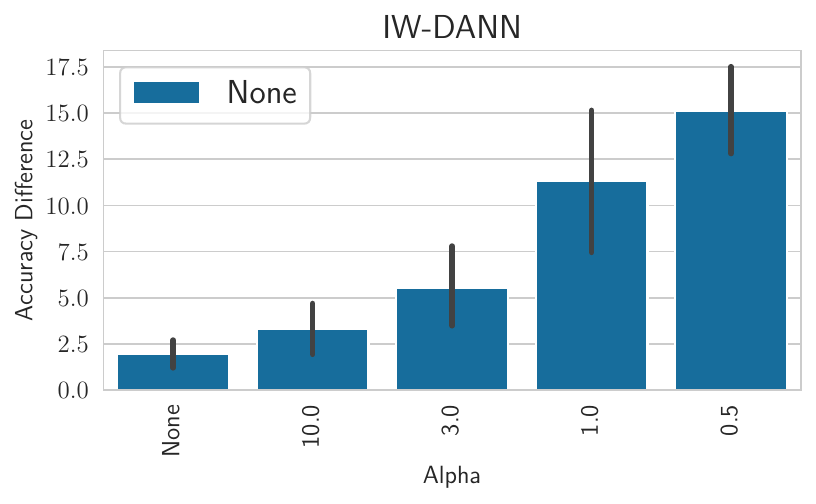}
    \includegraphics[width=0.24\linewidth]{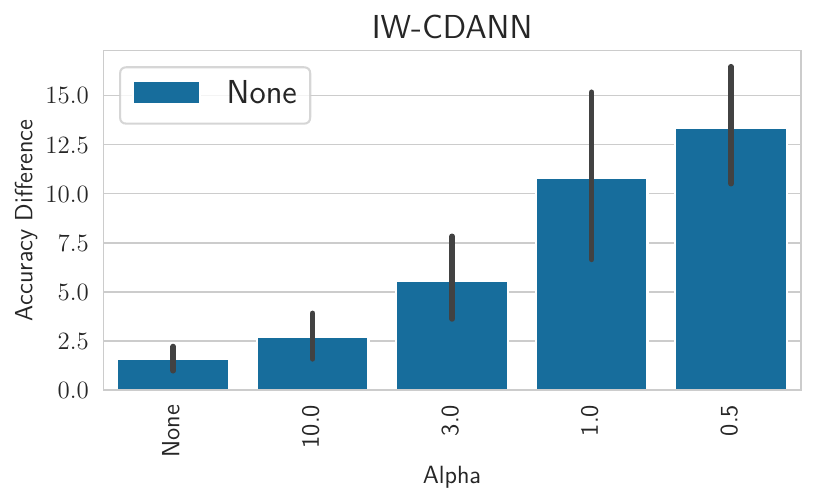}
    \includegraphics[width=0.24\linewidth]{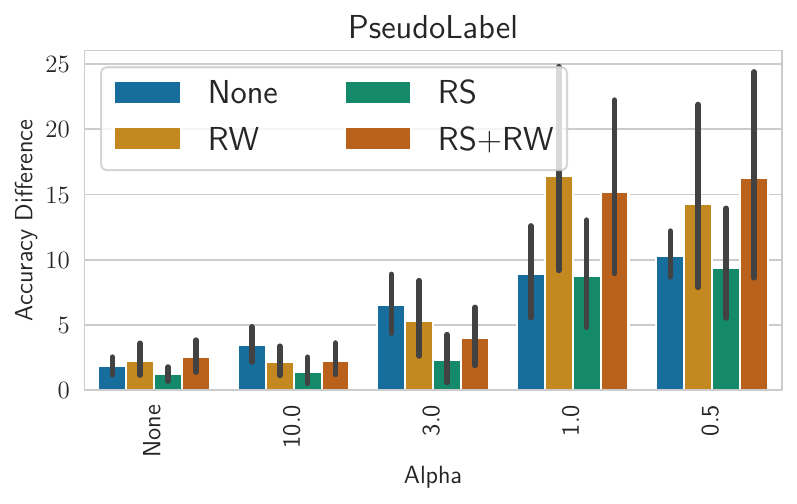}
    \caption{{\emph{Accuracy difference between using source and target performance as early stopping criteria for different DA methods aggregated across all distribution shift pairs in tabular datasets.} We observe that as the shift severity increases (i.e., as $\alpha$ decreases), the accuracy difference increases for all the methods.}}
    \label{fig:oracle_selection_tabular}
\end{figure}

\section{{Results on Individual Datasets}} 
\label{app:individual_dataset}
\begin{figure}[H]
 \centering
    \includegraphics[width=0.5\linewidth]{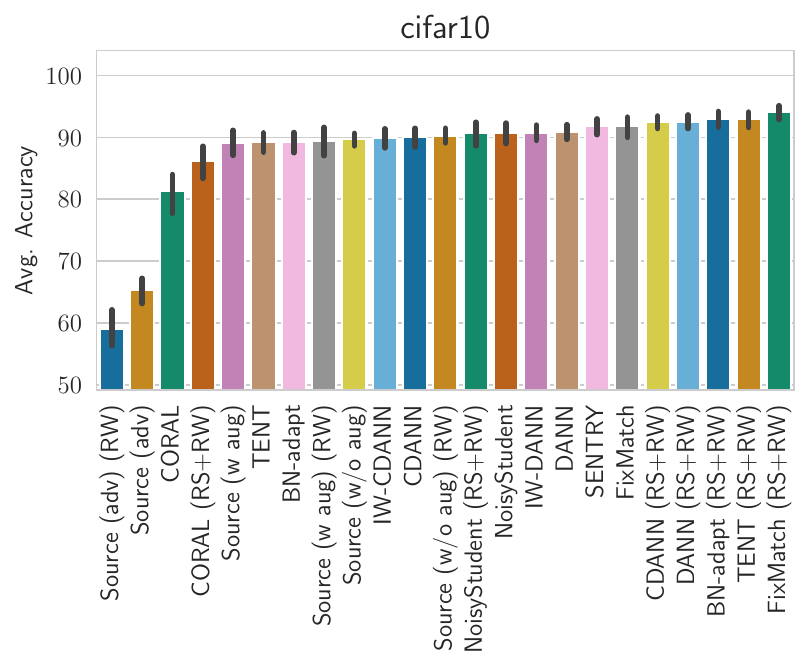} \\
    \includegraphics[width=0.5\linewidth]{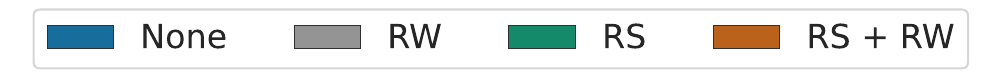} \\
    \includegraphics[width=0.32\linewidth]{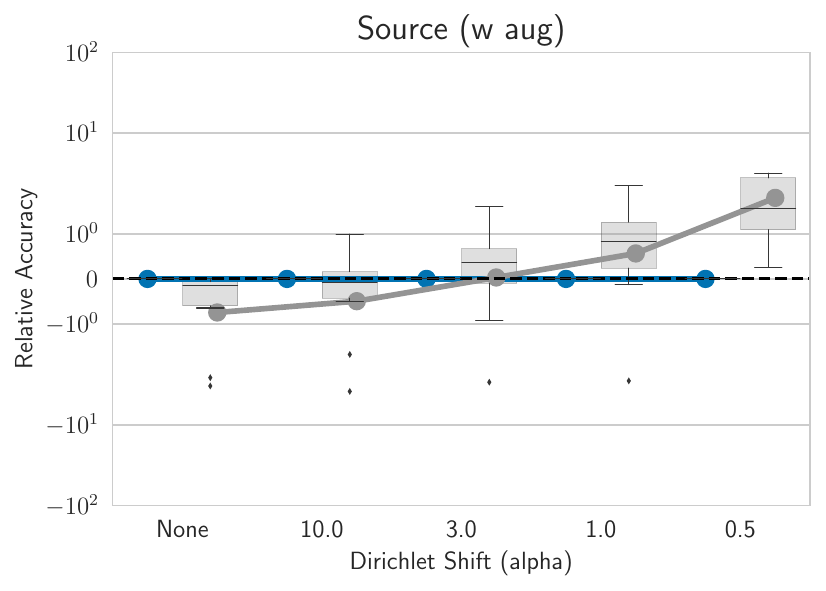} \hfil 
    \includegraphics[width=0.32\linewidth]{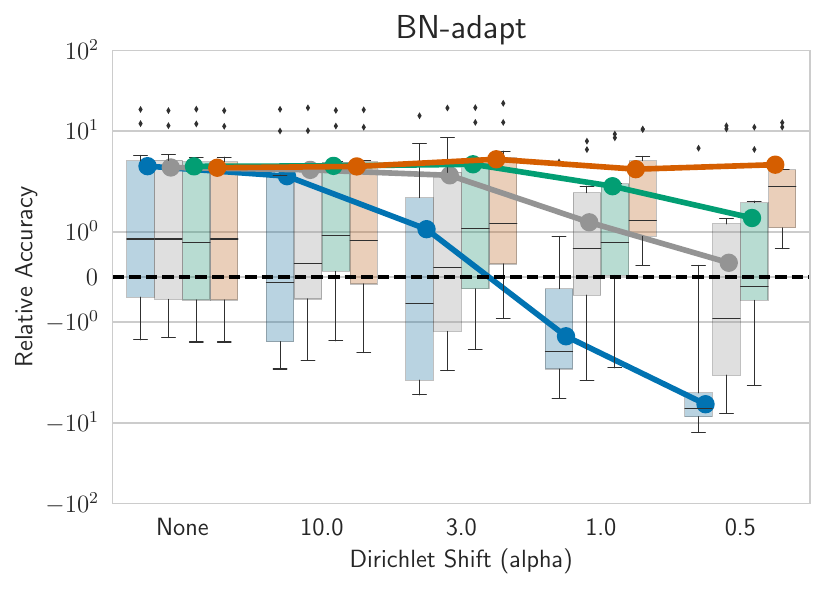}\hfil 
    \includegraphics[width=0.32\linewidth]{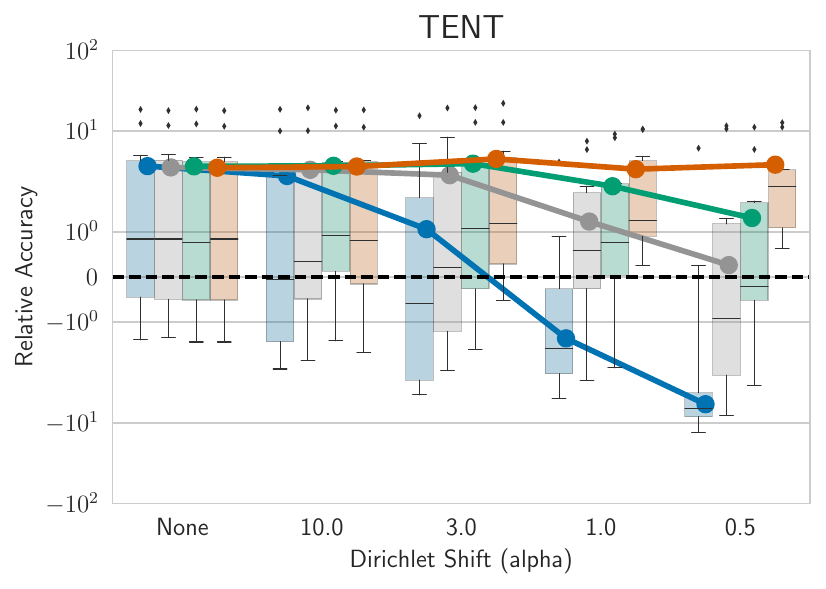}
    \includegraphics[width=0.32\linewidth]{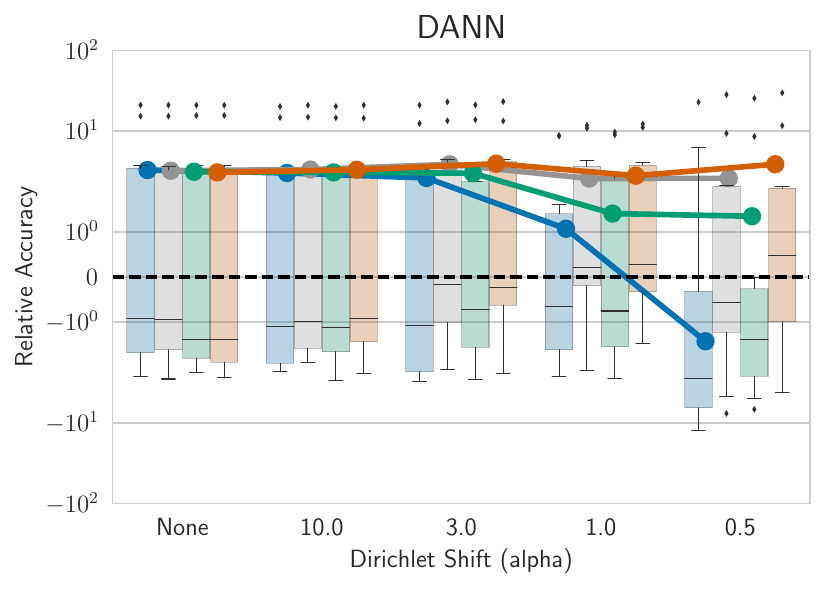} \hfil 
    \includegraphics[width=0.32\linewidth]{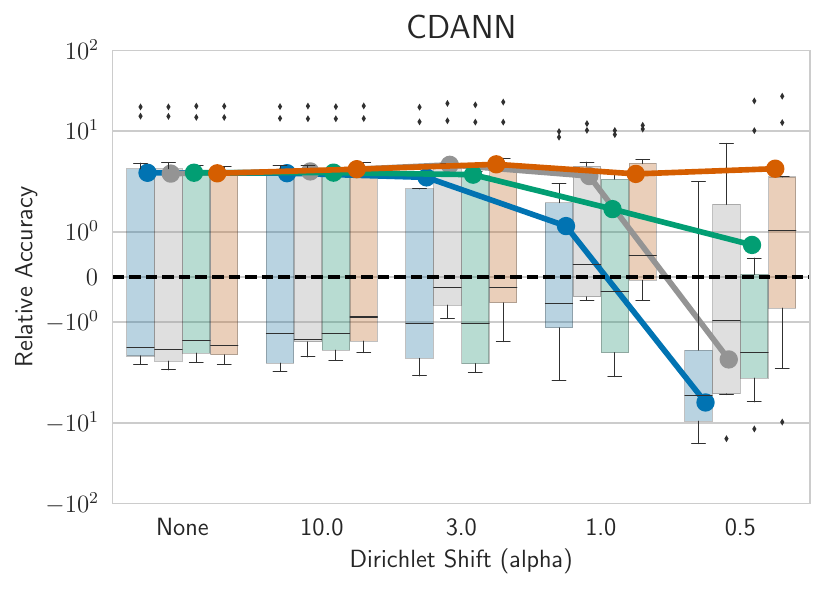}\hfil 
    \includegraphics[width=0.32\linewidth]{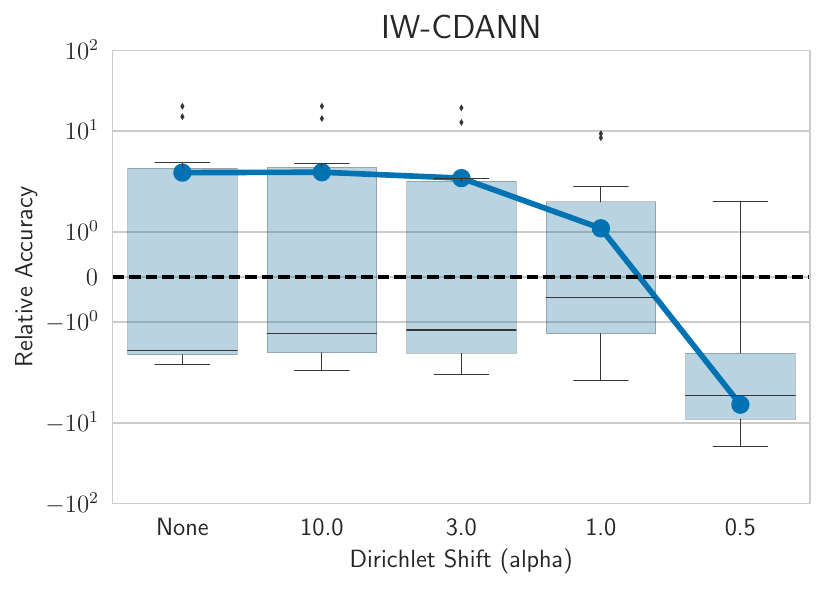}
    \includegraphics[width=0.32\linewidth]{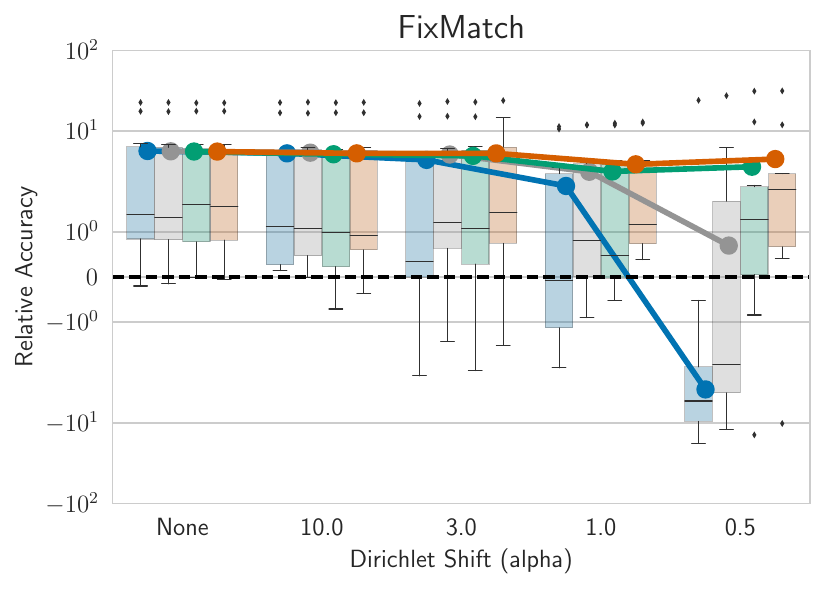} \hfil 
    \includegraphics[width=0.32\linewidth]{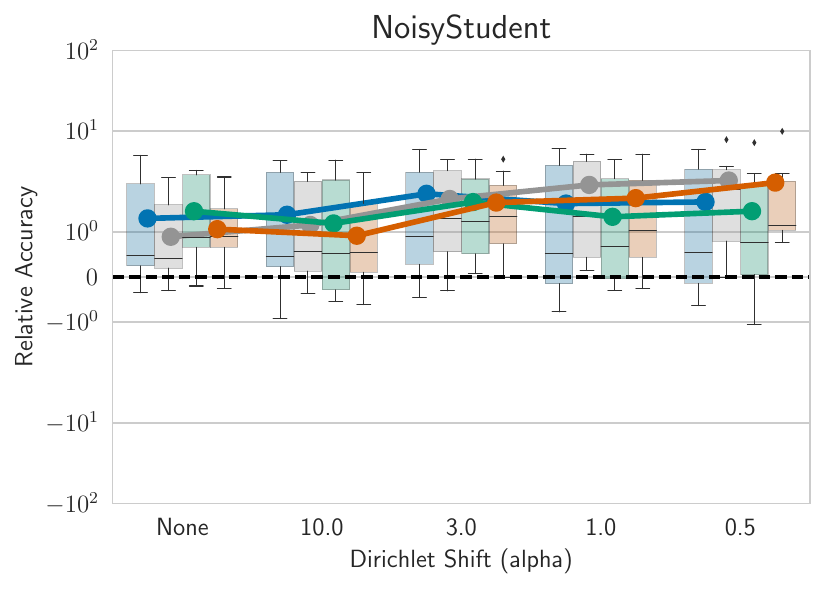}\hfil 
    \includegraphics[width=0.32\linewidth]{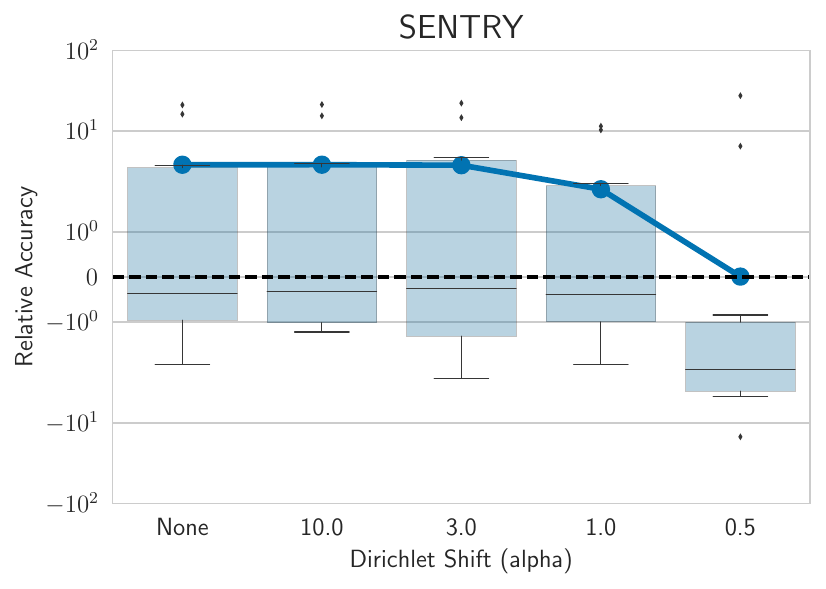}

    \caption{{CIFAR10. Relative performance and accuracy plots for different DA algorithms across various shift pairs in CIFAR10.}}
\end{figure}

\begin{figure}[H]
 \centering
    \includegraphics[width=0.5\linewidth]{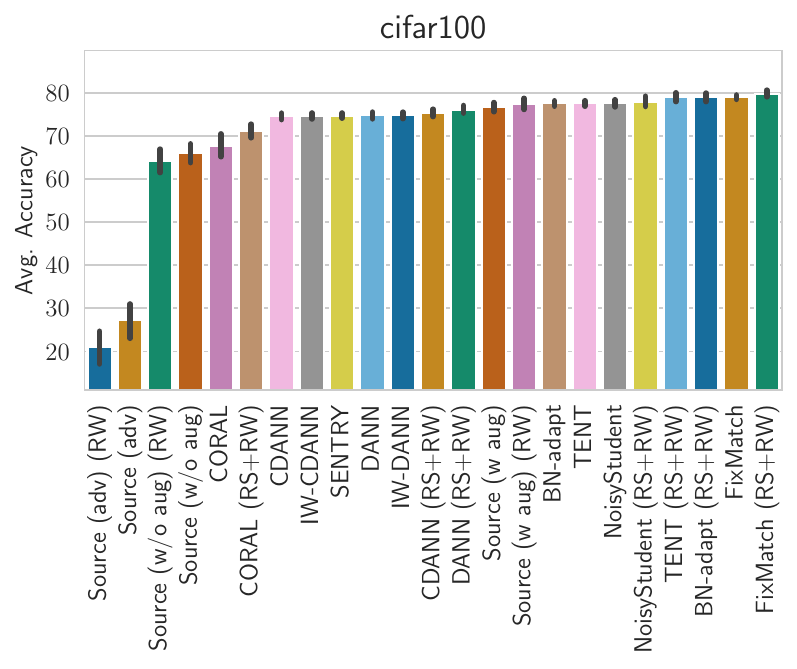} \\
    \includegraphics[width=0.5\linewidth]{figures/legend.pdf} \\
    \includegraphics[width=0.32\linewidth]{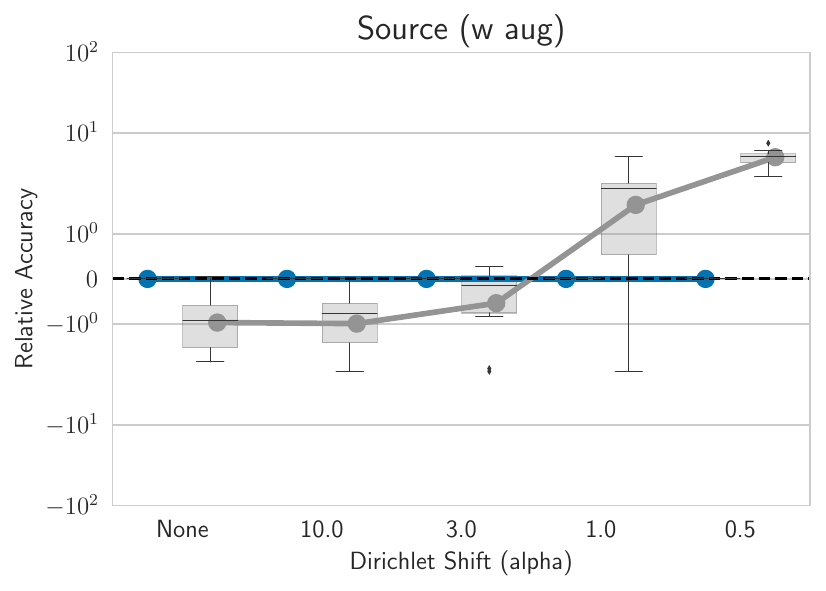} \hfil 
    \includegraphics[width=0.32\linewidth]{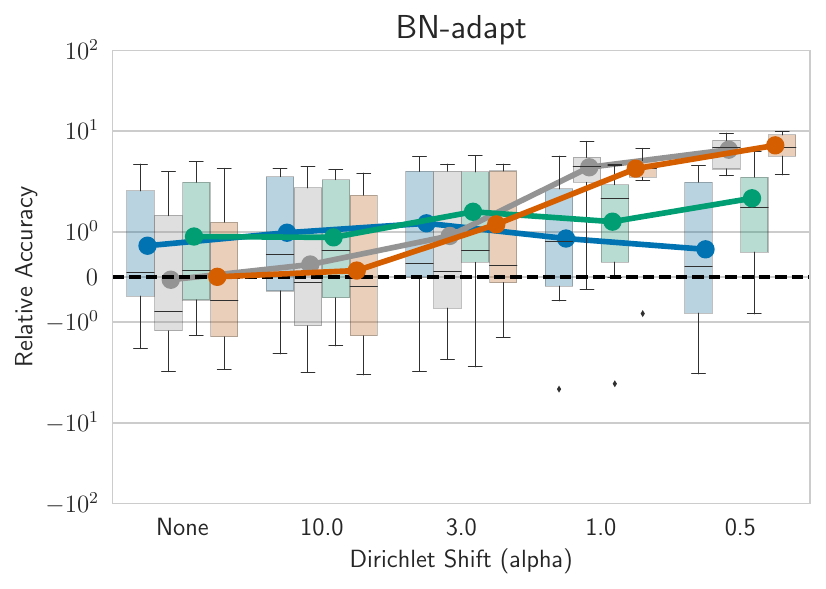}\hfil 
    \includegraphics[width=0.32\linewidth]{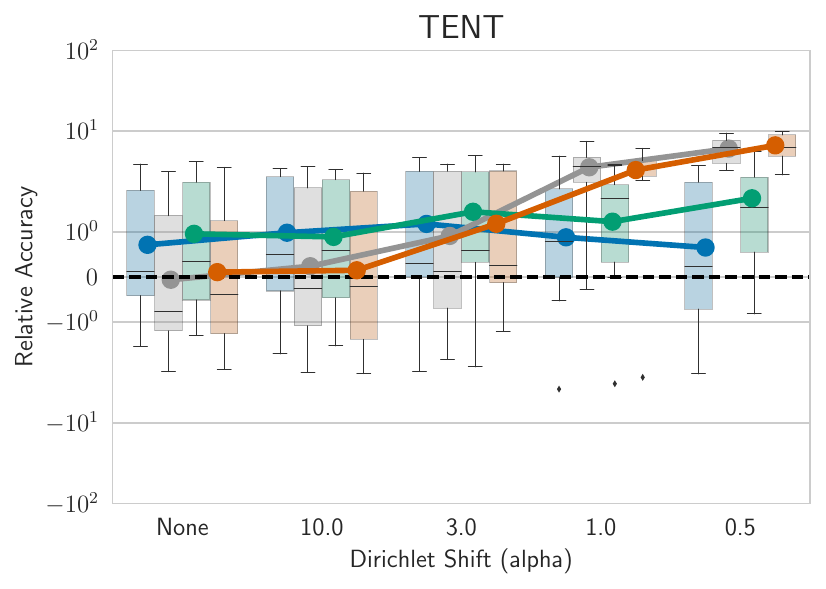}
    \includegraphics[width=0.32\linewidth]{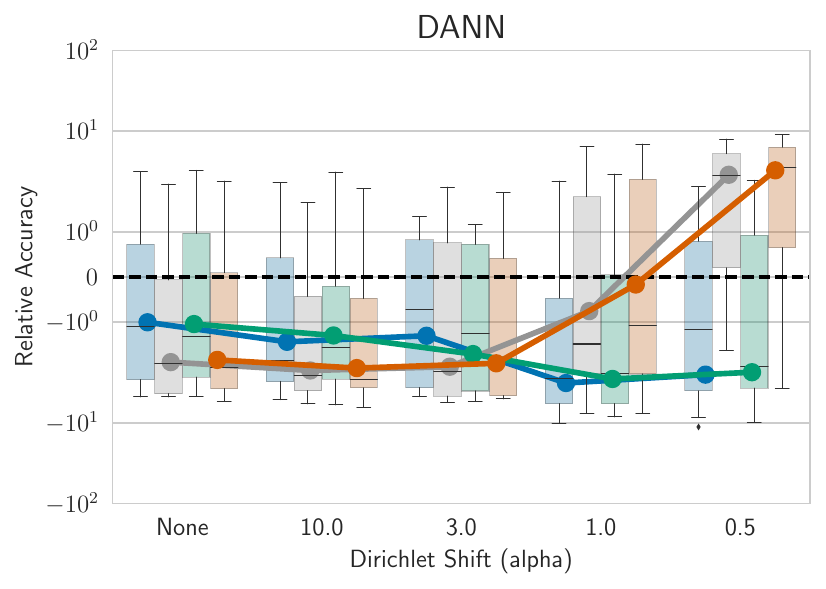} \hfil 
    \includegraphics[width=0.32\linewidth]{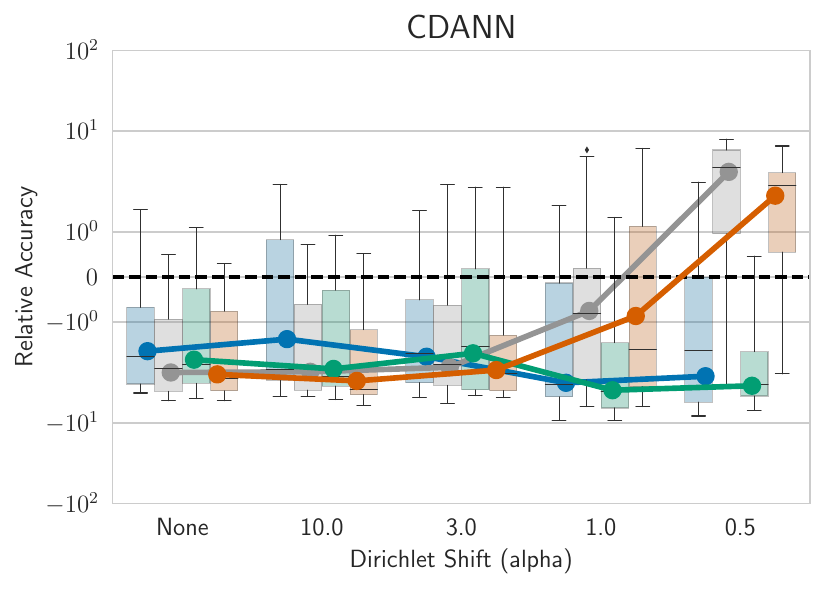}\hfil 
    \includegraphics[width=0.32\linewidth]{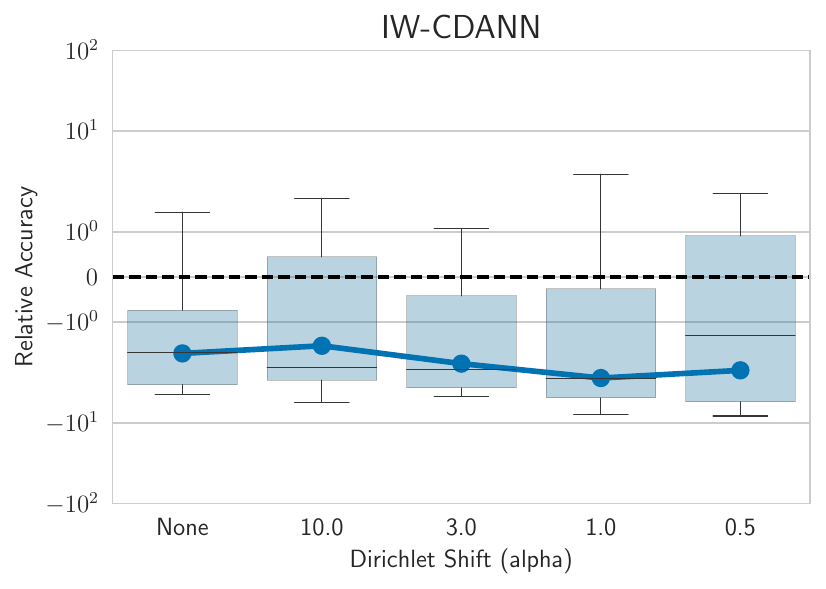}
    \includegraphics[width=0.32\linewidth]{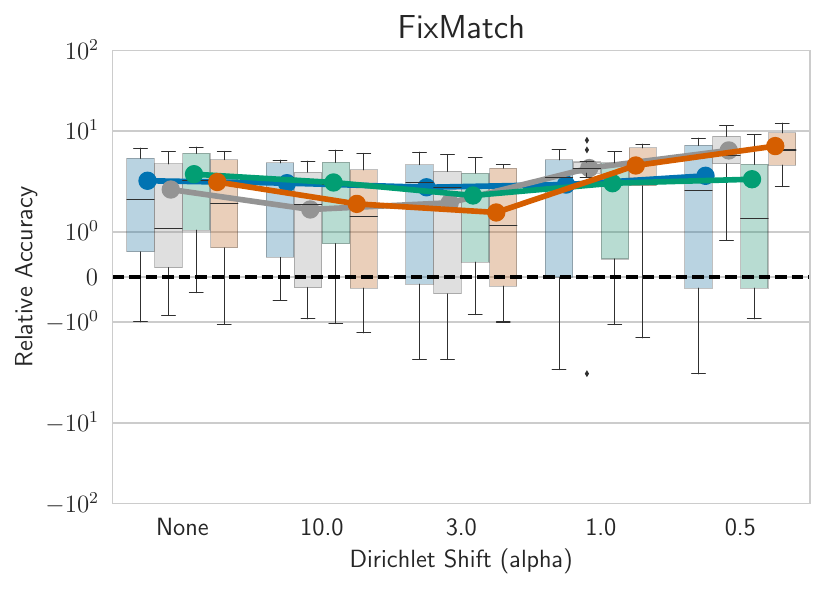} \hfil 
    \includegraphics[width=0.32\linewidth]{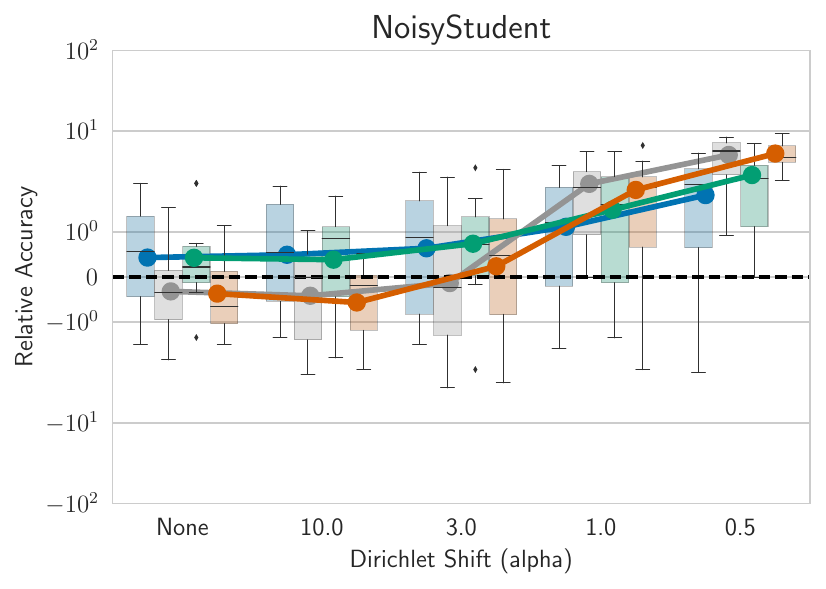}\hfil 
    \includegraphics[width=0.32\linewidth]{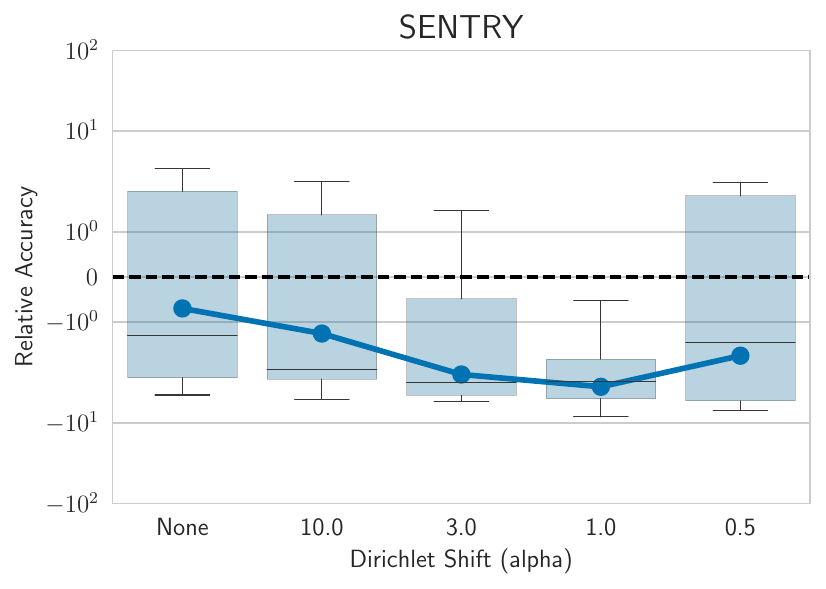}

    \caption{{CIFAR100. Relative performance and accuracy plots for different DA algorithms across various shift pairs in CIFAR100.}}
\end{figure}

\begin{figure}[H]
 \centering
    \includegraphics[width=0.5\linewidth]{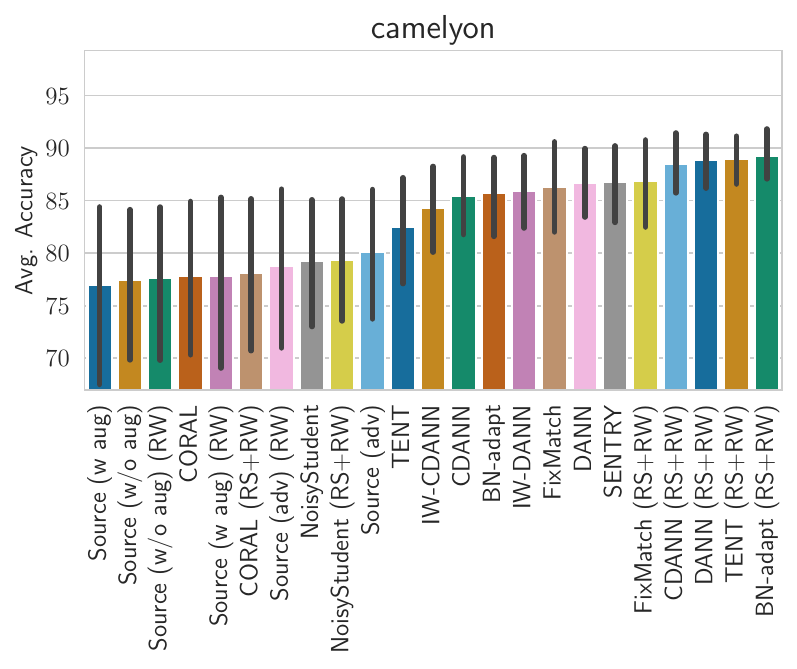} \\
    \includegraphics[width=0.5\linewidth]{figures/legend.pdf} \\
    \includegraphics[width=0.32\linewidth]{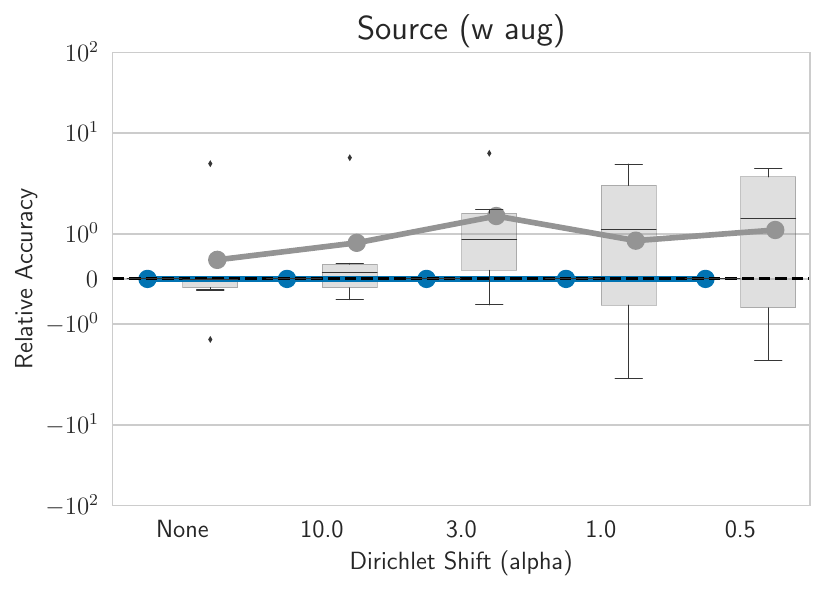} \hfil 
    \includegraphics[width=0.32\linewidth]{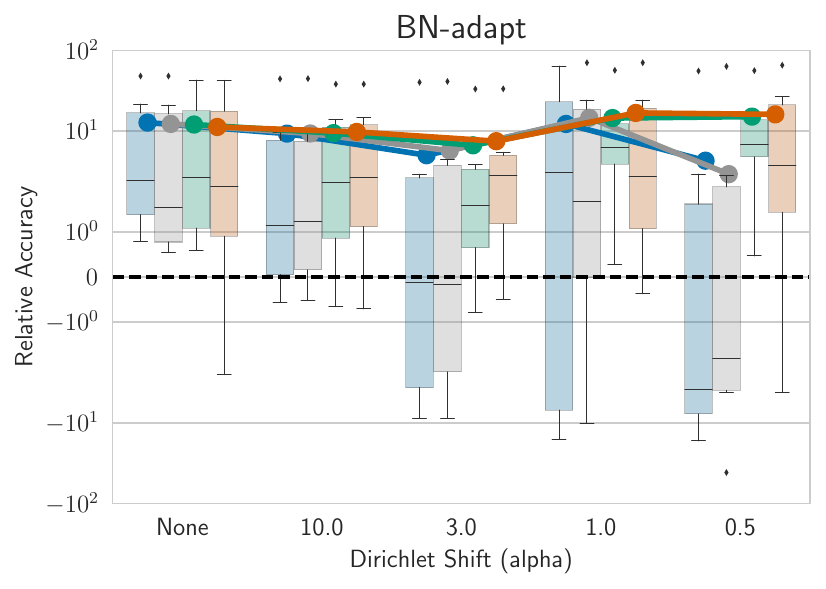}\hfil 
    \includegraphics[width=0.32\linewidth]{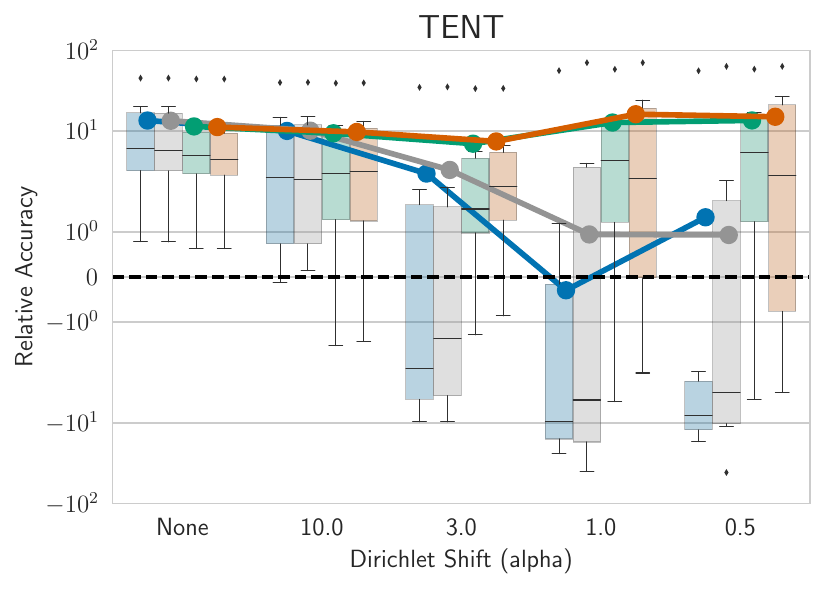}
    \includegraphics[width=0.32\linewidth]{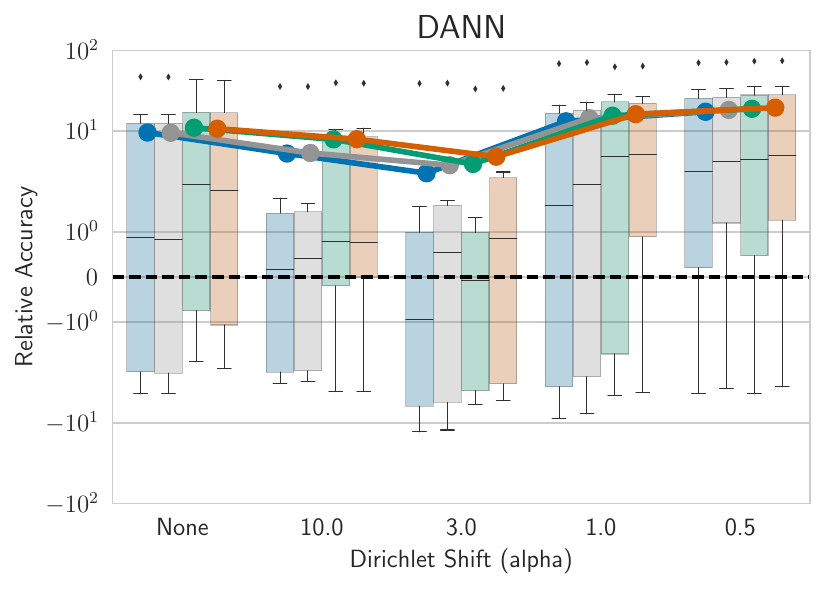} \hfil 
    \includegraphics[width=0.32\linewidth]{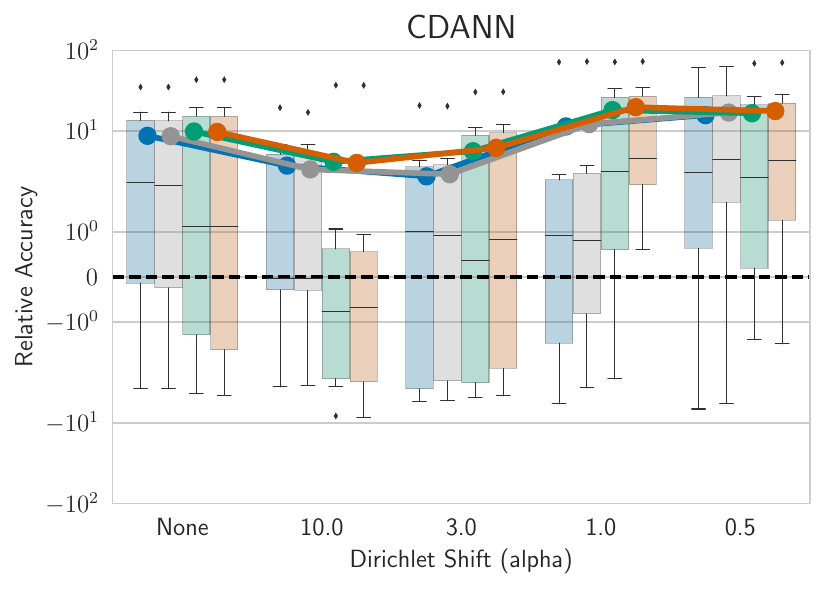}\hfil 
    \includegraphics[width=0.32\linewidth]{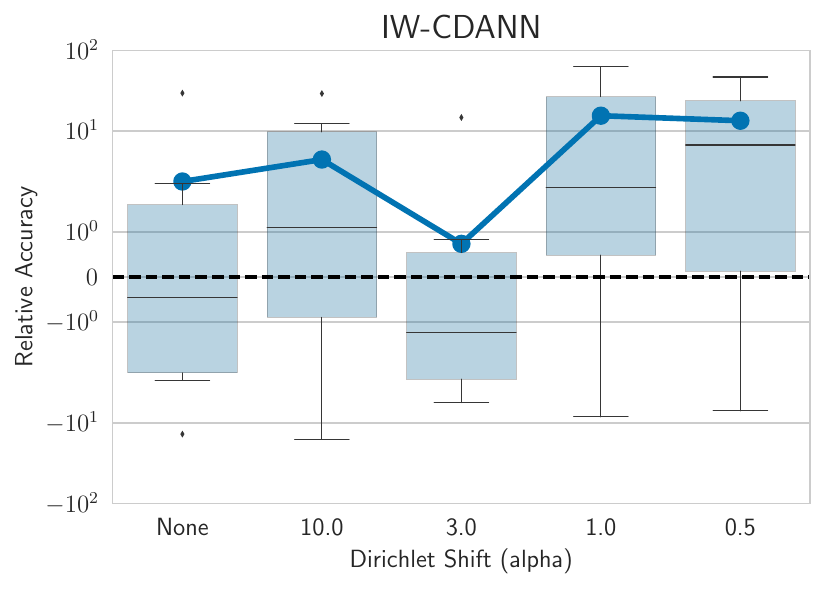}
    \includegraphics[width=0.32\linewidth]{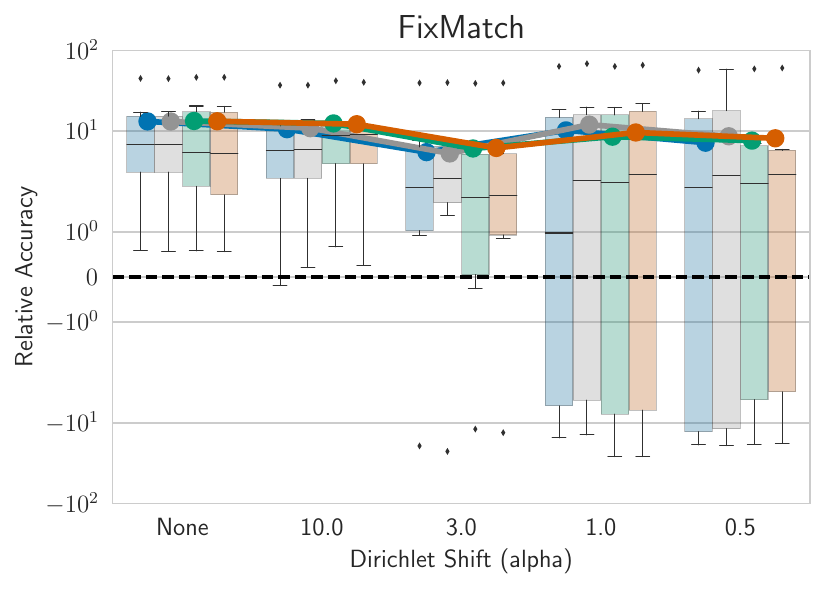} \hfil 
    \includegraphics[width=0.32\linewidth]{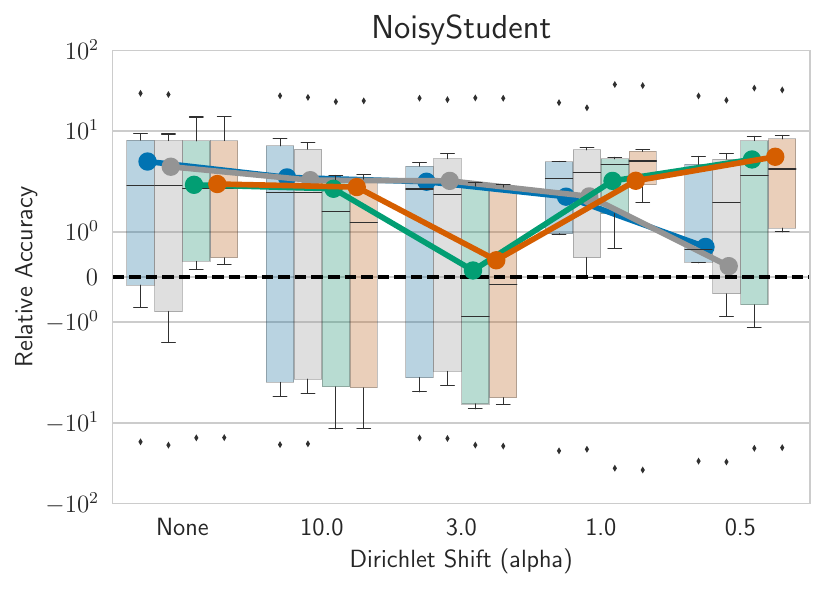}\hfil 
    \includegraphics[width=0.32\linewidth]{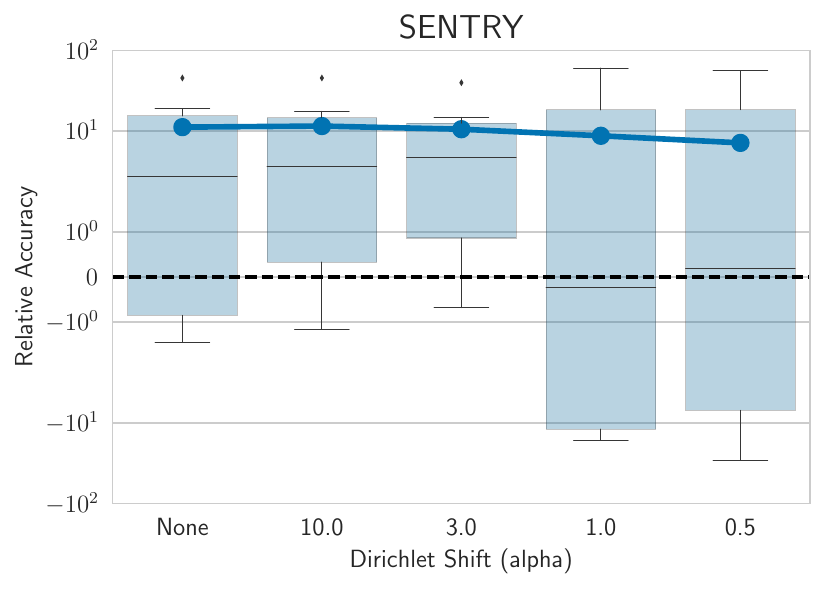}

    \caption{{Camelyon. Relative performance and accuracy plots for different DA algorithms across various shift pairs in Camelyon.}}
\end{figure}

\begin{figure}[H]
 \centering
    \includegraphics[width=0.5\linewidth]{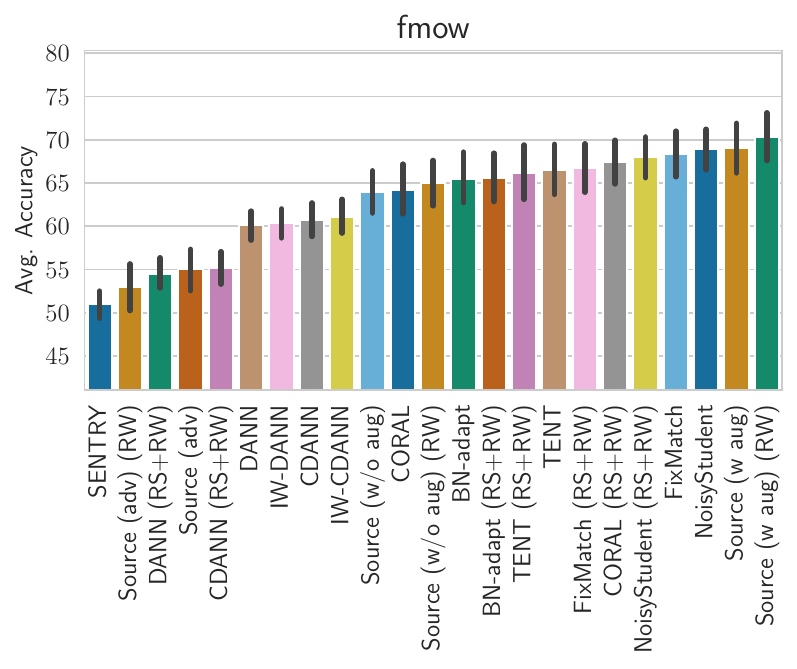} \\
    \includegraphics[width=0.5\linewidth]{figures/legend.pdf} \\
    \includegraphics[width=0.32\linewidth]{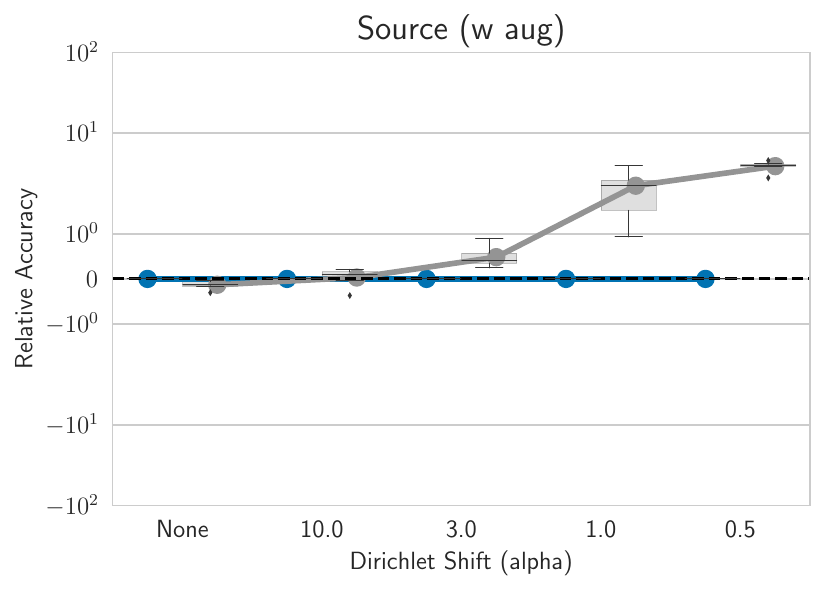} \hfil 
    \includegraphics[width=0.32\linewidth]{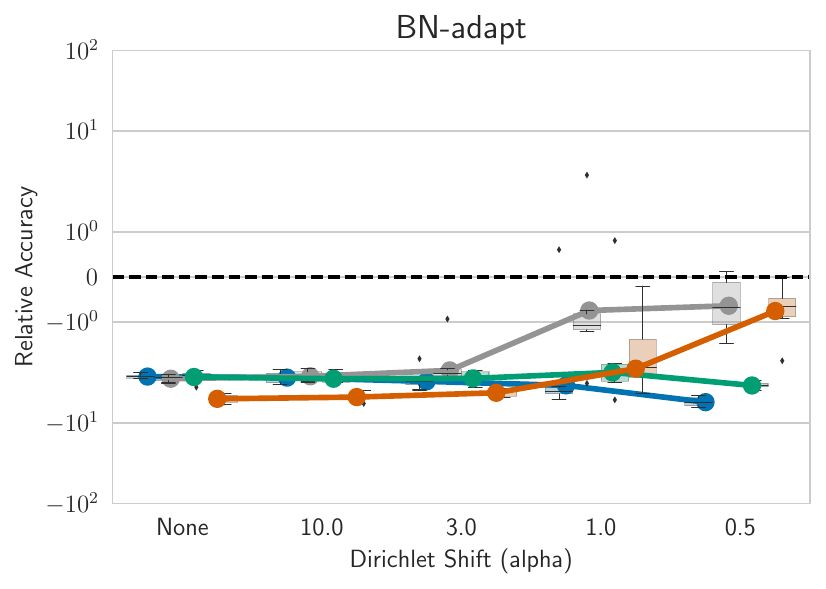}\hfil 
    \includegraphics[width=0.32\linewidth]{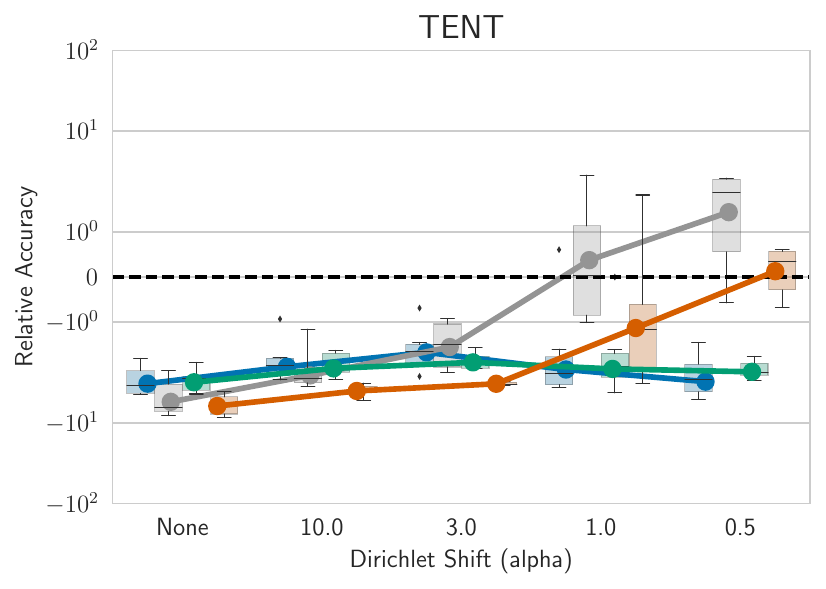}
    \includegraphics[width=0.32\linewidth]{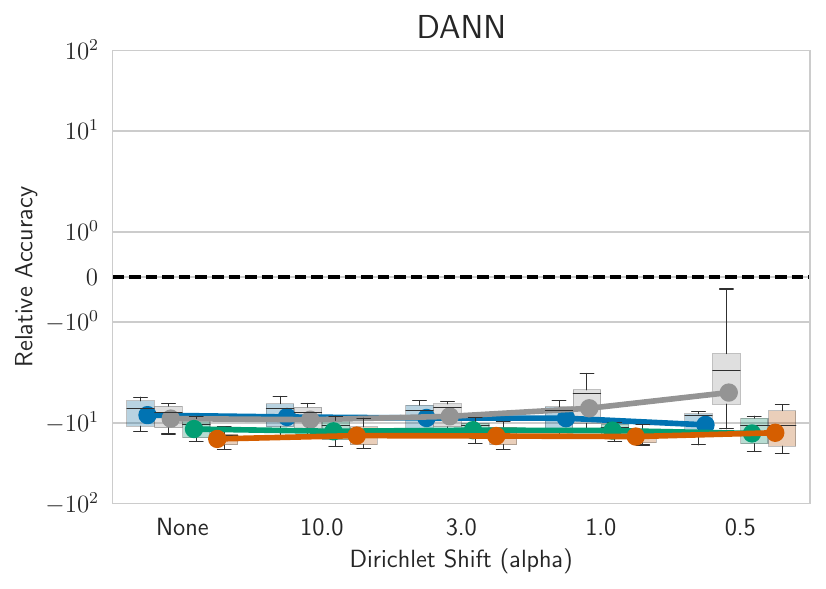} \hfil 
    \includegraphics[width=0.32\linewidth]{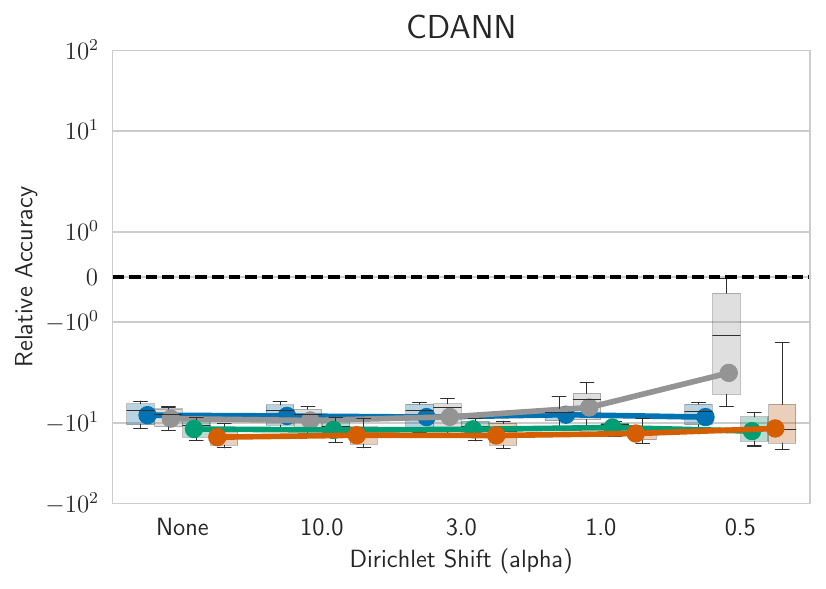}\hfil 
    \includegraphics[width=0.32\linewidth]{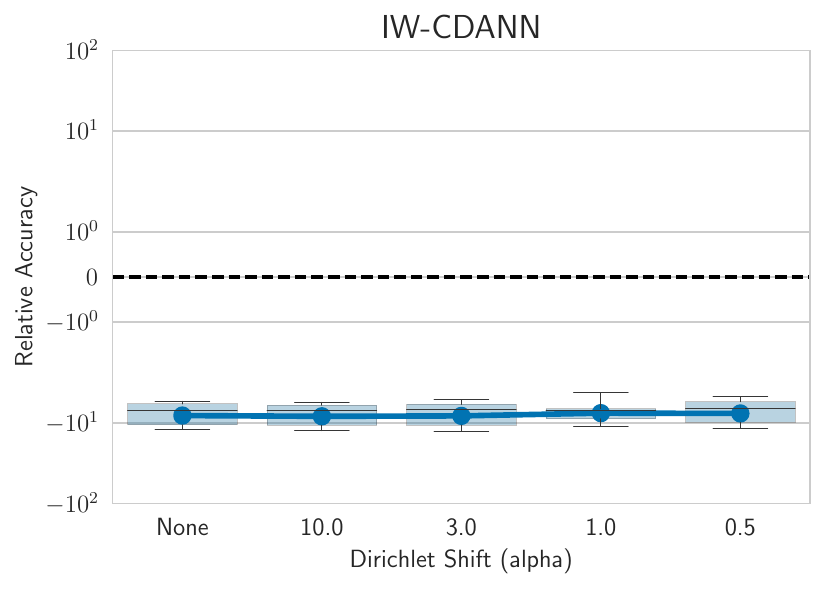}
    \includegraphics[width=0.32\linewidth]{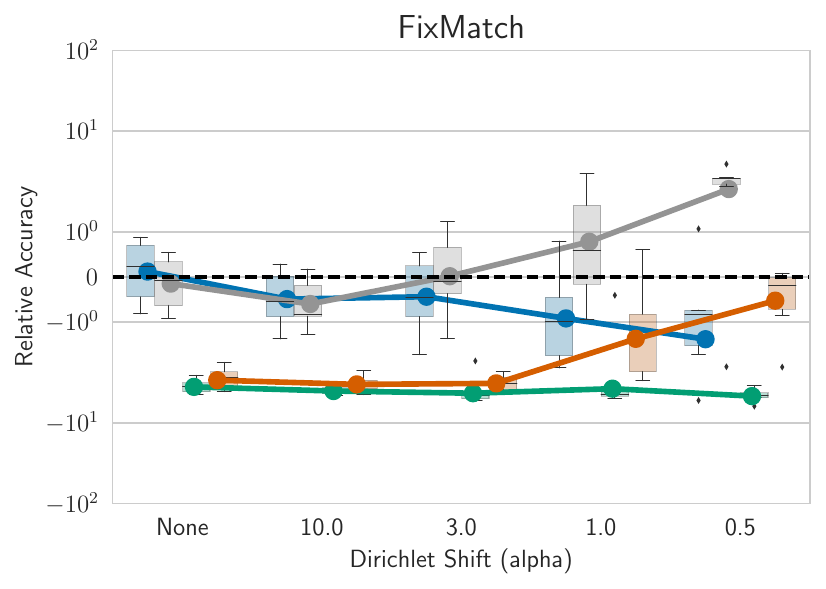} \hfil 
    \includegraphics[width=0.32\linewidth]{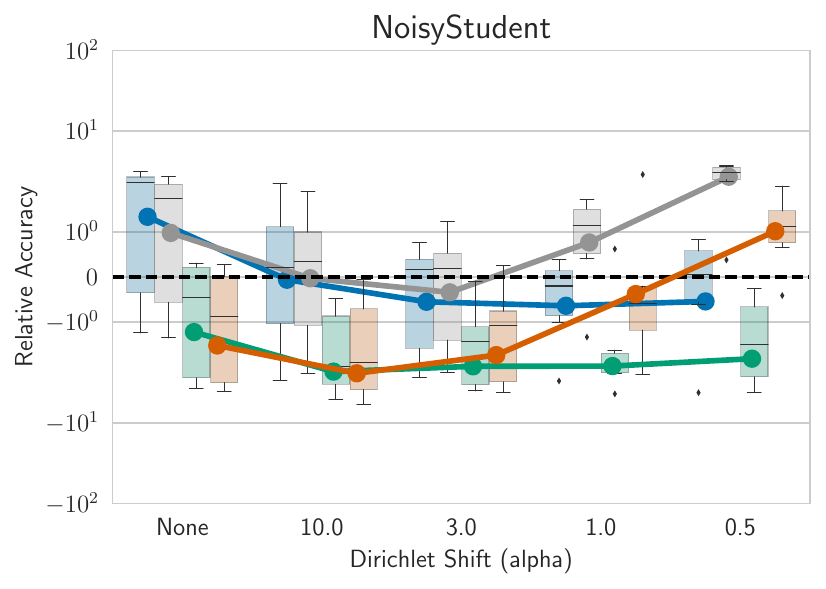}\hfil 
    \includegraphics[width=0.32\linewidth]{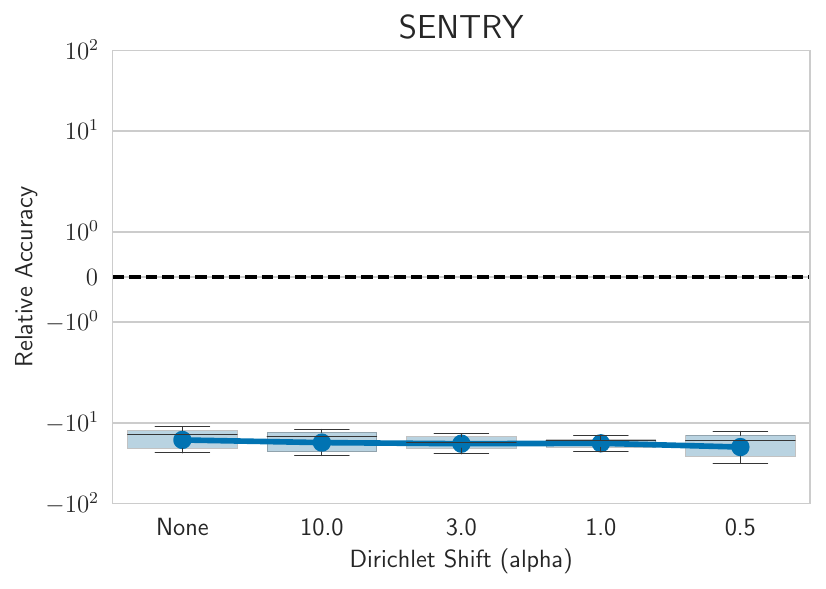}

    \caption{{FMoW. Relative performance and accuracy plots for different DA algorithms across various shift pairs in FMoW.}}
\end{figure}

\begin{figure}[H]
 \centering
    \includegraphics[width=0.5\linewidth]{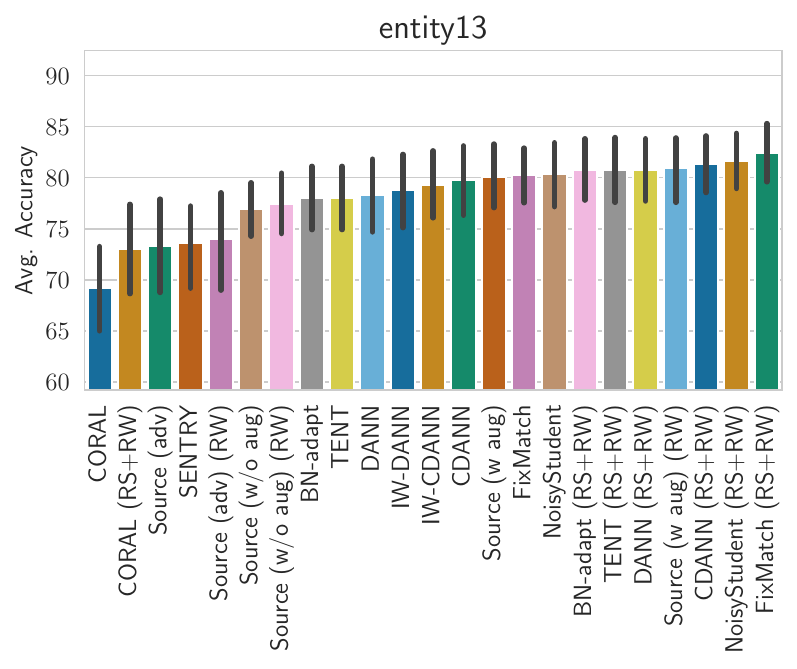} \\
    \includegraphics[width=0.5\linewidth]{figures/legend.pdf} \\
    \includegraphics[width=0.32\linewidth]{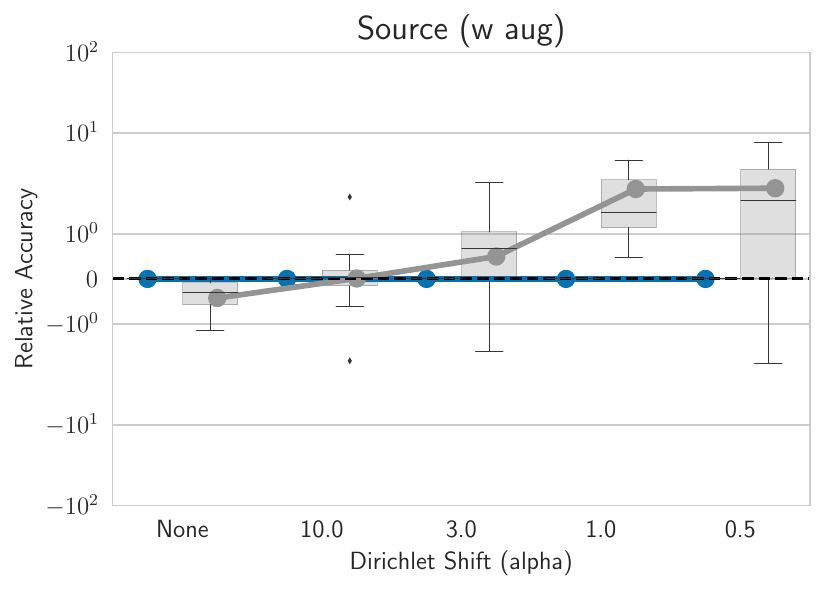} \hfil 
    \includegraphics[width=0.32\linewidth]{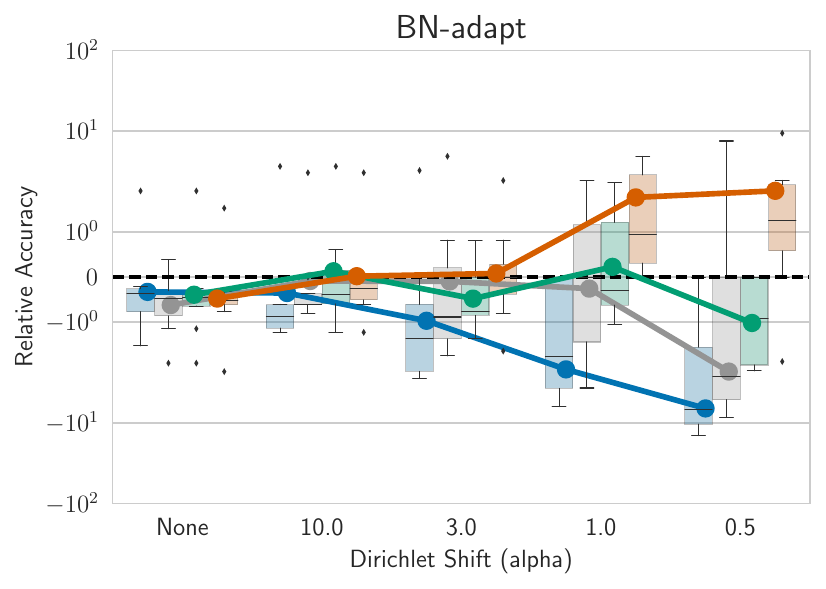}\hfil 
    \includegraphics[width=0.32\linewidth]{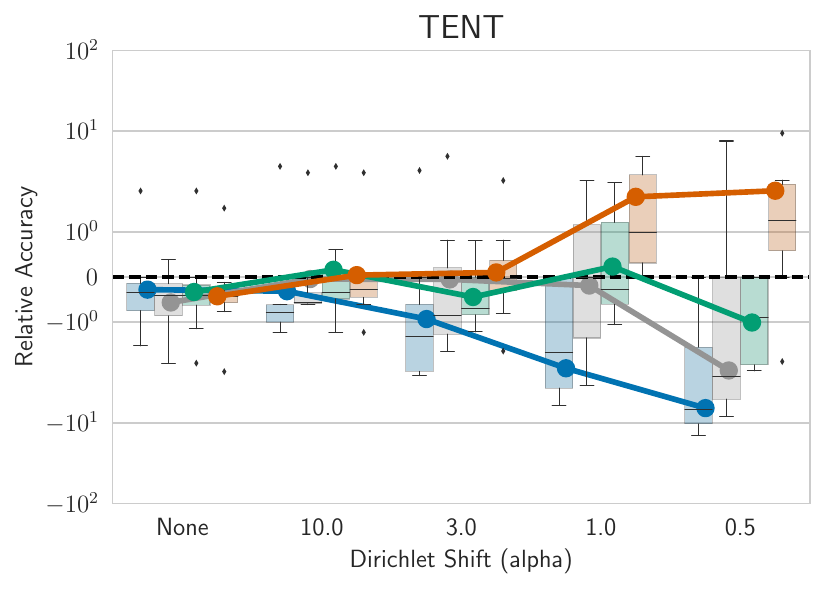}
    \includegraphics[width=0.32\linewidth]{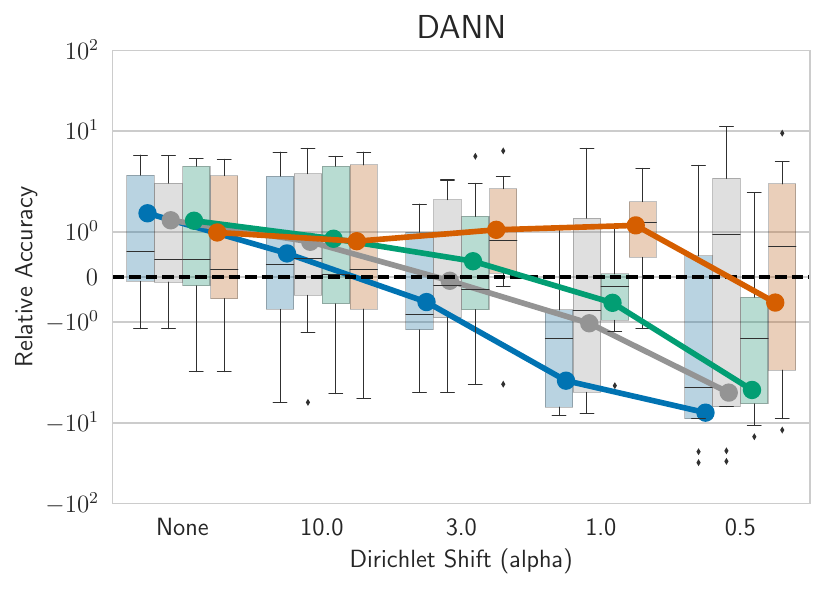} \hfil 
    \includegraphics[width=0.32\linewidth]{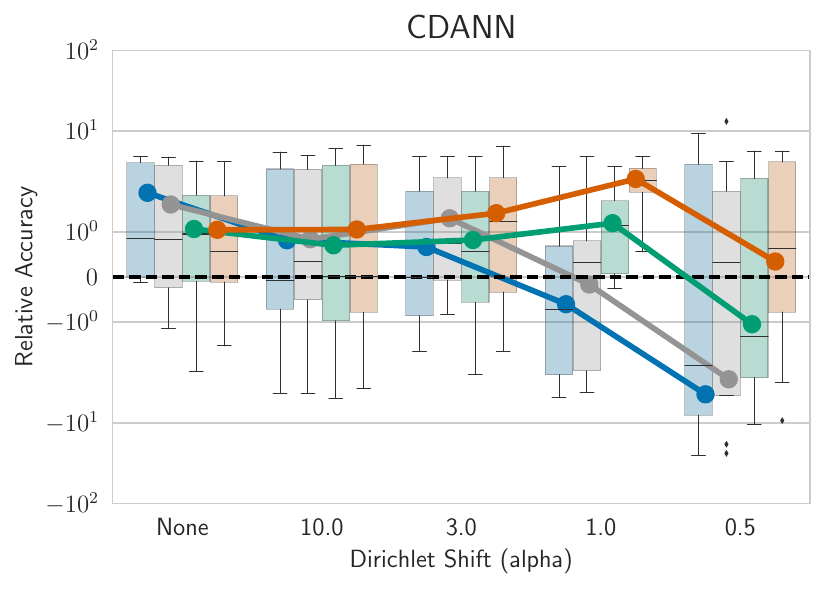}\hfil 
    \includegraphics[width=0.32\linewidth]{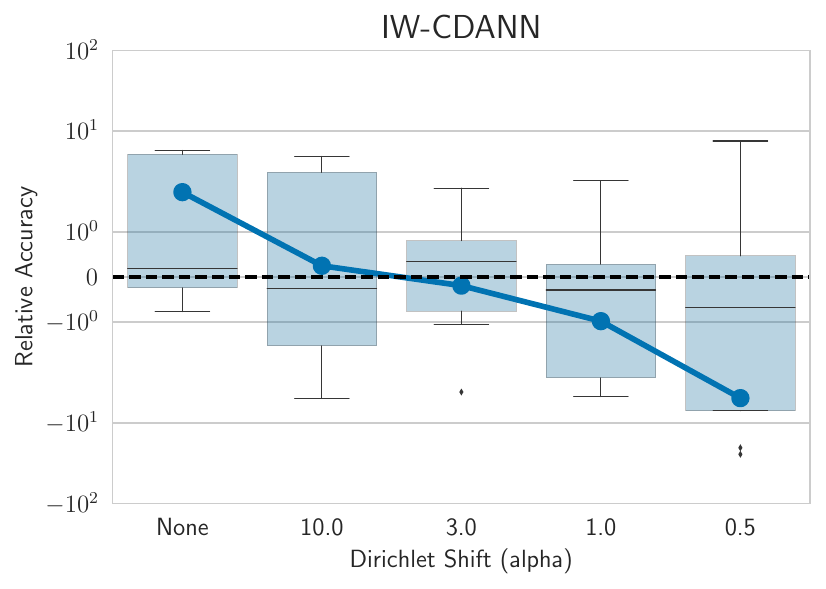}
    \includegraphics[width=0.32\linewidth]{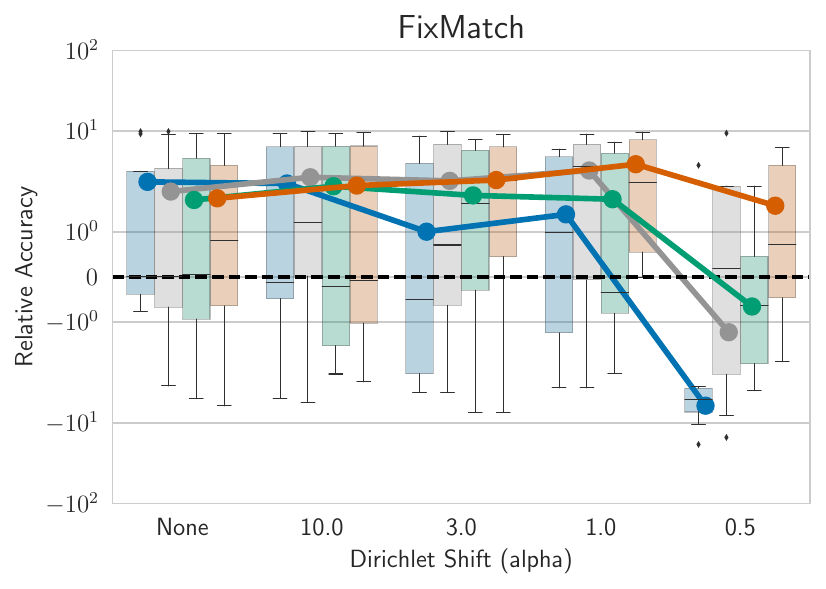} \hfil 
    \includegraphics[width=0.32\linewidth]{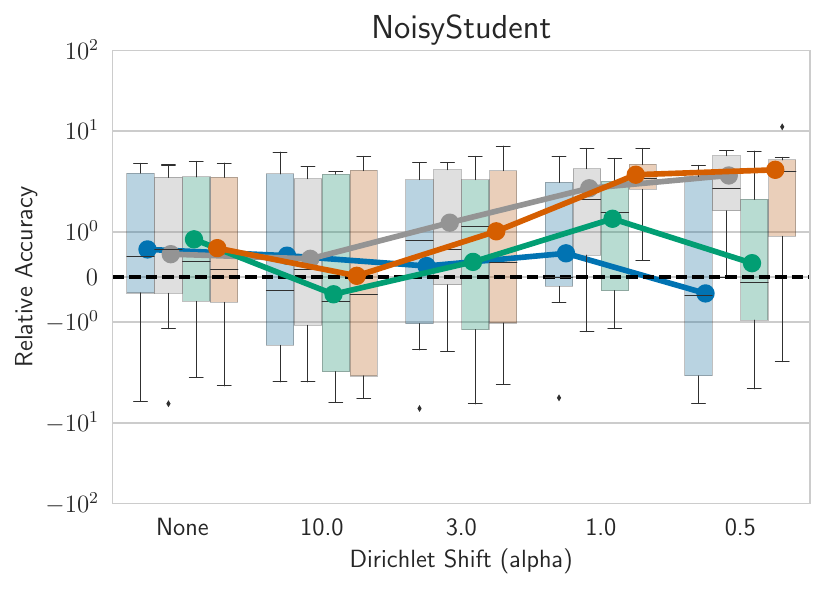}\hfil 
    \includegraphics[width=0.32\linewidth]{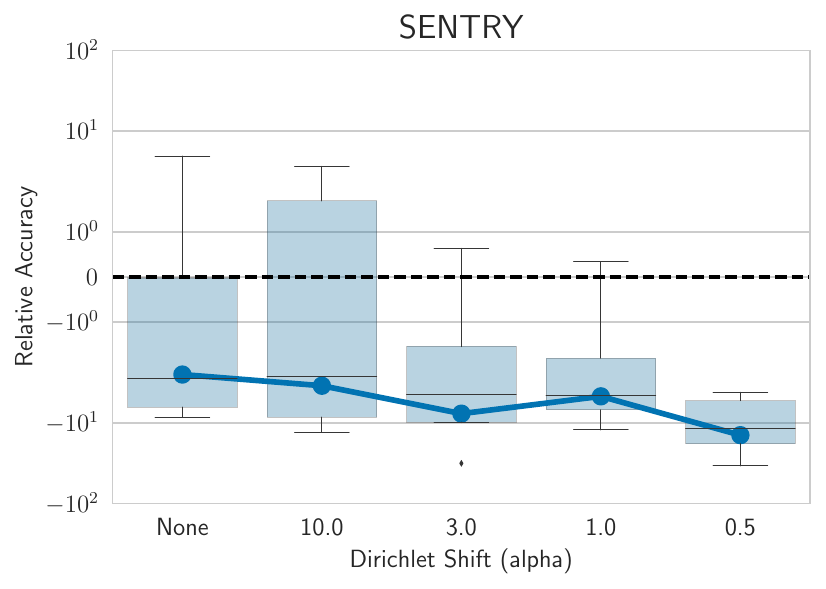}

    \caption{{Entity13. Relative performance and accuracy plots for different DA algorithms across various shift pairs in Entity13.}}
\end{figure}

\begin{figure}[H]
 \centering
    \includegraphics[width=0.5\linewidth]{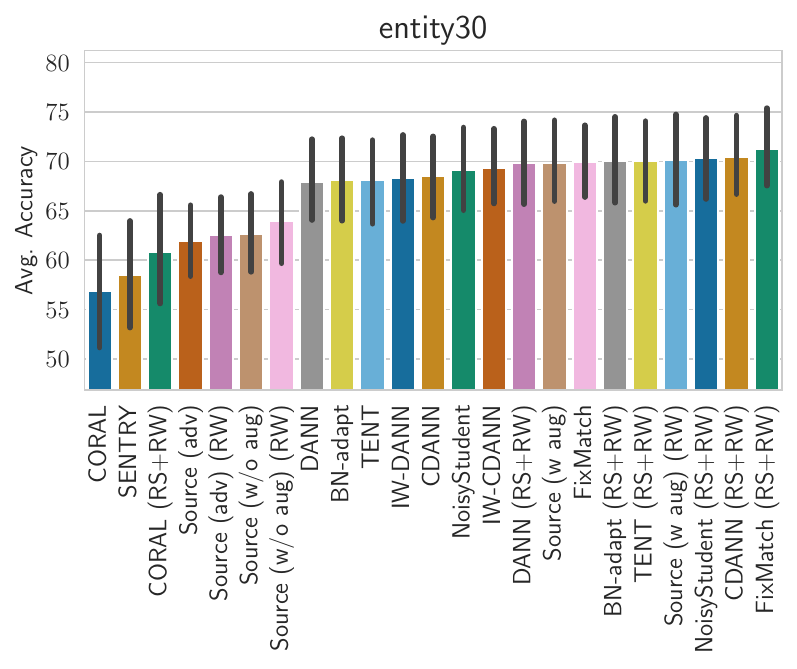} \\
    \includegraphics[width=0.5\linewidth]{figures/legend.pdf} \\
    \includegraphics[width=0.32\linewidth]{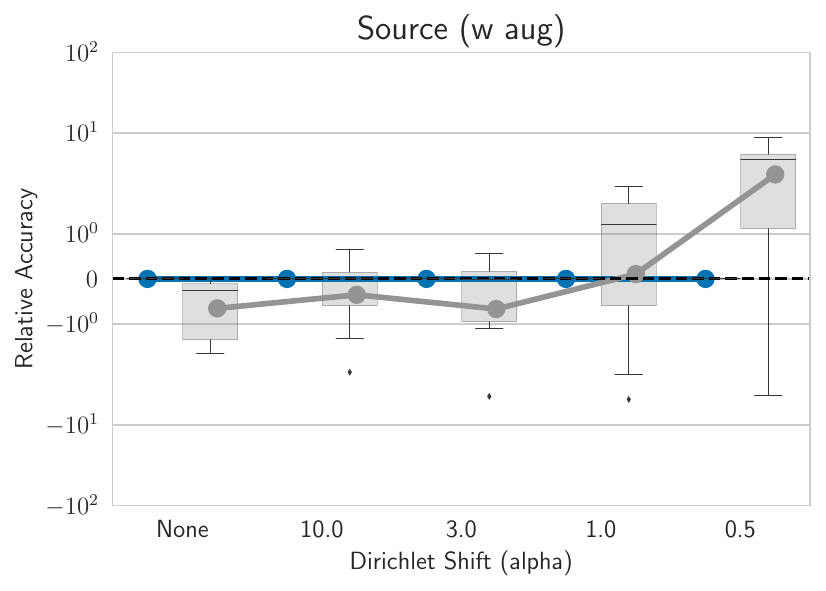} \hfil 
    \includegraphics[width=0.32\linewidth]{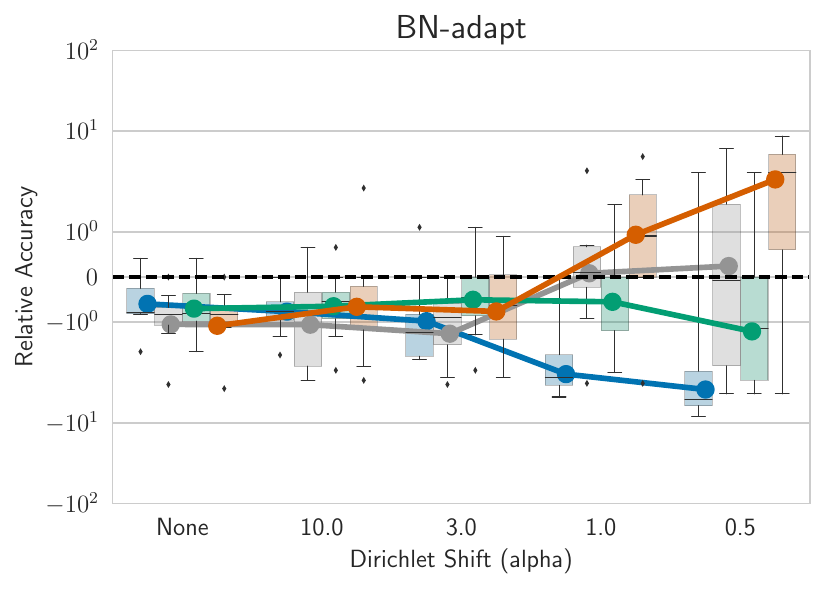}\hfil 
    \includegraphics[width=0.32\linewidth]{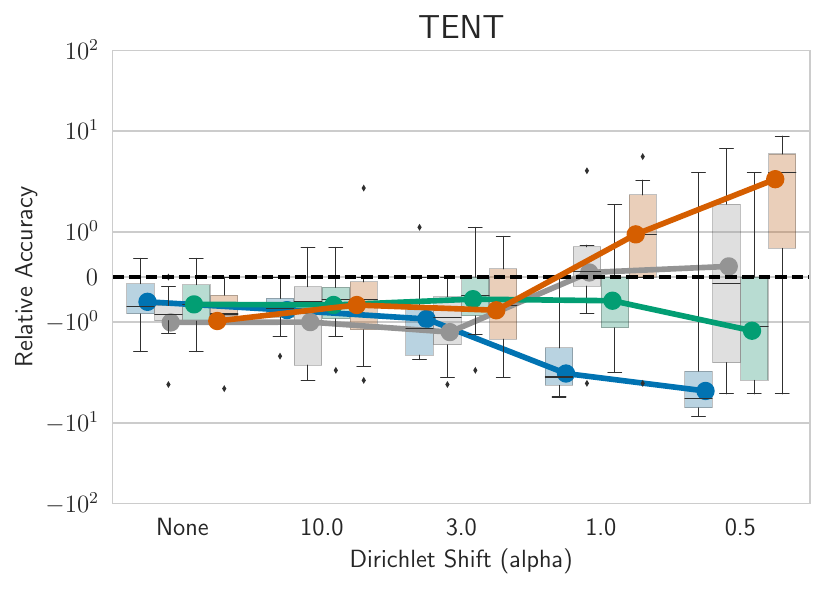}
    \includegraphics[width=0.32\linewidth]{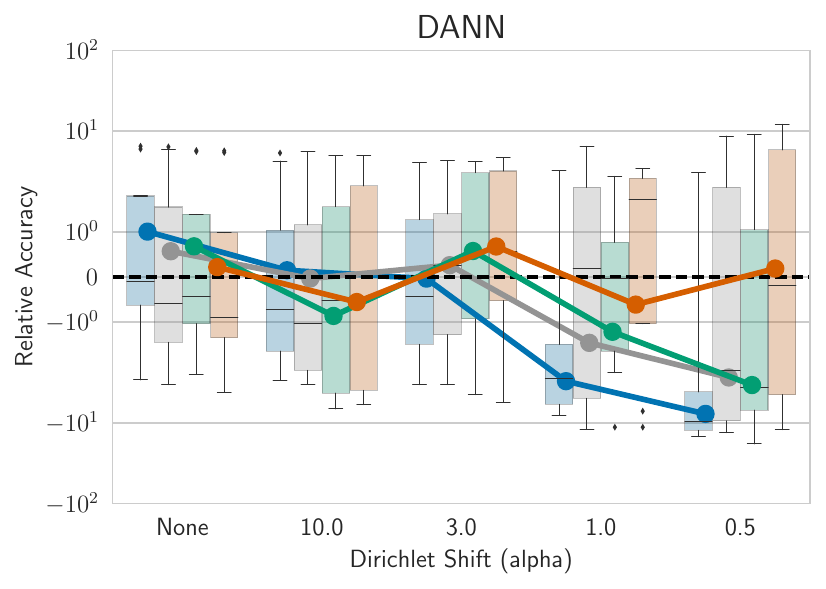} \hfil 
    \includegraphics[width=0.32\linewidth]{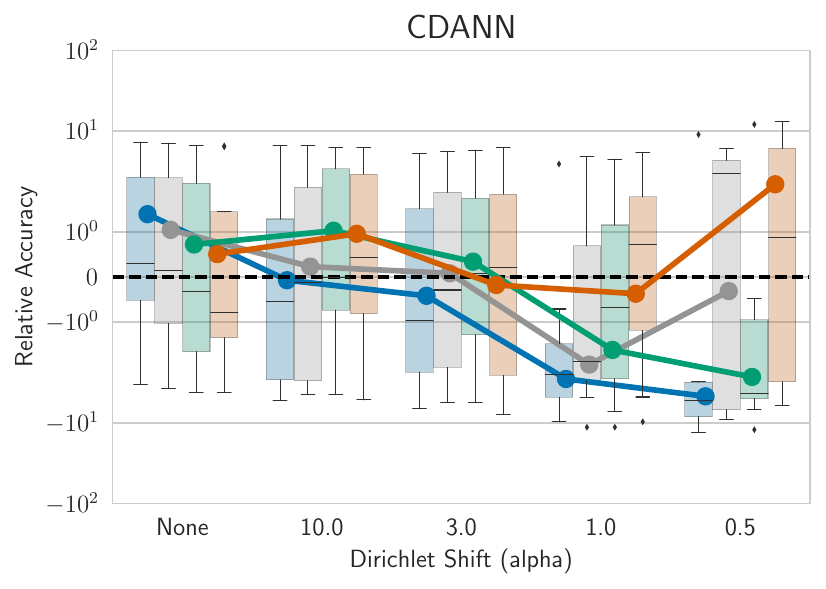}\hfil 
    \includegraphics[width=0.32\linewidth]{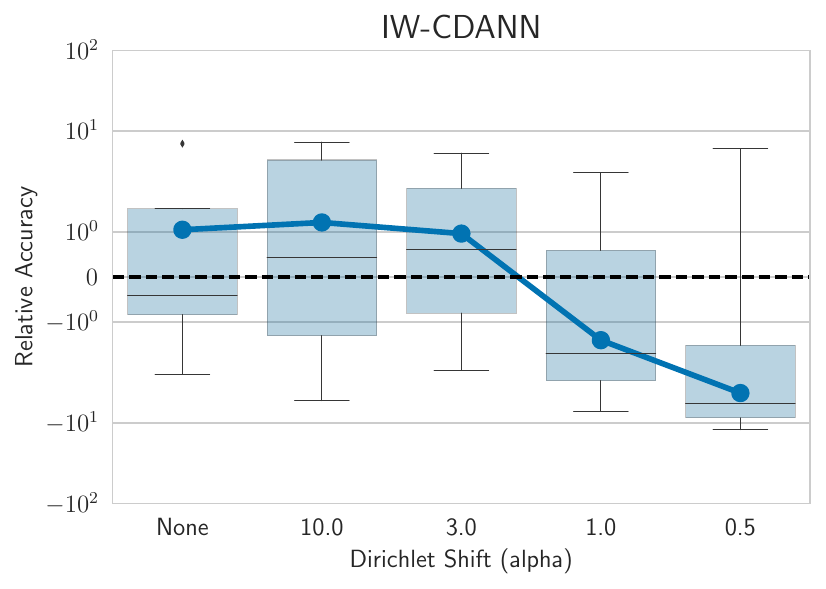}
    \includegraphics[width=0.32\linewidth]{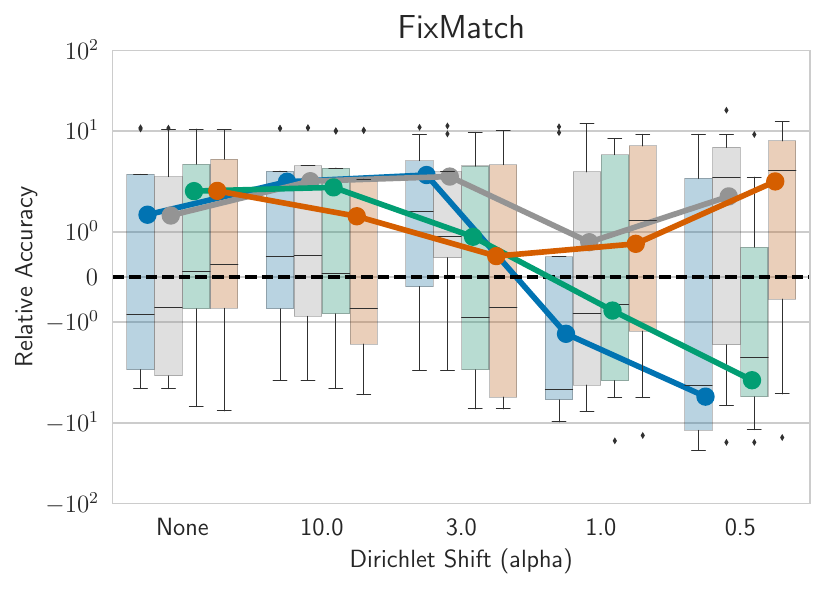} \hfil 
    \includegraphics[width=0.32\linewidth]{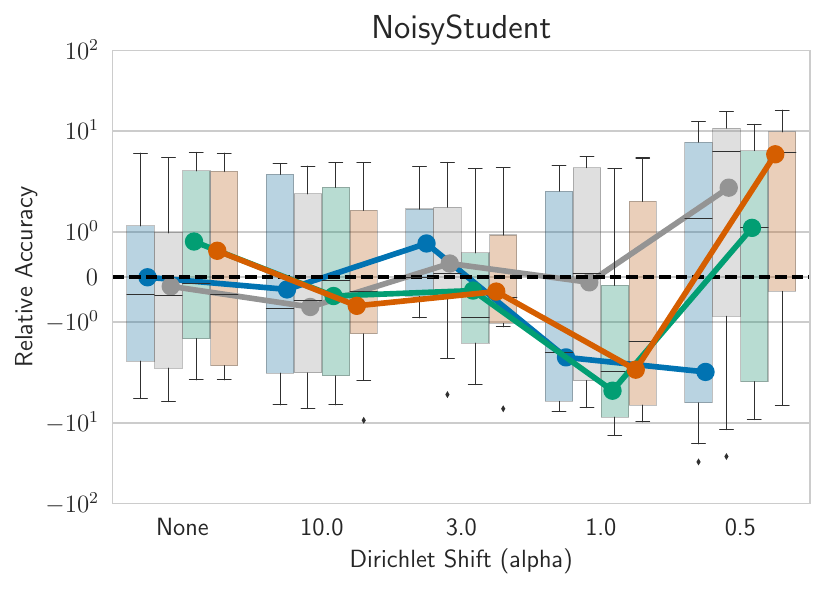}\hfil 
    \includegraphics[width=0.32\linewidth]{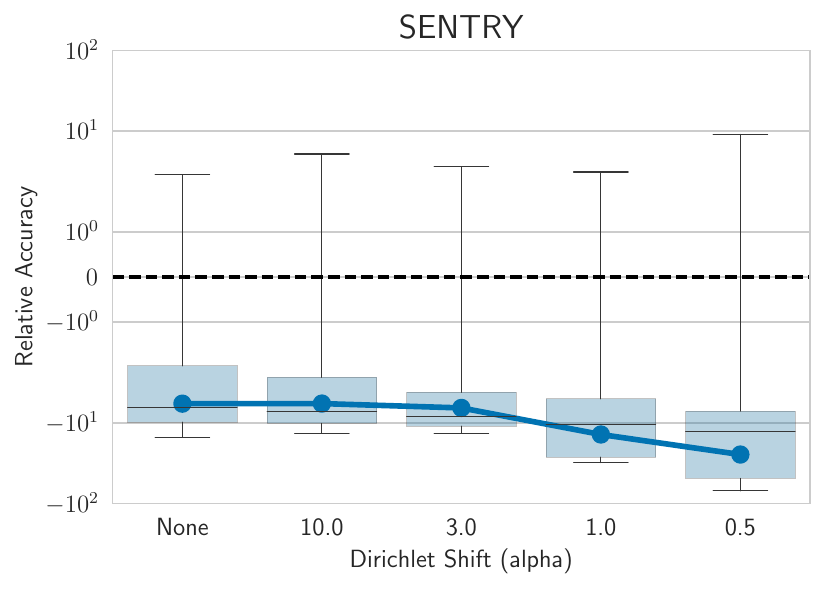}

    \caption{{Entity30. Relative performance and accuracy plots for different DA algorithms across various shift pairs in Entity30.}}
\end{figure}

\begin{figure}[H]
 \centering
    \includegraphics[width=0.5\linewidth]{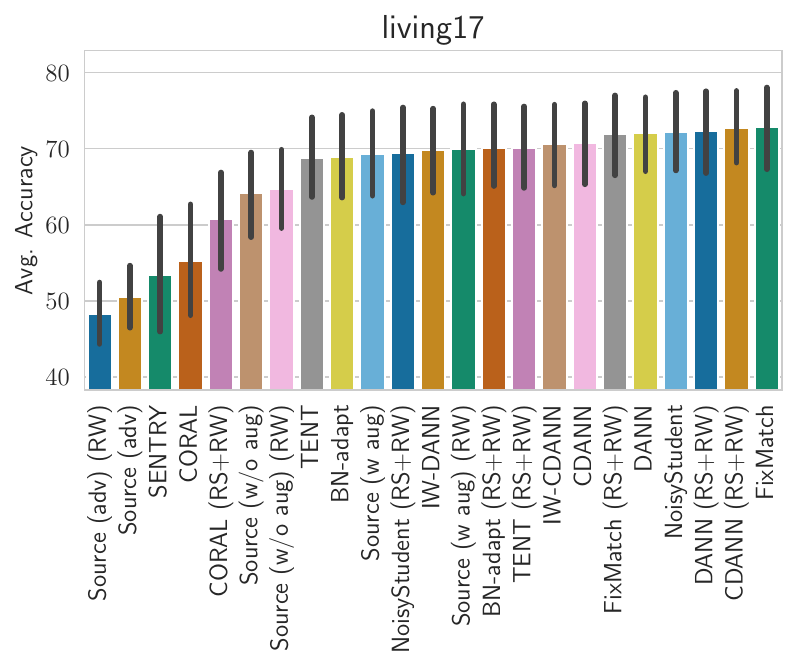} \\
    \includegraphics[width=0.5\linewidth]{figures/legend.pdf} \\
    \includegraphics[width=0.32\linewidth]{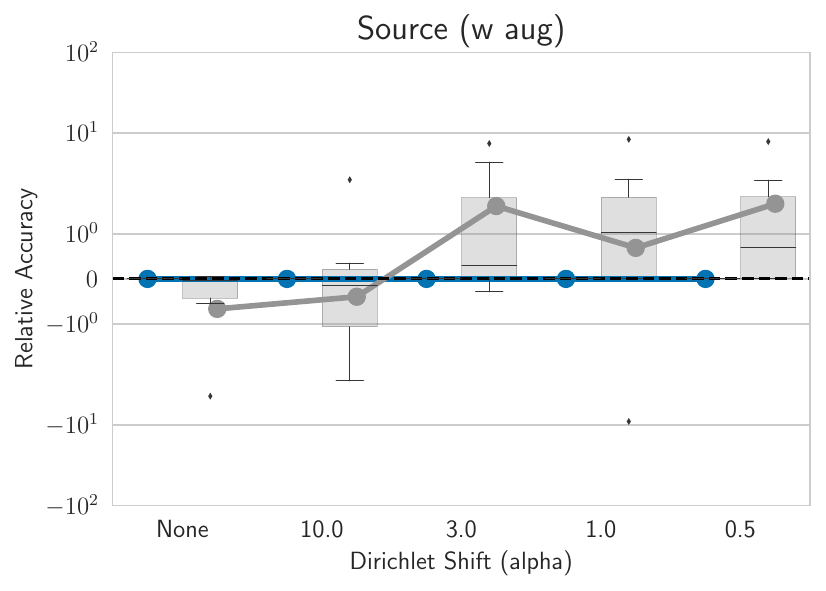} \hfil 
    \includegraphics[width=0.32\linewidth]{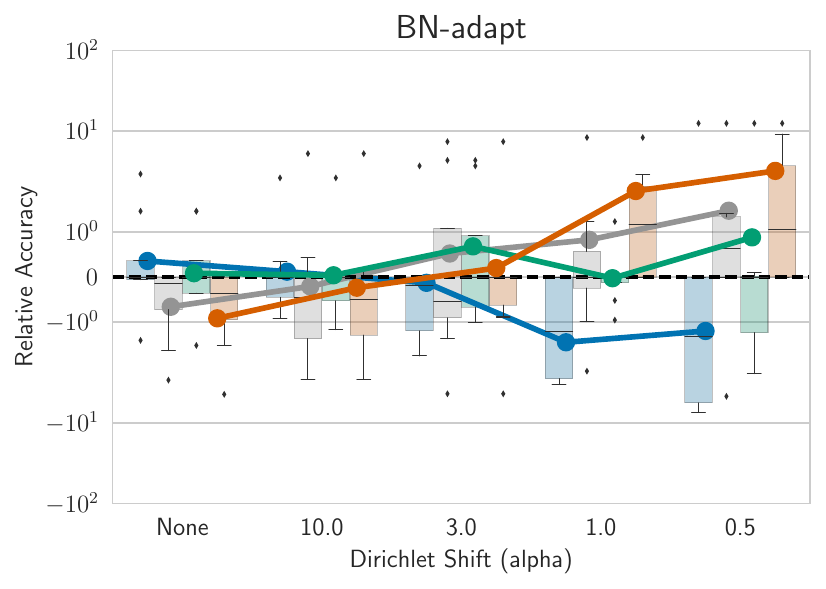}\hfil 
    \includegraphics[width=0.32\linewidth]{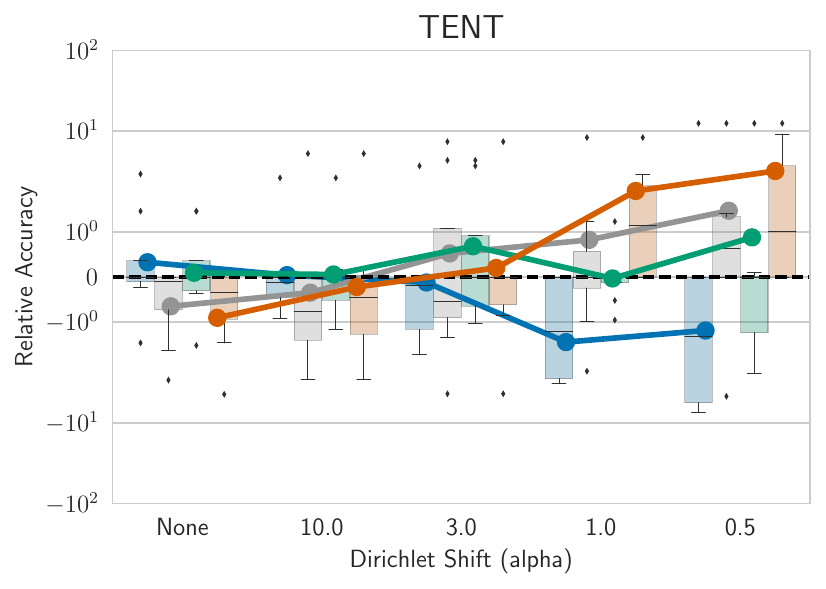}
    \includegraphics[width=0.32\linewidth]{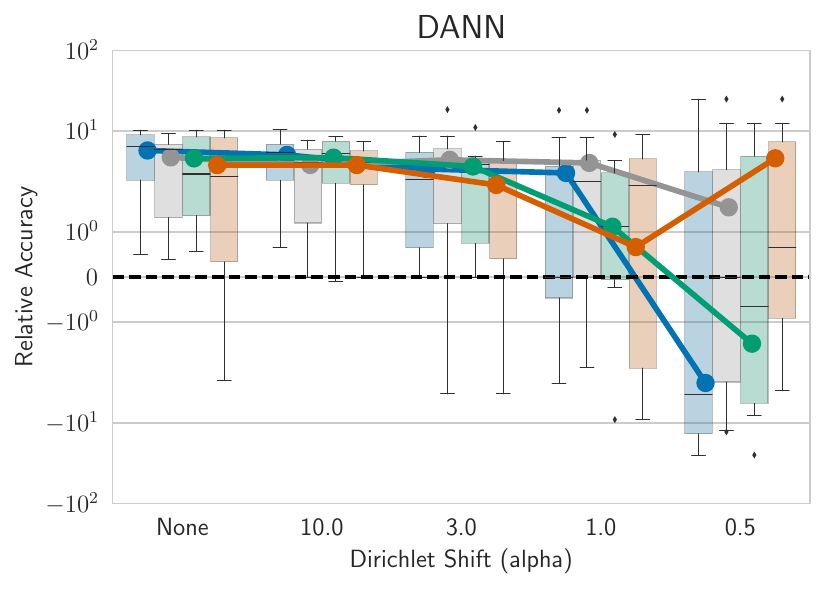} \hfil 
    \includegraphics[width=0.32\linewidth]{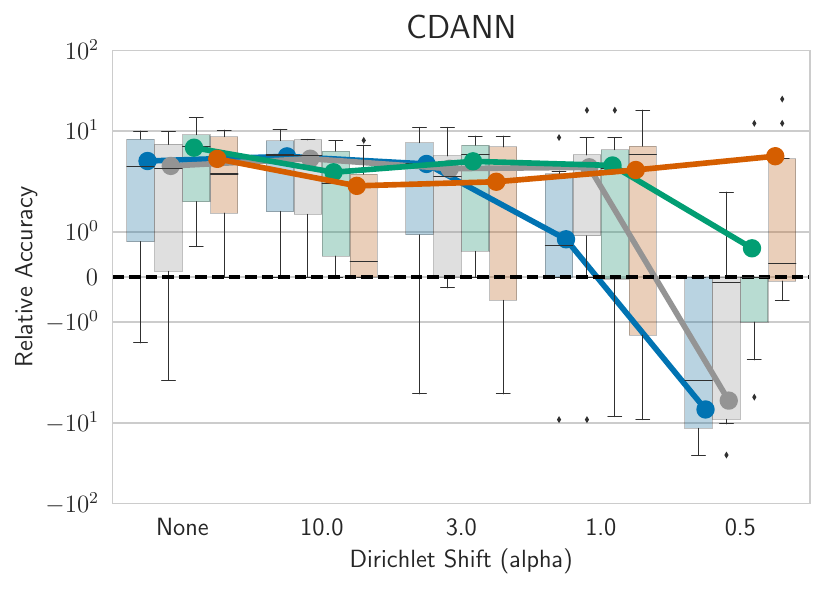}\hfil 
    \includegraphics[width=0.32\linewidth]{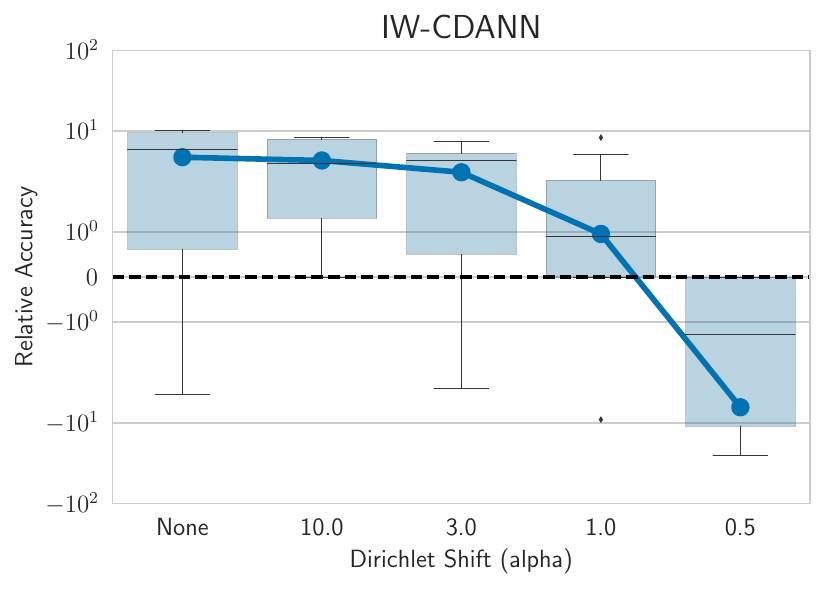}
    \includegraphics[width=0.32\linewidth]{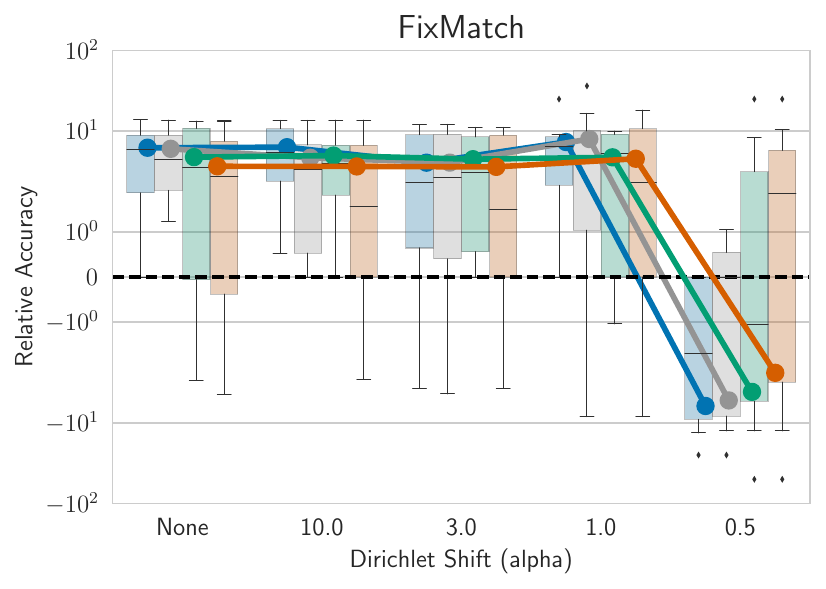} \hfil 
    \includegraphics[width=0.32\linewidth]{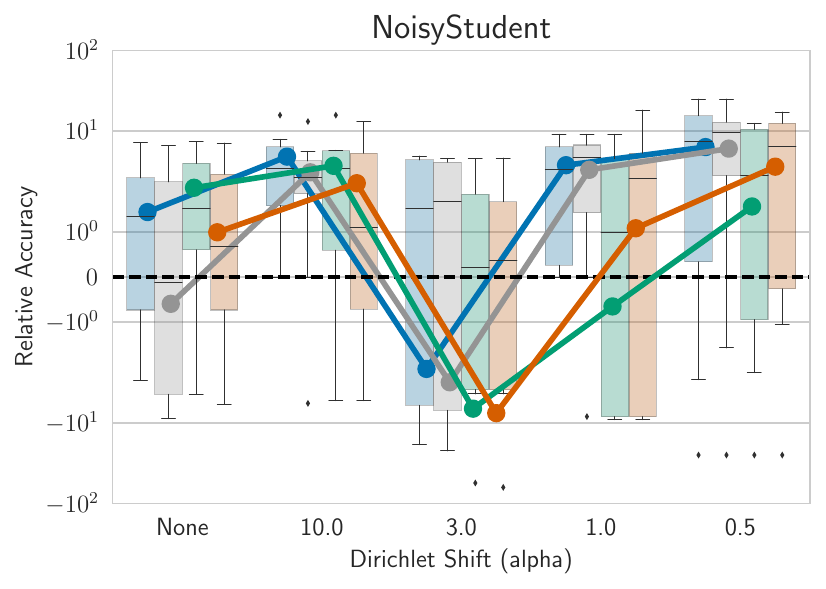}\hfil 
    \includegraphics[width=0.32\linewidth]{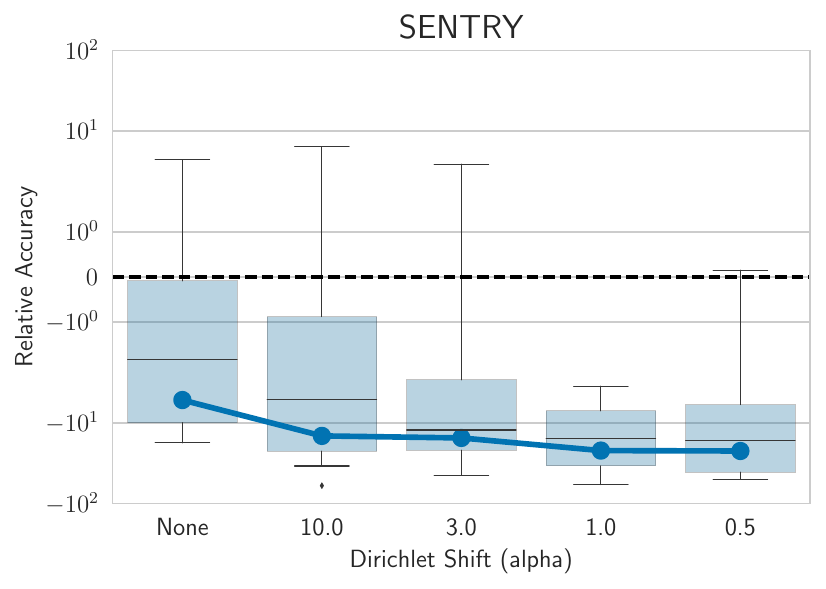}

    \caption{{Living 17. Relative performance and accuracy plots for different DA algorithms across various shift pairs in Living17.}}
\end{figure}

\begin{figure}[H]
 \centering
    \includegraphics[width=0.5\linewidth]{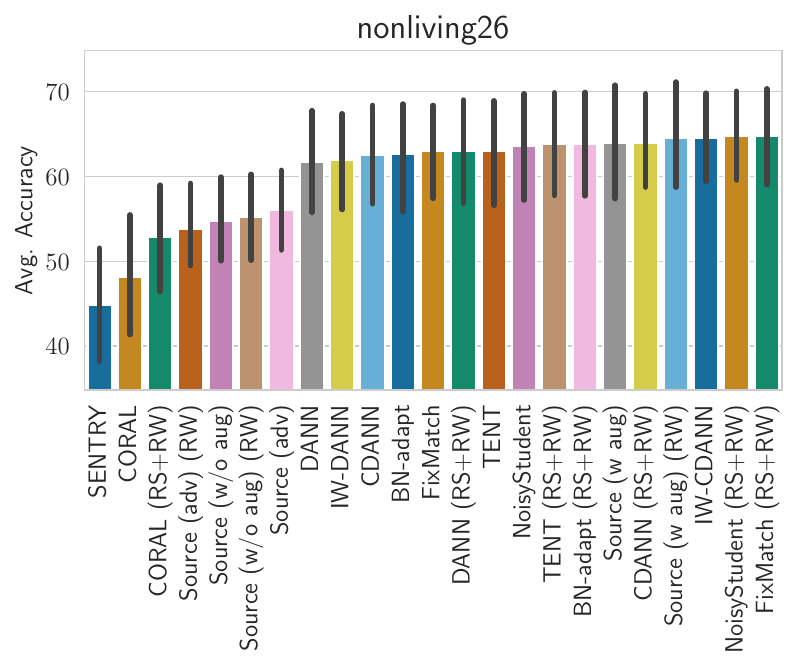} \\
    \includegraphics[width=0.5\linewidth]{figures/legend.pdf} \\
    \includegraphics[width=0.32\linewidth]{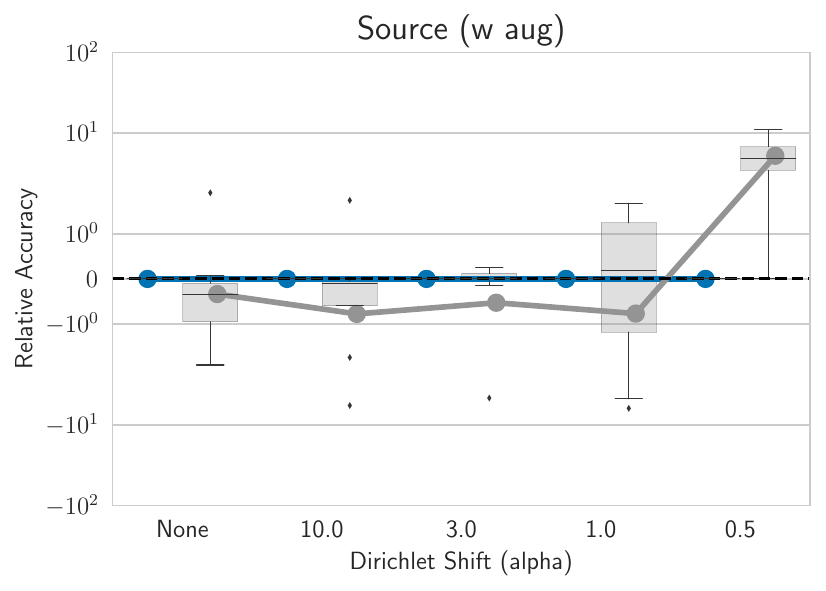} \hfil 
    \includegraphics[width=0.32\linewidth]{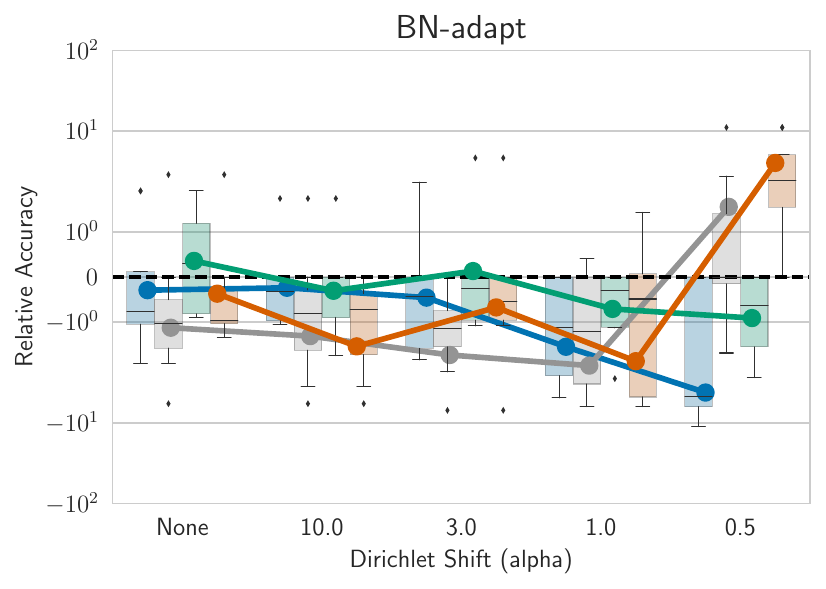}\hfil 
    \includegraphics[width=0.32\linewidth]{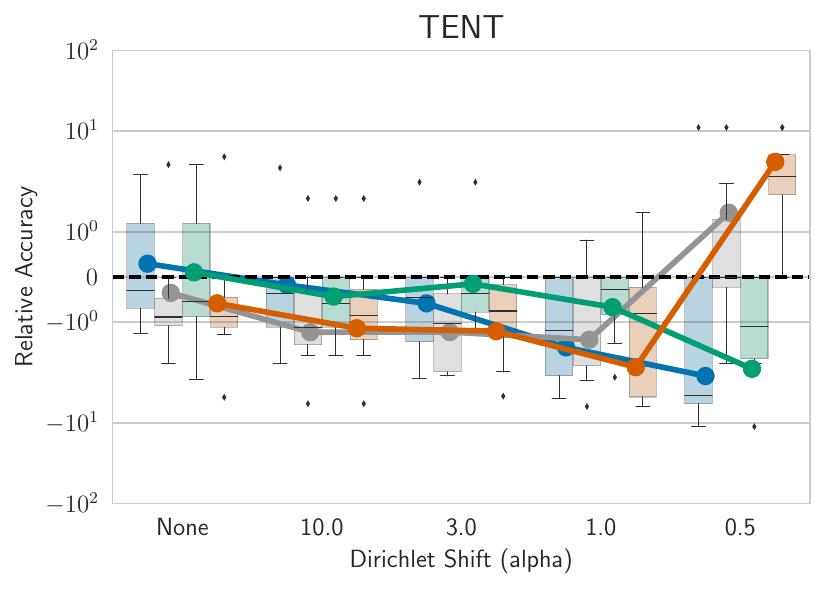}
    \includegraphics[width=0.32\linewidth]{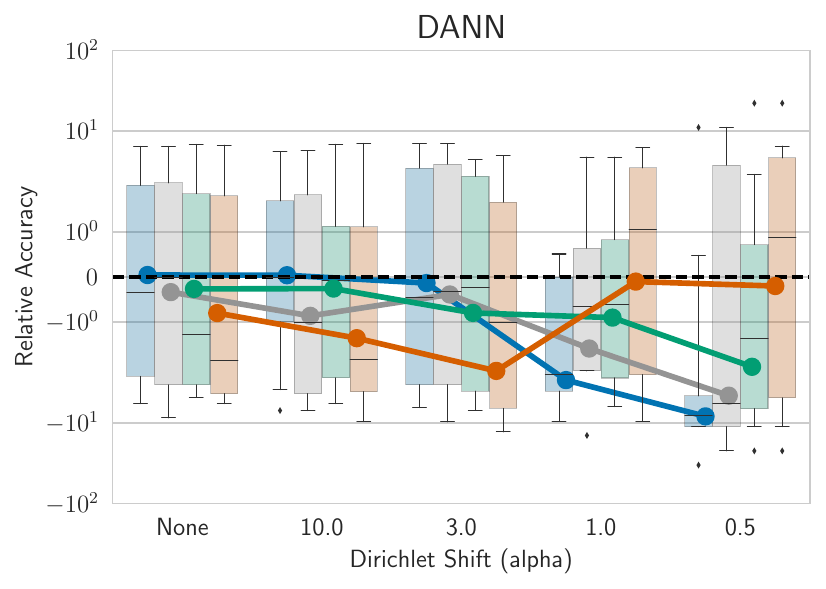} \hfil 
    \includegraphics[width=0.32\linewidth]{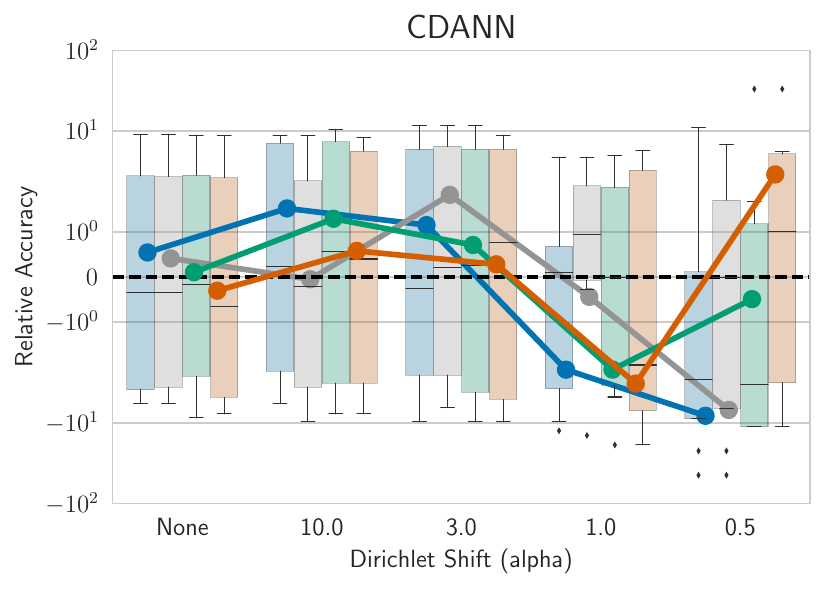}\hfil 
    \includegraphics[width=0.32\linewidth]{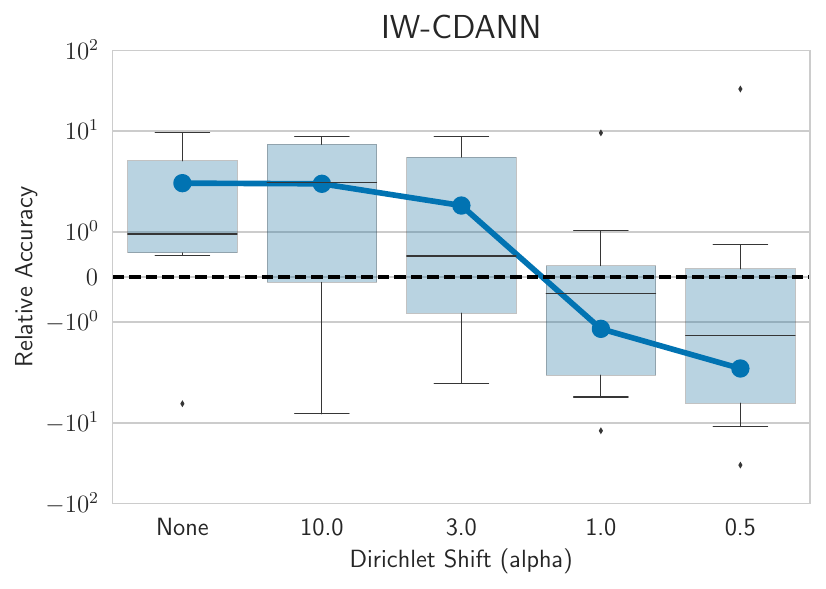}
    \includegraphics[width=0.32\linewidth]{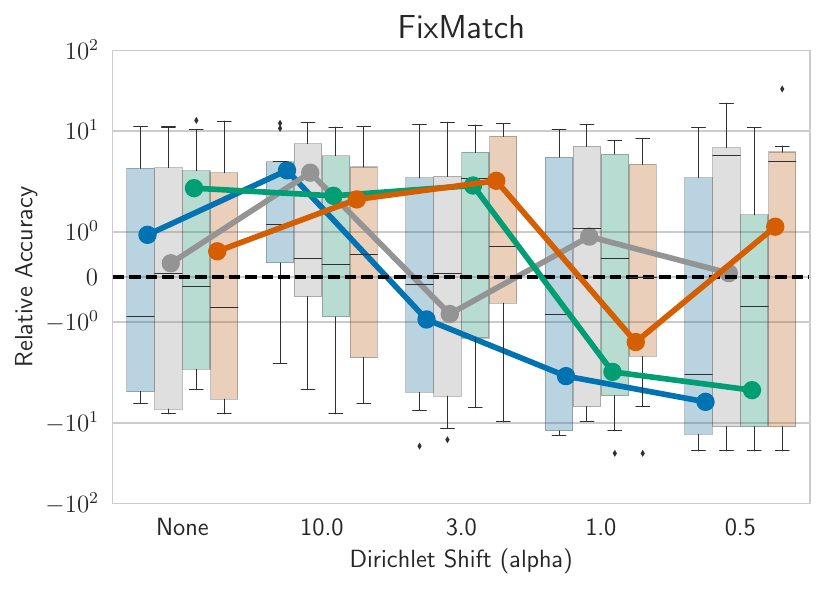} \hfil 
    \includegraphics[width=0.32\linewidth]{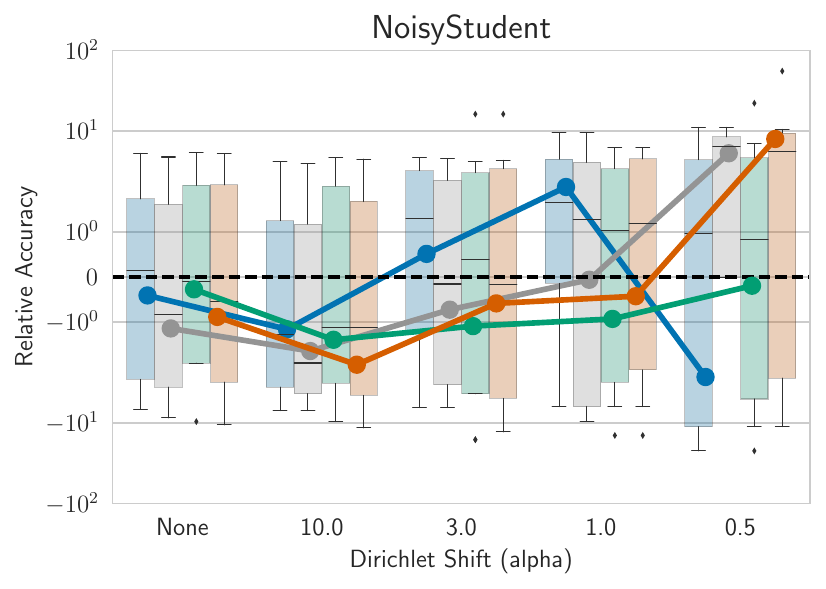}\hfil 
    \includegraphics[width=0.32\linewidth]{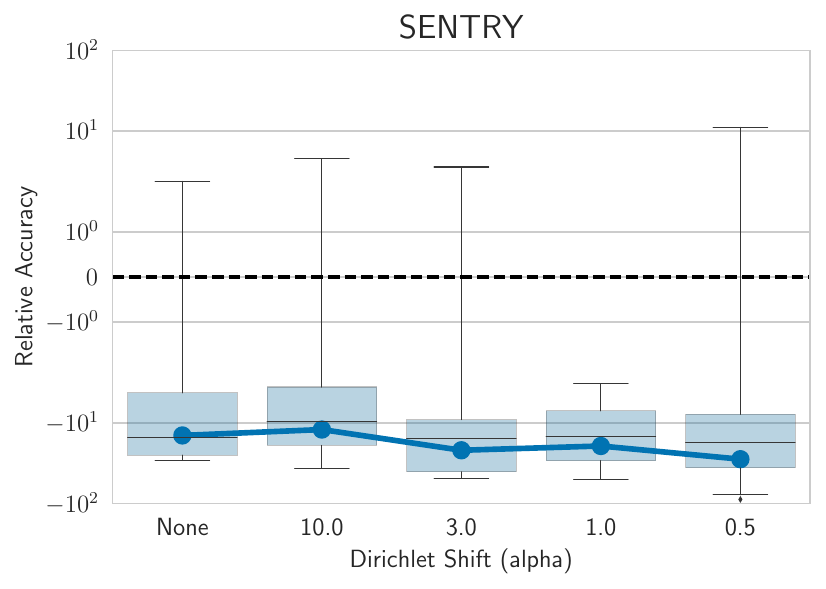}

    \caption{{Nonliving 26. Relative performance and accuracy plots for different DA algorithms across various shift pairs in Nonliving26.}}
\end{figure}

\begin{figure}[H]
 \centering
    \includegraphics[width=0.5\linewidth]{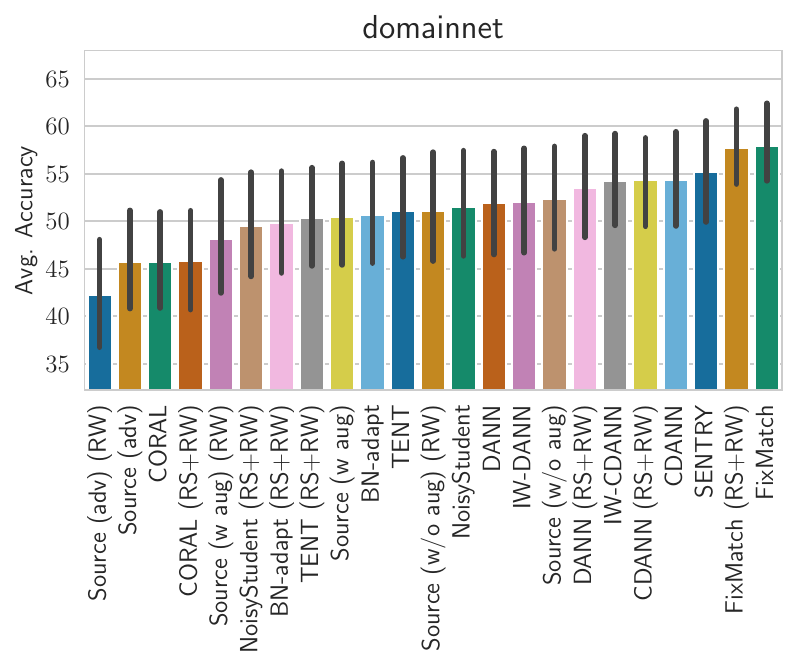} \\
    \includegraphics[width=0.5\linewidth]{figures/legend.pdf} \\
    \includegraphics[width=0.32\linewidth]{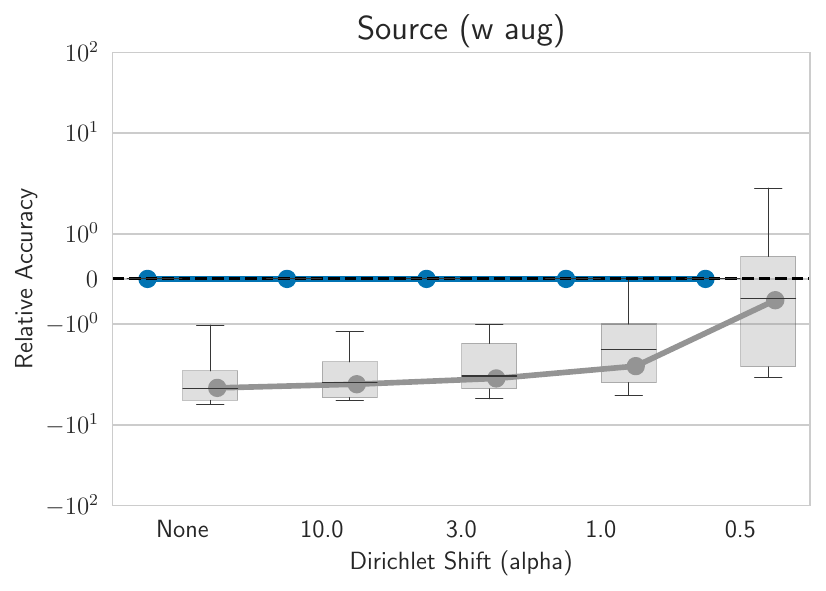} \hfil 
    \includegraphics[width=0.32\linewidth]{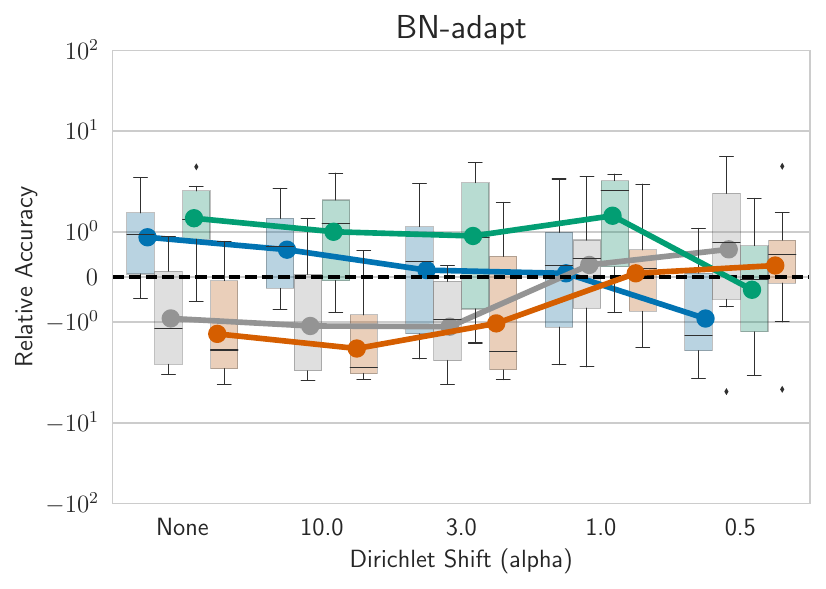}\hfil 
    \includegraphics[width=0.32\linewidth]{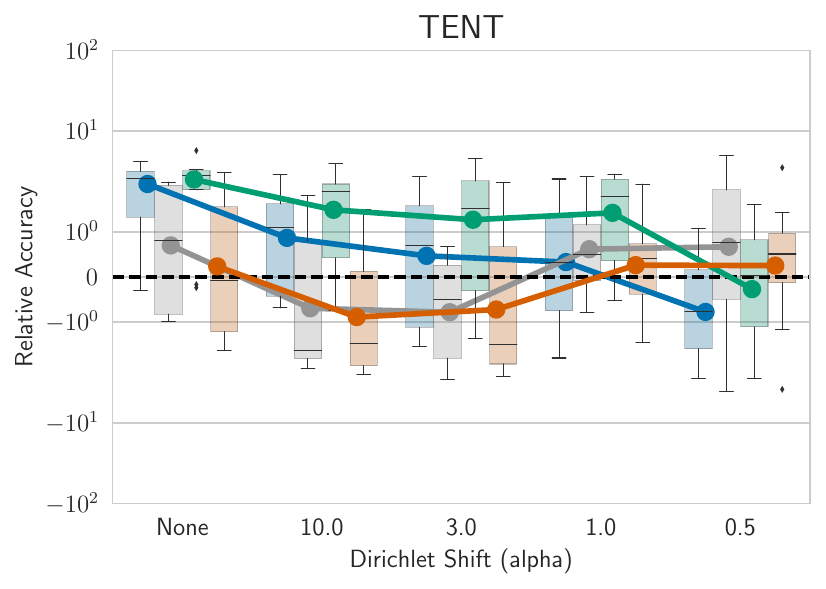}
    \includegraphics[width=0.32\linewidth]{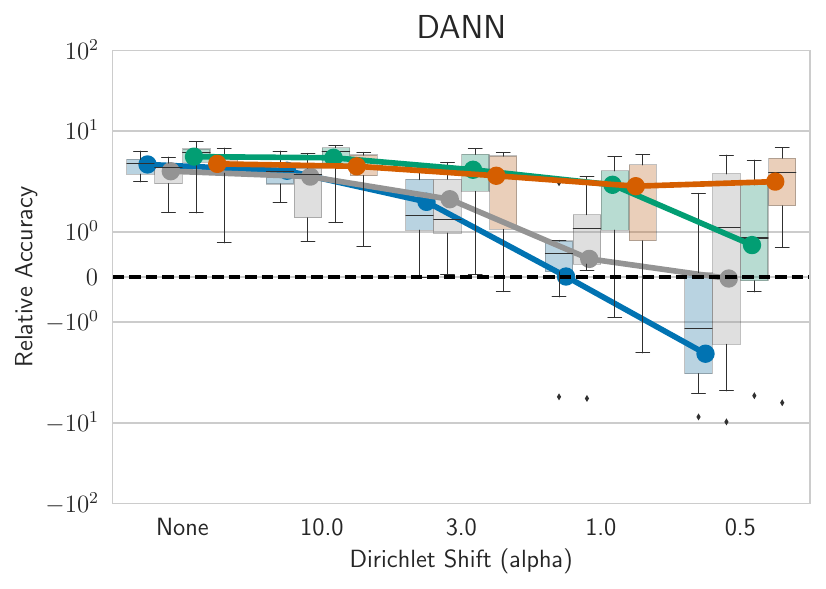} \hfil 
    \includegraphics[width=0.32\linewidth]{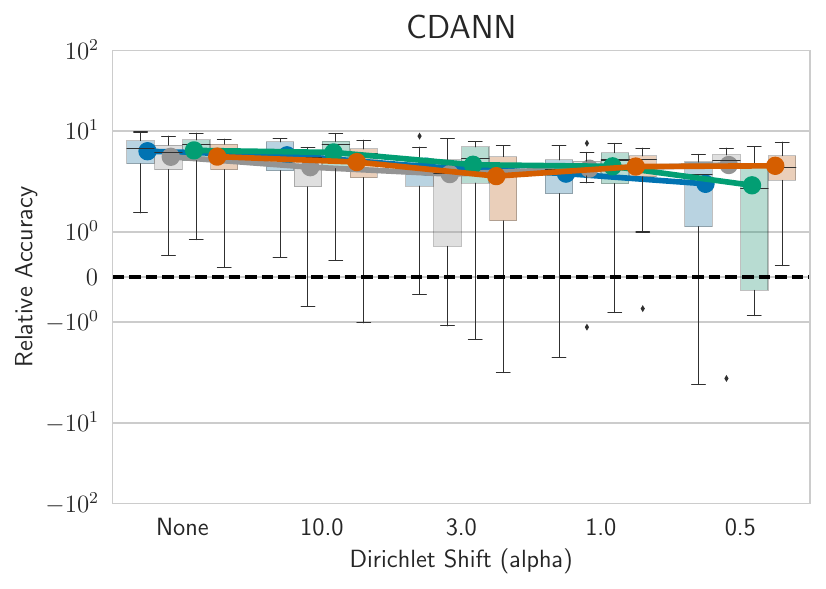}\hfil 
    \includegraphics[width=0.32\linewidth]{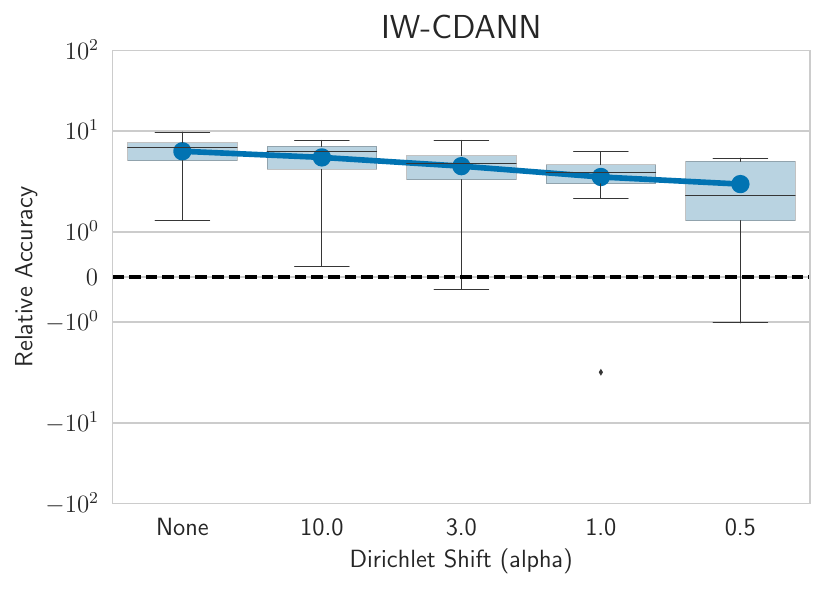}
    \includegraphics[width=0.32\linewidth]{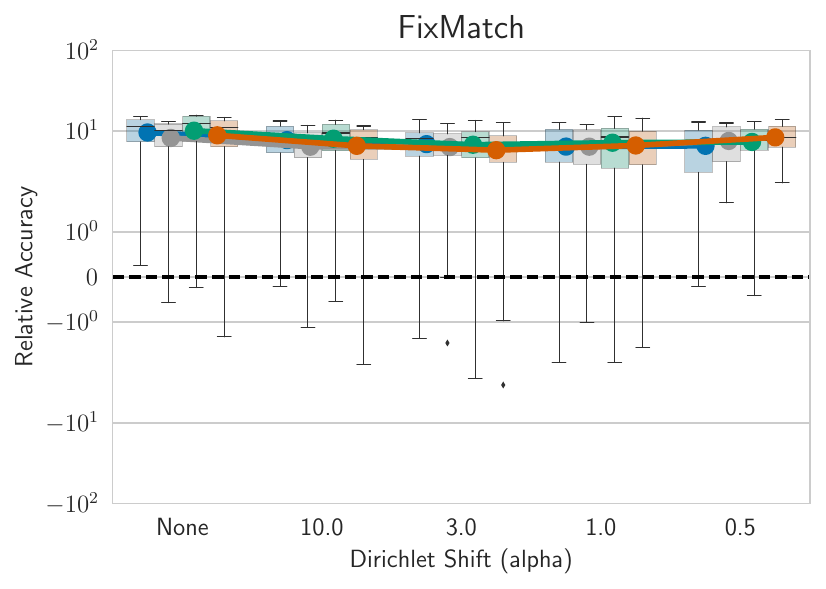} \hfil 
    \includegraphics[width=0.32\linewidth]{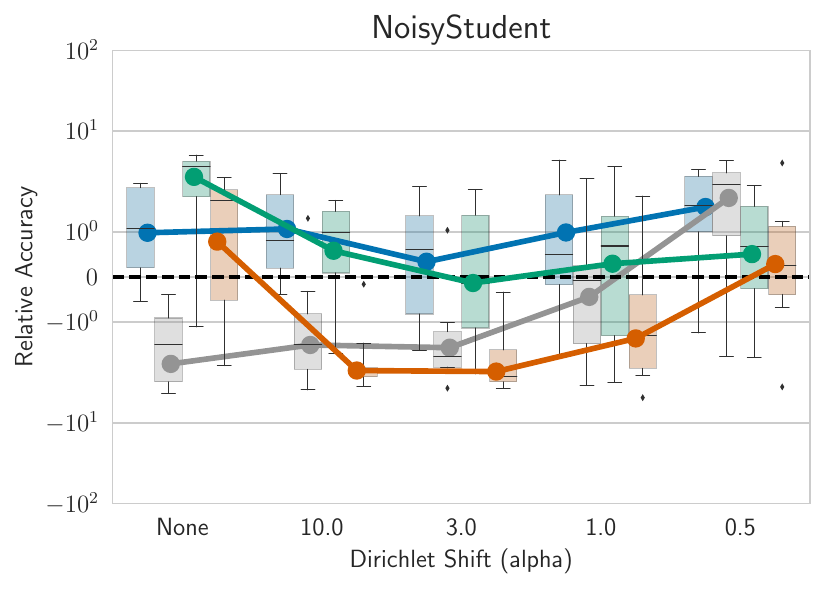}\hfil 
    \includegraphics[width=0.32\linewidth]{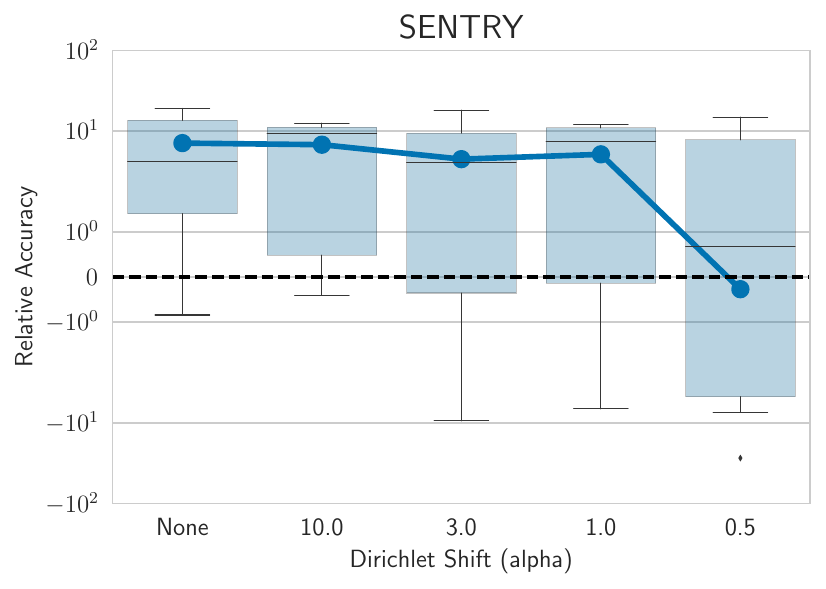}

    \caption{{DomainNet. Relative performance and accuracy plots for different DA algorithms across various shift pairs in DomainNet.}}
\end{figure}

\begin{figure}[H]
 \centering
    \includegraphics[width=0.5\linewidth]{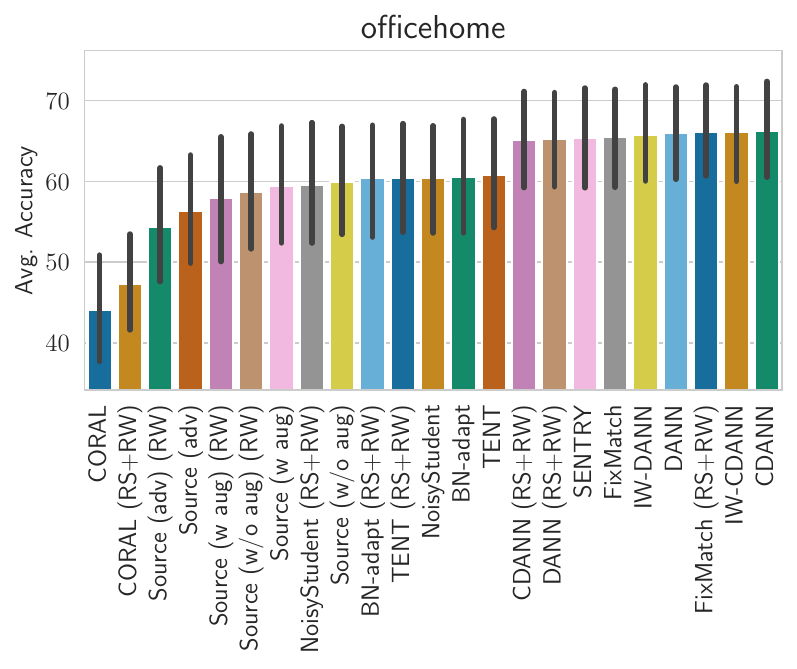} \\
    \includegraphics[width=0.5\linewidth]{figures/legend.pdf} \\
    \includegraphics[width=0.32\linewidth]{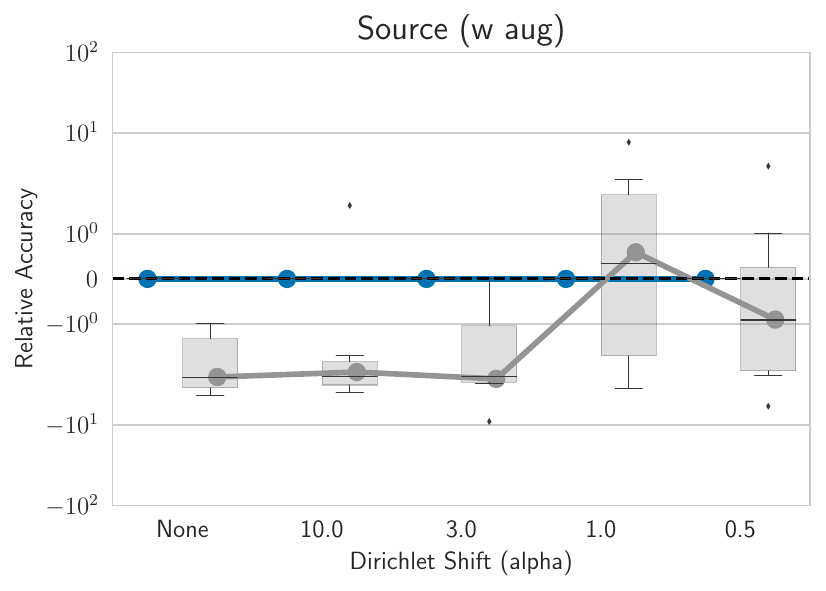} \hfil 
    \includegraphics[width=0.32\linewidth]{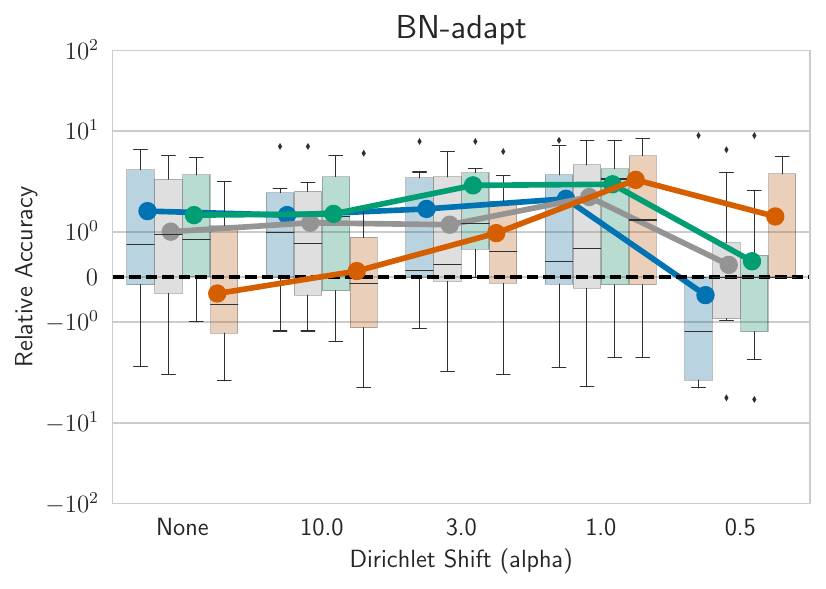}\hfil 
    \includegraphics[width=0.32\linewidth]{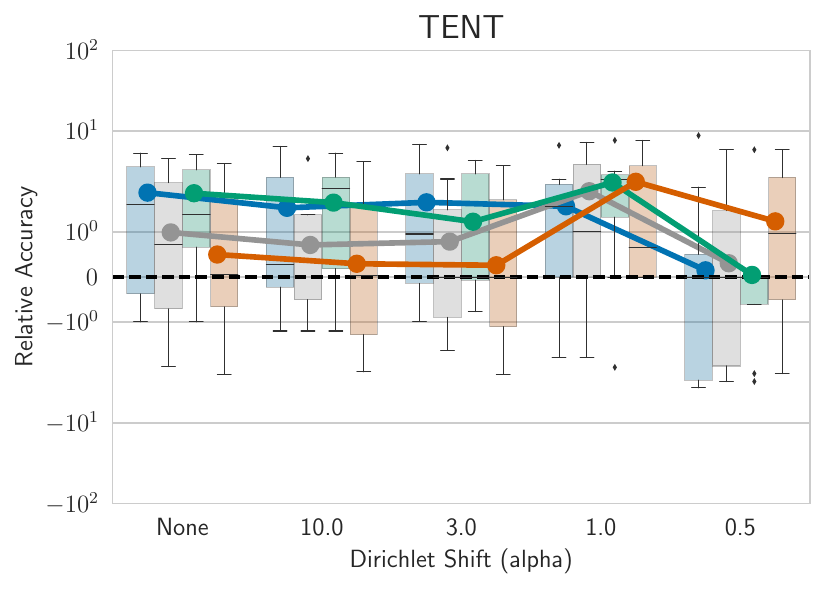}
    \includegraphics[width=0.32\linewidth]{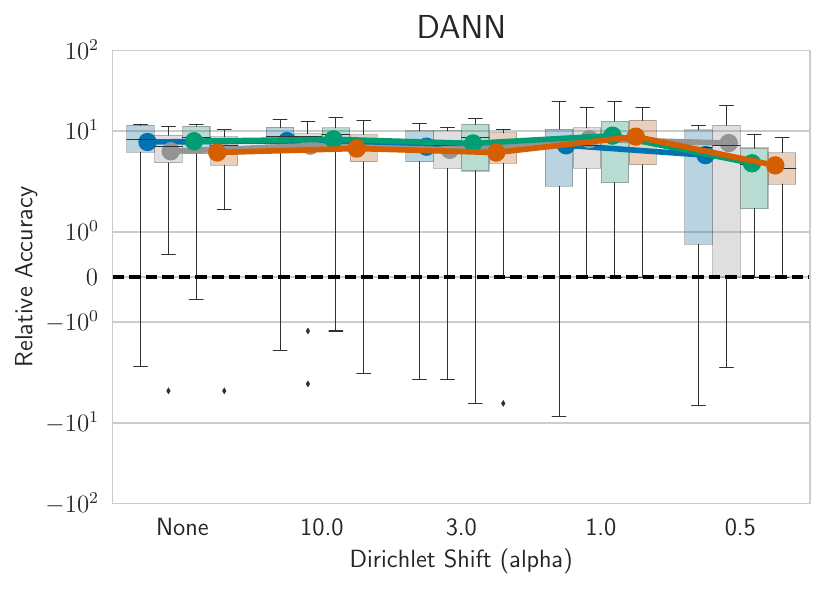} \hfil 
    \includegraphics[width=0.32\linewidth]{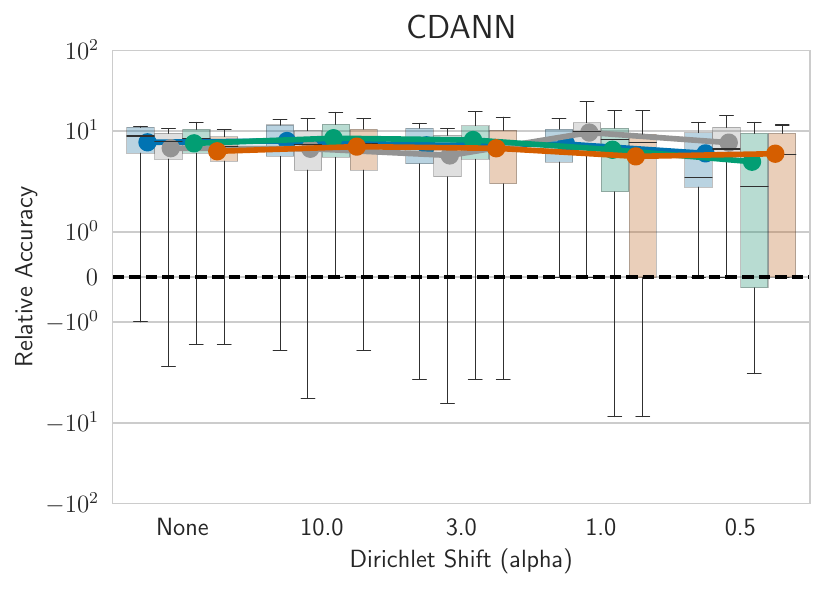}\hfil 
    \includegraphics[width=0.32\linewidth]{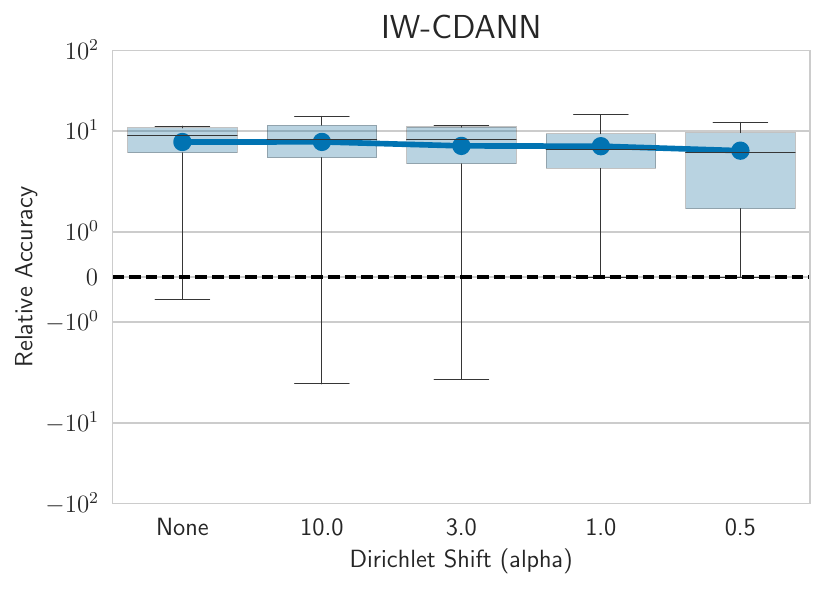}
    \includegraphics[width=0.32\linewidth]{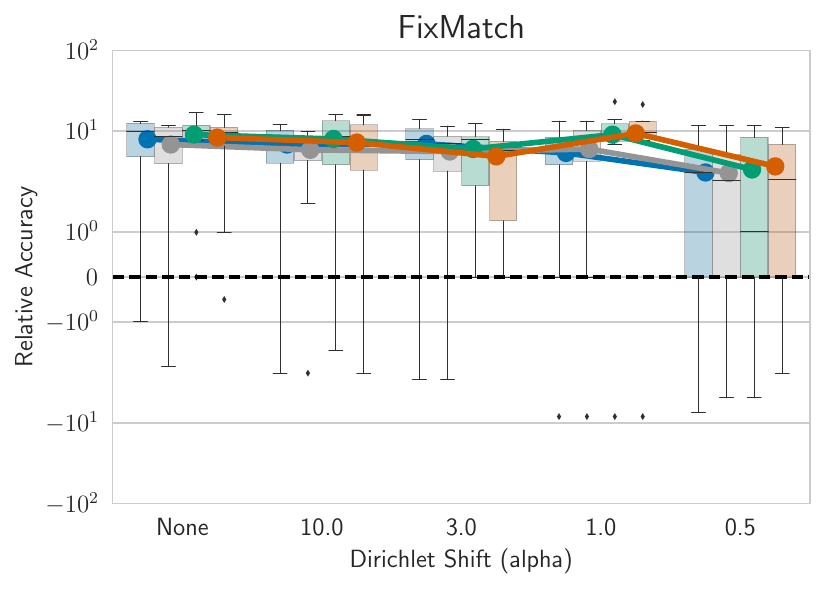} \hfil 
    \includegraphics[width=0.32\linewidth]{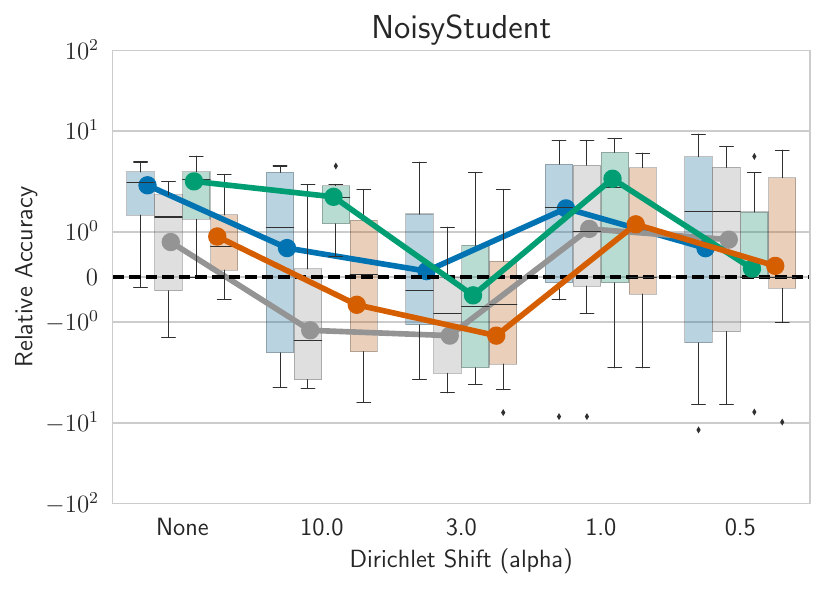}\hfil 
    \includegraphics[width=0.32\linewidth]{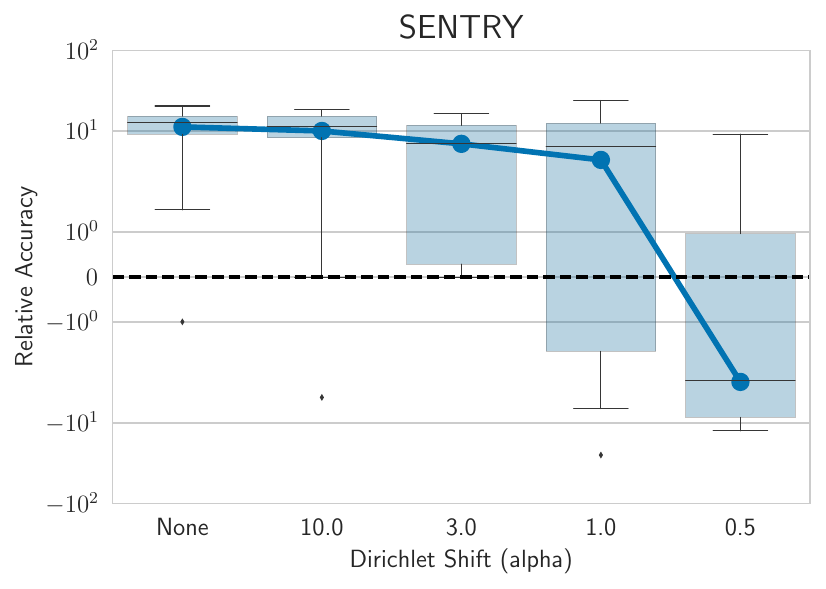}

    \caption{{Officehome. Relative performance and accuracy plots for different DA algorithms across various shift pairs in Officehome.}}
\end{figure}

\begin{figure}[H]
 \centering
    \includegraphics[width=0.5\linewidth]{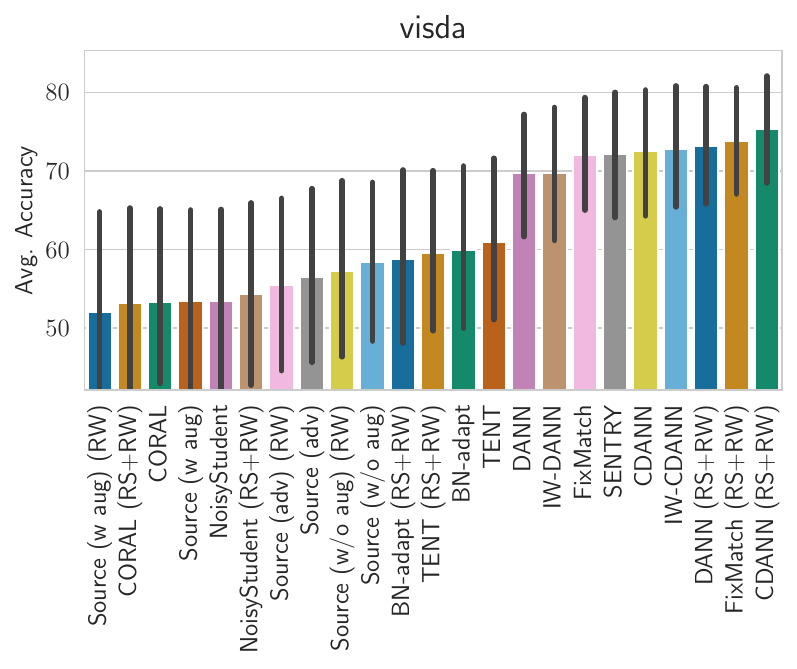} \\
    \includegraphics[width=0.5\linewidth]{figures/legend.pdf} \\
    \includegraphics[width=0.32\linewidth]{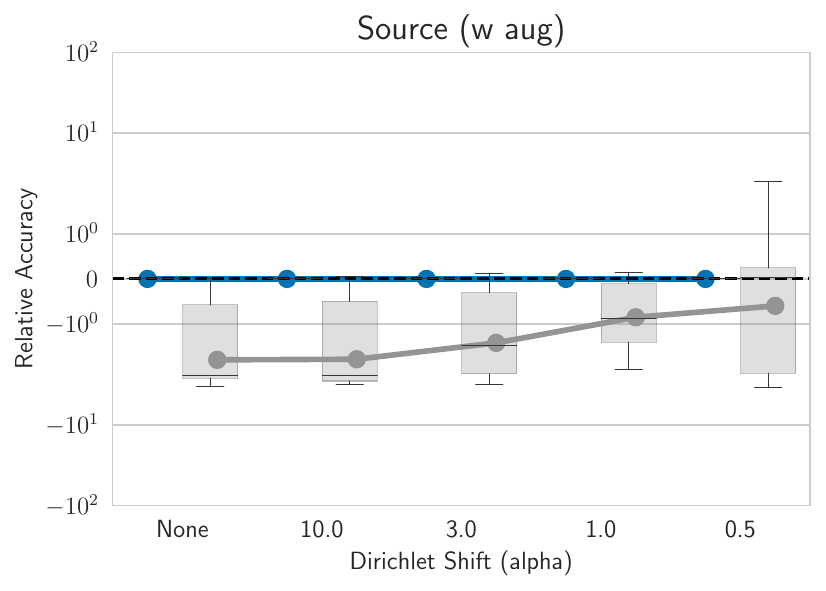} \hfil 
    \includegraphics[width=0.32\linewidth]{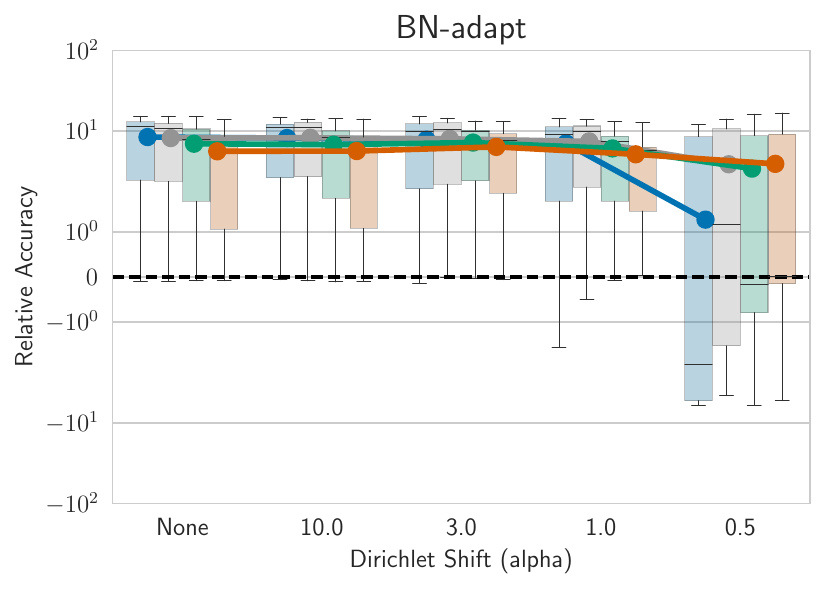}\hfil 
    \includegraphics[width=0.32\linewidth]{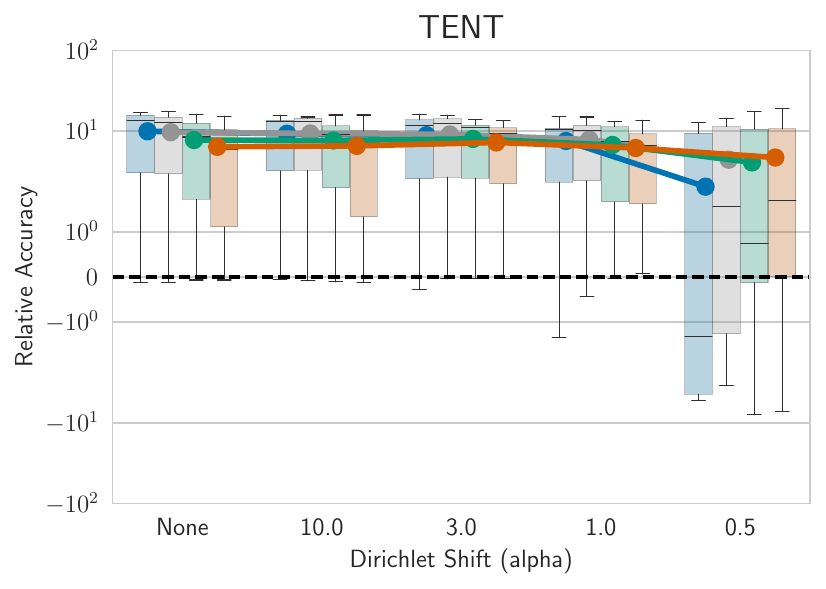}
    \includegraphics[width=0.32\linewidth]{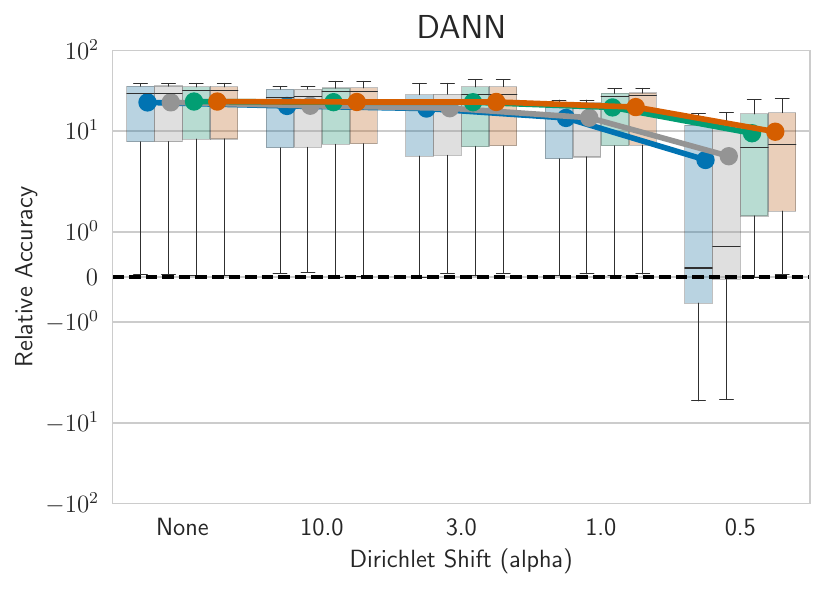} \hfil 
    \includegraphics[width=0.32\linewidth]{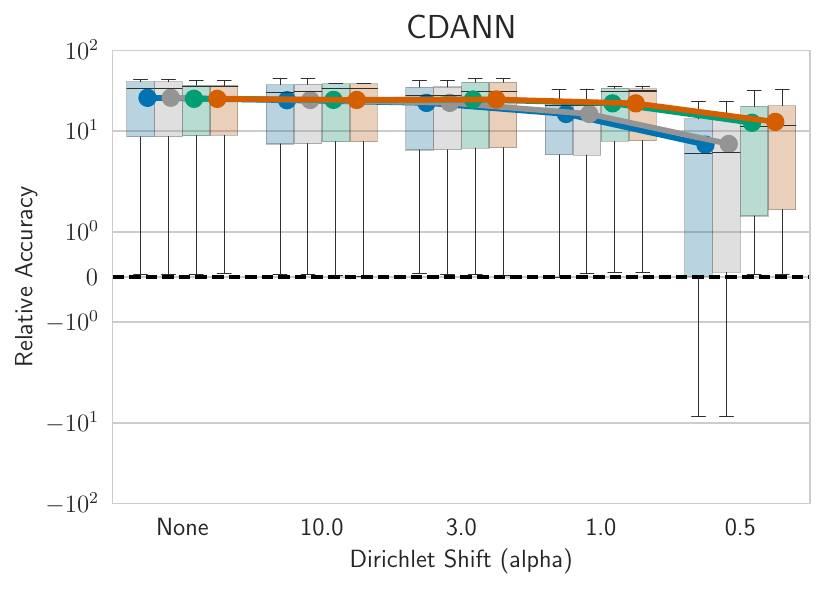}\hfil 
    \includegraphics[width=0.32\linewidth]{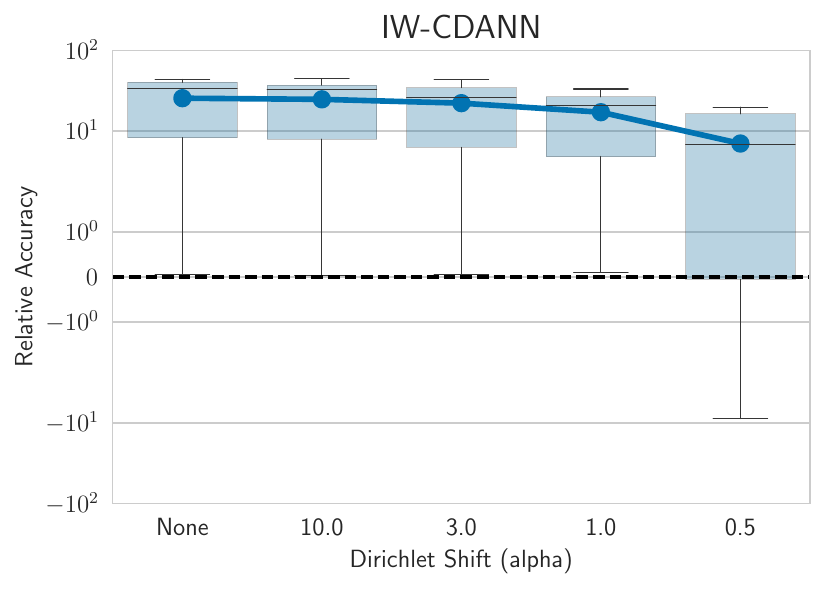}
    \includegraphics[width=0.32\linewidth]{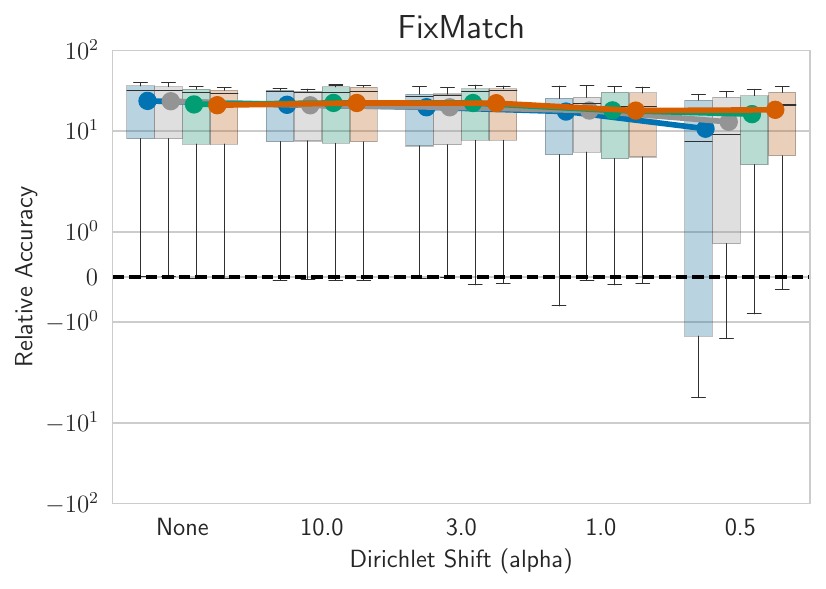} \hfil 
    \includegraphics[width=0.32\linewidth]{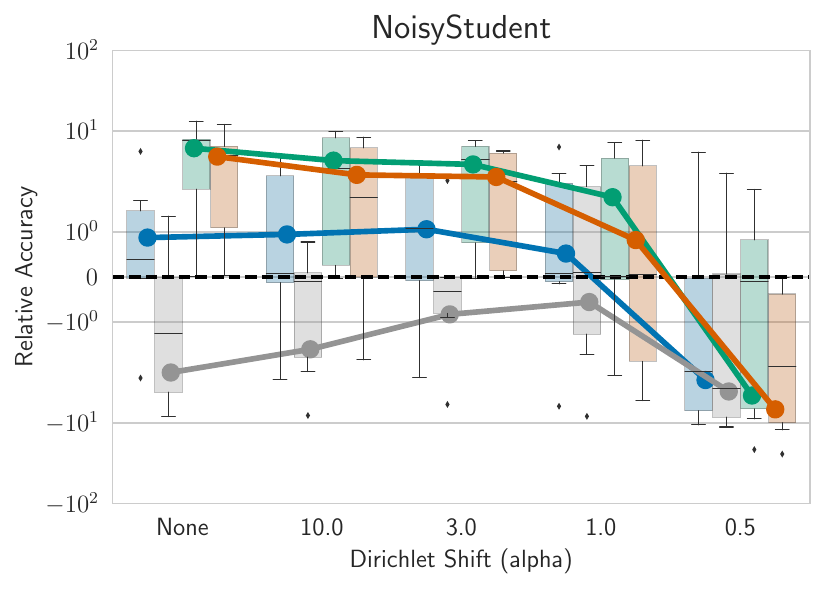}\hfil 
    \includegraphics[width=0.32\linewidth]{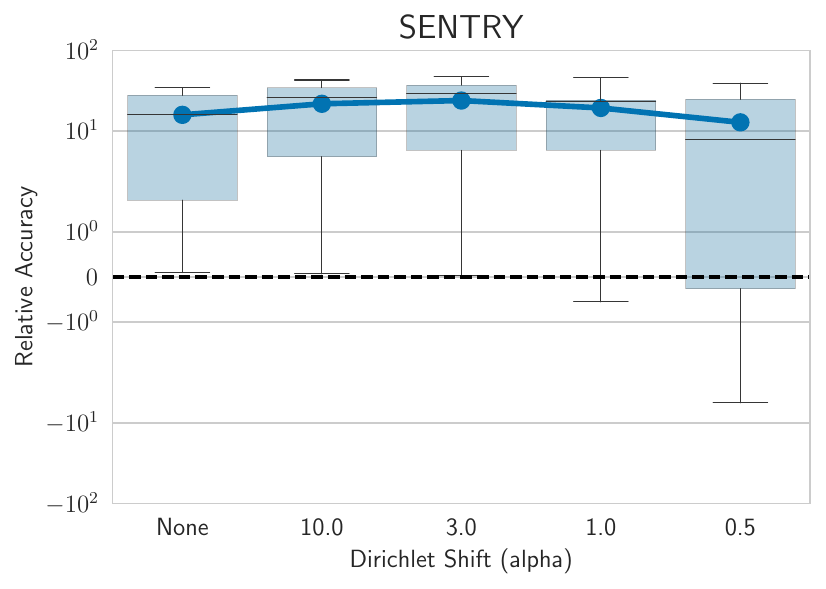}

    \caption{{Visda. Relative performance and accuracy plots for different DA algorithms across various shift pairs in Visda.}}
\end{figure}

\begin{figure}[H]
 \centering
    \includegraphics[width=0.5\linewidth]{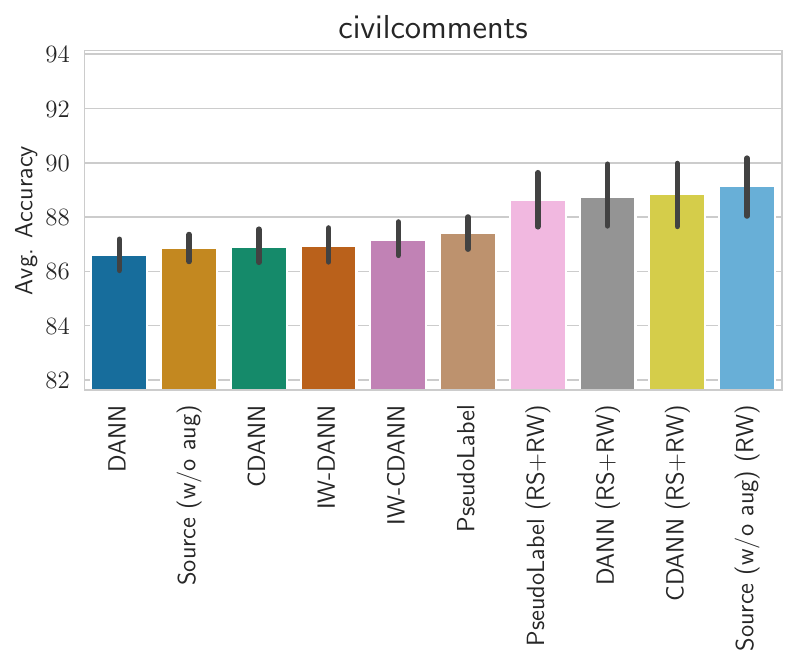} \\
    \includegraphics[width=0.5\linewidth]{figures/legend.pdf} \\
    \includegraphics[width=0.32\linewidth]{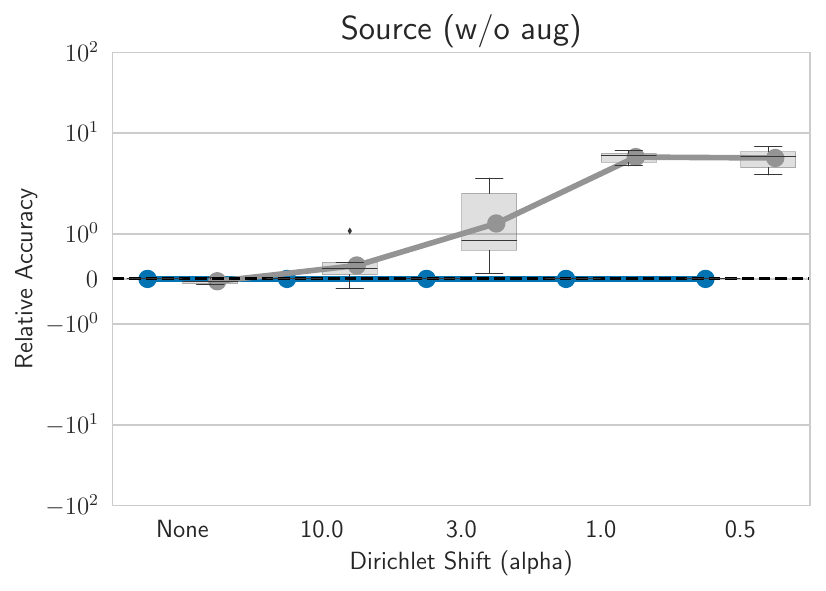} \hfil 
    \includegraphics[width=0.32\linewidth]{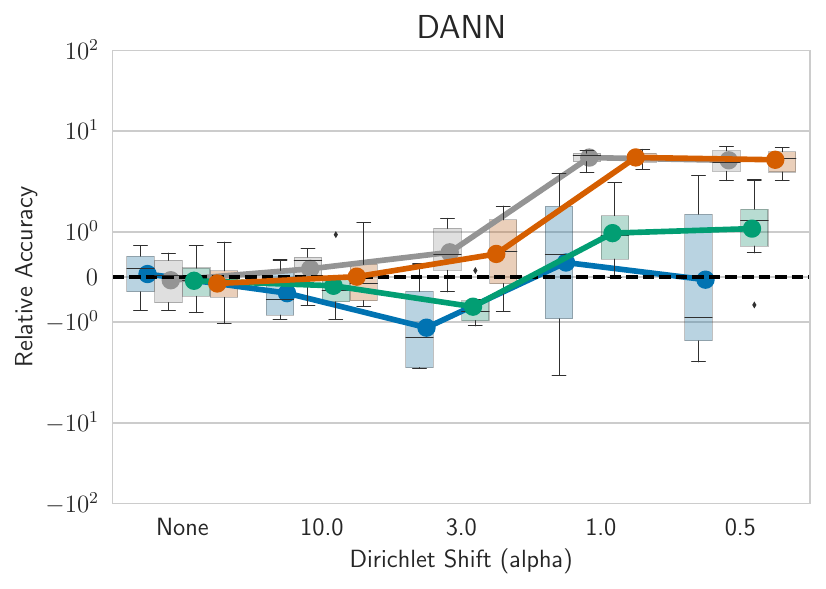} \hfil 
    \includegraphics[width=0.32\linewidth]{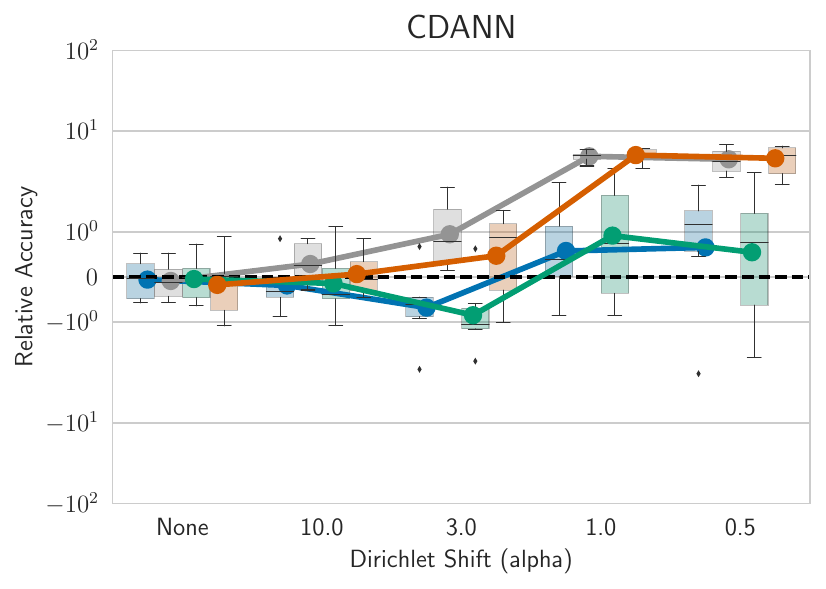}

    \includegraphics[width=0.32\linewidth]{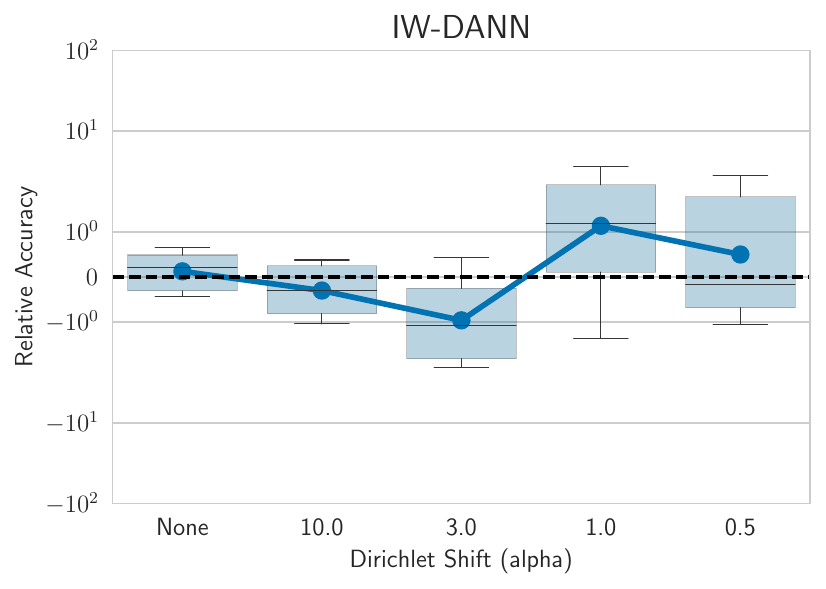}\hfil
    \includegraphics[width=0.32\linewidth]{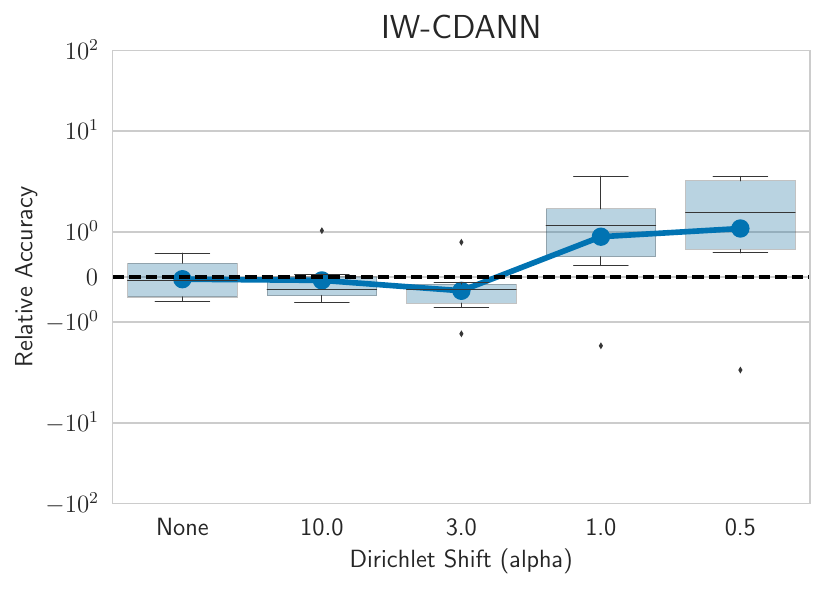}\hfil
    \includegraphics[width=0.32\linewidth]{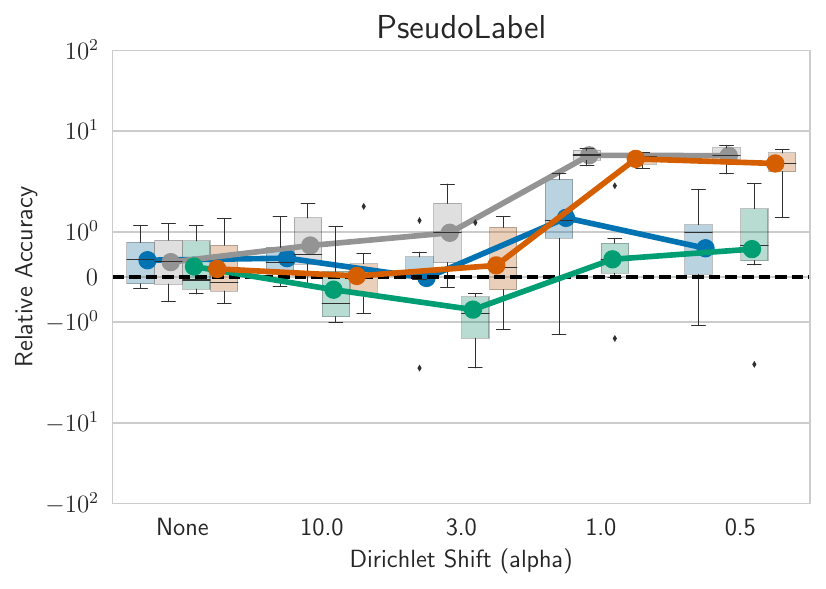}

    \caption{{Civilcomments. Relative performance and accuracy plots for different DA algorithms across various shift pairs in Civilcomments.}}
\end{figure}

\begin{figure}[H]
 \centering
    \includegraphics[width=0.5\linewidth]{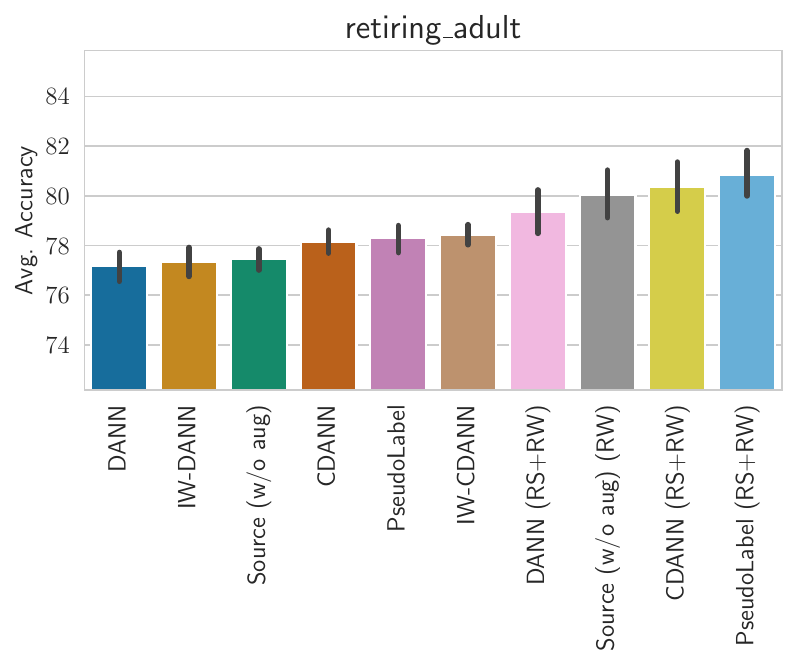} \\
    \includegraphics[width=0.5\linewidth]{figures/legend.pdf} \\
    \includegraphics[width=0.32\linewidth]{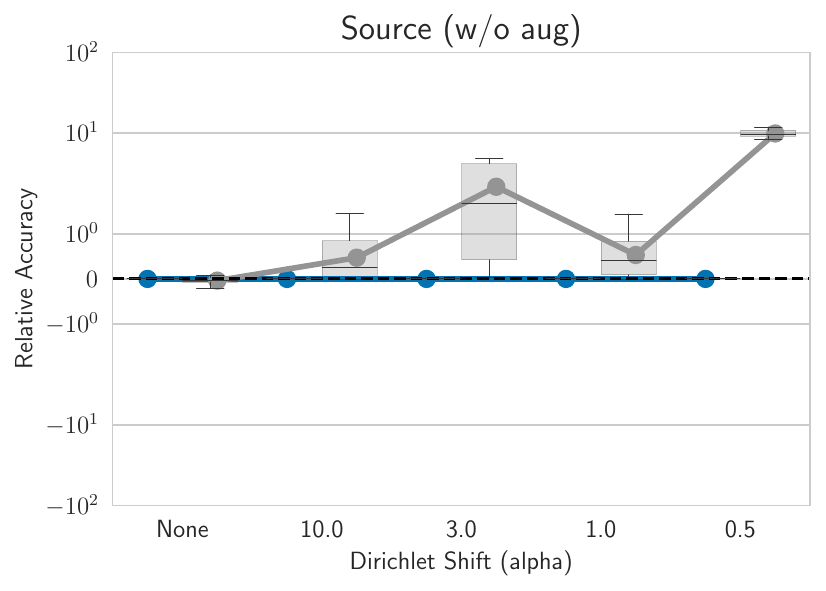} \hfil 
    \includegraphics[width=0.32\linewidth]{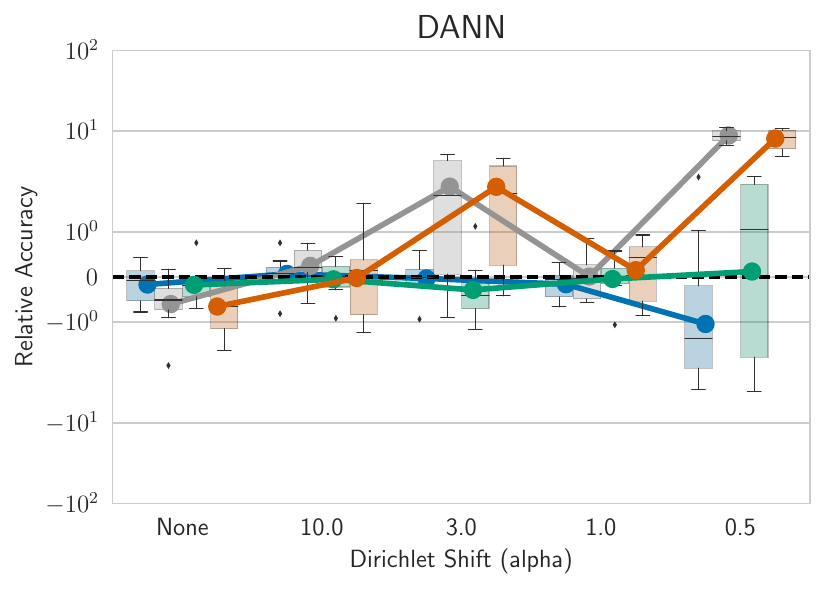} \hfil 
    \includegraphics[width=0.32\linewidth]{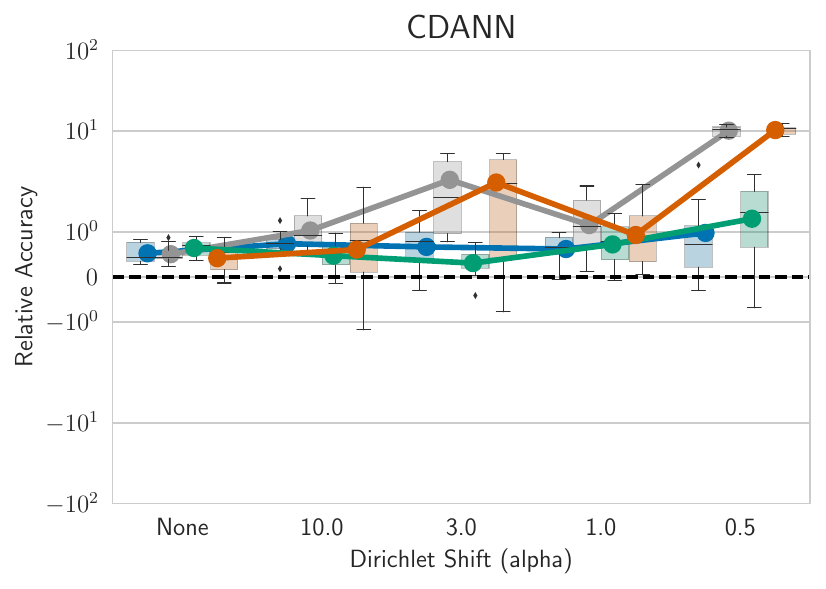}

    \includegraphics[width=0.32\linewidth]{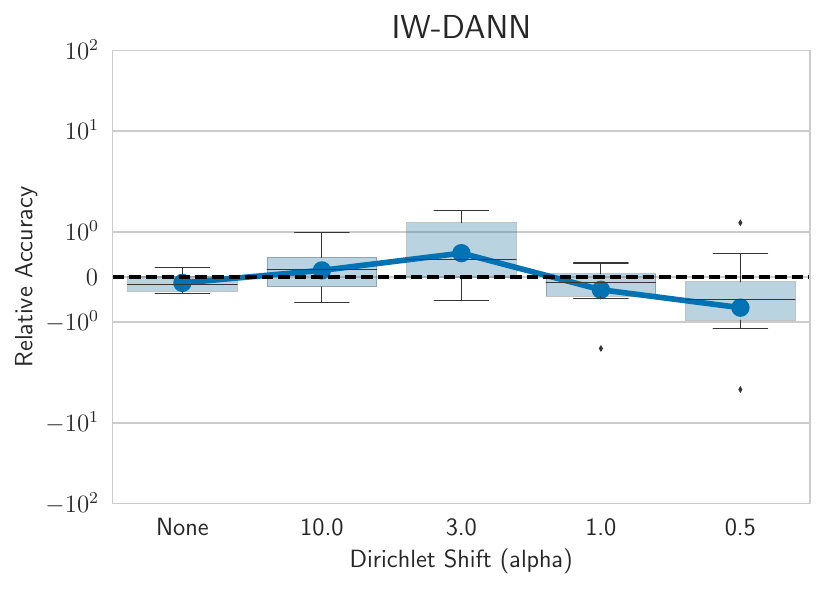}\hfil
    \includegraphics[width=0.32\linewidth]{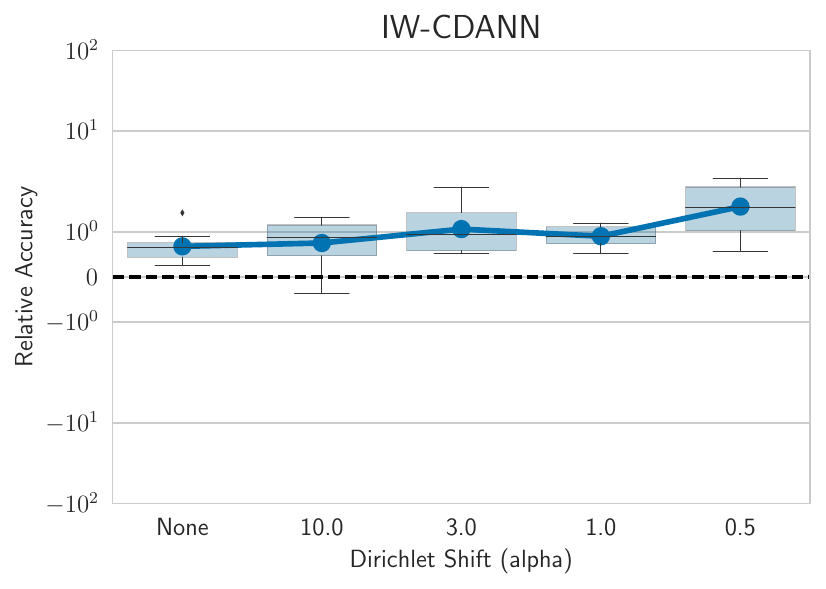}\hfil
    \includegraphics[width=0.32\linewidth]{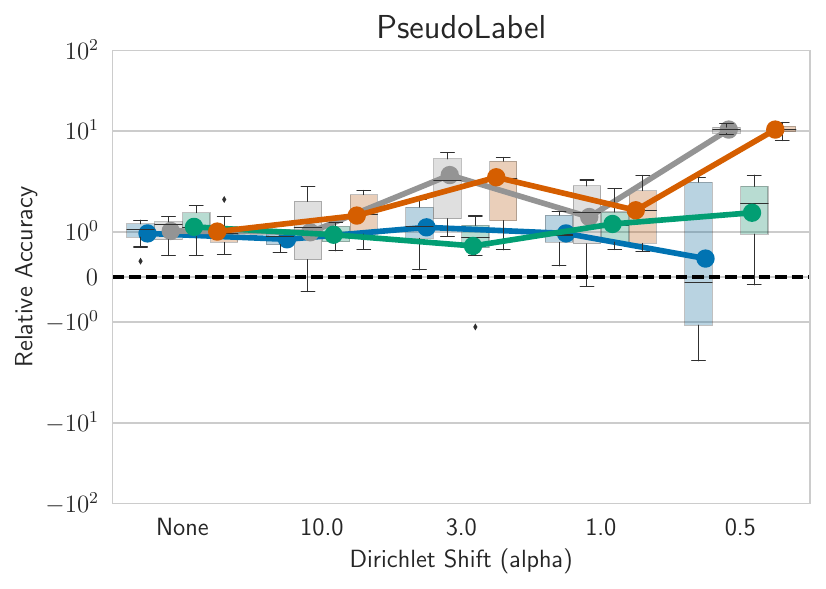}

    \caption{{Retiring Adults. Relative performance and accuracy plots for different DA algorithms across various shift pairs in Retiring Adults.}}
\end{figure}

\begin{figure}[H]
 \centering
    \includegraphics[width=0.5\linewidth]{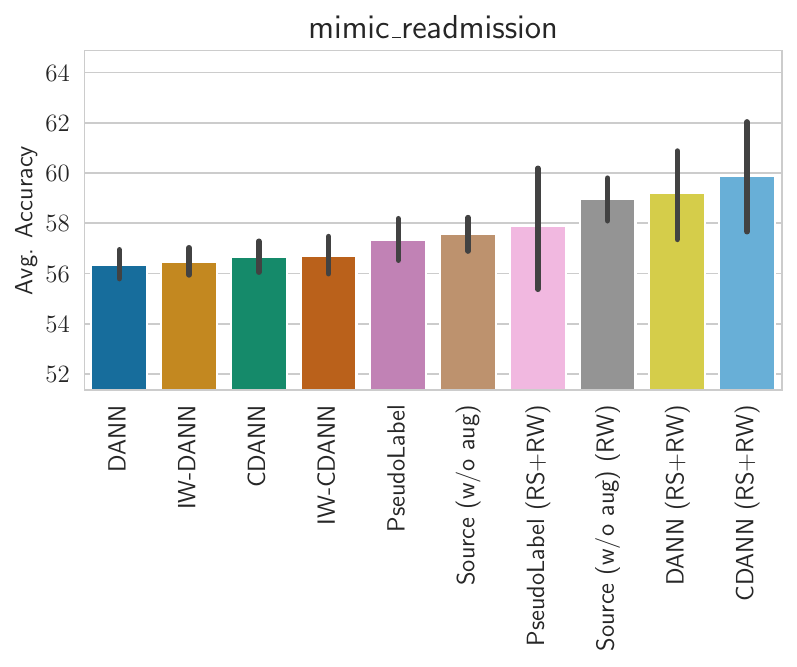} \\
    \includegraphics[width=0.5\linewidth]{figures/legend.pdf} \\
    \includegraphics[width=0.32\linewidth]{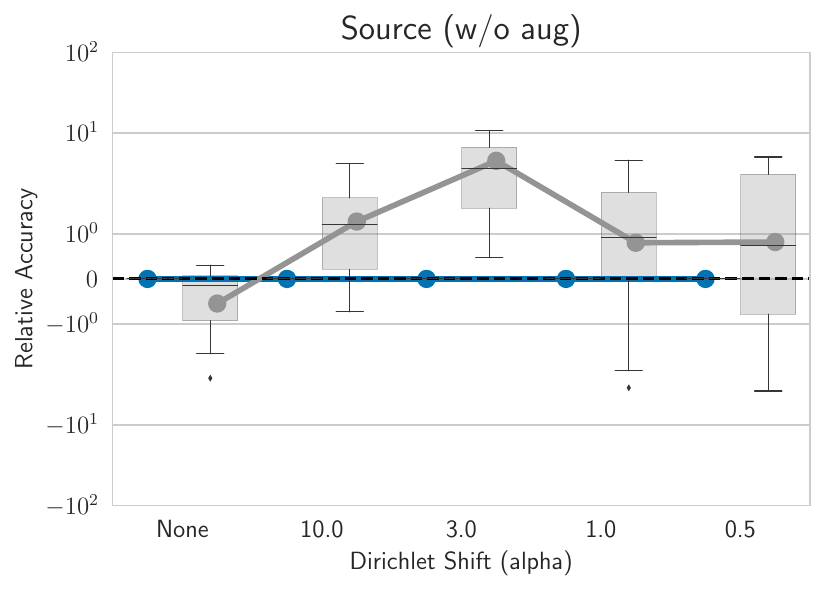} \hfil 
    \includegraphics[width=0.32\linewidth]{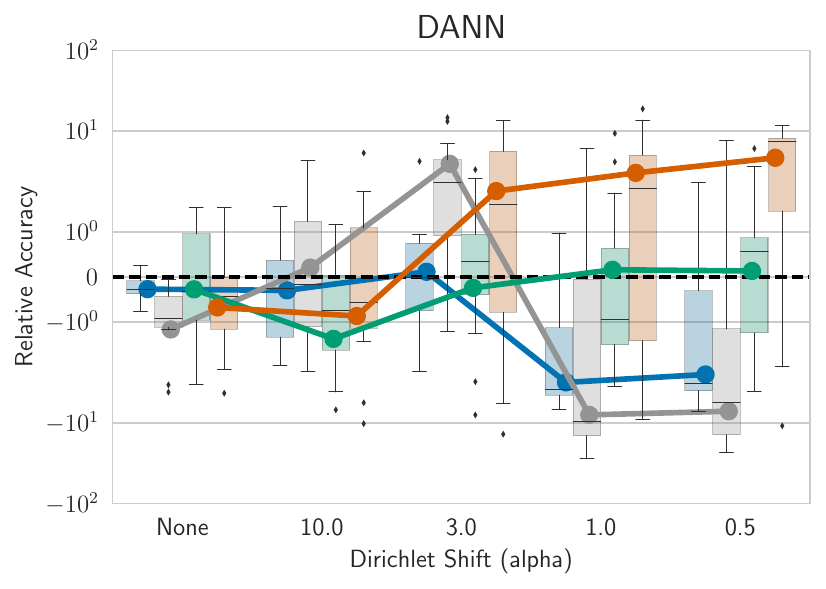} \hfil 
    \includegraphics[width=0.32\linewidth]{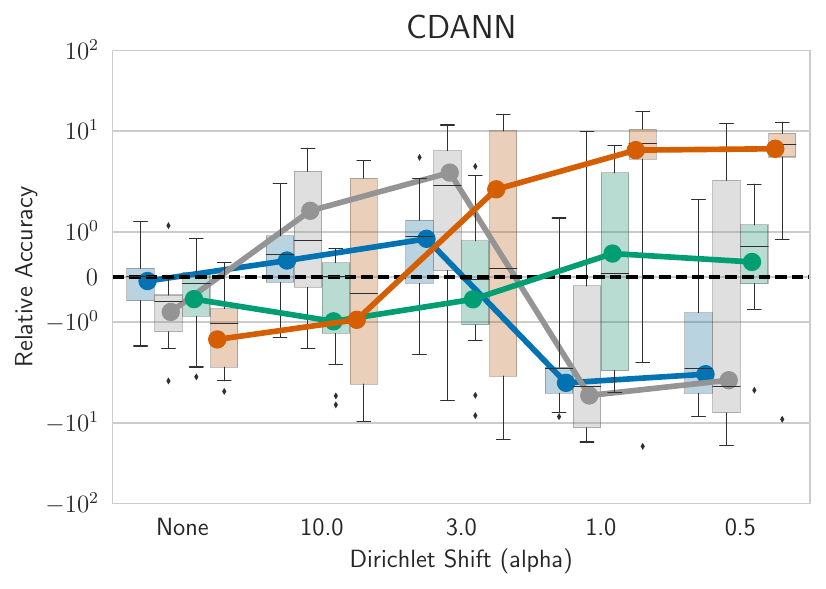}

    \includegraphics[width=0.32\linewidth]{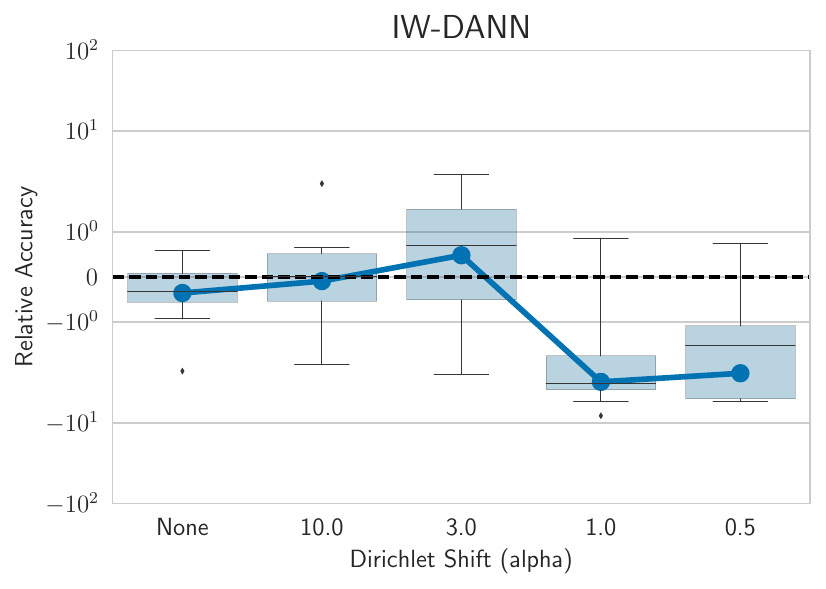}\hfil
    \includegraphics[width=0.32\linewidth]{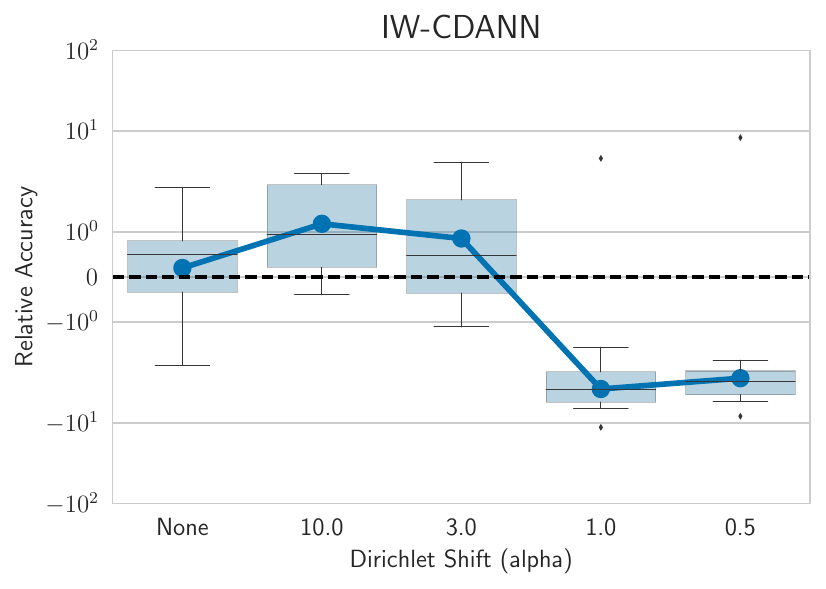}\hfil
    \includegraphics[width=0.32\linewidth]{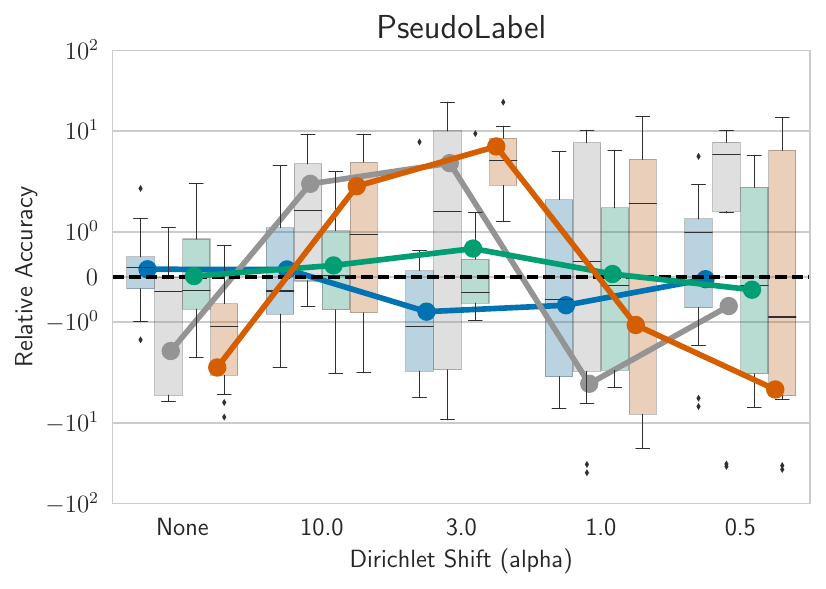}

    \caption{{Mimic Readmissions. Relative performance and accuracy plots for different DA algorithms across various shift pairs in Mimic Readmissions.}}
\end{figure}

\clearpage

\section{{Aggregate Accuracy with Different DA methods on Each Dataset}}
\label{app:tabular_results}

\begin{table}[h!]
    \centering
    \begin{adjustbox}{width=0.7\columnwidth,center}
    \footnotesize
     \setlength{\tabcolsep}{2.8pt}
     \renewcommand{\arraystretch}{1.3}
    \begin{tabular}{lcccccc}
\toprule
      Dataset &  Source &  DANN &  IW-DANN &  CDANN &  IW-CDANN &  PseudoLabel \\
\midrule
Civilcomments &   86.85 & 86.62 &    86.95 &  86.91 &     87.16 &         87.4 \\
\bottomrule
\end{tabular}

    \end{adjustbox}
    
    \vspace{5pt}
       \begin{adjustbox}{width=0.8\columnwidth,center}
    \footnotesize
     \setlength{\tabcolsep}{2.8pt}
     \renewcommand{\arraystretch}{1.3}
    \begin{tabular}{llcclcccclcccclcccc}
    \toprule
        \multirow{3}{*}{Dataset} & {} &  \multicolumn{2}{c}{\textbf{Source}} & {} &  \multicolumn{4}{c}{\textbf{DANN}} & {} &  \multicolumn{4}{c}{\textbf{CDANN}} & {} & \multicolumn{4}{c}{\textbf{PseudoLabel}} \\
         {} & {}  & \multirow{2}{*}{\parbox{0.5cm}{\centering None}}  & {} \multirow{2}{*}{\parbox{0.5cm}{\centering RW}} {} & {} & \multirow{2}{*}{\parbox{0.5cm}{\centering None}} & {} \multirow{2}{*}{\parbox{0.5cm}{\centering RW}} {} & {} \multirow{2}{*}{\parbox{0.5cm}{\centering RS}} & \multirow{2}{*}{\parbox{0.6cm}{\centering RS+ RW}} & {} & \multirow{2}{*}{\parbox{0.5cm}{\centering None}} & {} \multirow{2}{*}{\parbox{0.5cm}{\centering RW}} {} & {} \multirow{2}{*}{\parbox{0.5cm}{\centering RS}} & \multirow{2}{*}{\parbox{0.6cm}{\centering RS+ RW}} & {} & \multirow{2}{*}{\parbox{0.5cm}{\centering None}} & {} \multirow{2}{*}{\parbox{0.5cm}{\centering RW}} {} & {} \multirow{2}{*}{\parbox{0.5cm}{\centering RS}} & \multirow{2}{*}{\parbox{0.6cm}{\centering RS+ RW}}  \\
         & & & & & && & & & \\
\midrule
Civilcomments & {} &    86.8 &         89.1 & {} &  86.6 &           88.8 &         87.1 &              88.8 & {} &   86.9 &        89.0 &      86.9 &           88.9 & {} &         87.4 &           89.3 &         86.9 &              88.6 \\
\bottomrule
\end{tabular}

    \end{adjustbox}
    \caption{\update{\emph{Results with different DA methods on NLP datasets aggregated across target label marginal shifts}.}}
\end{table}

\begin{table}[h!]
    \centering
    \begin{adjustbox}{width=0.7\columnwidth,center}
    \footnotesize
     \setlength{\tabcolsep}{2.8pt}
     \renewcommand{\arraystretch}{1.3}
    \begin{tabular}{lcccccc}
\toprule
          Dataset &  Source &  DANN &  IW-DANN &  CDANN &  IW-CDANN &  PseudoLabel \\
\midrule
   Retiring Adult &   77.44 & 77.17 &    77.35 &  78.15 &     78.44 &        78.30 \\
Mimic Readmission &   57.57 & 56.36 &    56.48 &  56.67 &     56.71 &        57.35 \\
\bottomrule
\end{tabular}

    \end{adjustbox}
    
    \vspace{5pt}
    \begin{adjustbox}{width=0.8\columnwidth,center}
    \footnotesize
     \setlength{\tabcolsep}{2.8pt}
     \renewcommand{\arraystretch}{1.3}
    \begin{tabular}{llcclcccclcccclcccc}
    \toprule
        \multirow{3}{*}{Dataset} & {} &  \multicolumn{2}{c}{\textbf{Source}} & {} &  \multicolumn{4}{c}{\textbf{DANN}} & {} &  \multicolumn{4}{c}{\textbf{CDANN}} & {} & \multicolumn{4}{c}{\textbf{PseudoLabel}} \\
         {} & {}  & \multirow{2}{*}{\parbox{0.5cm}{\centering None}}  & {} \multirow{2}{*}{\parbox{0.5cm}{\centering RW}} {} & {} & \multirow{2}{*}{\parbox{0.5cm}{\centering None}} & {} \multirow{2}{*}{\parbox{0.5cm}{\centering RW}} {} & {} \multirow{2}{*}{\parbox{0.5cm}{\centering RS}} & \multirow{2}{*}{\parbox{0.6cm}{\centering RS+ RW}} & {} & \multirow{2}{*}{\parbox{0.5cm}{\centering None}} & {} \multirow{2}{*}{\parbox{0.5cm}{\centering RW}} {} & {} \multirow{2}{*}{\parbox{0.5cm}{\centering RS}} & \multirow{2}{*}{\parbox{0.6cm}{\centering RS+ RW}} & {} & \multirow{2}{*}{\parbox{0.5cm}{\centering None}} & {} \multirow{2}{*}{\parbox{0.5cm}{\centering RW}} {} & {} \multirow{2}{*}{\parbox{0.5cm}{\centering RS}} & \multirow{2}{*}{\parbox{0.6cm}{\centering RS+ RW}}  \\
         & & & & & && & & & \\
\midrule
   Retiring Adults & {} &    77.4 &         80.0 & {} &  77.2 &           79.5 &         77.4 &              79.4 & {} &   78.1 &        80.5 &      78.1 &           80.4 & {} &         78.3 &           80.8 &         78.5 &              80.8 \\
Mimic Readmissions & {} &    57.6 &         59.0 & {} &  56.4 &           55.1 &         57.3 &              59.2 & {} &   56.7 &        56.8 &      57.4 &           59.9 & {} &         57.4 &           57.7 &         57.7 &              57.9 \\
\bottomrule
\end{tabular}
    \end{adjustbox}
    
    \caption{\update{\emph{Results with different DA methods on tabular datasets aggregated across target label marginal shifts}.}}
\end{table}

\begin{table}[h!]
    \centering
    \begin{adjustbox}{width=0.8\columnwidth,center}
    \footnotesize
     \setlength{\tabcolsep}{0.6pt}
     \renewcommand{\arraystretch}{1.3}
    \begin{tabular}{lcccccccccccl}
    \toprule
    \multirow{2}{*}{\parbox{1.0cm}{\centering \textbf{Dataset}}}  &  \multirow{2}{*}{  \parbox{1.0cm}{\centering Source (wo aug)}} &  \multirow{2}{*}{  \parbox{1.0cm}{\centering Source (w aug)}}  &  \multirow{2}{*}{  \parbox{1.0cm}{\centering BN-adapt}} &  \multirow{2}{*}{  \parbox{1.0cm}{\centering TENT}} &  \multirow{2}{*}{  \parbox{1.0cm}{\centering DANN}} & \multirow{2}{*}{  \parbox{1.0cm}{\centering IW-DAN}} &  \multirow{2}{*}{  \parbox{1.0cm}{\centering CDAN}} &  \multirow{2}{*}{  \parbox{1.0cm}{\centering IW-CDAN}} &  \multirow{2}{*}{  \parbox{1.0cm}{\centering Fix-Match}} &  \multirow{2}{*}{  \parbox{1.0cm}{\centering Noisy-Student}} & \multirow{2}{*}{  \parbox{1.0cm}{\centering Sentry}} \\
    & & & & & & & & & \\

\midrule
    CIFAR-10  &             89.69 &           89.14 &     89.21 & 89.20 & 90.86 &    90.78 &  90.00 &     89.93 &     91.87 &         90.72 &   91.83 \\
   CIFAR-100 &             65.99 &           76.69 &     77.57 & 77.58 & 74.80 &    74.81 &  74.57 &     74.66 &     79.03 &         77.60 &   74.74 \\
       FMoW &             64.00 &           68.99 &     65.52 & 66.55 & 60.11 &    60.33 &  60.79 &     61.05 &     68.37 &         68.90 &   51.06 \\
   Camelyon &             77.42 &           76.95 &     85.70 & 82.48 & 86.66 &    85.89 &  85.45 &     84.27 &     86.29 &         79.29 &   86.81 \\
  Domainnet &             52.37 &           50.50 &     50.66 & 51.12 & 51.91 &    52.05 &  54.40 &     54.29 &     57.96 &         51.49 &   55.16 \\
   Entity13 &             76.93 &           80.07 &     77.99 & 78.04 & 78.26 &    78.75 &  79.74 &     79.28 &     80.25 &         80.37 &   73.58 \\
   Entity30 &             62.61 &           69.83 &     68.09 & 68.09 & 67.90 &    68.36 &  68.51 &     69.34 &     69.95 &         69.10 &   58.51 \\
   Living17 &             64.13 &           69.30 &     68.84 & 68.82 & 72.12 &    69.87 &  70.72 &     70.65 &     72.86 &         72.16 &   53.44 \\
Nonliving26 &             54.75 &           63.95 &     62.60 & 63.02 & 61.69 &    61.99 &  62.53 &     64.51 &     62.98 &         63.60 &   44.82 \\
 Officehome &             59.89 &           59.45 &     60.59 & 60.82 & 66.05 &    65.79 &  66.19 &     66.15 &     65.48 &         60.47 &   65.37 \\
      Visda &             58.47 &           53.41 &     59.98 & 60.96 & 69.69 &    69.79 &  72.55 &     72.80 &     72.02 &         53.51 &   72.23 \\
      \midrule
       \textbf{Avg}     &             66.02 &           68.94 &     69.70 & 69.70 & 70.92 &    70.77 &  71.40 &     71.54 &     73.37 &         69.75 &   66.14 \\
\bottomrule
\end{tabular}

    \end{adjustbox}
    
    \caption{\update{\emph{Results with different DA methods on vision datasets aggregated across target label marginal shifts}.
    While no single DA method performs consistently across different datasets, FixMatch seems to provide the highest aggregate improvement over a source-only classifier in our testbed.}}
\end{table}

\begin{table}[t]
    \centering
    \begin{adjustbox}{width=0.8\columnwidth,center}
    \footnotesize
     \setlength{\tabcolsep}{3pt}
     \renewcommand{\arraystretch}{1.3}
    \begin{tabular}{llcclcccclcccclcccc}
    \toprule
        \multirow{3}{*}{Dataset} & {} &  \multicolumn{2}{c}{\textbf{Source}} & {} &  \multicolumn{4}{c}{\textbf{BN-adapt}} & {} &  \multicolumn{4}{c}{\textbf{CDANN}} & {} & \multicolumn{4}{c}{\textbf{FixMatch}} \\
         {} & {}  & \multirow{2}{*}{\parbox{0.5cm}{\centering None}}  & {} \multirow{2}{*}{\parbox{0.5cm}{\centering RW}} {} & {} & \multirow{2}{*}{\parbox{0.5cm}{\centering None}} & {} \multirow{2}{*}{\parbox{0.5cm}{\centering RW}} {} & {} \multirow{2}{*}{\parbox{0.5cm}{\centering RS}} & \multirow{2}{*}{\parbox{0.6cm}{\centering RS+ RW}} & {} & \multirow{2}{*}{\parbox{0.5cm}{\centering None}} & {} \multirow{2}{*}{\parbox{0.5cm}{\centering RW}} {} & {} \multirow{2}{*}{\parbox{0.5cm}{\centering RS}} & \multirow{2}{*}{\parbox{0.6cm}{\centering RS+ RW}} & {} & \multirow{2}{*}{\parbox{0.5cm}{\centering None}} & {} \multirow{2}{*}{\parbox{0.5cm}{\centering RW}} {} & {} \multirow{2}{*}{\parbox{0.5cm}{\centering RS}} & \multirow{2}{*}{\parbox{0.6cm}{\centering RS+ RW}}  \\
         & & & & & && & & & \\
         \midrule
    CIFAR-10 & {} &            89.1 &                 89.4 & {} &      89.2 &           91.4 &         92.1 &              92.9 & {} &   90.0 &        91.3 &      91.4 &           92.5 & {} &      91.9 &           93.1 &         93.6 &              94.1 \\
   CIFAR-100 & {} &            76.7 &                 77.5 & {} &      77.6 &           78.8 &         77.9 &              79.0 & {} &   74.6 &        75.8 &      74.1 &           75.3 & {} &      79.0 &           79.6 &         79.1 &              79.8 \\
       FMoW& {} &            69.0 &                 70.3 & {} &      65.5 &           67.2 &         66.2 &              65.6 & {} &   60.8 &        61.9 &      57.0 &           55.2 & {} &      68.4 &           69.4 &         64.9 &              66.7 \\
   Camelyon & {} &            77.0 &                 77.9 & {} &      85.7 &           85.9 &         88.5 &              89.3 & {} &   85.5 &        85.8 &      87.9 &           88.5 & {} &      86.3 &           87.0 &         86.6 &              86.8 \\
  Domainnet & {} &            50.5 &                 48.2 & {} &      50.7 &           50.1 &         51.4 &              49.8 & {} &   54.4 &        54.2 &      54.7 &           54.3 & {} &      58.0 &           57.5 &         58.4 &              57.8 \\
   Entity13 & {} &            80.1 &                 80.9 & {} &      78.0 &           79.4 &         79.8 &              80.7 & {} &   79.7 &        80.2 &      80.6 &           81.4 & {} &      80.3 &           81.9 &         81.4 &              82.4 \\
   Entity30 & {} &            69.8 &                 70.1 & {} &      68.1 &           69.2 &         69.1 &              70.0 & {} &   68.5 &        69.6 &      69.4 &           70.5 & {} &      70.0 &           71.6 &         70.1 &              71.2 \\
   Living17 & {} &            69.3 &                 69.9 & {} &      68.8 &           69.7 &         69.6 &              70.1 & {} &   70.7 &        71.3 &      72.9 &           72.7 & {} &      72.9 &           72.8 &         72.3 &              71.9 \\
Nonliving26 & {} &            63.9 &                 64.5 & {} &      62.6 &           63.0 &         63.7 &              63.9 & {} &   62.5 &        62.9 &      63.8 &           64.0 & {} &      63.0 &           64.7 &         63.9 &              64.8 \\
 Officehome & {} &            59.4 &                 57.9 & {} &      60.6 &           60.5 &         60.9 &              60.4 & {} &   66.2 &       66.3 &      66.1 &           65.1 & {} &      65.5 &           64.9 &         66.5 &              66.1 \\
      Visda & {} &            53.4 &                 52.1 & {} &      60.0 &           60.6 &         59.5 &              58.8 & {} &   72.6 &        72.6 &      75.3 &           75.3 & {} &      72.0 &           72.5 &         73.5 &              73.8 \\
           \midrule 
         \textbf{Avg}   & {} &            68.9 &                 69.0 & {} &      69.7 &           70.5 &         70.8 &              70.9 & {} &   71.4 &   
            72.0 &      72.1 &           72.3 & {} &      73.4 &           74.1 &         73.7 &              74.1 \\
    \bottomrule
\end{tabular}
    
    \end{adjustbox}
    \vspace{5pt}

     \begin{adjustbox}{width=0.7\columnwidth,center}
    \footnotesize
     \setlength{\tabcolsep}{3pt}
     \renewcommand{\arraystretch}{1.3}
     \begin{tabular}{llcccclcccclcccc}
            \toprule
                \multirow{3}{*}{Dataset} & {} &  \multicolumn{4}{c}{\textbf{TENT}} & {} &  \multicolumn{4}{c}{\textbf{DANN}} & {} & \multicolumn{4}{c}{\textbf{NoisyStudent}} \\
                 {} & {}  & \multirow{2}{*}{\parbox{0.5cm}{\centering None}} & {} \multirow{2}{*}{\parbox{0.5cm}{\centering RW}} {} & {} \multirow{2}{*}{\parbox{0.5cm}{\centering RS}} & \multirow{2}{*}{\parbox{0.6cm}{\centering RS+ RW}} & {} & \multirow{2}{*}{\parbox{0.5cm}{\centering None}} & {} \multirow{2}{*}{\parbox{0.5cm}{\centering RW}} {} & {} \multirow{2}{*}{\parbox{0.5cm}{\centering RS}} & \multirow{2}{*}{\parbox{0.6cm}{\centering RS+ RW}} & {} & \multirow{2}{*}{\parbox{0.5cm}{\centering None}} & {} \multirow{2}{*}{\parbox{0.5cm}{\centering RW}} {} & {} \multirow{2}{*}{\parbox{0.5cm}{\centering RS}} & \multirow{2}{*}{\parbox{0.6cm}{\centering RS+ RW}}  \\
                 & & & &&& & & & \\
                 \midrule
    CIFAR-10 & {} &  89.2 &       91.4 &     92.1 &          92.9 & {} &  90.9 &       92.3 &     91.5 &          92.6 & {} &          90.7 &               90.8 &             90.6 &                  90.7 \\
   CIFAR-100& {} &  77.6 &       78.8 &     78.0 &          79.0 & {} &  74.8 &       75.9 &     74.8 &          76.1 & {} &          77.6 &               78.0 &             77.9 &                  78.0 \\
       FMoW & {} &  66.6 &       67.4 &     66.7 &          66.1 & {} &  60.1 &       61.6 &     56.4 &          54.5 & {} &          68.9 &               69.8 &             67.1 &                  68.0 \\
   Camelyon & {} &  82.5 &       82.7 &     87.8 &          88.9 & {} &  86.7 &       87.3 &     88.4 &          88.8 & {} &          79.3 &               79.1 &             79.2 &                  79.3 \\
  Domainnet & {} &  51.1 &       50.6 &     51.8 &          50.3 & {} &  51.9 &       52.1 &     53.6 &          53.5 & {} &          51.5 &               49.8 &             51.3 &                  49.5 \\
   Entity13 & {} &  78.0 &       79.5 &     79.8 &          80.8 & {} &  78.3 &       79.4 &     79.7 &          80.8 & {} &          80.4 &               81.5 &             80.6 &                  81.7 \\
   Entity30 & {} &  68.1 &       69.2 &     69.1 &          70.1 & {} &  67.9 &       69.2 &     69.0 &          69.8 & {} &          69.1 &               70.1 &             69.3 &                  70.3 \\
   Living17 & {} &  68.8 &       69.7 &     69.6 &          70.1 & {} &  72.1 &       73.0 &     71.8 &          72.3 & {} &          72.2 &               71.1 &             69.3 &                  69.4 \\
Nonliving26 & {} &  63.0 &       63.4 &     63.3 &          63.8 & {} &  61.7 &       62.4 &     63.1 &          63.0 & {} &          63.6 &               64.3 &             63.2 &                  64.8 \\
 Officehome & {} &  60.8 &       60.4 &     60.9 &          60.4 & {} &  66.1 &       66.1 &     66.5 &          65.3 & {} &          60.5 &               59.5 &             60.8 &                  59.5 \\
      Visda & {} &  61.0 &       61.5 &     60.3 &          59.6 & {} &  69.7 &       69.9 &     73.1 &          73.2 & {} &          53.5 &               51.5 &             55.7 &                  54.3 \\
    \midrule
    \textbf{Avg} & {} &  69.7 &       70.4 &     70.8 &          71.1 & {} &  70.9 &       71.7 &     71.6 &          71.8 & {} &          69.7 &               69.6 &             69.5 &                  69.6 \\
\bottomrule
\end{tabular}
    \end{adjustbox}
     \caption{\update{\emph{Results with DA methods paired with re-sampling (RS) and re-weighting (RW) correction (with RLLS estimate) aggregated across target label marginal shifts for vision datasets.}  
    RS and RW seem to help for all datasets and they both together significantly improve aggregate performance over no correction for all DA methods.}}
\end{table}

\clearpage

\section{Description of Deep Domain Adaptation Methods}
\label{app:methods}

In this section, we summarize deep DA methods compared in our $\rlsbench$ testbed. We also discuss how each method combines with our meta-algorithm to handle shift in class proportion. 

\subsection{Source only training}

We consider empirical risk minimization on the labeled source data 
as a baseline. Since this simply ignores the unlabeled target data, we call this as source only training. 
As mentioned in the main paper, we perform source only training with and without data augmentations. Formally, we minimize the following ERM loss: 
\begin{equation}
    L_{\text{source only}} (f) = \frac{1}{n}\sum_{i=1}^n \ell ( f(T(x_i), y_i) ) \,,
\end{equation}
where $T$ is the stochastic data augmentation operation \update{for vision datasets} and $\ell$ is a loss function. 
\update{For NLP and tabular datasets, $T$ is  the identity function.} 
Throughout the paper, we use cross-entropy loss minimization. Unless specified otherwise, we use strong augmentations as the data augmentation technique \update{for vision datasets. For NLP and tabular datasets, we do not use any data augmentation.} 

As mentioned in the main paper, we do not include re-sampling results with a source only model as it is trained only on source data and we observed no differences with just balancing the source data (as for most datasets source is already balanced) in our experiments. 
After obtaining a classifier $f$, we can first estimate the target label marginal and then adjust the classifier $f$ with post-hoc re-weighting with importance ratios $w_t(y) = \wh p_t(y)/ \wh p_s(y)$.

\paragraph{Adversarial training of a source only model} Along with standard training of a source only model with data augmentation, we experiment with adversarially robust models~\citep{madry2017towards}. To train adversarially robust models, we replace the standard ERM objective with a robust risk minimization objective: 
\begin{equation}
    L_{\text{source only (adv)}} (f) = \frac{1}{n} \sum_{i = 1} ^n \ell(R(T(x_i), y_i), y_i ) \,, 
\end{equation}
where $R(\cdot)$ performs the adversarial augmentation. In our paper, we use targeted Projected Gradient Descent (PGD) attacks with $\ell_2$ perturbation model.

\subsection{Domain-adversarial training methods}

Domain-adversarial trianing methods aim to learn domain invariant feature representations. 
These methods aimed at practical problems with non-overlapping support 
and are motivated by theoretical results showing that the gap between in- and
out-of-distribution performance depends on some measure of divergence between the source and
target distributions~\citep{ben2010theory, ganin2016domain}. 
While simultaneously minimizing the source error, these methods align the representations between source and target distribution. To perform alignment, these methods penalize divergence 
between feature representations across domains, encouraging the
model to produce feature representations that are similar across domain. 

Before describing these methods, we first define some notation. Consider a model $f = g \circ h$, where $h : \calX \to \Real^d$ is the featurizer that maps the inputs to some $d$ dimensional feature space, and the head $g: \Real^d \to \Delta^{k-1}$ maps the features to the prediction space. 
Following \citet{sagawa2021extending}, with all of our domain invariant methods, we use strong augmentations with source and target data \update{for vision datasets. For NLP and tabular datasets, we do not use any data augmentation.} 

\paragraph{DANN} DANN was proposed in \citet{ganin2016domain}. 
DANN approximates the divergence 
between feature representations of source and target domain 
by leveraging a domain discriminator
classifier. Domain discriminator $f_d$ aims to discriminate between source and target domains.  
Given a batch of inputs from source and target, this deep network $f_d$ classifies whether the examples are from the source data or target data. In particular, the following loss function is used: 
\begin{equation}
    L_{\text{domain disc.}} (f_d) = \frac{1}{n} \sum_{i =1 }^n \ell ( f_d( h(T(x_i))), 0 ) + \frac{1}{m} \sum_{i =n+1 }^{n+m} \ell(f_d(h(T(x_i))), 1) \,,
\end{equation}
where $\{x_1, x_2, \ldots, x_n\}$ are $n$ source examples and  $\{x_{n+1}, \ldots, x_{m+n}\}$  are $m$ target examples. Overall, the following loss function is used to optimize models with DANN: 
\begin{equation}
    L_{\text{DANN}} (h, g, f_d) =  L_{\text{source only}} (g\circ h) - \lambda L_{\text{domain disc.}} (f_d) \,.  
\end{equation}

$L_{\text{DANN}} (h, g, f_d)$ is maximized with respect to the domain discriminator classifier and $L_{\text{DANN}} (h, g, f_d)$ minimized with respect to the underlying featurize and the source classifier. This is achieved by gradient reversal layer in practice. To train, three networks, we use three different learning rate $\eta_f, \eta_g,$ and $\eta_{f_d}$. We discuss these hyperparameter details in \appref{app:setup_details}. 
We adapted our DANN implementation from \citet{sagawa2021extending} and Transfer learning library~\citep{jiang2022transferability}. 

\paragraph{CDANN} Conditional Domain adversarial neural network is a variant of DANN~\citep{long2018conditional}. Here the domain discriminator is conditioned on the classifier $g$'s prediction. In particular, instead of training the domain discriminator on the representation output of $h$, these methods operate on the outer product between the feature presentation $h(x)$ at an input $x$ and the classifier's probabilistic prediction $f = g \circ h(x)$ (i.e., $h(x) \otimes f(x)$). Thus instead of training the domain discriminator classifier $f_d$ on the $d$ dimensional input space, they train it on $d\times k$ dimensional space. In particular, the following loss function is used:  
\begin{equation}
    L_{\text{CDAN domain disc.}} (f_d, g, h) = \frac{1}{n} \sum_{i =1 }^n \ell ( f_d( f \otimes h(T(x_i))), 0 ) + \frac{1}{n} \sum_{i =n+1 }^{n+m} \ell(f_d(  f \otimes h(T(x_i))), 1) \,,
\end{equation}
where $\{x_1, x_2, \ldots, x_n\}$ are $n$ source examples and  $\{x_{n+1}, \ldots, x_{m+n}\}$  are $m$ target examples. The overall loss is the same as DANN where $L_{\text{domain disc.}} (f_d)$ is replaced with 
$ L_{\text{CDAN domain disc.}} (f_d, g, h)$. 

We adapted our implementation for CDANN from Transfer learning library~\citep{jiang2022transferability}. 

To adapt DANN and CDANN to our meta algorithm, at each epoch we can perform re-balancing of source and target data as in Step 1 and 4 of \algoref{alg:LSBench}. After obtaining the classifier $f$, we can use this classifier 
to first obtain an estimate of the target label marginal and then perform re-weighting adjustment with the obtained estimate.

\paragraph{IW-DANN and IW-CDANN} \citet{tachet_domain_2020} proposed training with importance re-weighting correction with DANN and CDANN objectives to accommodate for the shift in the target label proportion. 
In particular, at every epoch of training they first estimate 
the importance ratio $\wh w_t$ (with BBSE on training source and training target data) and then re-weight the  domain discriminator objective and ERM objective. %
In particular, the domain discriminator loss for IW-DANN can be written as: 

\begin{equation}
    L^{\wh w}_{\text{domain disc.}} (f_d) = \frac{1}{n} \sum_{i =1 }^n \wh w(y_i) \ell ( f_d( h(T(x_i))), 0 ) + \frac{1}{n} \sum_{i =n+1 }^{n+m} \ell(f_d(h(T(x_i))), 1) \,,
\end{equation}
where we multiply the source loss with importance weights. Similarly, we can re-write the source only training objective with importance re-weighting as follows: 
\begin{equation}
    L_{\text{source only}}^{\wh w} (f) = \frac{1}{n}\sum_{i=1}^n \wh w(y_i) \ell ( f(T(x_i), y_i) ) \,.
\end{equation}

Overall, the following objective is used to optimize models with IW-DANN: 
\begin{equation}
    L_{\text{IW-DANN}} (h, g, f_d) =  L^{\wh w}_{\text{source only}} (g\circ h) - \lambda L^{\wh w}_{\text{domain disc.}} (f_d) \,, 
\end{equation}
where the importance weights are updated after every epoch with classifier obtained in previous step. Similarly,  with using importance re-weights with the CDANN objective, we obtain IW-CDANN objective.

In population, IW-CDANN and IW-DANN correction matches the correction with our meta-algorithm for DANN and CDANN. However, the behavior this importance re-weighting correction can be different from our meta-algorithm for over-parameterized models with finite data~\citep{byrd2019effect}. Recent empirical and theoretical findings have highlighted that importance re-weighting have minor to no effect on overparameterized models when trained for several epochs~\citep{byrd2019effect, xu2021understanding}. On the other hand, with finite samples, re-sampling (when class labels are available) has shown  different and promising empirical behavior~\citep{an2020resampling, idrissi2022simple}.  
This may highlight the differences in the behavior of IW-CDANN (or IW-DANN)  with our meta algorithm on CDANN (or DANN).

We refer to the implementation provided by the authors~\citep{tachet_domain_2020}. 

\subsection{Self-training methods}

Self-training methods leverage unlabeled data by `pseudo-labeling' unlabeled examples with the
classifier's own predictions and training on them as if they were labeled examples.
Recent self-training methods also often make use of consistency regularization, for example, encouraging the model to make similar predictions on augmented versions of unlabeled
example. 
In our work, we experiment with the following methods: 

\paragraph{\update{PseudoLabel}} \update{\citep{lee2013pseudo} proposed PseudoLabel that leverages unlabeled examples with classifier's own prediction. This algorithm dynamically generates psuedolabels and overfits on them in each batch. 
In particular, while pseudolabels are generated on unlabeled examples, the loss is computed with respect to the same label. 
PseudoLabel only overfits to the assigned label if the confidence of the prediction is greater than some threshold $\tau$.} 

\update{Refer to $T$ as the data-augmentation technique (i.e., identity for NLP and tabular datasets and strong augmentation for vision datasets). Then, PseudoLabel uses the following loss function: 
\begin{align*}
    L_{\text{PseudoLabel}} (f) &= \frac{1}{n}\sum_{i=1}^n \ell ( f(T(x_i), y_i) ) + \frac{\lambda_t}{m} \sum_{i=n+1}^{m+n} \ell ( f(T(x_i), \wt y_i) ) \cdot \indict{\max_y f_y(T (x_i)) \ge \tau }\,,  
\end{align*}
where $\wt y_i = \argmax_y f_y(T(x_i))$. 
PseudoLabel increases $\lambda_t$ between labeled and unlabeled losses over epochs,
initially placing 0 weight on  unlabeled loss and then linearly increasing the unlabeled loss weight until 
it reaches
the full value of hyperparameter $\lambda$ at some threshold step.
We fix the step at which $\lambda_t$ reaches its
maximum value $\lambda$ be 40\% of the total number of training steps, matching the implementation to \citep{sohn2020fixmatch, sagawa2021extending}.}

\paragraph{FixMatch} \citet{sohn2020fixmatch} proposed FixMatch as a variant of the simpler Pseudo-label method~\citep{lee2013pseudo}. This algorithm dynamically generates psuedolabels and overfits on them in each batch. FixMatch employs consistency regularization on the  unlabeled data. 
In particular, while pseudolabels are generated on a
weakly augmented view of the unlabeled examples, the loss is computed with respect to predictions
on a strongly augmented view. The intuition behind such an update is to encourage a model to make predictions on weakly augmented data consistent with the strongly augmented example. Moreover, FixMatch only overfits to the assigned labeled with weak augmentation if the confidence of the prediction with strong augmentation is greater than some threshold $\tau$. 

Refer to $T_{\text{weak}}$ as the weak-augmentation and $T_{\text{strong}}$ as the strong-augmentation function. Then, FixMatch uses the following loss function: 
\begin{align*}
    L_{\text{FixMatch}} (f) &= \frac{1}{n}\sum_{i=1}^n \ell ( f(T_{\text{strong}}(x_i), y_i) ) \\ 
    &+ \frac{\lambda}{m} \sum_{i=n+1}^{m+n} \ell ( f(T_{\text{strong}}(x_i), \wt y_i) ) \cdot \indict{\max_y f_y(T_{\text{strong}} (x_i)) \ge \tau }\,,  
\end{align*}
where $\wt y_i = \argmax_y f_y(T_{\text{weak}} (x_i))$. 
We adapted our implementation from \citet{sagawa2021extending} which matches the implementation of \citet{sohn2020fixmatch} except for one detail. While \citet{sohn2020fixmatch} augments labeled examples with weak augmentation, \citet{sagawa2021extending} proposed to strongly augment the labeled source examples. 

\paragraph{NoisyStudent} \citet{xie2020self} proposed a different variant of Pseudo-labeling. Noisy Student generates pseudolabels, fixes them, and then trains the model (from scratch) until convergence before generating new pseudolabels.
Contrast it with FixMatch and PseudoLabel which dynamically generate pseudolabels.  
The first set of pseudolabels are obtained by training an initial teacher model only on the source labeled data. Then in each iteration, randomly initialized models fit the labeled source data and pseudolabeled target data with pseudolabels assigned by the converged model in the previous iteration. Noisy student objective can be summarized as: 

\begin{align*}
    L_{\text{NoisyStudent}} (f^N) = \frac{1}{n}\sum_{i=1}^n \ell ( f^N(T_{\text{strong}}(x_i), y_i) ) +  \frac{1}{m} \sum_{i=n+1}^{m+n} \ell ( f^N(T_{\text{strong}}(x_i), \wt y_i) )\,,  
\end{align*}
where $\wt y_i = \argmax_y f^{N-1}_y(T_{\text{weak}} (x_i))$ is computed with the classifier obtained at $N-1$ step.  Note that the randomly initialized model at each iteration uses a dropout of $p=0.5$ in the penultimate layer.  We adopted our implementation of NoisyStudent to \citet{sagawa2021extending}. To initialize the initial teacher model, we use the source-only model trained with strong augmentations without dropout.

\paragraph{SENTRY} \citet{prabhu2021sentry} proposed a different variant of pseudolabeling method. This method is aimed to tackle DA under relaxed label shift scenario. a SENTRY incorporates a target instance based on its predictive consistency under a committee of strong image transformations. In particular, SENTRY makes N strong augmentations of an unlabeled target example and makes a prediction on those. If the majority of the committee matches the prediction on the sample example with weak-augmentation then entropy is minimized on that example, otherwise the entropy is maximized. Moreover, the authors employ an 'information-entropy' objective aimed to match the prediction at every example with the estimated target label marginal. Overall the SENTRY objective is defined as follows: 
\begin{align*}
L_{\text{SENTRY}}(f) &=  \frac{1}{n}\sum_{i=1}^n \ell ( f(T_{\text{strong}}(x_i), y_i) )  + \frac{1}{m} \sum_{i=n+1}^{m+n} \sum_{j=1}^k f_k(y = j| x_i) \log(\wt p_t(y = j)) \\ &\quad + \lambda_{\text{unsup}} \frac{1}{m} \sum_{i=n+1}^{m+n} \sum_{j=1}^k  - f_k(y = j| x_i) \log(f_k(y = j| x_i)) \cdot (2 l(x) - 1) \,, 
\end{align*}
where $l(x) \in \{0,1\}$ is majority vote output of the committee consistency. For more details, we refer the reader to \citet{prabhu2021sentry}.  Additionally, at each training epoch, SENTRY balances the source data and pseudo-balances the target data. We adopted our implementation with the official implementation in \citet{prabhu2021sentry} with minor differences. In particular, to keep the implementation consistent with all the other DA methods, we train with the objective above from scratch instead of training sequentially after a initialization with source-only classifier as in the original paper~\citep{prabhu2021sentry}.

\update{Since Fix-Match, NoisyStuent, and Sentry use strong data-augmentations in their implementation, the applicability of these algorithms is restricted to vision datasets. For NLP and tabular datasets, we only train models with PseudoLabel as it doesn't rely on any augmentation technique.}

\subsection{Test-time training methods} 

These take an already trained source model and adapt a few parameters (e.g. batch norm parameters, batch norm statistics) on the unlabeled target data with an aim to improve target performance. \update{Hence, we restrict these methods to vision datasets with architectures that use batch norm.}
These methods are computationally cheaper than other DA methods in the suite as they adapt a classifier on-the-fly.  
We include the following methods in our experimental suite:

\paragraph{BN-adapt} \citet{li2016revisiting} proposed batch norm adaptation. More recently, \citet{schneider2020improving} showed gains with BN-adapt on common corruptions benchmark. Batch norm adaptation is applicable for deep models with batch norm parameters. With this method we simply adapt the Batchnorm statistics, in particular, mean and std of each batch norm layer.

\paragraph{TENT} \citet{wang2021tent} proposed optimizing batch norm parameters to minimize the entropy of the predictor on the unlabeled target data. In our implementation of TENT, we perform BN-adapt before learning batch norm parameters.

\paragraph{CORAL} \citet{sun2016return} proposed CORAL to adapt a model trained on the source to target by whitening the feature representations. In particular, say $\wh \Sigma_s$ is the empirical covariance of the target data representations and $\Sigma_s$ is the empirical covariance of the source data representations, CORAL adjusts a linear layer $g$ on target by re-training the final layer on the outputs: $\Sigma_t^{1/2} \Sigma_s^{-1/2} h(x)$. DARE~\citep{rosenfeld2022domain} simplified the procedure and showed that this is equivalent to training a linear head $h$ on $\Sigma_s^{-1/2} h(x)$ and whitening target data representations with  $\Sigma_t^{-1/2} h(x)$ before input to the classifier. We choose to implement the latter procedure as it is cheap to train a single classifier in multi-domain datasets.

With our meta-algorithm, before adapting the source-only classifier 
with test time adaptation methods, 
we use it to perform the re-sampling correction. 
After obtaining the adapted classifier, 
we estimate target label marginal and use it to 
adjust the classifier with re-weighting.

\section{Hyperparameter and Architecture Details}
\label{app:setup_details}

\subsection{Architecture and Pretraining Details}

For all datasets, we used the same architecture across different algorithms: 

\begin{itemize}
    \item CIFAR-10: Resnet-18~\citep{he2016deep} pretrained on Imagenet
    \item CIFAR-100: Resnet-18~\citep{he2016deep} pretrained on Imagenet
    \item Camelyon: Densenet-121~\citep{huang2017densely} \emph{not} pretrained on Imagenet as per the suggestion made in \citep{wilds2021}
    \item FMoW: Densenet-121~\citep{huang2017densely} pretrained on Imagenet
    \item BREEDs (Entity13, Entity30, Living17, Nonliving26): Resnet-18~\citep{he2016deep} \emph{not} pretrained on Imagenet as per the suggestion in \citep{santurkar2020breeds}. The main rationale is to avoid pre-training on the superset dataset where we are simulating sub-population shift. 
    \item Officehome: Resnet-50~\citep{he2016deep} pretrained on Imagenet
    \item Domainnet:  Resnet-50~\citep{he2016deep} pretrained on Imagenet
    \item Visda:  Resnet-50~\citep{he2016deep} pretrained on Imagenet
    \item \update{Civilcomments: Pre-trained DistilBERT-base-uncased~\citep{Sanh2019DistilBERTAD}} 
    \item \update{Retiring Adults: We use an MLP with $2$ hidden layers and $100$ hidden units in both of the hidden layer}
    \item \update{Mimic Readmissions: We use the transformer architecture described in \citet{yao2022wildtime}\footnote{https://github.com/huaxiuyao/Wild-Time/.}}
\end{itemize}

Except for Resnets on CIFAR datasets, we used the standard pytorch implementation~\citep{gardner2018gpytorch}. For Resnet on cifar, we refer to the implementation here: \url{https://github.com/kuangliu/pytorch-cifar}. For all the architectures, whenever applicable, we add antialiasing~\citep{zhang2019shiftinvar}. 
We use the official library released with the paper. 

For imagenet-pretrained models with standard architectures, 
we use the publicly available models here: \url{https://pytorch.org/vision/stable/models.html}. 
For imagenet-pretrained models on the reduced input size images (e.g. CIFAR-10), we 
train a model on Imagenet on reduced input size from scratch. We include the model with our 
publicly available repository. For bert-based models, we use the publicly available models here: \url{https://huggingface.co/docs/transformers/}.

\subsection{Hyperparameters}

First, we tune learning rate and $\ell_2$ regularization parameter 
by fixing batch size for each dataset that correspond to maximum we can fit to 15GB GPU memory. 
We set the number of epochs for training as per the 
suggestions of the authors of respective benchmarks. 
Note that we define the number of epochs as a 
full pass over the labeled training source data.  
We summarize learning rate, batch size, number of epochs, and 
$\ell_2$ regularization parameter used in our study in \tabref{table:hyperparameter_dataset}.

\begin{table}[h!]
    \begin{adjustbox}{width=\columnwidth,center}
    \centering
    \small
    \tabcolsep=0.12cm
    \renewcommand{\arraystretch}{1.5}
    \begin{tabular}{lccccccccc}
    \toprule    
    Dataset && Epoch && Batch size && $\ell_2$ regularization && Learning rate \\
    \midrule
    CIFAR10 && 50 && 200 && 0.0001 (chosen from $\{0.0001, 0.001, $1e-5$, 0.0\}$) && 0.01  (chosen from $\{0.001, 0.01, 0.0001\}$) \\ 
    CIFAR100 && 50 && 200 && 0.0001 (chosen from $\{0.0001, 0.001, $1e-5$, 0.0\}$) && 0.01  (chosen from $\{0.001, 0.01, 0.0001\}$) \\ 
    Camelyon && 10 && 96 && 0.01 (chosen from $\{0.01, 0.001, 0.0001, 0.0\}$) && 0.03 (chosen from $\{0.003, 0.3, 0.0003, 0.03\}$) \\ 
    FMoW && 30 && 64 && 0.0 (chosen from $\{0.0001, 0.001, $1e-5$, $0.0$\}$) && 0.0001 (chosen from $\{0.001, 0.01, 0.0001\}$)  \\
    Entity13 && 40 && 256 && 5e-5 (chosen from $\{$5e-5, 5e-4, 1e-4, 1e-5$\}$) && 0.2 (chosen from $\{0.1, 0.5, 0.2, 0.01, 0.0\}$)\\
    Entity30 && 40 && 256 &&  5e-5 (chosen from $\{$5e-5, 5e-4, 1e-4, 1e-5$\}$) && 0.2 (chosen from $\{0.1, 0.5, 0.2, 0.01, 0.0\}$) \\
    Living17 && 40 && 256 &&  5e-5 (chosen from $\{$5e-5, 5e-4, 1e-4, 1e-5$\}$) && 0.2 (chosen from $\{0.1, 0.5, 0.2, 0.01, 0.0\}$)  \\ 
    Nonliving26 && 40 && 256 && 0  5e-5 (chosen from $\{$5e-5, 5e-4, 1e-4, 1e-5$\}$) && 0.2 (chosen from $\{0.1, 0.5, 0.2, 0.01, 0.0\}$)  \\ 
    Officehome && 50 && 96 && 0.0001 (chosen from $\{0.0001, 0.001, $1e-5$, 0.0\}$) &&0.01 (chosen from $\{0.001, 0.01, 0.0001\}$)   \\
    DomainNet && 15 &&  96 && 0.0001 (chosen from $\{0.0001, 0.001, $1e-5$, 0.0\}$)  && 0.01 (chosen from $\{0.001, 0.01, 0.0001\}$)  \\
    Visda &&  10 && 96  && 0.0001 (chosen from $\{0.0001, 0.001, $1e-5$, 0.0\}$) && 0.01 (chosen from $\{0.001, 0.01, 0.0001\}$) \\
    Civilcomments && 5 && 32  && 0.01 (chosen from $\{0.01, 0.001, 0.0001, 0.0\}$) && 2e-5 (chosen from $\{2e-4, 2e-5\}$)\\
    Retiring Adults && 50 && 200 && 0.0001 (chosen from $\{0.01, 0.001, 0.0001, 0.0\}$) && 0.01 (chosen from $\{0.001, 0.01, 0.0001\}$) \\
    Mimic Readmissions && 100 && 128 && 0.0 (chosen from $\{0.01, 0.001, 0.0001, 0.0\}$) && 5e-4 (chosen from $\{0.005,0.0001 0.0005\}$) \\
    \bottomrule 
    \end{tabular}  
    \end{adjustbox}  
    \caption{Details of the learning rate and batch size considered in our $\rlsbench$} \label{table:hyperparameter_dataset}
 \end{table}

For each algorithm, we use the hyperparameters reported in the initial papers.
For domain-adversarial methods (DANN and CDANN), we refer to the suggestions made in 
Transfer Learning Library~\citep{jiang2022transferability}.
We tabulate hyperparameters  for each algorithm next: 

\begin{itemize}
    \item \textbf{DANN, CDANN, IW-CDANN and IW-DANN ~~} As per Transfer Learning Library suggestion, we use a learning rate multiplier of $0.1$ for the featurizer when initializing with a pre-trained network and $1.0$ otherwise. We default to a penalty weight of $1.0$ for all datasets with pre-trained initialization. 
    \item \textbf{FixMatch~~} We use the lambda is 1.0 and use threshold $\tau$ as 0.9.  
    \item \textbf{NoisyStudent~~} We repeat the procedure for $2$ iterations and use a drop level of $p=0.5$. 
    \item \textbf{SENTRY~~}  We use $\lambda_{\text{src}} = 1.0$, $\lambda_{\text{ent}} = 1.0$, and $\lambda_{\text{unsup}} = 0.1$. We use a committee of size 3. 
    \item \update{\textbf{PsuedoLabel~~} We use the lambda is 1.0 and use threshold $\tau$ as 0.9.}
\end{itemize}

Recent works~\citep{deng2021labels, guillory2021predicting, chen2021detecting, jiang2021assessing, baek2022agreement, garg2022ATC} have proposed numerous heuristics to predict classifier performance under distribution shift. Analyzing the usefulness of these heuristics for hyperparameter selection is an interesting avenue for future work.

\subsection{Compute Infrastructure}

Our experiments were performed across a combination of Nvidia T4, A6000, P100 and V100 GPUs. 
Overall, to run the entire $\lsbench$ suite on a T4 GPU machine with 8 CPU cores
we would approximately need $70k$ GPU hours of compute.  

\subsection{Data Augmentation}

In our experiments, we leverage data augmentation techniques that 
encourage robustness to some variations between domains \update{for vision datasets}. 

For weak augmentation, we leverage random horizontal flips and random crops of pre-defined size.
For strong augmentation, we apply the following transformations sequentially: random horizontal flips, random crops of pre-defined size, augmentation with Cutout~\citep{devries2017improved}, and RandAugment~\citep{cubuk2020randaugment}.  For the exact
implementation of RandAugment, we directly use the implementation of \citet{sohn2020fixmatch}. 
The pool of operations includes: autocontrast, brightness, color jitter, contrast, equalize, posterize, rotation,
sharpness, horizontal and vertical shearing, solarize, and horizontal and vertical translations. We
apply N = 2 random operations for all experiments.

\section{Comparison with SENTRY on officehome dataset with different hyperparameters}
\label{app:sentry_more}
In this section, we shed more light on the discrepancy observed 
between SENTRY results reported in the original paper~\citep{prabhu2021sentry} and 
our implementation. 

We note that for the main experiments on Officehome dataset, we used a 
batch size of $96$ for all methods including SENTRY. 
However, SENTRY reported results with a batch size of 16 in their work. 
Hence, we re-run the SENTRY algorithm with a batch size of $16$. 
To investigate the impact of the decreased batch size, 
we make a comparison with FixMatch (the best algorithm on Officehome in our runs) 
by re-running it with the decreased batch size.

In \tabref{table:officehome_results} we report results on individual shift pairs in officehome. 
We observe that SENTRY improves over FixMatch for the default minor shift in the label distribution in the officehome dataset. 
However, as the shift severity increases we observe that SENTRY performance degrades. 
Overall, we observe that RS-FixMatch performs similar or superior to SENTRY on 3 out of 4 shift pairs in officehome.

\begin{table}[ht]
    
    \small
    \setlength{\tabcolsep}{5.2pt}
     \renewcommand{\arraystretch}{1.3}
    \begin{subtable}{\linewidth}\centering
    \begin{tabular}{lcccccc}
    \toprule
      Algorithm &  Alpha = None &  Alpha = 10.0 &  Alpha = 3.0 &  Alpha = 1.0 &  Alpha = 0.5 &  Avg \\
    \midrule
       FixMatch &          92.5 &          95.2 &         98.0 &        100.0 &        100.0 & 97.1 \\
    RS-FixMatch &          92.5 &          96.4 &         98.0 &        100.0 &        100.0 & 97.4 \\
         SENTRY &          93.0 &          94.0 &         98.0 &         83.3 &         87.5 & 91.2 \\
    \bottomrule
    \end{tabular}
    \caption{Product to Product (in-distribution)}
    \end{subtable}
    \vspace{5pt}

    \vspace{5pt}

    \begin{subtable}{\linewidth}\centering
   \begin{tabular}{lrrrrrr}
    \toprule
      Algorithm &  Alpha = None &  Alpha = 10.0 &  Alpha = 3.0 &  Alpha = 1.0 &  Alpha = 0.5 &  Avg \\
    \midrule
       FixMatch &          71.4 &          71.5 &         70.7 &         73.1 &         75.5 & 72.4 \\
    RS-FixMatch &          74.7 &          74.0 &         72.1 &         73.1 &         70.4 & 72.9 \\
         SENTRY &          78.1 &          78.0 &         75.1 &         71.7 &         65.3 & 73.6 \\
    \bottomrule
    \end{tabular}

    \caption{Product to Real}
    \end{subtable}
    \vspace{5pt}

    \begin{subtable}{\linewidth}\centering
    \begin{tabular}{lrrrrrr}
    \toprule
      Algorithm &  Alpha = None &  Alpha = 10.0 &  Alpha = 3.0 &  Alpha = 1.0 &  Alpha = 0.5 &  Avg \\
    \midrule
       FixMatch &          41.5 &          44.0 &         44.2 &         48.4 &         39.4 & 43.5 \\
    RS-FixMatch &          45.5 &          44.8 &         43.6 &         50.0 &         37.4 & 44.2 \\
         SENTRY &          45.8 &          46.5 &         41.4 &         40.3 &         27.3 & 40.3 \\
    \bottomrule
    \end{tabular}
    \caption{Product to ClipArt}
    \end{subtable}
    \vspace{5pt}
    
    \begin{subtable}{\linewidth}\centering
   \begin{tabular}{lrrrrrr}
    \toprule
      Algorithm &  Alpha = None &  Alpha = 10.0 &  Alpha = 3.0 &  Alpha = 1.0 &  Alpha = 0.5 &  Avg \\
    \midrule
       FixMatch &          54.4 &          51.3 &         54.7 &         57.3 &         55.9 & 54.7 \\
    RS-FixMatch &          57.2 &          53.6 &         55.9 &         57.3 &         58.8 & 56.6 \\
         SENTRY &          63.7 &          62.0 &         62.1 &         65.3 &         55.9 & 61.8 \\
    \bottomrule
    \end{tabular}

    \caption{Product to Art}
    \end{subtable}
    \vspace{5pt}
    
    \caption{Officehome results with batch size $16$ instead of $96$ used throughout our experiments.}\label{table:officehome_results}
 \end{table}

More generally, across our runs, we also observed  model training with SENTRY
to be unstable. Investigating further, we observe that the maximization objective to enforce 
consistency cause instabilities.
This behavior is specifically prevalent for experiments where we don't use initiale the underlying
model with pre-trained weights (for example, in BREEDs datasets).

\clearpage

\end{document}